\documentclass[
	a4paper,
	11pt,
	titlepage,
	bibliography=totoc,
	open=right, 
	captions=tableheading, 
	BCOR=12mm, 
]{scrbook}

\usepackage{xcolor}
\definecolor{aqua}{HTML}{009ee3}
\definecolor{softgray}{HTML}{706f6f}
\definecolor{vir1of4}{RGB}{68,1,84}
\definecolor{vir2of4}{RGB}{49,104,142}
\definecolor{vir3of4}{RGB}{53,183,121}
\definecolor{vir4of4}{RGB}{253,231,37}

\definecolor{vir0}{RGB}{68,1,84}
\definecolor{vir1}{RGB}{72,24,106}
\definecolor{vir2}{RGB}{71,45,123}
\definecolor{vir3}{RGB}{66,64,134}
\definecolor{vir4}{RGB}{59,82,139}
\definecolor{vir5}{RGB}{51,99,141}
\definecolor{vir6}{RGB}{44,114,142}
\definecolor{vir7}{RGB}{38,130,142}

\definecolor{ml2rblu}{rgb}{0.02,0.27,0.45}
\definecolor{ml2ryel}{rgb}{0.98,0.72,0.18}
\definecolor{ml2rgrn}{rgb}{0.50,0.71,0.18}
\definecolor{ml2rtrq}{rgb}{0.00,0.57,0.57}

\usepackage{fontspec}

\setkomafont{captionlabel}{\bfseries\sffamily}
\addtokomafont{caption}{\small}

\usepackage[pass]{geometry}
\usepackage[parfill]{parskip}

\usepackage{hyperref}
\hypersetup{hidelinks}

\usepackage[
	backend=biber,
	style=authoryear,
    minnames=1,
    maxnames=1,
    maxbibnames=99,
	defernumbers=true,
	dashed=false,
	url=false,
	doi=true,
	eprint=false,
	autolang=other,
]{biblatex}
\DeclareSourcemap{
	\maps[datatype=bibtex, overwrite]{
		\map{
			\step[fieldset=editor, null]
            \step[fieldset=isbn, null]
            \step[fieldset=issn, null]
			\step[fieldset=series, null]
			\step[fieldset=volume, null]
			\step[fieldset=number, null]
			\step[fieldset=address, null]
			\step[fieldsource=groups, final]
			\step[fieldset=keywords, fieldvalue={,}, append]
			\step[fieldset=keywords, origfieldval, append]
			\step[fieldsource=keywords, match=\regexp{\A,}, replace={}]
		}
		\map{
			\pernottype{book} 
			\step[fieldset=publisher, null] 
		}
	}
}
\defbibenvironment{description}
{\list
	{\printtext[labelnumberwidth]{%
			\printfield{labelprefix}%
			\printfield{labelnumber}}}
	{\setlength{\labelwidth}{\labelnumberwidth}%
		\setlength{\leftmargin}{\labelwidth}%
		\setlength{\labelsep}{\biblabelsep}%
		\addtolength{\leftmargin}{\labelsep}%
		\setlength{\itemsep}{10.0\baselineskip}%
        \setlength{\topsep}{10.0\baselineskip}%
		\setlength{\parsep}{\bibparsep}}%
	}
{\endlist}
{\item}
\makeatletter
\@removefromreset{footnote}{chapter}
\makeatother

\addtotoclist{loor}
\newcommand\onlineresource[3][]{%
	\footnote{\url{#2}\label{#3}#1\addxcontentsline{loor}{section}[\thefootnote]{\nolinkurl{#2}}}
}
\newcommand{\listofloorname}{List of Footnotes and Online~Resources}
\newcommand{\listofonlineresources}{\listoftoc{loor}}
\setuptoc{loor}{chapteratlist}

\usepackage{bm}
\usepackage{amsmath}
\usepackage[capitalise]{cleveref}
\usepackage{amsthm}
\usepackage{thmtools}
\declaretheoremstyle[
headfont=\bfseries\sffamily,
notefont=\normalfont\bfseries,
bodyfont=\normalfont\itshape,
parent=chapter
]{definition}
\declaretheorem[style=definition]{definition}

\declaretheoremstyle[
headfont=\bfseries\sffamily,
notefont=\normalfont\bfseries,
bodyfont=\normalfont,
parent=chapter
]{theorem}
\declaretheorem[style=theorem]{theorem}
\declaretheorem[style=theorem]{proposition}
\declaretheorem[style=theorem]{lemma}
\declaretheorem[style=theorem]{corollary}
\declaretheoremstyle[
headfont=\itshape\sffamily,
notefont=\normalfont\itshape,
bodyfont=\normalfont,
parent=chapter
]{remark}

\declaretheorem[style=remark]{example}
\declaretheoremstyle[
headfont=\bfseries\sffamily,
notefont=\normalfont\bfseries,
bodyfont=\normalfont\itshape
]{rq}
\declaretheorem[style=rq,name=Research Question]{research_question}

\crefformat{research_question}{Research Question~#2#1#3}

\usepackage[chapter]{algorithm}
\usepackage{algorithmic}

\usepackage{etoolbox}
\makeatletter
\patchcmd{\@chapter}
{\chaptermark{#1}}
{\chaptermark{#1}%
	\addtocontents{loa}{\protect\addvspace{15pt}}}
{}{}
\makeatother
\AtEndEnvironment{algorithm}{%
	\addtocontents{loa}{\protect\addvspace{5pt}}%
}

\usepackage[inline]{enumitem}
\usepackage{pgfplotstable}
\usepackage{booktabs}
\usepackage{multirow}
\usepackage{makecell}

\usepackage{pifont}

\usepackage{amsmath,amssymb,amsfonts}
\usepackage{dsfont} 

\makeatletter
\newcommand*\bigcdot{\mathpalette\bigcdot@{.5}}
\newcommand*\bigcdot@[2]{\mathbin{\vcenter{\hbox{\scalebox{#2}{$\m@th#1\bullet$}}}}}
\makeatother 

\usepackage{tikz} 
\usetikzlibrary{calc, positioning, snakes, calligraphy}
\usetikzlibrary{calc,arrows.meta,positioning,quotes}
\colorlet{plan}{blue!66!black} 

\usepackage{soul}
\usepackage{lipsum}

\usepackage{bm, bbm}
\usepackage{mathtools}
\usepackage{physics}
\usepackage{siunitx}
\usepackage{subcaption}
\usepackage{xspace}
\usepackage{hyperref} 

\usepackage{amsmath,amsfonts,bm}

\def\eqref#1{equation~\ref{#1}}

\def\1{\bm{1}}

\def\eps{{\epsilon}}

\def\vtheta{{\bm{\theta}}}
\def\vlambda{{\bm{\lambda}}}
\def\va{{\bm{a}}}
\def\vb{{\bm{b}}}
\def\vc{{\bm{c}}}
\def\vd{{\bm{d}}}

\def\vm{{\bm{m}}}
\def\vn{{\bm{n}}}

\def\vp{{\bm{p}}}

\def\vr{{\bm{r}}}
\def\vs{{\bm{s}}}

\def\vu{{\bm{u}}}
\def\vv{{\bm{v}}}
\def\vw{{\bm{w}}}
\def\vx{{\bm{x}}}
\def\vy{{\bm{y}}}
\def\vz{{\bm{z}}}

\def\evlambda{{\lambda}}

\def\eva{{a}}
\def\evb{{b}}
\def\evc{{c}}
\def\evd{{d}}

\def\evw{{w}}
\def\evx{{x}}
\def\evy{{y}}

\def\mA{{\bm{A}}}
\def\mB{{\bm{B}}}
\def\mC{{\bm{C}}}
\def\mD{{\bm{D}}}
\def\mE{{\bm{E}}}

\def\mG{{\bm{G}}}
\def\mH{{\bm{H}}}
\def\mI{{\bm{I}}}
\def\mJ{{\bm{J}}}

\def\mL{{\bm{L}}}
\def\mM{{\bm{M}}}
\def\mN{{\bm{N}}}
\def\mO{{\bm{O}}}
\def\mP{{\bm{P}}}
\def\mQ{{\bm{Q}}}
\def\mR{{\bm{R}}}
\def\mS{{\bm{S}}}
\def\mT{{\bm{T}}}
\def\mU{{\bm{U}}}
\def\mV{{\bm{V}}}
\def\mW{{\bm{W}}}
\def\mX{{\bm{X}}}
\def\mY{{\bm{Y}}}
\def\mZ{{\bm{Z}}}
\def\mLambda{{\bm{\Lambda}}}

\def\mLambda{{\bm{\Lambda}}}
\def\mSigma{{\bm{\Sigma}}}

\DeclareMathAlphabet{\mathsfit}{\encodingdefault}{\sfdefault}{m}{sl}
\SetMathAlphabet{\mathsfit}{bold}{\encodingdefault}{\sfdefault}{bx}{n}
\newcommand{\tens}[1]{\bm{\mathsfit{#1}}}

\def\tM{{\tens{M}}}

\def\tW{{\tens{W}}}

\def\tTheta{{\bm{\Theta}}}

\def\gA{{\mathcal{A}}}

\def\gC{{\mathcal{C}}}
\def\gD{{\mathcal{D}}}
\def\gE{{\mathcal{E}}}

\def\gG{{\mathcal{G}}}
\def\gH{{\mathcal{H}}}

\def\gM{{\mathcal{M}}}
\def\gN{{\mathcal{N}}}

\def\gQ{{\mathcal{Q}}}

\def\gV{{\mathcal{V}}}
\def\gW{{\mathcal{W}}}
\def\gX{{\mathcal{X}}}
\def\gY{{\mathcal{Y}}}

\def\sN{{\mathbb{N}}}

\def\sR{{\mathbb{R}}}

\def\emLambda{{\Lambda}}
\def\emA{{A}}
\def\emB{{B}}

\def\emD{{D}}

\def\emJ{{J}}

\def\emP{{P}}

\def\emU{{U}}

\def\emW{{W}}
\def\emX{{X}}

\def\emSigma{{\Sigma}}

\newcommand{\etens}[1]{\mathsfit{#1}}

\def\etW{{\etens{W}}}

\newcommand{\R}{\mathbb{R}}

\newcommand{\bracks}[1]{\left(#1\right)}

\DeclareMathOperator*{\argmax}{arg\,max}

\usepackage{import}
\DeclareMathOperator{\tvec}{vec}
\DeclareMathOperator{\trt}{tr}
\newcommand{\tsym}{\text{sym}}
\newcommand{\trw}{\text{rw}}
\DeclareMathOperator{\tspan}{span}
\DeclareMathOperator{\tdiag}{diag}
\DeclareMathOperator*{\plim}{plim}

\title{Towards Understanding and Avoiding Limitations of Convolutions on Graphs}
\author{Andreas Roth}
\date{2025}

\begin{document}
\sloppy	
	\frontmatter
	
	\makeatletter
	\begin{titlepage}
		\centering
		\Large
		\vspace*{\fill}
		\vspace*{\fill}
		\vspace*{\fill}
		{\usekomafont{title}\huge\@title} \\
		\vspace*{\fill}
		\vspace*{\fill}
		{\bfseries Dissertation} \par\smallskip
		{zur Erlangung des Grades eines} \par\bigskip
		{Doktors der Naturwissenschaften} \par\bigskip 
		{der Technischen Universit\"at Dortmund \\
			an der Fakult\"at für Informatik} \par\bigskip
		{von} \par\bigskip
		{\@author} \\
		\vspace*{\fill}
		\vspace*{\fill}
		{Dortmund} \par\smallskip
		{\@date}
	\end{titlepage}
	\makeatother
	
	\thispagestyle{empty}
	\vspace*\fill
	\begin{center}
		\begin{tabular}{rl}
			\phantom{Prof. Dr.-Ing. Gernot A. Fink} & \phantom{Tag der m\"undlichen Pr\"ufung:} \\
			Dekan: & Prof. Dr. Jens Teubner \\
			Gutachter*innen: & Jun.-Prof. Dr. Thomas Liebig \\
            & Assoz. Prof. Dr. Nils Morten Kriege
		\end{tabular}
	\end{center}

	\clearpage
\vspace*{1.3cm}
\hspace*{1.1cm}
\begin{minipage}{\textwidth - 2.4cm}

\begin{center}
    {\normalsize\bfseries Abstract}
\end{center}
\vspace{2.2em}

\small
Many challenging tasks involve data with an inherent graph structure. Examples of such graph-structured data include molecules, road networks, and transaction records. As such data becomes increasingly available, the need for methods that can effectively leverage graph-structured data grows rapidly.
While message-passing neural networks (MPNNs) have shown promising results, their impact on real-world applications remains limited.
Although various phenomena have been identified as possible limitations to the performance of MPNNs, their theoretical foundations remain poorly understood, leading to fragmented research efforts. In this thesis, we provide an in-depth theoretical analysis and identify several key properties limiting the performance of MPNNs. Building on these findings, we propose several frameworks that address these shortcomings.
\vspace{\baselineskip}

We identify and analyze two properties exhibited by many MPNNs: shared component amplification (SCA), where each message-passing iteration amplifies the same components across all feature channels, and component dominance (CD), where a single component gets increasingly amplified as more message-passing steps are applied. 
These properties lead to the observable phenomenon of rank collapse of node representations, which we identify as a generalization of the established over-smoothing phenomenon. By generalizing and decomposing over-smoothing, we enable a deeper understanding of MPNNs, more targeted solutions, the identification of the relevance of each property, and more precise communication within the field. \vspace{\baselineskip}

To avoid SCA, we show that utilizing multiple computational graphs or edge relations is necessary. 
The multi-relational split (MRS) framework transforms any existing MPNN into one that leverages multiple edge relations. We identify properties on multiple edge relations that enable the resulting MPNN to avoid SCA. Additionally, we introduce the spectral graph convolution for multiple feature channels (MIMO-GC), which naturally uses multiple computational graphs. A localized version, LMGC, serves as a framework for constructing MPNNs that approximate the MIMO-GC while inheriting its beneficial properties. We show that LMGCs can avoid SCA and achieve injectivity on multisets.
\vspace{\baselineskip}

To address CD, we demonstrate a close connection between MPNNs and the PageRank algorithm. This connection allows us to transfer insights and modifications from PageRank to MPNNs. Based on personalized PageRank, we propose a variant of MPNNs that allows for infinitely many message-passing iterations, while preserving initial node features. 
\vspace{\baselineskip}

Collectively, these results contribute to a more comprehensive understanding of MPNNs, allowing future research to be more targeted. Our findings also establish frameworks for constructing MPNNs that exhibit favorable properties.

\end{minipage}
\normalsize
\restoregeometry
\cleardoublepage
\vspace*{1.3cm}
\hspace*{1.1cm}
\begin{minipage}{\textwidth - 2.4cm}

\begin{center}
    {\normalsize\bfseries Acknowledgments}
\end{center}
\vspace{2.2em}

\small

The past four years have shaped me profoundly — both personally and professionally — far beyond the acquisition of technical knowledge.
I am deeply grateful for the time I have had: the opportunity to pursue research that truly interests me, the many exciting experiences, and to meet so many amazing people in Dortmund and on various occasions around the world. 
\vspace{\baselineskip}

I would like to sincerely thank Thomas Liebig, who made it possible for me to pursue my PhD. You gave me the freedom to explore topics that sparked my curiosity and supported me by allowing me to fully dedicate my time and energy to research. The opportunity to follow my interests and attend various conferences has influenced both my personal growth and academic journey.
\vspace{\baselineskip}

I am also grateful to the many researchers who shared their perspectives with me and, often unknowingly, influenced my research: Francesco Di Giovanni, Konstantin Rusch, Adrián Arnaiz-Rodríguez, Pascal Esser, Nils Kriege, MoonJeong Park, Sohir Maskey, Christian Koke, and the many reviewers of my work over the years, whose feedback helped me better understand diverse viewpoints. 
Thank you as well to everyone I had the pleasure of interacting with at conferences, making each event uniquely memorable.
A special thanks goes out to Xeniya Reich, with whom I studied countless related papers during the early phase of my PhD, laying the foundation for much of what followed. I am also thankful to Gilbert Strang and Ulrike von Luxburg for making their excellent resources on mathematical foundations publicly available — I benefited from them greatly.
I am also grateful to everyone involved in our irregular monthly graph meeting: Frank Bause, Alice Moallemy-Oureh, Pascal Welke, Max Thiessen, Fabian Jogl, Sagar Malhotra, Tamara Drucks, and all others who have been involved over time.
\vspace{\baselineskip}

To my co-authors — Thomas Liebig, Franka Bause, Pascal Welke, Konstantin Wüstefeld, Alice Moallemy-Oureh, Veronica Lachi, Nils Kriege, Frank Weichert, and Cedric Sanders — thank you for your collaboration. Working with you over the past four years has provided invaluable insights, improved our research outcomes, and served as a constant source of motivation.
\vspace{\baselineskip}

I would like to thank my family — especially my parents, Karin and Gerd — for their unconditional support throughout all this time. A heartfelt thank you also to my brother, Stefan, who always listened with patience and shared his valuable experiences of the general research process. 
\vspace{\baselineskip}

Finally, I want to thank the love of my life, Hanna. Your love and support motivate me every single day. I look forward to having you by my side — now and always.
\end{minipage}
\normalsize
\restoregeometry
\clearpage
	\tableofcontents
	
	\mainmatter
    \cleardoublepage
    \chapter{Introduction}
\section{Motivation}

There are numerous emerging challenges across the world, from global issues such as the 17 Sustainable Development Goals outlined by the United Nations, to more specialized problems faced by individuals and organizations. 
Humans can address many of these challenges more effectively when suitable algorithms are available to support them. Such algorithms can assist in understanding complex phenomena, contribute to the development of viable solutions, or enable the application of solutions to recurring situations. 
In many domains, large volumes of data are available. This data can range from historical observations and simulations to data collected from real-world environments or digital systems. In recent years, the amount of available data has increased substantially across various tasks. Additionally, the available computational power has advanced considerably~\parencite{schmidhuber2015deep}. 
These developments underscore the importance of designing algorithms that can effectively leverage large amounts of data and computational power to address challenges of interest.

Machine learning offers a broad set of techniques well-suited to utilize available data and computational power effectively. For visual and language data, machine learning techniques have achieved substantial progress~\parencite{he2016deep,vaswani2017attention,roth2021adata}, resulting in impactful real-world applications~\parencite{openai2023gpt4,touvron2023llama}. These data types exhibit regular structures, with language being inherently sequential, as one word follows another. Visual data, such as images or videos, exhibits consistent spatial directionalities.

For many other challenges, data does not have such a regular structure. For example, in chemical processes, data may represent molecules, and the goal might be to identify their properties or interactions with other molecules automatically. 
A molecule can generally consist of any number of atoms without a clear ordering among them. The bonds between atoms likewise do not follow a regular structure. 
Similarly, social networks consist of individual users and their interactions, which can vary significantly across different users. 
Other examples of irregular data include knowledge graphs~\parencite{hogan2021knowledge} or transaction records~\parencite{you2022roland,fey2024position}. 
Any such irregular structure can be represented as a graph that consists of entities and arbitrary interactions between them. Machine learning techniques for graph-structured data have not achieved the same level of success as those used in vision or language applications~\parencite{bechler2025position}.

This thesis investigates the causes of the performance discrepancy of methods for graph-based machine learning compared to other machine learning methods for regular data. By theoretically analyzing the properties of graph-based machine learning methods that contribute to this discrepancy, we aim to advance our understanding of existing methods. This understanding also allows the design of targeted improvements to these methods that avoid the detrimental properties of existing methods.
Based on these theoretical findings, we propose several directions to improve the effectiveness of machine learning methods for graph-structured data.

\section{Research Questions}
\label{sec:1:rq}
Our first goal is to gain a deeper understanding of machine learning methods for graph-structured data, focusing specifically on differentiable approaches known as graph neural networks (GNNs).
While neural networks in general are not yet well understood, this is especially true for GNNs. GNNs commonly rely on an iterative message-passing framework where each iteration combines the representations of each node with those of its neighbors. 
In these models, it has been observed that increasing the number of message-passing steps often results in performance degradation. 
However, the specific properties of message-passing operations that cause this performance degradation, as well as the conditions under which it can be prevented, remain poorly understood.
Our first research question seeks to improve our understanding of the message-passing framework and the reason behind the performance degradation:

\begin{research_question}
\label{rq:1}
How do established message-passing operations affect the representations, and why does repeatedly applying such operations result in a performance degradation? 
\end{research_question}

A widely considered explanation for this performance degradation is over-smoothing, a phenomenon that refers to the observation that repeatedly applying message-passing operations causes node representations to become increasingly similar to one another. Although over-smoothing has been extensively studied both empirically and theoretically, its details remain unclear. This has led to multiple, sometimes contradictory, definitions and interpretations of over-smoothing. We are interested in examining how these different perspectives emerged, and how our understanding can be clarified and improved.

Better insights into the reasons for the performance degradation are not only theoretically valuable but also practically relevant for practitioners who seek to avoid this phenomenon.
More broadly, a better understanding of the message-passing process is essential for the long-term development of GNNs, as future research efforts can build on a more solid theoretical foundation.

Due to the lack of a unified explanation for the causes behind the performance degradation and over-smoothing, the proposed improvements target symptoms rather than the currently unknown reasons behind the phenomena.
The missing theoretical foundation of many recently proposed methods motivates our second research question as follows:

\begin{research_question} 
\label{rq:2}
How can our improved understanding of message-passing operations be used to streamline the design and selection of methods to exhibit favorable properties?
\end{research_question}

While many promising techniques have been proposed, utilizing and designing specific methods that exhibit properties we understand and demand for a challenge of interest is crucial. We aim to investigate how general frameworks can be constructed to support the design of principled methods and allow future theoretical and empirical studies within a provided framework.

\section{Main Contributions}
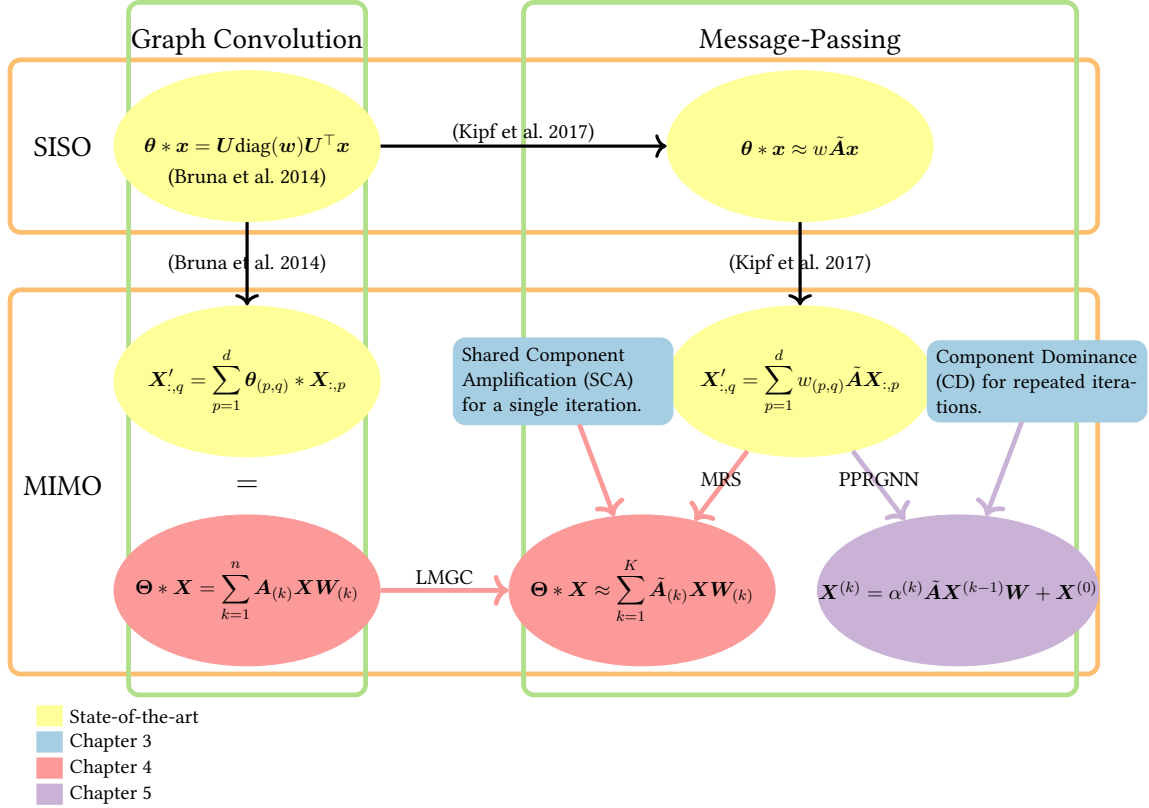
\begin{figure*}
\definecolor{vibrantpink}{HTML}{fb9a99} 
\centering
\begin{tikzpicture}[x=\textwidth/4, y=\textwidth/4]
\definecolor{vibrantblue}{HTML}{a6cee3} 
\definecolor{vibrantgreen}{HTML}{b2df8a} 
\definecolor{vibrantpink}{HTML}{fb9a99} 
\definecolor{vibrantorange}{HTML}{fdbf6f} 
\definecolor{vibrantpurple}{HTML}{cab2d6} 
\definecolor{vibrantsand}{HTML}{ffff99} 
\usetikzlibrary{shapes}
\pgfdeclarelayer{background}
\pgfsetlayers{background,main}
\tikzset{
dot/.style = {circle, minimum size=#1,
              inner sep=0pt, outer sep=0pt, draw=black},
dot/.default =13pt  
}
    \tikzstyle{box} = [draw, draw=vibrantgreen, thick,rounded corners,line width=1.8pt]

    \def\ysiso{0.7}
    \def\ymimo{-0.2}
    \def\xgc{-1.1}
    \def\xmp{1.00}
    
    \draw[box,draw=vibrantorange] (-2, 0.37) rectangle (2.13, 1.03); 
    \node at (-1.80, \ysiso) {SISO}; 
        
    \draw[box,draw=vibrantorange] (-2, -1.32) rectangle (2.13, 0.15); 
    \node at (-1.80, \ymimo-0.4) {MIMO}; 
    \node[scale=1.0] at (\xgc,\ymimo -0.4) {$=$};

    \draw[box] (\xgc -0.45, -1.4) rectangle (\xgc + 0.45, 1.25); 
    \node at (\xgc, 1.1) {Graph Convolution};
    \node[fill=vibrantsand,draw=vibrantsand,minimum width=3.5cm,minimum height=2cm,draw,ellipse] (scgc) at (\xgc,\ysiso) {};
    \node[scale=0.7] at (\xgc, \ysiso) {$\vtheta*\vx=\mU \mathrm{diag}(\vw)\mU^\top\vx$};
    \node[color=black] at (\xgc, \ysiso-0.12) {\scriptsize\parencite{bruna2014spectral}};

    \node[fill=vibrantsand,draw=vibrantsand,minimum width=3.5cm,minimum height=2cm,draw,ellipse] (mcgc) at (\xgc,\ymimo) {};
    \node[scale=0.7] at (\xgc, \ymimo) {$\displaystyle\mX^\prime_{:,q} = \sum_{p=1}^d\vtheta_{(p,q)}*\mX_{:,p}$};
    \node[fill=vibrantpink,draw=vibrantpink,minimum width=3.5cm,minimum height=2cm,draw,ellipse] (mimogc) at (\xgc,\ymimo-0.8) {};
    \node[scale=0.7] at (\xgc, \ymimo - 0.8) {$\displaystyle\tTheta*\mX = \sum_{k=1}^n \mA_{(k)}\mX\mW_{(k)}$};

    \draw[box] (\xmp - 1.05, -1.4) rectangle (\xmp + 1.05, 1.25);   
    \node at (\xmp, 1.1) {Message-Passing};
    \node[fill=vibrantsand,draw=vibrantsand,minimum width=3.5cm,minimum height=2cm,draw,ellipse] (scmpnn) at (\xmp,\ysiso) {};    
    \node[scale=0.7] at (\xmp, \ysiso) {$\vtheta * \vx \approx w \tilde{\mA}\vx$}; 

    \coordinate (x) at (\xmp,-0.15);
        \node[fill=vibrantsand,draw=vibrantsand,minimum width=3.5cm,minimum height=2cm,draw,ellipse] (a) at (\xmp,\ymimo) {}; 
    \node[scale=0.7] at (\xmp,\ymimo) {$\displaystyle\mX^\prime_{:,q} = \sum_{p=1}^dw_{(p,q)}\tilde{\mA}\mX_{:,p}$};


    \node[fill=vibrantpink,draw=vibrantpink,minimum width=3.5cm,draw,ellipse,minimum height=2cm] (lmcgc) at (\xmp-0.6,\ymimo-0.8) {};
    \node[scale=0.7] at (\xmp-0.6,\ymimo -0.8) {$\displaystyle\tTheta * \mX \approx \sum_{k=1}^K \tilde{\mA}_{(k)}\mX\mW_{(k)}$}; 

    \node[fill=vibrantpurple,draw=vibrantpurple,minimum width=3.7cm,draw,ellipse,minimum height=2cm] (pprgnn) at (\xmp+0.6,\ymimo-0.8) {};
    \node[scale=0.7] at (\xmp+0.6,\ymimo -0.8) {$\mX^{(k)} = \alpha^{(k)}\tilde{\mA}\mX^{(k-1)}\mW + \mX^{(0)}$}; 

    \node [box, draw=vibrantblue, text width=3.8cm, scale=0.7, fill=vibrantblue] (sca) at (\xmp - 0.9, \ymimo) {Shared Component Amplification (SCA) for a single iteration.};
    \node [box, draw=vibrantblue, text width=3.8cm, scale=0.7, fill=vibrantblue] (cd) at (\xmp + 0.9, \ymimo) {Component Dominance (CD) for repeated iterations.};


    \draw[->,line width=1.2pt] (scgc) to (scmpnn);
    \node[color=black] at ($(\xmp,\ysiso+0.05)!0.5!(\xgc,\ysiso+0.05)$){\scriptsize\parencite{kipf2017semi}};
    \draw[->,line width=1.2pt] (scmpnn) to (a);
    \node[color=black] at ($(\xmp,\ymimo)!0.5!(\xmp,\ysiso)$){\scriptsize\parencite{kipf2017semi}};

    \draw[vibrantpink,->,line width=1.8pt] (sca) to (lmcgc);
    \draw[vibrantpink,->,line width=1.8pt] (a) to (lmcgc);    
    \node[color=black] at ($(\xmp, \ymimo)!0.5!(\xmp-0.6,\ymimo -0.75)$){\scriptsize MRS};

    \draw[vibrantpurple,->,line width=1.8pt] (cd) to (pprgnn);
    \draw[vibrantpurple,->,line width=1.8pt] (a) to (pprgnn);    
    \node[color=black] at ($(\xmp, \ymimo)!0.5!(\xmp+0.6,\ymimo -0.75)$){\scriptsize PPRGNN};


    \draw[black,->,line width=1.2pt] (scgc) to (mcgc);
    \node[color=black] at ($(\xgc,\ymimo)!0.5!(\xgc,\ysiso)$){\scriptsize\parencite{bruna2014spectral}};
    
    \draw[vibrantpink,draw=vibrantpink,->,line width=1.8pt] (mimogc) to node[above, color=black,scale=0.7] {}  (lmcgc);
    \node[color=black] at ($(\xgc,\ymimo -0.75)!0.5!(\xmp-0.6,\ymimo -0.75)$){\scriptsize LMGC};

    \def\ylegend{-1.58}
    \def\xlegend{-1.8}

  \draw[vibrantpurple, very thick, line width=8pt] (\xlegend-0.1,\ylegend-0.2) -- (\xlegend,\ylegend-0.2);
  \node[anchor=west,scale=0.7] at (\xlegend,\ylegend-0.2) {Chapter 5};

  \draw[vibrantpink,draw=vibrantpink, thick, line width=8pt] (\xlegend-0.1,\ylegend-0.1) -- (\xlegend,\ylegend-0.1);
  \node[anchor=west,scale=0.7] at (\xlegend,\ylegend-0.1) {Chapter 4};

  \draw[vibrantblue, thick, line width=8pt] (\xlegend-0.1,\ylegend+0.0) -- (\xlegend,\ylegend+0.0);
  \node[anchor=west,scale=0.7] at (\xlegend,\ylegend+0.0) {Chapter 3};

    \draw[vibrantsand, thick, line width=8pt] (\xlegend-0.1,\ylegend+0.1) -- (\xlegend,\ylegend+0.1);
  \node[anchor=west,scale=0.7] at (\xlegend,\ylegend+0.1) {State-of-the-art};

\end{tikzpicture}
\caption{Schematic overview of this thesis and the connections between our contributions.}
\label{fig:contributions}
\end{figure*}
Following our outlined research questions, this thesis presents several key contributions to address these. We present a visual guidance through our contributions and chapters in \Cref{fig:contributions}.

We conduct an in-depth study on graph convolutions and message-passing operations in GNNs, specifically examining the underlying properties that lead to the over-smoothing phenomenon (\Cref{rq:1}). As many works employ different and partially contradictory definitions, we clarify the differences in the current state-of-the-art research on over-smoothing as a first contribution~\parencite{roth2023rank}. We identify that an overlapping phenomenon of vanishing norms, causing representations to approach the zero state the more message-passing iterations are applied, has led to these contrary definitions~\parencite{roth2023rank}.
We also provide insight into the effect of message-passing operations on node representations from a spectral perspective~\parencite{roth2025what}.
We then conduct a detailed theoretical study on the effect of message-passing methods on the representations. We identify two fine-grained phenomena, namely shared component amplification (SCA) that occurs for a single message-passing step, and component dominance (CD)~\parencite{roth2023rank,roth2025what} that arises for the repeated application of message-passing steps. We find that the joint occurrence of these phenomena results in the general issue of rank collapse, which refers to node representations becoming increasingly close to a low-rank state. Over-smoothing occurs as a special case for particular message-passing methods. These findings extend the current theoretical understanding of GNNs and the conditions under which various phenomena arise~\parencite{roth2023rank}. 
We also simplify our theoretical study by demonstrating that many message-passing operations can be viewed as a special case of power iteration, for which extensive theoretical insights are available~\parencite{roth2024simplifying}.

Based on our enhanced intuition and theoretical understanding of the effect of graph convolutions, we propose several frameworks for message passing that exhibit favorable properties (\Cref{rq:2}).
First, we consider SCA, which we identify as always occurring for a single computational graph. We show that SCA can only be avoided when performing message-passing with multiple computational graphs, or equivalently, multigraphs~\parencite{roth2023rank,roth2024preventing,roth2025what}.
We introduce two directions to obtain multiple computational graphs within the MPNN framework.
The first approach starts with existing methods and outlines how multiple computational graphs can be integrated into any existing method~\parencite{roth2024preventing}.
With our proposed multi-relational splits, we decompose the computational graph of existing GNNs into multiple computational graphs by assigning each edge to one of several computational graphs.
The second approach uses a novel approximation of the spectral graph convolution defined in graph Fourier space~\parencite{roth2025what}. Previous work defined the graph convolution in graph Fourier space and proposed approximations in the SISO case, and applied it to the MIMO case afterwards. We instead define the MIMO graph convolution in graph Fourier space and directly approximate it in the MIMO case. The MIMO graph convolution naturally utilizes multiple computational graphs, with each amplifying a different component in the data. Our novel approximation inherits this property and introduces a framework for message-passing that closely follows the MIMO graph convolution. We prove that both methods for obtaining multiple computational graphs solve SCA.

We relate CD as the second theoretical shortcoming related to over-smoothing in GNNs to the PageRank algorithm~\parencite{page1999pagerank} as it exhibits a phenomenon closely related to CD~\parencite{roth2022transforming}. For PageRank, this shortcoming was mitigated by an adaptation called personalized PageRank (PPR), which extends the algorithm with a teleportation term that ensures a localization even as the number of iterations approaches infinity~\parencite{page1999pagerank}. Inspired by this, we propose a modification of MPNNs based on PPR that similarly prevents CD even in the limit of infinitely many message-passing iterations~\parencite{roth2022transforming}. We refer to our method as the personalized PageRank GNN (PPRGNN).

\section{Outline and Covered Publications}
This thesis combines and presents five individual peer-reviewed publications led by the author of this thesis. Here, we outline the structure of this thesis and refer to the publications each chapter is based on. The structure of this thesis and the connection between the chapters are also visualized in \Cref{fig:contributions}.

\subsubsection{Chapter 2: Fundamentals of Graph Machine Learning}
\Cref{sec:fundamentals} provides an overview of the fundamental concepts of graph structures and machine learning for graphs. This will be necessary to understand the details of our contributions in the following chapters. The overview includes basic graph-theoretical definitions and properties, such as the graph isomorphism problem and the graph Fourier transform (\Cref{sec:fundamentals:graph_theory}).
We also provide details on machine learning techniques for graph-structured data (\Cref{sec:fundamentals:graph_ml}). There, we particularly focus on the definition of the graph convolution and message-passing neural networks.

\subsubsection{Chapter 3: Extending Our Understanding of Graph Convolutions}
In \Cref{sec:understanding}, we strive to find a better theoretical and intuitive understanding of graph convolutions and their localized approximations. We study the effect of message-passing operations on node representations and particularly examine the reasons behind the over-smoothing phenomenon. We provide a spectral intuition (\Cref{sec:understanding:fourier}), study the theoretical properties in-depth (\Cref{sec:understanding:rank}), and then simplify the analysis for more accessible and intuitive theoretical properties (\Cref{sec:understanding:rank:power}). Here, we also show that two finer-grained phenomena underlie over-smoothing: shared component amplification (SCA) and component dominance (CD). 
This chapter is mainly based on the following two peer-reviewed publications: 
\begin{itemize}
    \item \textbf{Andreas Roth}, Thomas Liebig (2023). “Rank Collapse Causes Over-Smoothing and Over-Correlation in Graph Neural Networks.” In: Learning on Graphs Conference (LoG).
    \item \textbf{Andreas Roth} (2024). “Simplifying the Theory on Over-Smoothing.” In: Lernen, Wissen, Daten, Analysen (LWDA).
\end{itemize}

\subsubsection{Chapter 4: Preventing Shared Component Amplification With Multiple Computational Graphs}
As one of two phenomena underlying over-smoothing, we show that message-passing operations can only prevent SCA when they utilize multiple computational graphs. We propose two methods for obtaining multiple computational graphs in this chapter. First, we demonstrate how the computational graph of existing approaches can be split into multiple computational graphs by assigning each edge to one of several relation types (\Cref{sec:4:splitting}). Second, we propose a novel way of approximating and localizing the spectral graph convolution defined in graph Fourier space (\Cref{sec:uniformity:mimo-gc}). We prove that both directions can effectively prevent SCA. This chapter is based on the following publications:
\begin{itemize}
    \item \textbf{Andreas Roth}, Franka Bause, Nils M. Kriege, Thomas Liebig (2024). “Preventing Representational Rank Collapse in MPNNs by Splitting the Computational.” In: Learning on Graphs Conference (LoG).
    \item \textbf{Andreas Roth}, Thomas Liebig (2025). “What can we learn from MIMO Graph Convolutions?” In: International Joint Conferences on Artificial Intelligence (IJCAI).
\end{itemize}

\subsubsection{Chapter 5: Preventing Component Dominance based on Personalized PageRank}
As identified in \Cref{sec:understanding}, the second issue underlying over-smoothing is what we refer to as component dominance. In \Cref{chap:5}, we establish the connection between this phenomenon in MPNNs and the PageRank algorithm, which exhibits a closely related phenomenon. As personalized PageRank was proposed as a solution to this issue, we show how MPNNs can adapt this solution. This chapter is based on the following peer-reviewed publication:
\begin{itemize}
    \item \textbf{Andreas Roth}, Thomas Liebig (2022). “Transforming PageRank into an Infinite-Depth Graph Neural Network.” In: Joint European conference on machine learning and knowledge discovery in databases (ECMLPKDD) (\textbf{best paper award}).
\end{itemize}

\subsubsection{Chapter 6: Summary and Outlook}
To conclude this thesis, \Cref{chap:6} summarizes our findings, puts them into greater context, and outlines promising directions for future work.

\subsubsection{Additional Publications}
The author additionally contributed to several peer-reviewed publications that are outside the scope of this thesis. These works consider further challenges of gradient-based machine learning:
\begin{itemize}
    \item \textbf{Andreas Roth}*, Konstantin Wüstefeld*\def\thefootnote{*}\footnotetext{These authors contributed equally to the corresponding publication.}\def\thefootnote{\arabic{footnote}}, Frank Weichert (2021). “A Data-Centric Augmentation Approach for Disturbed Sensor Image Segmentation.” In: Journal of Imaging 7.10.
    \item \textbf{Andreas Roth}, Thomas Liebig (2022). “Forecasting Unobserved Node-States with Spatio-Temporal Graph Neural Networks.” In: ICDM Workshop on Machine Learning on Graphs (MLoG).
    \item \textbf{Andreas Roth}, Thomas Liebig (2023). “Distilling Influences to Mitigate Prediction Churn in Graph Neural Networks.” In: Asian Conference on Machine Learning (ACML).    
    \item Cedric Sanders*, \textbf{Andreas Roth}*, Thomas Liebig (2023). “Curvature-based Pooling within Graph Neural Networks.” In: ECMLPKDD Workshop on Mining and Learning with Graphs (MLG).
    \item Veronica Lachi*, Alice Moallemy-Oureh*, \textbf{Andreas Roth}*, Pascal Welke* (2025). “Expressive Pooling for Graph Neural Networks.” In: Transactions on Machine Learning Research (TMLR).    
\end{itemize}

\clearpage
    \cleardoublepage
    \chapter{Fundamentals of Graph Machine Learning}
\label{sec:fundamentals}
In this chapter, we provide a brief introduction to graph machine learning and its related background topics. We establish our notation and provide necessary background knowledge underlying the following chapters. These fundamentals will help to understand the current state-of-the-art, our considered challenges, and our contributions. We will first introduce required graph theoretical basics in \Cref{sec:fundamentals:graph_theory}, including relevant spectral properties, particularly the concept of the graph Fourier transform. 
Building on this, we will introduce the topic of machine learning for graph-structured data, with a focus on differentiable approaches and the convolution on graphs in \Cref{sec:fundamentals:graph_ml}. Based on this, we demonstrate how message-passing neural networks (MPNNs) are derived from graph convolution and how machine learning tasks based on graph-structured data are typically addressed using these methods.

\section{Notation}
The notation used in this thesis closely follows that of \textcite{goodfellow2016deep}. A set is denoted by a calligraphic symbol, such as $\gA = \{\,a_1,a_2,\dots\,\}$, with special notation reserved for the set of natural numbers $\sN$ and the set of real numbers $\sR$. For $n\in\sN$, we use $[n] = \{\,1,\dots,n\,\}$ to denote the set of natural numbers from $1$ to $n$.
A multiset is written as $\gA = \{\{\,a_1,\dots,a_1,a_2,\dots\,\}\}$, where elements may appear multiple times. Vectors are denoted in bold lowercase, e.g., $\va$, with the $i$-th element written as $\eva_i$. Matrices are denoted in bold uppercase, e.g., $\mA$, with the $i$-th row written as $\mA_{i,:}$, the $j$-th column as $\mA_{:,j}$, and the element in the $i$-th row and $j$-th column as $\emA_{i,j}$. Further notation that is specific to individual sections or methods is introduced as needed.

\section{Graph Theory}
\label{sec:fundamentals:graph_theory}
A graph $\gG=(\gV,\gE)$ is defined as a finite set of nodes $\gV = \{v_1,\dots,v_n\}$ and a set of edges $\gE \subseteq \gV \times \gV$ that represents connections between pairs of nodes. An attributed graph $\gG=(\gV,\gE,x)$ is a graph extended by a function $x\colon \gV \to \R^d$ that assigns a $d$-dimensional vector $x(v_i)\in\R^d$ of real numbers as attributes to every node $v_i\in\gV$. 
Unless otherwise indicated, we assume graphs to be undirected, i.e., $(v_i,v_j)\in\gE$ whenever $(v_j,v_i)\in\gE$. Graphs that do not satisfy this condition are called directed graphs. For every $(v_i,v_j)\in\gE$, we refer to $v_i$ and $v_j$ as neighboring or adjacent nodes. The set of nodes adjacent to node $v_i\in\gV$ is represented by $\gN_{i} = \{\,v_j \mid (v_j,v_i)\in \mathcal{E}\,\}$. The degree of node $v_i$ is defined as the number of neighboring nodes $d_i = \left|\gN_{i}\right|$, where $|\cdot|$ indicates the number of elements in the set. A graph is bipartite when there exist $\gV_1,\gV_2\subset\gV$ with $\gV_1\cup\gV_2 = \gV$ and $\gV_1\cap\gV_2 = \emptyset$ such that for every edge $(v_i,v_j)\in\gE$ the two corresponding nodes $v_i$ and $v_j$ are in a different subset, i.e., we have either $v_i\in\gV_1$ and $v_j\in\gV_2$ or $v_i\in\gV_2$ and $v_j\in\gV_1$. Due to the properties we will introduce in \Cref{sec:fundamentals:spectral}, we assume graphs to be non-bipartite.

Graphs are typically processed by their corresponding matrix representations for simpler notation and hardware acceleration.
A graph $\gG = (\gV,\gE)$ with $n = |\gV|$ can be equivalently described by its adjacency matrix $\mA\in\{0,1\}^{n\times n}$ where $\emA_{i,j} = 1$ if $(v_i,v_j)\in \gE$ and $\emA_{i,j} = 0$ otherwise. The degree matrix $\mD\in\mathbb{N}^{n\times n}$ is a diagonal matrix that contains the node degrees $\emD_{i,i} = d_i$ on its diagonal. For attributed graphs, the function $x\colon \gV\to\R^d$ is represented as an equivalent matrix $\mX\in\R^{n\times d}$ where every row $\mX_{i,:} = x(v_i)$ corresponds to the attributes of the corresponding node $v_i$. Analogously to the set notation, we refer to $\gG = (\mA,\mX)$ as a graph in matrix notation.

\subsection{Graph Isomorphism}
\label{sec:2:isomorphism}
A key challenge of graph theory is determining whether two graphs $\gG_1=\bracks{\gV_1,\gE_1}$ and $\gG_2=\bracks{\gV_2,\gE_2}$ are isomorphic~\parencite{luks1982isomorphism}. Two graphs are said to be isomorphic when there exists a bijective function $\phi\colon \gV_1\to \gV_2$ mapping each node of graph $\gG_1$ to exactly one node of graph $\gG_2$ such that $\bracks{\phi\bracks{v_i},\phi\bracks{v_j}}\in \gE_2$ if and only if $\bracks{v_i,v_j}\in \gE_1$ for every $v_i,v_j\in\gV_1$. We denote two isomorphic graphs as $\gG_1 \simeq \gG_2$.
For attributed graphs, we additionally require $x_1\bracks{v_i} = x_2\bracks{\phi\bracks{v_i}}$ for all $v_i\in \gV_1$. There are currently no polynomial-time algorithms known for this task~\parencite{babai2016graph}.

When representing graphs in their matrix notation $\gG = (\mA,\mX)$, the ordering of the nodes is arbitrary. Thus, two isomorphic graphs $\gG_1=\bracks{\mA_1,\mX_1}$ and $\gG = \bracks{\mA_2,\mX_2}$ may have a different ordering of rows and columns in $\mA$ and $\mX$. Deciding whether two graphs with corresponding adjacency matrices $\mA_1$ and $\mA_2$ are isomorphic is equivalent to determining whether there exists a permutation matrix $\mP$ such that $\mP\mA_1\mP^\top = \mA_2$. A permutation matrix $\mP$ contains a single one in each row and column, with all other entries being zero. Applying such a permutation matrix to an adjacency matrix corresponds to modifying the ordering of the nodes.
For matrix representations $\mX_1$ and $\mX_2$ of node attributes, we additionally require the permutation matrix $\mP$ to satisfy $\mP\mX_1 = \mX_2$.
A function that maps graphs $\gG$ to a vector space and is invariant under graph isomorphisms can be equivalently expressed as a function over matrix representations of the graphs that is invariant under permutations of node orderings. Otherwise, such a function $f$ would assign different outputs to isomorphic graphs. We define this property as follows:

\begin{definition}[Permutation Invariance]
\label{def:perm_inv}
    A function $f$ is \emph{permutation invariant} under node orderings if
    \begin{equation}
        f(\mA,\mX) = f\bracks{\mP\mA\mP^\top,\mP\mX}
    \end{equation}
    for all $n\in\mathbb{N}$, permutation matrices $\mP\in\{0,1\}^{n\times n}$, adjacency matrices $\mA\in\{0,1\}^{n\times n}$, and feature matrices $\mX\in\R^{n\times d}$.
\end{definition}

The methods considered in this thesis are typically stated to operate on matrix representations of graphs. All of these methods are permutation invariant and can be equivalently reformulated to operate directly on the original graph representation.

\subsection{Spectral Graph Theory}
\label{sec:fundamentals:spectral}
Key to the methods considered in this work is the field of spectral graph theory~\parencite{chung1997spectral,luxburg2007a}. Based on the matrix notation of a graph, spectral graph theory involves properties of different types of graph Laplacians and their spectra. We will describe graph Laplacians and their necessary properties for this work in the following. 

\paragraph{Graph Laplacians}
We now introduce graph Laplacians and some of their properties. This part is mainly based on \textcite{luxburg2007a}.
Consider an undirected and non-bipartite graph $\gG = (\mA,\mX)$ of $n$ nodes in matrix notation with adjacency matrix $\mA\in\{0,1\}^{n\times n}$ and degree matrix $\mD\in\mathbb{N}^{n\times n}$. The unnormalized graph Laplacian is defined as the matrix
\begin{equation}
    \mL = \mD - \mA\, .
\end{equation}
When applied to a vector $\vx\in\R^n$, each value of $\mL\vx$ represents the sum of the differences between the state of a node and all of its neighbors, i.e., 
\begin{equation}
\label{eq:2:laplacian_diff}
    \bracks{\mL\vx}_i = \sum_{v_j\in\gN_i} \evx_i - \evx_j\, .
\end{equation}
We will also rely on two normalized graph Laplacians. For this, we utilize two normalized adjacency matrices. The symmetrically normalized adjacency matrix $\mA_\tsym = \mD^{-1/2}\mA\mD^{-1/2}$ scales every entry $\emA_{i,j} = \frac{1}{\sqrt{d_i\cdot d_j}}$ by the degrees of the corresponding nodes. The random walk normalized adjacency matrix $\mA_\trw = \mD^{-1}\mA$ normalizes the values of each row to sum to one, i.e., we have $\emA_{i,j} = \frac{1}{d_i}$. The corresponding symmetrically normalized graph Laplacian is defined as
\begin{equation}
    \mL_\tsym = \mI - \mA_\tsym\, ,
\end{equation}
and the random walk normalized graph Laplacian as
\begin{equation}
\mL_\trw = \mI - \mA_\trw\, .
\end{equation}
Similarly to \Cref{eq:2:laplacian_diff}, applying these to a vector $\vx$ results in the weighted difference between the state of a node and its neighboring states.
These graph Laplacians have several key properties that we will rely on throughout this thesis. All three graph Laplacians $\mL,\mL_\tsym$, and $\mL_\trw$ are positive semi-definite matrices. Given a vector $x\in\R^n$, the Dirichlet sum or Dirichlet energy
\begin{equation}
    \vx^\top\mL\vx = \sum_{v_i\in \gV}\sum_{v_j\in \gN_i} (\evx_i - \evx_j)^2
\end{equation}
describes the sum of squared differences in values between adjacent nodes~\parencite{luxburg2007a}. The Dirichlet energy for the normalized graph Laplacians is defined equivalently.
As $\mL$ and $\mL_\text{sym}$ are symmetric matrices, their eigenvectors are orthogonal. 

We denote the $i$-th largest eigenvalue of a matrix $\mP$ as $\lambda_i^{(\mP)}$.
Utilizing the assumed undirected property, the smallest eigenvalues $\lambda_n^{(\mL)}=\lambda_n^{(\mL_\tsym)}=\lambda_n^{(\mL_\trw)}=0$ of the graph Laplacians $\mL,\mL_\tsym,\mL_\trw$ are zero with associated eigenvector $\mathbf{1}$ for $\mL$ and $\mL_\trw$, and $\mD^{1/2}\mathbf{1}$ for $\mL_\tsym$~\parencite{luxburg2007a}. 

Given the non-bipartite property, we further have $\lambda_i\in[0,2)$ for $\mL_\tsym$ and $\mL_\trw$ and all $i\in[n]$.
The random walk normalized graph Laplacian $\mL_\trw$ and the corresponding normalized adjacency matrix $\mA_\trw$ have the same eigenvectors with the associated eigenvalues shifted as $\lambda_{n-i+1}^{(\mA_\trw)} = 1 - \lambda_i^{(\mL_\trw)}$. This is equivalent for the symmetrically normalized graph Laplacian $\mL_\tsym$ and the symmetrically normalized adjacency matrix $\mA_\tsym$. These properties are further exploited to construct the graph Fourier transform, which we will introduce next.

\paragraph{Graph Fourier Transform}
\label{sec:fundamentals:graph_fourier}
The Fourier transform $F$ maps a given function $f$ defined on $\R$ into another function $g = F(f)$ defined on $\R$. The function $g$ is typically interpreted as the frequency domain representation of $f$. The Fourier transform exploits the periodicity and regularity of $\R$ by decomposing $f$ into sinusoidal components with varying frequencies. Evaluating the Fourier transformed function $g(a)$ at some frequency $a$ represents the magnitude of this frequency for the function $f$.
In the discrete case, the discrete Fourier transform is defined over finite sequences that are equally spaced.

For graphs, the aim is to find a frequency domain representation of a graph signal $x\colon\gV\to\R$, also referred to as node attributes. However, as graphs do not exhibit any regularity in general, the discrete Fourier transform cannot be applied to $x$. 
Instead, the graph Fourier transform 
\begin{equation}
F(\vx) = \mU^\top\vx
\end{equation}
is defined as the projection of the vectorized node attributes $\vx\in\R^n$ onto the orthogonal eigenvectors $\mU\in\R^{n\times n}$ of $\mL_\tsym$~\parencite{sandryhaila2013discrete,shuman2013the}.
The eigenvectors $\mU$ of the graph Laplacian $\mL_\tsym$ are used as the graph Fourier base for the graph Fourier transform, as they exhibit similar properties to sinusoidal components.
The corresponding eigenvalues $\mLambda_\tsym$ of $\mL_\tsym$ represent the frequencies.
The eigenvector corresponding to the smallest eigenvalue or frequency $\lambda_1$ is $\mU_{:,1} = \mD^{1/2}\mathbf{1}$ for $\mL_{\text{sym}}$, i.e., it has similar values for all nodes. For larger frequencies $\lambda_i$, the corresponding eigenvectors $\mU_{:,i}$ increasingly oscillate between adjacent nodes. The eigenvector associated with the second smallest eigenvalue, also referred to as the Fiedler vector~\parencite{fiedler1973algebraic}, is used for graph cutting due to $(\emU_\tsym)_{i,2}$ being $>0$ for all nodes in one of the partitions and $<0$ for all nodes in the other partition~\parencite{fiedler1973algebraic}. The eigenvectors associated with higher frequencies increasingly oscillate across adjacent nodes~\parencite{sandryhaila2013discrete,shuman2013the}.
Assuming $\mU$ is orthonormal, the inverse graph Fourier transform 
\begin{equation}
F^{-1}(F(\vx)) = \mU F(\vx) = \mU\mU^\top\vx = \vx
\end{equation} 
is performed by multiplying a frequency domain representation with the graph Fourier basis $\mU$.

The fast Fourier transform reduces the runtime of the discrete Fourier transform to $\mathcal{O}(n \log n)$ for sequences of length $n$. However, as it exploits the regularity of the space, the same technique cannot be applied to the graph Fourier transform. There is currently no subquadratic algorithm known to compute the graph Fourier transform. As the graph Fourier basis $\mU$ depends on the structure of a given graph, it also cannot be applied across distinct graphs.  

\section{Graph Machine Learning}
\label{sec:fundamentals:graph_ml}
In machine learning, we are generally interested in making predictions based on previously observed data. While the field of machine learning encompasses a multitude of different tasks and methods, we here focus on the foundations of machine learning for graph-structured data. We describe the task of classifying graphs, motivate the use of the convolution operation on graphs, and demonstrate how it can be efficiently approximated by message-passing neural networks.

\subsection{Graph Classification}
Classifying attributed graphs into a set of classes $\gY$ is similar to classification tasks for other domains. The key is to find functions that map from a set of graphs to the set of classes $\gY$. We start by defining such functions as graph classifiers:

\begin{definition}[Graph Classifier]
Let $\gH$ denote a set of attributed graphs, and let $\gY$ be a finite set of discrete class labels.
A \emph{graph classifier} is a function $f \colon \gH\to \gY$ that assigns a label $f(\gG)\in\gY$ to each graph $\gG\in\gH$.
\end{definition}

As described in \Cref{def:perm_inv}, a graph classifier $f$ is typically defined as a function on the matrix form of an attributed graph $\gG=(\mA,\mX)$ that is invariant under node permutations.

The performance of a given graph classifier $f$ is computed using a suitable metric, e.g., accuracy or the F1-score. 
This is typically achieved in the supervised learning setting, where a dataset $\gD = \{(\gG_i,y_i)\mid i\in[m]\}$ of $m$ graphs for which the label is known is given. 
The dataset can be split into training, validation, and testing graphs, and the classifier can be optimized over a family of classifiers using the training data. The validation graphs are utilized to ensure good generalization to the underlying distribution of graphs, and the testing graphs are used to approximate the performance of a selected graph classifier over the true distribution.

\paragraph{Challenges in Graph Machine Learning}
The particular challenge of graph classification is that well-performing graph classifiers typically need to consider both the structure of the data and the data itself. For other domains, all inputs to a classifier typically have a shared structure, e.g., vectors with the same number of features, text with a sequential structure, or images with a regular grid structure. 
There are no such regularities for graphs. Each graph can have a different number of nodes and any connectivity between these nodes. Thus, each node may have a different number of neighboring nodes, and there is no inherent ordering or directionality for neighboring nodes. 
Such differences in the data structure are key to many graph classification tasks. For example, in molecular datasets, a graph may correspond to a protein, where nodes represent the atoms and edges represent the bonds between them. The behavior of proteins changes drastically when their atoms or bonds are modified~\parencite{pérez2015activity}. As such, the data structure of a graph is a key feature.

\paragraph{Expressivity}
\label{sec:2:expressivity}
Ideally, considered graph classifiers can distinguish all graphs from the considered set of graphs. This would allow us to find a graph classifier that assigns all graphs to their correct class. 
Based on the graph isomorphism problem (Section~\ref{sec:2:isomorphism}), any graph classifier $f$ with polynomial runtime cannot distinguish some graphs $\gG_1,\gG_2\in\gH$ and must map them to the same label $y\in \gY$. The expressivity of graph classifiers typically describes their ability to distinguish non-isomorphic graphs.  

An established method for testing graph isomorphism is the Weisfeiler-Leman (WL) test~\parencite{weisfeiler1968reduction}.
Given a set of colors $\gC$, the WL test initially assigns the same color $c_i^{(0)}$ to every node $v_i\in\gV$.
The test then iteratively refines the color of each node $v_i\in\gV$ using the update rule
\begin{equation}
c_i^{(k)} = \phi\bracks{c_i^{(k-1)},\left\{\left\{\,c_j^{(k-1)} \mid v_j \in \gN_i\,\right\}\right\}}\, ,
\end{equation}
where $\phi$ is an injective function and $c_i^{(k-1)}$ is the color of node $v_i\in\gV$ at iteration $k-1$. The new color of each node is unique for different multisets of neighboring colors or previous colors.
Two graphs are non-isomorphic if the multisets of node colors differ at any iteration $i$. However, when the multisets of colors for two graphs remain identical across all iterations, it cannot be concluded that these are isomorphic. Despite this limitation, it is computationally efficient and can distinguish a large class of non-isomorphic graphs.
Graph classifiers are often compared to the WL test in terms of their expressivity~\parencite{shervashidze2011weisfeiler,xu2019how,morris2019weisfeiler,kriege2020a} and methods are constructed to improve the expressivity~\parencite{bouritsas2023improving,sanders2023curvature,lachi2025expressive}.

\paragraph{Graph Kernels}
Graph kernels are a classical and effective approach to graph classification~\parencite{gartner2003on,kashima2003marginalized,shervashidze2011weisfeiler,kriege2020a}. A graph kernel defines a similarity measure between pairs of graphs. To classify a graph, similarity scores are computed with respect to all training graphs and used as feature vectors for a vector-based classifier that operates on vector inputs. A common approach is to combine graph kernels with support vector machines (SVMs)~\parencite{cortes1995support}, allowing for the classification of graphs based on these feature vectors. 
Different graph kernels vary in their definition of similarity, and most can be grouped into the following four design paradigms~\parencite{kriege2020a}. The first paradigm is referred to as neighborhood aggregation-based graph kernels. These assign labels to nodes based on their local neighborhood. Similar labels indicate a high similarity between those nodes. The Weisfeiler-Leman kernel~\parencite{shervashidze2011weisfeiler} and its modifications~\parencite{hido2009a,morris2017glocalized} are prominent examples. The second paradigm defines assignments between the nodes in two graphs~\parencite{frohlich2005optimal,johansson2015learning,woznica2010adaptive,kriege2016on}.
Another paradigm considers sets of subgraphs and measures the similarity of two graphs by comparing these sets~\parencite{horvath2004cyclic,shervashidze2009efficient,costa2010fast}.
As the last paradigm, graph kernels based on walks and paths were proposed. These include shortest-path kernels~\parencite{borgwardt2005shortest,hermansson2015generalized} and random walk kernels~\parencite{gartner2003on,kashima2003marginalized,vishwanathan2010graph,kriege2014explicit,kriege2019aunifying}.
Extensions of these kernels address challenges such as handling continuous node attributes~\parencite{kriege2012subgraph,feragen2013scalable,orsini2015graph,morris2016faster}, improving computational efficiency~\parencite{kriege2019computing}, and analyzing the theoretical expressivity of kernel-based graph classifiers~\parencite{gartner2003on,johansson2014global,johansson2015learning,oneto2017measuring,kriege2018a}, as discussed in \Cref{sec:2:expressivity}.

Despite their strong performance on benchmark datasets~\parencite{cai2018asimple,kriege2020a}, this thesis focuses on differentiable methods, which have shown better scalability and performance for domains for which large amounts of data are available~\parencite{he2016deep,vaswani2017attention}.





\subsection{Differentiable Classifiers}
Due to the improved availability of larger amounts of data and computing resources~\parencite{hu2021ogblsc}, graph classifiers that can effectively benefit from these developments are desired. Differentiable classifiers have emerged as a promising method across various domains, which can benefit more effectively from large amounts of data and computing resources~\parencite{lecun2015deep}. Here, graph classifiers produce a continuous output, e.g., class probabilities $\hat{\vy}\in[0,1]^{|\gY|}$, from which a single class can be predicted using the class with maximal probability. For notational simplicity, we will further define graph classifiers to take the matrix notation of a graph as input directly, i.e., it is a function $f(\mA,\mX) = \hat{\vy}$ taking the adjacency matrix $\mA$ and a matrix of node representations $\mX$ as arguments. 

Based on the continuous class probabilities $\hat{\vy} = f(\mA,\mX)$ and the true label $y\in\left[\left|\gY\right|\right]$ in numerical form, we compute a differentiable loss $\ell\bracks{y,\hat{\vy}}$, where $\ell$ for graph classification is typically chosen to be the cross-entropy loss
\begin{equation}
    \ell\bracks{y,\hat{\vy}} = - \sum_{c\in[|\gY|]} \mathbbm{1}\bracks{{c = y}}\log\bracks{\hat{\evy}_c}  \, ,
\end{equation}
where $\mathbbm{1}(\cdot)$ is the indicator function that is one when the condition provided as its argument is true, and zero otherwise. 
A differentiable classifier contains a set of parameters $\gW$, for which the partial derivative $\frac{\partial \ell\bracks{y,\hat{\vy}}}{\partial w}$ can be computed for each parameter $w\in\gW$. A differentiable classifier typically consists of a sequence of differentiable operations, so that the partial derivative can be decomposed into a sequence of partial derivatives of each operation under the use of the chain rule.
Gradient descent-based optimization approaches then update each parameter $w$ as
\begin{equation}
    w^\prime = w - \lambda \frac{\partial \ell(y,\hat{\vy})}{\partial w}
\end{equation}
where $\lambda\in\R$ is called the learning rate. These updated parameters $\gW^\prime$ are then used within a new graph classifier, and the process is iterated for different graphs. Updating the parameters once is referred to as an optimization step.
An epoch refers to iterating over all graphs in the dataset once. The full training process typically consists of multiple epochs.
While this optimization procedure uses a single graph per optimization step, variations like batch gradient descent use the average partial derivative over multiple graphs for each optimization step.
This end-to-end optimization or learning has been shown to scale effectively to large amounts of available data in other domains, such as computer vision~\parencite{he2016deep} or natural language processing~\parencite{vaswani2017attention}.

A differentiable graph classifier is typically split into two parts. The final part is a function
\begin{equation}
\label{eq:2:diff_classifier}
    \omega(\mX) = \hat{\vy}
\end{equation}
that typically takes a matrix of node representations $\mX$ as the sole argument of a permutation invariant function $\omega$ mapping $\mX$ to class probabilities $\hat{\vy}\in[0,1]^{|\gY|}$.
This permutation invariant function is equivalent to operating on multisets $\{\{\,x(v_i) \mid v_i\in\gV\,\}\}$ for a graph signal $x$ in function form.
For example, all rows of $\mX$ may be summed into a single vector. Then, a differentiable function taking a vector as an argument, such as a multi-layer perceptron (MLP)~\parencite{amari1967a}, can be applied.

As a classifier $\omega$ operates only on the node representations $\mX$ and not on the structural representation $\mA$ of the graph, the node representations $\mX$ need to capture structural properties of the graph and information about interactions between node attributes.
Initially available node representations $\mX$ may not capture these properties. Thus, considerable research has been dedicated to obtaining more informative node representations $\mX^\prime$ that capture the structural information and interactions between node representations of a given attributed graph. This can be achieved using the graph convolution and its approximations, which we will introduce next.

\subsubsection{Convolution on Graphs}
\label{sec:fundamentals:graph_conv}
The convolution is a classic operation in signal processing and probability theory. In machine learning, it has been particularly successful as a part of differentiable classifiers in computer vision. It is an operation on two functions that produces a third function by shifting one of the functions.
In the graph domain, the two input functions typically correspond to a graph signal $x:\gV\to\R$ and a filter $\theta : \gV\to\R$ for nodes $\gV$ of a graph $\gG=(\gV,\gE,x)$. The graph convolution $*$ produces a function $\theta * x \colon \gV\to\R$. 
For notational simplicity, we will use the equivalent matrix notation $\vx^\prime = \vtheta * \vx$ where $\vx^\prime,\vx,\vtheta\in\R^{|\gV|}$ for the graph convolution. 
While the convolution is typically defined using the shift operation of a given domain, there is no shift operation available for graphs, and the convolution does not directly apply to graphs. Instead, the graph convolution is defined based on the convolution theorem~\parencite{oppenheim2013discrete}.
It states that the Fourier transform of the convolution $*$ of two functions is equal to the element-wise product of the Fourier transformed functions. 
However, the Fourier transform is also not applicable to graphs, as detailed in \Cref{sec:fundamentals:graph_fourier}.
Instead, the Fourier transform is defined as the graph Fourier transform and the graph convolution in graph Fourier space accordingly as 
\begin{equation}
F(\vtheta * \vx) = F(\vtheta) \odot F(\vx)\, ,
\end{equation}
where $F$ is the graph Fourier transform~\parencite{hammond2011wavelets,bruna2014spectral}.

As described in Section~\ref{sec:fundamentals:graph_fourier}, the graph Fourier transform $F(\vx)=\mU^\top\vx$ is defined as the projection of $\vx$ onto the eigenvectors $\mU$ of the graph Laplacian $\mL_\tsym$ of $\gG$~\parencite{bruna2014spectral}. 
Using the inverse graph Fourier transform $F^{-1}(\vx) = \mU\vx$ (see Section~\ref{sec:fundamentals:graph_fourier}), the graph convolution is computed as
\begin{equation}
\begin{split}
  \vtheta * \vx &= \mU\bracks{\mU^\top\vtheta \odot\mU^\top\vx} \\
  &= \mU\tdiag(\vw)\mU^\top\vx
  \end{split}
\end{equation}
where $\vw =\mU^\top\vtheta$ is the vector containing the filter coefficients in graph Fourier space. Due to this definition utilizing spectral properties of the graph Laplacian, it is also called the spectral graph convolution~\parencite{bruna2014spectral,henaff2015deep}.
The graph convolution is differentiable, i.e., the gradient $\frac{\partial (\vtheta * \vx)}{\partial \vtheta}$ with respect to the filter $\vtheta$ can be directly computed.

In Section~\ref{sec:2:isomorphism}, we discussed permutation invariance as a property for functions operating on matrices of graphs. However, while we considered there functions mapping from the graph domain to the vector domain, here we consider functions mapping from the graph domain to the graph domain.
As such, the arbitrary ordering of the vectors $\vx$ and $\vx^\prime = \vtheta * \vx$ should align. This property of a function is called permutation equivariance.
Similar to the definition of permutation invariance in Definition~\ref{def:perm_inv}, permutation equivariance is generally defined as the existence of a permutation matrix that permutes the input and output in the same way:
\begin{definition}[Permutation Equivariance]
A function $f$ is \emph{permutation equivariant} under node orderings if
\begin{equation}
    \mP f(\mA,\mX) = f\bracks{\mP\mA\mP^\top,\mP\mX}\, .
\end{equation}
for all $n\in\mathbb{N}$ and a $d\in\mathbb{N}$, permutation matrices $\mP\in\{0,1\}^{n\times n}$, adjacency matrices $\mA\in\{0,1\}^{n\times n}$, and node representation matrices $\mX\in\R^{n\times d}$.
\end{definition}

The graph convolution is a permutation-equivariant function
\begin{equation}
    \psi(\mA,\vx) = \vtheta * \vx\, .
\end{equation}
The output $\vx^\prime = \psi(\mA,\vx)$ of a graph convolution can be viewed as a new graph signal or vector of node representations that captures structural information about the graph and interactions between node attributes.
These new node representations $\vx^\prime$ can then be used as the argument for further graph convolutions or a differentiable graph classifier $\omega$, as given by \Cref{eq:2:diff_classifier}. It is also end-to-end differentiable, allowing the computation of the gradients $\frac{\partial \omega(\psi(\mA,\vx))}{\partial \vtheta}$ and optimization by some form of gradient descent.

However, this convolution is typically not applied for multiple reasons.
Most importantly, the graph Fourier transform $F$ depends on the graph $\gG$. 
The graph Fourier basis vectors and the number of such vectors are typically distinct across graphs.
Therefore, the filter $\vtheta$ is not applicable across distinct graphs.
Another disadvantage of this graph convolution is the necessity to compute the eigendecomposition for each graph to obtain the graph Fourier transform, which has quadratic complexity. As $\mU$ is a dense matrix, applying the graph Fourier transform $F$ also has quadratic runtime complexity. 
Thus, approximations of the spectral graph convolution are typically considered.

\subsubsection{Message-Passing Neural Networks}
To reduce the computational complexity of the graph convolution, the message-passing scheme was derived as a localized approximation of the graph convolution~\parencite{kipf2017semi}. It is based on the idea of updating the representations of each node by incorporating the node representations of all adjacent nodes. 
Given a graph with adjacency matrix $\mA$ and node representations $\mX$, the message-passing scheme updates the node representations $\mX^\prime = g(\mA,\mX)$ by combining the representation of each node with all neighboring node representations. It is generally defined as a node-wise function
\begin{equation}
\label{eq:2:message-passing}
    \bracks{g(\mA,\mX)}_{i,:} = 
     \phi\bracks{\mX_{i,:},\bigoplus\left\{\left\{\,\alpha_{(i,j)}\cdot\psi\bracks{\mX_{i,:},\mX_{j,:}}\mid v_j\in \gN_i\,\right\}\right\}}
\end{equation}
where $\psi$ and $\phi$ are functions on pairs of node features, $\alpha_{(i,j)}\in\R$ is an edge-specific scalar, and $\oplus$ is a permutation invariant set aggregation function, e.g., sum or mean. 
Message-passing operations are localized operations due to the aggregation over the set of neighboring nodes $\gN_i$, rather than over all nodes, as in the spectral graph convolution. It can also be applied across different graphs~\parencite{hamilton2017inductive}.
By this definition, any such message-passing function $g$ is permutation equivariant. 
Similar to the graph convolution, the output $\mX^\prime = g(\mA,\mX)$ of \Cref{eq:2:message-passing} represents an updated node representation matrix that is supposed to contain structural properties and information about interactions of node representations.

\paragraph{Examples}
To provide intuition for the message-passing scheme, we now present three established examples. We assume the node representation matrices have shapes $\mX\in\R^{n\times d}$ and $\mX^\prime\in\R^{n\times d^\prime}$.
First, the graph convolutional network (GCN)~\parencite{kipf2017semi} was proposed as a localized approximation of the spectral graph convolution that updates node representations as
\begin{equation}
\mX^\prime_{i,:} = \sum_{v_j\in\gN_i} \frac{1}{\sqrt{d_id_j}}\mX_{j,:}\mW
\end{equation}
where $\mX_{j,:}\in\R^{1\times d}$ and $\mW\in\R^{d\times d^\prime}$ is a feature transformation.

The SAGE convolution from GraphSAGE~\parencite{hamilton2017inductive} utilizes a second feature transformation to control the effect of the previous node representations on the resulting ones. This leads to the updated node representations
\begin{equation}
    \mX^\prime_{i,:} = \mX_{i,:}\mW_{(1)} + \sum_{v_j\in\gN_i}\frac{1}{d_i}\mX_{j,:}\mW_{(2)}\, ,
\end{equation}
where $\mW_{(1)},\mW_{(2)}\in\R^{d\times d^\prime}$ are two feature transformation matrices of the same shape.

The graph attention network (GAT)~\parencite{velickovic2017graph} learns edge weights as a function of the adjacent node representations. The updated node representations are then obtained as
\begin{equation}
\mX^\prime_{i,:} = \sum_{v_j\in\gN_i} a_{(i,j)}\mX_{j,:}\mW
\end{equation}
where $a_{(i,j)}$ are the attention coefficients that satisfy $a_{(i,j)}>0$ for every edge $\bracks{v_i,v_j}$ and $\sum_{v_j\in\gN_i} a_{(i,j)} = 1$ for all $i\in[n]$. GAT is often employed with multiple attention heads, which refers to computing multiple attention-based updates with independent parameters and aggregating the results.

\paragraph{Message-Passing Neural Networks for Graph Classification}
As a single message-passing step can only produce node representations that are affected by the direct neighbors of each node, these are typically applied iteratively.  
Such a sequence 
\begin{equation}
\rho\bracks{\mA,\mX} := g_k\bracks{\mA,g_{k-1}\bracks{\dots g_1\bracks{\mA,\mX}\dots}}
\end{equation} 
of $k$ message-passing iterations $g_1,\dots,g_k$ produces the final node representations.
A message-passing neural network (MPNN) is then defined as a graph classifier
\begin{equation}
    f(\mA,\mX) = \omega\bracks{\rho\bracks{\mA,\mX}}
\end{equation}
combining $k$ iterations of message-passing with the classifier presented in \Cref{eq:2:diff_classifier}. Here, $\rho$ is a permutation equivariant function and $\omega$ is a permutation invariant function.

While many choices for the function $\rho$ and individual message-passing functions $g_1,\dots,g_k$ have been proposed, finding the optimal choices and identifying their limitations or beneficial properties is a key aspect of many research efforts. We will discuss various choices and their properties in the following chapters.

\subsubsection{Further Machine Learning Tasks on Graphs}
Apart from the graph classification task, which aims to predict a single value for each graph, many other tasks are of interest.
Predicting values for individual nodes is of interest for many applications. For example, when a graph corresponds to a road network, we might be interested in predicting traffic at multiple locations~\parencite{roth2022forecasting}.
These node-level tasks include node classification and node regression, where a discrete class or a continuous value is predicted for every node, respectively.
The graph convolution and MPNNs can be easily adapted to such tasks.

By applying a permutation equivariant function $\rho(\mA,\mX)$, the updated node representations $\mX^\prime = \rho(\mA,\mX)$ can contain information about the structural properties of each node and their combination with node attributes. The vector of representations $\mX_{i,:}^\prime$ of each node $i$ is then used to perform node-level prediction tasks. That means, based on a function $h\colon \mathbb{R}^d\to \mathcal{Y}$, we can assign a class or continuous prediction $\hat{y}_{(i)} = h\bracks{\mX_{i,:}^\prime}$ to every node $v_i$. 

Similarly to these node-level prediction tasks, edge-level prediction tasks can be performed. For each node combination $\bracks{v_i,v_j}$ of interest, a classifier $e\colon \mathbb{R}^{d^\prime}\times\mathbb{R}^{d^\prime}\to\mathcal{Y}$ can be applied to the representations of the two nodes, i.e., $\hat{\vy}_{(i,j)} = e\bracks{\mX_{i,:}^\prime,\mX_{j,:}^\prime}$. Many other tasks are similarly defined based on transformed node representations $\mX^\prime = \rho\bracks{\mA,\mX}$. This concludes the foundational background required for the following sections.

\clearpage
    \cleardoublepage
    \chapter{Extending The Understanding of Graph Convolutions}
\label{sec:understanding}

\textit{Many message-passing neural networks (MPNNs) exhibit a degradation of performance on several tasks as the number of message-passing iterations increases. Several related phenomena - such as over-smoothing, over-correlation, over-squashing, and vanishing gradients - have been proposed, but their interrelations and underlying theoretical foundations remain poorly understood. 
In this chapter, we build upon prior work on over-smoothing and present an in-depth theoretical analysis of widely used MPNNs.
We identify two key properties of a commonly used class of MPNNs that fundamentally limit their performance. The first, shared component amplification, refers to the same component of the representations being amplified in all feature channels during each iteration. The second, component dominance, describes how, with an increased number of iterations, a single component of the node representations gets dominantly amplified while all other components are progressively suppressed.
Our analysis provides a deeper theoretical and intuitive understanding of several known phenomena, enabling the development of methods that avoid these properties.
We validate our analysis through a series of experiments that empirically confirm our theoretical findings and expand upon them.
}

\section{Introduction}
As localized approximations of the spectral graph convolution, message-passing neural networks (MPNNs) have achieved promising results in various applications, such as traffic forecasting~\parencite{roth2022forecasting} or molecular prediction~\parencite{stark2022equibind}. However, it has also been observed that the performance of MPNNs degrades for various tasks when increasing the number of message-passing iterations~\parencite{li2018deeper}. For many tasks, the best performance is achieved after only two such iterations~\parencite{kipf2017semi}. For other tasks, a single iteration is already detrimental to the performance~\parencite{rusch2022graph}. Initially, this performance degradation was observed for the graph convolutional network (GCN)~\parencite{kipf2017semi}, with various other methods exhibiting similar degradation~\parencite{li2018deeper,nt2019revisiting}. 

Studying the causes behind this performance degradation is crucial for multiple reasons. Many tasks require the combination of representations that are not directly connected or even far apart within a graph~\parencite{gilmer2017neural,alon2021on}. Multiple iterations also allow for more feature transformations and non-linearities, potentially allowing for more complex features.
A better understanding also ideally provides insights into the severity of this phenomenon for the performance of shallow MPNNs with few iterations. 

Many previous works studied this performance degradation empirically or theoretically, such as over-smoothing~\parencite{li2018deeper,nt2019revisiting,oono2020graph,cai2020anote,giovanni2023understanding,rusch2023survey}, over-squashing~\parencite{alon2021on,topping2022understanding}, and over-correlation~\parencite{jin2022feature}. 
However, the theoretical details of these phenomena remain poorly understood. 
As a result, current research on performance degradation is scattered and many different approaches to mitigate aspects of these phenomena have been proposed. 
Several works attribute the performance degradation to a phenomenon referred to as over-smoothing, which refers to the representations of different nodes becoming increasingly similar with an increased number of layers~\parencite{li2018deeper,oono2020graph}.
Understanding the mathematical properties of MPNNs leading to over-smoothing has been an active area of research~\parencite{li2018deeper,nt2019revisiting,oono2020graph,cai2020anote}.
However, previous studies have not managed to provide comprehensive theoretical insights, even for linearized cases. Multiple different definitions and intuitions for over-smoothing currently coexist, with the role of learnable parameters remaining unclear. 
Therefore, current methods to mitigate over-smoothing are built on an incomplete theoretical understanding and tackle different aspects of the phenomenon.
It is unclear which of the many aspects related to the performance degradation are most promising for focusing future research on.

In this chapter, we improve the theoretical understanding of MPNNs. In \Cref{sec:understanding_sota}, we provide an extensive overview of current insights into over-smoothing and over-correlation. We include both empirical and theoretical studies.
In \Cref{sec:understanding:fourier}, we conduct an empirical analysis of convolutional filters in the graph Fourier domain. As MPNNs were derived as localized approximations of the spectral graph convolution defined using the graph Fourier transform, we compare their expressible convolutional filters. To clarify conflicting definitions of over-smoothing in previous work, we find that a vanishing norm of representations overshadows potential insights of previous studies in \Cref{sec:understanding:zero}.

In \Cref{sec:understanding:rank}, we present our main theoretical study. We identify two key properties of a general class of MPNNs. We refer to the first property as shared component amplification (SCA) that describes a property of a single message-passing iteration. We find that the same component gets amplified for any feature transformation and across all feature channels.
As the second property, we identify that a single component is increasingly dominantly amplified, with all other components being relatively filtered out. We term this property component dominance (CD).
These properties lead to a phenomenon that we call rank collapse, where representations become closer to a lower-dimensional subspace. We further show that the observed phenomenon of over-smoothing occurs as a special case for particular aggregation functions. We also introduce a property that allows message-passing methods to prevent the underlying phenomena. This property requires message-passing methods to use multiple distinct aggregation functions and feature transformations. 

In \Cref{sec:3:summary}, we summarize our extensive findings, provide definitions for the varying phenomena, and show how over-smoothing and rank collapse can be prevented by methods that directly change the underlying properties.
We empirically support our theoretical findings with various experiments in \Cref{sec:3:eval}. 
\begin{figure*}[tb]
    \centering
    \def\svgwidth{\textwidth}
    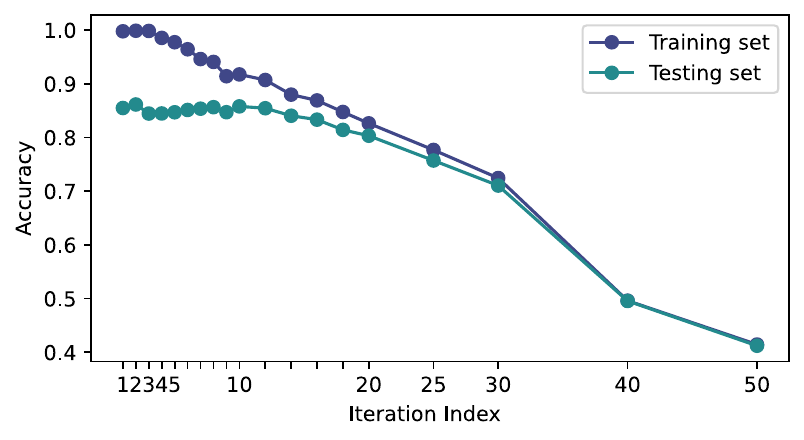
    \caption[Training and testing accuracy for varying numbers of iterations.]{Training and testing accuracy achieved on the Cora~\parencite{sen2008collective} dataset for training GCN models with different numbers of iterations. This experiment is based on \textcite{kipf2017semi}.}
    \label{fig:3:degradation}
\end{figure*}

\section{Related Work}
\label{sec:understanding_sota}
While a consensus in previous work on over-smoothing is that node representations become more similar with an increasing number of iterations~\parencite{li2018deeper,oono2020graph,rusch2023survey}, the details are less clear. Here, we detail the available insights into over-smoothing from previous works. We review available studies and provide proofs of key theoretical statements that our work builds upon. 

Most theoretical studies consider an attributed graph $\gG = (\mA,\mX)$ in matrix form, where $\mA\in\R^{n\times n}$ is the adjacency matrix representing a graph with $n$ nodes, and $\mX\in\R^{n\times d}$ is the matrix containing $d$ attributes for each node. Unless otherwise indicated, we assume $\gG$ is an undirected graph that is connected and non-bipartite. Previous works also mainly consider the graph convolutional network (GCN)~\parencite{kipf2017semi} as an approximation to the spectral graph convolution. A message-passing iteration of the GCN in matrix notation is defined as
\begin{equation}
    \mX^{(k)} = \sigma\bracks{\mA_\tsym\mX^{(k-1)}\mW^{(k)}}\, ,
\end{equation}
where the initial node representations $\mX^{(0)}$ are typically node-wise obtained by applying a node-wise transformation to $\mX$, $\mA_\tsym\in\R^{n\times n}$ is the symmetrically normalized adjacency matrix. The matrix $\mW^{(k)}\in\R^{d\times d}$ contains the learnable parameters and serves as a feature transformation. $\sigma$ is a non-linear activation function, typically assumed to be the identity function for theoretical analysis. 

\begin{figure*}[tb]

     \centering
     \begin{subfigure}[t]{0.49\textwidth}
         \centering
        \def\svgwidth{\textwidth}
     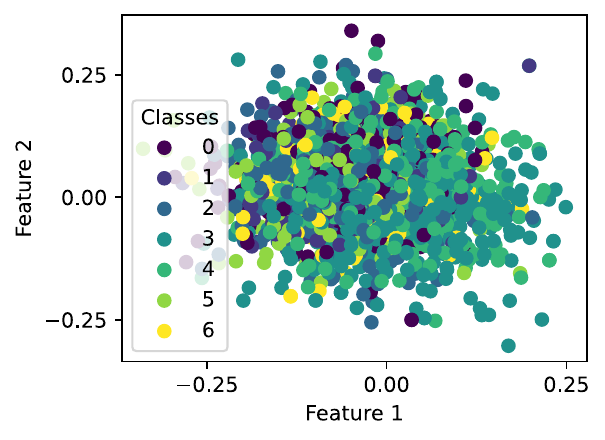
     \caption{Initial representations.}
     \end{subfigure}
     \hfill
     \begin{subfigure}[t]{0.49\textwidth}
         \centering
         \def\svgwidth{\textwidth}
         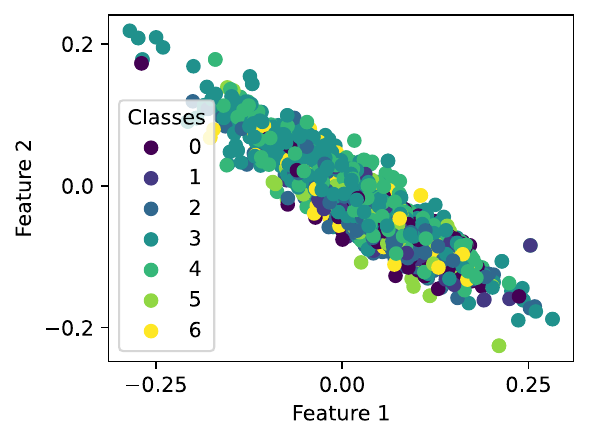
         \caption{After one iteration.}
    \end{subfigure}         
         \begin{subfigure}[t]{0.49\textwidth}
         \centering
         \def\svgwidth{\textwidth}
         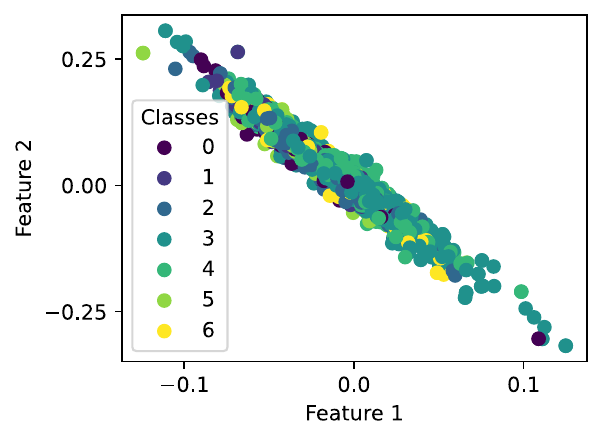
         \caption{After two iterations.}
    \end{subfigure}  
         \begin{subfigure}[t]{0.49\textwidth}
         \centering
         \def\svgwidth{\textwidth}
         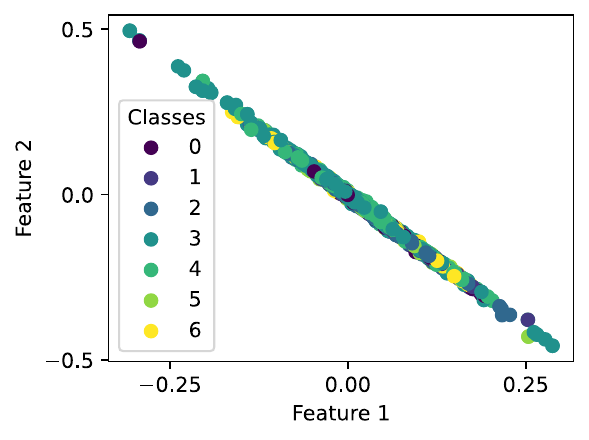
         \caption{After three iterations.}
    \end{subfigure}  
    \caption[Node features after applying GCN iterations.]{Two-dimensional node features $\mX^{(k)}\in\R^{n\times d}$ after applying up to three GCN iterations $\mX^{(k)} = \mA_{\tsym}\mX^{(k-1)}\mW^{(k)}$ where $\mA_\tsym\in\R^{n\times n}$ is given by the Cora dataset~\parencite{sen2008collective}. $\mX^{(0)}$ is given by a linear transformation of the provided node features for Cora. All parameters are randomly initialized. Each dot represents the state of a single node. This experiment is based on \textcite{li2018deeper}.}
    \label{fig:lap_smoothing}
\end{figure*}

\begin{figure*}[tb]
    \centering
    \def\svgwidth{\textwidth}
    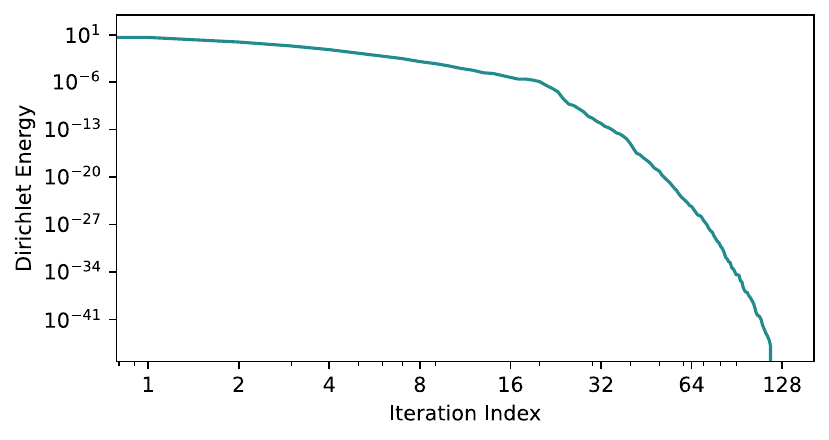
    \caption[Dirichlet energy for the GCN over $128$ iterations.]{Dirichlet energy $E\bracks{\mX^{(k)}} = \trt\bracks{\bracks{\mX^{(k)}}^\top\mL_\tsym\mX^{(k)}}$ for $\mX^{(k)} = \phi\bracks{\mA_\tsym\mX^{(k-1)}\mW^{(k)}}$ where $\phi$ is the ReLU activation function, each $\mW^{(k)}$ is randomly initialized and $k$ indicates the iteration number. The Cora dataset was used for initial features and as graph structure~\parencite{sen2008collective}. Average values over $20$ random initializations are shown. This experiment is based on \textcite{oono2020graph,cai2020anote}.}
    \label{fig:3:motivation:dir_zero}
\end{figure*}

\subsection{Over-Smoothing}
\label{sec:understanding:state:oversmoothing}

Here, we provide an extensive review of available insights into over-smoothing. We also present proofs of the theoretical findings underlying our subsequent analysis.
When the GCN was initially proposed~\parencite{kipf2017semi}, the authors conducted an ablation study that showed the optimal performance was achieved after two iterations of the GCN. Further iterations were shown to decrease performance for both the training and testing sets. To provide an initial visual intuition for this performance degradation of the GCN with an increasing number of iterations, we reproduce their findings and visualize the achieved training and testing accuracy in \Cref{fig:3:degradation}. It confirms that the obtained accuracy for both the training and testing sets degrades with an increased number of iterations. As both sets exhibit similar performance degradation, \textcite{kipf2017semi} argue that overfitting is not the primary cause behind this phenomenon.

\paragraph{The Effect of the Aggregation}
\textcite{li2018deeper} initially introduced the term over-smoothing by studying limitations of the GCN. First, they visualized two-dimensional node representations after applying a varying number of randomly initialized GCN iterations. We reproduce their experiment and provide the visualization in \Cref{fig:lap_smoothing}. They empirically observed that node representations become increasingly aligned with more iterations.

They theoretically studied the effect of the aggregation step $\mA_\tsym$ of the GCN without the feature transformation and non-linear activation function. They relate the aggregation to Laplacian smoothing of node representations and state the following key theorem:
\begin{theorem}[Theorem 1 from \textcite{li2018deeper}]
\label{theorem:lap_smoothing}
Let $\mA\in\R^{n\times n}$ be the adjacency matrix of an undirected and non-bipartite graph. Then, for any $\vx\in\R^n$,
    \begin{align}
    \lim_{k\to\infty}(\mA_{\tsym})^k\vx&=c\cdot \mD^{-1/2}\mathbf{1}\\
    \lim_{k\to\infty}(\mA_{\trw})^k\vx&=d\cdot \mathbf{1}
    \end{align}
     where $\mathbf{1}\in\R^{n}$ and $c,d\in\R$.
\end{theorem}
\begin{proof}
    We will start with the proof for $\mA_\tsym$ and show the necessary changes for $\mA_\trw$ afterwards.
    Consider the decomposition $\vx = \evb_1\mU_{:,1} + \dots + \evb_n\mU_{:,n}$ of $\vx$ as a linear combination of the eigenvectors $\mU$ of $\mA_\tsym$ with coefficients $\vb = \mU^\top\vx\in\R^n$.
    Applying $\mA_\tsym$ to the decomposed form results in $\mA_\tsym\vx = \mA_\tsym \evb_1\mU_{:,1} + \dots + \mA_\tsym \evb_n\mU_{:,n}$. Each $\mU_{:,i}$ is an eigenvector of $\mA_\tsym$ with corresponding eigenvalue $\lambda_i$ by definition. This simplifies the application of $\mA_\tsym$ to 
    \begin{equation}
        \mA_\tsym\vx = \lambda_1\evb_1\mU_{:,1} + \dots + \lambda_n\evb_n\mU_{:,n}\,.
    \end{equation}
    Following properties known from spectral graph theory (\Cref{sec:fundamentals:spectral}), the eigenvalues satisfy the properties $\lambda_1=1$ and $|\lambda_i|<1$ for all $i>1$ and any connected non-bipartite undirected graph. In the limit case, this simplifies to 
    \begin{equation}
    \lim_{k\to\infty}\lambda_i^k = \begin{cases}1, & \text{if $i = 1$} \\ 0, & \text{otherwise}\, .\end{cases}
    \end{equation}
    Combining these properties, we have
    \begin{equation}
        \lim_{k\to\infty} \mA_{\tsym}^k\vx = \evb_1\mU_{:,1}\, .
    \end{equation}
    As known in spectral graph theory (\Cref{sec:fundamentals:spectral}), the corresponding eigenvector $\mU_{:,1}$ of $\mA_\tsym$ is a degree-proportional vector $\mD^{1/2}\mathbf{1}$.

    The proof for $\mA_\trw$ is equivalent, as only the eigenvectors and eigenvalues change. For $\mA_\trw$, we also have $\lambda_1 = 1$ and $|\lambda_i|<1$ for $i>1$. It is also known that any $\mA_\trw$ under the above assumptions has eigenvector $\mU_{:,1} = \mathbf{1}$ (\Cref{sec:fundamentals:spectral}).
\end{proof}

Recalling from \Cref{sec:fundamentals:spectral}, we refer to $\vb = \mU^\top\vx$ as the (spectral) components of $\vx$. We denote the component of $\vx$ corresponding to eigenvector $\mU_{:,i}$ by $\evb_i$.
\textcite{li2018deeper} find components corresponding to eigenvectors of $\mA_\tsym$ to be amplified based on their respective eigenvalue.
They further argue that smoothing is a key reason for the success of the GCN as the smoothing property can ease classification tasks. 
At the same time, they argue that excessive smoothing combines representations that should not be similar, making optimization more challenging. A similar study was later conducted by Keriven~\parencite{keriven2022not}, showing that some smoothing is beneficial for some downstream tasks, while excessive iterations are detrimental. 
\textcite{nt2019revisiting} show that the aggregations $\mA_\tsym$ and $\mA_\trw$ act like low-pass filters that suppress high-frequency components from the node representations, while low-frequency components are less affected. For other aggregation functions $\tilde{\mA}$, it has been shown that high-frequency components can be amplified~\parencite{bo2021beyond,luan2022revisiting}.

\paragraph{The Effect of the Feature Transformation}
While the effect of aggregating features with $\mA_\tsym$ or $\mA_\trw$ is understood, the impact of feature transformations was not considered by \textcite{li2018deeper}. 
\textcite{oono2020graph} consider the distance of node representations to a smooth subspace spanned by the eigenvector $\mU_{:,1}$ corresponding to the largest eigenvalue of $\mA_\tsym$ in magnitude. When applying a feature transformation $\mW^{(k)}$ to node representations, they prove that the distance of the resulting node representations to this subspace can be upper bounded by the largest singular value $\sigma_1$ of $\mW^{(k)}$. When $\sigma_1$ is sufficiently small, representations converge exponentially fast in the number of iterations to the smooth subspace.

Cai and Wang~\parencite{cai2020anote} propose to utilize the graph Dirichlet energy as a metric to quantify over-smoothing. It is defined as
\begin{equation}
\begin{split}
\label{eq:dir_spatial}
E(\mX) &= \trt\bracks{\mX^\top\mL_\tsym\mX} \\
&= \frac{1}{2}\sum_{v_i\in\gV}\sum_{v_j\in\gN_i} \norm{\frac{\mX_{i,:}}{\sqrt{d_i}} - \frac{\mX_{j,:}}{\sqrt{d_j}}}_2^2
\end{split}
\end{equation}
as detailed in \Cref{sec:fundamentals:spectral}. \Cref{eq:dir_spatial} is the spatial form of the Dirichlet energy.
Due to its origin in spectral graph theory, the Dirichlet energy has an intuitive spectral interpretation. 
The graph Laplacian can be decomposed into $\mL_\tsym = \mU\bracks{\mathbf{I}_n -\mLambda}\mU^\top$, where $\mU$ is the orthonormal matrix of eigenvectors and $\mLambda$ is the diagonal matrix of eigenvalues of $\mA_\tsym$.
The null space of $\mL_\tsym$ is spanned by the eigenvector $\mU_{:,1} = \mD^{1/2}\mathbf{1}$ with corresponding eigenvalue $0$, i.e., $\mL_\tsym\mD^{1/2}\mathbf{1} = \mathbf{0}$. In the graph Fourier domain, this vector corresponds to the lowest-frequency (i.e., the smoothest) component. 
The node representations can be expressed in the graph Fourier space as $\mB = \mU^\top\mX$. For every feature channel $c\in[d]$, $\emB_{i,c}$ indicates the strength of its component corresponding to eigenvector $\mU_{:,i}$. As $\mU$ is orthonormal, we have $\mU\mU^\top = \mI_n$. This leads to the equivalent spectral form of the Dirichlet energy 
\begin{equation}
\begin{split}
E(\mX) 
&= \sum_{i=1}^n \sum_{c=1}^d\bracks{1-\lambda_i} \emB_{i,c}^2\, .
\end{split}
\end{equation}

As $\lambda_1 = 1$, we have $1-\lambda_1=0$, and thus all components apart from the components of $\mX$ corresponding to eigenvector $\mU_{:,1}$ are taken into account. When $\mX$ consists of non-zero components only corresponding to $\mU_{:,1}$, the Dirichlet energy is zero.
Cai and Wang~\parencite{cai2020anote} study the effect of GCN iterations on the Dirichlet energy and state the following key proposition:

\begin{proposition}[Proposition 3.4 from \textcite{cai2020anote}]
\label{prop:cai_wang}
Let $\mX\in\R^{n\times d}$ be any matrix, $\phi$ be the ReLU or LeakyReLU activation function, and $\lambda_2$ be the second largest eigenvalue in magnitude of $\mA_\tsym\in\R^{n\times n}$. Further, let $\mW\in\R^{d\times d^\prime}$ be any matrix for some $d^\prime$ with $\sigma_1$ being its largest singular value in magnitude. Then,

    \begin{equation}
       E\bracks{\phi\bracks{\mA_\tsym\mX\mW}} \leq  \sigma_1^2\cdot \lambda_2\cdot E(\mX)\, .
    \end{equation}
\end{proposition}
\begin{proof}
    Equivalently to \parencite[Lemma 3.1]{cai2020anote}, we first prove the inequality
    \begin{equation}
        E\bracks{\mA_\tsym\mX} \leq  \lambda_2\cdot E(\mX)\, .
    \end{equation}
    Consider the decomposition $\mX = \mU\mB$, where $\mB = \mU^\top\mX$ is the graph Fourier transformed representation.
    We use the fact that $\mA_\tsym$ and $\mL$ have the same orthonormal eigenvectors $\mU$. Writing the Dirichlet energy in the spectral form results in 
    \begin{equation}
        E\bracks{\mA_\tsym\mX} = \trt\bracks{\mB^\top\mU^\top\mU\mLambda\mU^\top\mU\bracks{\mI_n-\mLambda}\mU^\top\mU\mLambda\mU^\top\mU\mB}\, .
    \end{equation}
    As $\mU^\top\mU = \mI_n$, this simplifies to
    \begin{equation}
        E\bracks{\mA_\tsym\mX} = \trt\bracks{\mB^\top\mLambda\bracks{\mI_n-\mLambda}\mLambda\mB}\, .
    \end{equation}
    As $\mLambda$ and $\mI_n$ are diagonal matrices, this further simplifies as a sum of scalars:
    \begin{equation}
        E\bracks{\mA_\tsym\mX} = \sum_{i=1}^n \sum_{c=1}^d\bracks{1-\lambda_i}\lambda_i^2\emB_{i,c}^2 
    \end{equation}
    As $1-\lambda_1=0$, this can then be upper bounded by
    \begin{equation}
        E\bracks{\mA_\tsym\mX} \leq \lambda_2^2\cdot E(\mX)
    \end{equation}
    
    Next, we show that $E(\mX\mW) \leq \sigma_1^2 E(\mX)$ \parencite[Lemma 3.2]{cai2020anote} using the spatial form of the Dirichlet energy:
    \begin{equation}
        E(\mX\mW) = \frac{1}{2}\sum_{v_i\in\gV}\sum_{v_j\in\gN_i} \norm{\frac{\mX_{i,:}\mW}{\sqrt{d_i}} - \frac{\mX_{j,:}\mW}{\sqrt{d_j}}}_2^2\, .
    \end{equation}
    By sub-multiplicativity of the norm, this is bounded by
    \begin{equation}
        E(\mX\mW) \leq \frac{1}{2}\sum_{v_i\in\gV}\sum_{v_j\in\gN_i} \norm{\frac{\mX_{i,:}}{\sqrt{d_i}} - \frac{\mX_{j,:}}{\sqrt{d_j}}}_2^2\norm{\mW}_2^2\, .
    \end{equation}
    By definition of the Dirichlet energy, we thus have
    \begin{equation}
        E(\mX\mW) \leq E(\mX)\norm{\mW}_2^2\, .
    \end{equation}
    As the 2-norm $\norm{\mW}_2 = \sigma_1$ is equal to the largest singular value of $\mW$, it follows that 
    \begin{equation}
        E(\mX\mW) \leq \sigma_1^2\cdot E(\mX)\, .
    \end{equation}

    To show that $E\bracks{\phi\bracks{\mX}} \leq E(\mX)$, the property that ReLU and LeakyReLU are 1-Lipschitz is used, i.e., $\norm{\phi(\mP) - \phi(\mQ)} \leq \norm{\mP-\mQ}$ for any matrices $\mP,\mQ$. These activation functions also satisfy that for any non-negative scalar $r$, it holds that $\phi(r\cdot\mP) = r\cdot\phi(\mP)$. Applying these two properties to the Dirichlet energy proves this part \parencite[Lemma 3.3]{cai2020anote}:
    \begin{equation}
        \begin{split}
        E\bracks{\phi(\mX)} &= \frac{1}{2}\sum_{v_i\in\gV}\sum_{v_j\in\gN_i} \norm{\frac{\phi\bracks{\mX_{i,:}}}{\sqrt{d_i}} - \frac{\phi\bracks{\mX_{j,:}}}{\sqrt{d_j}}}_2^2 \\
        &= \frac{1}{2}\sum_{v_i\in\gV}\sum_{v_j\in\gN_i} \norm{\phi\bracks{\frac{\mX_{i,:}}{\sqrt{d_i}}} - \phi\bracks{\frac{\mX_{j,:}}{\sqrt{d_j}}}}_2^2 \\
        &\leq \frac{1}{2}\sum_{v_i\in\gV}\sum_{v_j\in\gN_i} \norm{\frac{\mX_{i,:}}{\sqrt{d_i}} - \frac{\mX_{j,:}}{\sqrt{d_j}}}_2^2 = E(\mX) \\
    \end{split}
    \end{equation}
    
    Combining these three properties concludes this proof.
\end{proof}

Based on \Cref{prop:cai_wang}, the Dirichlet energy is decreasing in each iteration when $\sigma_1<\frac{1}{\lambda_2}$. We visualize the Dirichlet energy converging to zero with an increased number of iterations in \Cref{fig:3:motivation:dir_zero}.

As the Dirichlet energy can be upper-bounded by the singular values of the feature transformation $\mW$, \textcite{zhou2021dirichlet} showed that the Dirichlet energy can similarly be lower-bounded based on the smallest singular value.
They refer to the phenomenon when the Dirichlet energy becomes excessively large as an over-separation of node representations.

\paragraph{Normalization and Over-Separation}
Contrarily, \textcite{giovanni2023understanding} found that a large Dirichlet energy does not necessarily imply an over-separation of node representations. 
They consider the case
\begin{equation}
\label{eq:di_giovanni}
    \mX^{(k)} = \mA_\tsym\mX^{(k-1)}\mW
\end{equation}
without a non-linear activation function where $\mW\in\R^{d\times d}$ is a symmetric matrix that is shared across iterations.
As shown by \parencite{zhou2021dirichlet}, the Dirichlet energy of $\mX^{(k)}$ may grow unboundedly $\lim_{k\to\infty} E(\mX^{(k)}) = \infty$. \textcite{giovanni2023understanding} find that when considering the normalized node representations, the Dirichlet energy still converges to zero, i.e., $\lim_{k\to\infty} E\left(\frac{\mX^{(k)}}{\norm{\mX^{(k)}}_F}\right) = 0$ for \Cref{eq:di_giovanni}. 
Because the smooth component dominates all other components, they refer to this phenomenon as low-frequency dominated dynamics. They prove this is the case for \Cref{eq:di_giovanni} and any symmetric feature transformation $\mW\in\R^{d\times d}$, independently of its singular values:

\begin{proposition}[Theorem 5.3 from \textcite{giovanni2023understanding}]
\label{prop:lfd_symmetric}
Let $\mA\in\{0,1\}^{n\times n}$ be the adjacency matrix of an undirected and non-bipartite graph and $\mW\in\R^{d\times d}$ be a symmetric matrix. Then, for a.e. matrix $\mX\in\R^{n\times d}$,
\begin{equation}
    \lim_{k\to\infty}E\bracks{\frac{\mA_\tsym^k\mX\mW^k}{\norm{\mA_\tsym^k\mX\mW^k}_F}} = 0\, .
\end{equation}    
\end{proposition}
\begin{proof}
We present an alternative proof to the one used by \textcite{giovanni2023understanding}, which aligns better with our subsequent theoretical analysis.

We start by using the definition of the Dirichlet energy and the Frobenius norm:
    \begin{equation}
        E\bracks{\frac{\mA_\tsym^k\mX\mW^k}{\norm{\mA_\tsym^k\mX\mW^k}_F}} = \frac{\trt\bracks{\mW^k\mX^\top\mA_\tsym^k\mL_\tsym\mA_\tsym^k\mX\mW^k}}{\trt\bracks{\mW^k\mX^\top\mA_\tsym^k\mA_\tsym^k\mX\mW^k}}
    \end{equation}
    Using the property $\trt(\mP\top\mQ) = \tvec(\mP)^\top\tvec(\mQ)$ for any suitable matrices $\mP,\mQ$ we have
    \begin{equation}
    E\bracks{\frac{\mA_\tsym^k\mX\mW^k}{\norm{\mA_\tsym^k\mX\mW^k}_F}} = \frac{\tvec\bracks{\mL_\tsym\mA_\tsym^k\mX\mW^k}^\top\tvec\bracks{\mA_\tsym^k\mX^\top\mW^k}}{\tvec\bracks{\mA_\tsym^k\mX\mW^k}^\top\tvec\bracks{\mA_\tsym^k\mX^\top\mW^k}}
    \end{equation}
    Further using the property $\tvec(\mP\mQ\mR) = (\mR^\top\otimes\mP)\tvec(\mQ)$ for any suitable matrices $\mP,\mQ,\mR$, 
    \begin{equation}
        E\bracks{\frac{\mA_\tsym^k\mX\mW^k}{\norm{\mA_\tsym^k\mX\mW^k}_F}} = \frac{\tvec\bracks{\mX}^\top\bracks{\mW^{k}\otimes\mL_\tsym\mA_\tsym^k}^\top\bracks{\mW^k\otimes\mA_\tsym^k}\tvec(\mX)}{\tvec(\mX)^\top\bracks{\mW^{k}\otimes\mA_\tsym^k}^\top\bracks{\mW^k\otimes\mA_\tsym^k}\tvec(\mX)}
    \end{equation}
    Now let $\mA_\tsym = \mU\mLambda\mU^\top$ and $\mW=\mV\mSigma\mV^\top$ be the eigendecompositions of $\mA_\tsym$ and $\mW$. We then restate $\tvec(\mX)=(\mV\otimes\mU)\tvec(\mB)$ using the eigenbasis $\mV\otimes\mU$ where $\tvec(\mB) = (\mV\otimes\mU)^\top\tvec(\mX)$. This simplifies to
    \begin{equation}
    E\bracks{\frac{\mA_\tsym^k\mX\mW^k}{\norm{\mA_\tsym^k\mX\mW^k}_F}} = \frac{\tvec(\mB)^\top\bracks{\mSigma^{2k}\otimes\mLambda^{k}\bracks{\mI_n-\mLambda}\mLambda^{2k}}\tvec(\mB)}{\tvec(\mB)^\top\bracks{\mSigma\otimes\mLambda^{2k}}\tvec(\mB)}
    \end{equation}
    Due to diagonality of $\mSigma^{2k}\otimes\mLambda^{k}\bracks{\mI_n-\mLambda}\mLambda^{k}$ this simplifies to
    \begin{equation}
        E\bracks{\frac{\mA_\tsym^k\mX\mW^k}{\norm{\mA_\tsym^k\mX\mW^k}_F}} = \frac{\sum_{i=1}^{n}\sum_{r=1}^d \emB_{i,r}^2 \lambda_i^{2k}\sigma_r^{2k}\bracks{1-\lambda_i}}{\sum_{i=1}^{n}\sum_{r=1}^d \emB_{i,r}^2 \lambda_i^{2k}\sigma_r^{2k}}\, .
    \end{equation}

    We normalize the numerator and the denominator by dividing by the largest absolute eigenvalue $\sigma_{1}^{2k}$ of the parameter matrix. This leads to the identity
    \begin{equation}
         E\bracks{\frac{\mA_\tsym^k\mX\mW^k}{\norm{\mA_\tsym^k\mX\mW^k}_F}} = \frac{\sum_{i=1}^{n}\sum_{r=1}^d \emB_{i,r}^2 \lambda_i^{2k}\bracks{\frac{\sigma_r^{2k}}{\sigma_{1}^{2k}}}(1-\lambda_i)}{\sum_{i=1}^{n}\sum_{r=1}^d \emB_{i,r}^2 \lambda_i^{2k}\bracks{\frac{\sigma_r^{2k}}{\sigma_{1}^{2k}}}}\, .
    \end{equation}
    As $\left|\lambda_i\bracks{\frac{\sigma_r}{\sigma_{1}}}\right|<1$ for $i>1$, the, the numerator decays to zero as $k\to\infty$ and we have
    \begin{equation}
        \lim_{k\to\infty}E\bracks{\frac{\mA_\tsym^k\mX\mW^k}{\norm{\mA_\tsym^k\mX\mW^k}_F}} = 0
    \end{equation}
    assuming $\emB_{i,1} \neq 0$, which is satisfied for a.e. choice for $\mX$.
    
\end{proof}

Based on this finding, \textcite{giovanni2023understanding} also show cases in which a different frequency component can dominate the node representation.
\textcite{maskey2023fractional} build on this idea and show that the Dirichlet energy of the normalized representations also converges to zero for directed graphs when using the symmetrically normalized adjacency matrix.
Both of these works assume $\mW$ to be a symmetric matrix that is shared across iterations.

\begin{figure*}[tb]
     \centering
     \def\svgwidth{\textwidth}
     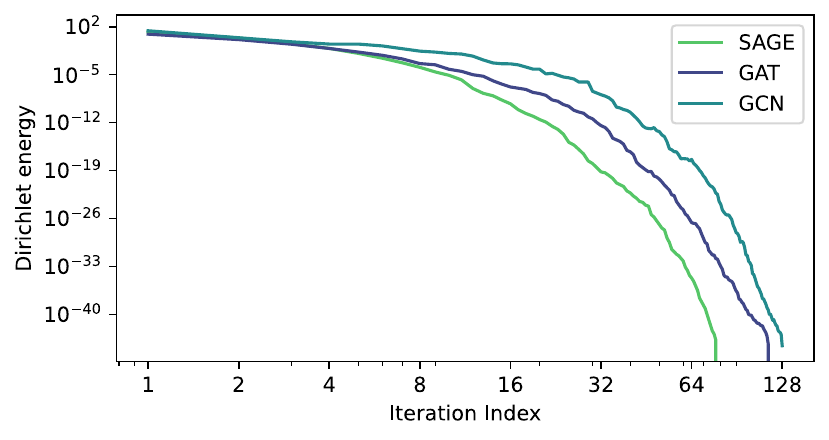
    \caption[Dirichlet energy using the unnormalized graph Laplacian over $128$ iterations.]{Dirichlet energy $E\bracks{\mX^{(k)}} = \trt\bracks{\bracks{\mX^{(k)}}^\top\mL\mX^{(k)}}$ using the unnormalized graph Laplacian $\mL$ for the node representations obtained by three message-passing methods on the Cora dataset~\parencite{sen2008collective}. As the kernel of $\mL$ is spanned by constant-valued vectors, this version of the Dirichlet energy is zero when $\mX^{(k)}$ only has non-zero components belonging to constant-valued vectors. This experiment is based on \textcite{rusch2022graph}.}
    \label{fig:constant_state}
\end{figure*}

\paragraph{Convergence to a Constant State}
\textcite{rusch2022graph} empirically observed that $\mX^{(k)}$ converges to a constant state for the GCN and various other established methods. We reproduced their experiments and visualize this phenomenon for several methods in \Cref{fig:constant_state}. In follow-up works, over-smoothing was defined accordingly as a convergence to a constant state~\parencite{rusch2023gradient,rusch2023survey}. Their definition utilizes a node similarity measure that is zero when all node representations are equal. Over-smoothing is then defined as an exponential convergence to this state over the number of iterations. 
Many of the following studies and methods were built on this definition~\parencite{wu2023demystifying,kothapalli2023a}.

This observation does not align with \Cref{theorem:lap_smoothing} and \Cref{prop:lfd_symmetric}, as these theoretical insights show that $\mX^{(k)}$ converges to a degree-proportional state for some of the same methods. It is currently unclear how these similar yet slightly different observations relate to one another. 
It also remains unclear whether the over-separation of node representations is a relevant phenomenon. Similarly, the complete theoretical implications of applying arbitrary feature transformations remain an open question. As the theoretical analyses mainly considered $\mA_\tsym$ as the aggregation matrix, the effect of further aggregation functions $\tilde{\mA}$ or specific choices for edge weights is also not fully understood.
Due to this incomplete understanding of over-smoothing and related phenomena, potential solutions to over-smoothing may address different aspects behind the umbrella term over-smoothing and may not address the underlying problem. 

\paragraph{Metrics to Quantify Over-Smoothing}

\textcite{chen2020measuring} propose to quantify over-smoothing by computing the average cosine distance between all pairs of nodes, which they call mean average distance (MAD). They consider smoothing as a natural process in GNNs, with smoothing between nodes belonging to the same class as desired. For nodes of different classes, this smoothing adds noise and is undesired. Accordingly, they compute the MAD separately for nodes with the same label and nodes with different labels. The difference between these values is referred to as MADgap.

\textcite{liu2020towards} study the over-smoothing phenomenon by introducing a novel metric called SMV that computes the average Euclidean distance between a pair of nodes. They consider normalized node representations, ensuring that the magnitude of the representations does not affect the metric.

The Dirichlet energy or Dirichlet sum as a metric for theoretically studying over-smoothing was proposed by \textcite{cai2020anote}. In following studies, the Dirichlet energy was also used to empirically identify methods suffering from the over-smoothing phenomenon~\parencite{zhou2021dirichlet}. \textcite{rusch2023survey} generalize the Dirichlet energy to node-similarity measures that aim to quantify the similarity of node representations to a constant state. The Dirichlet energy 
\begin{equation}
    E(\mX) = \trt\bracks{{\mX}^\top\mL\mX}
\end{equation}
defined with the unnormalized graph Laplacian $\mL$ is one metric that satisfies their criteria. 

\subsection{Over-Correlation}
One connected phenomenon that our study also connects to is over-correlation. \textcite{jin2022feature} found the pairwise correlation of feature channels to increase with increased depth. They observe this phenomenon in several message-passing methods by measuring the Pearson correlation coefficient between feature channels. They find that over-smoothing implies over-correlation, but the converse does not hold. As feature channels may be perfectly correlated but contain non-smooth values, over-correlation can arise independently of over-smoothing. They accordingly distinguish that over-smoothing corresponds to a node-wise smoothness, while over-correlation corresponds to a channel-wise similarity. They conclude that additional feature channels are redundant and provide no additional information. 

\textcite{guo2023contranorm} study the effective rank of representations of GNNs and transformers. They find that most singular values of node representations are closer to zero after a few iterations. Similarly to over-correlation, they find this phenomenon to occur even when representations are not similar, as measured by the cosine similarity. They refer to this phenomenon as a more general form of over-smoothing, which they call dimensional collapse.
The phenomenon of dimensional collapse was similarly observed for contrastive learning methods~\parencite{gao2019representation,jing2022understanding}. The theoretical details of over-correlation and dimensional collapse, as well as their relationship to over-smoothing, are not well understood.

\label{sec:understanding:fourier}
\begin{figure*}[tb]
     \centering
     \begin{subfigure}[t]{0.32\textwidth}
         \centering
        \def\svgwidth{\textwidth}
     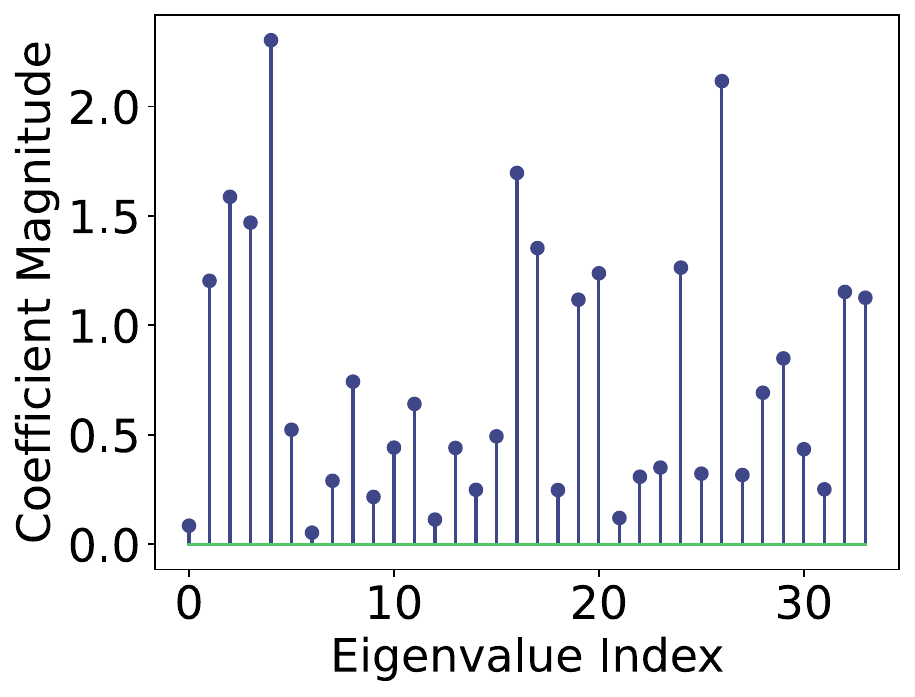
     \caption{Random filter.}
     \end{subfigure}
     \hfill
     \begin{subfigure}[t]{0.32\textwidth}
         \centering
         \def\svgwidth{\textwidth}
         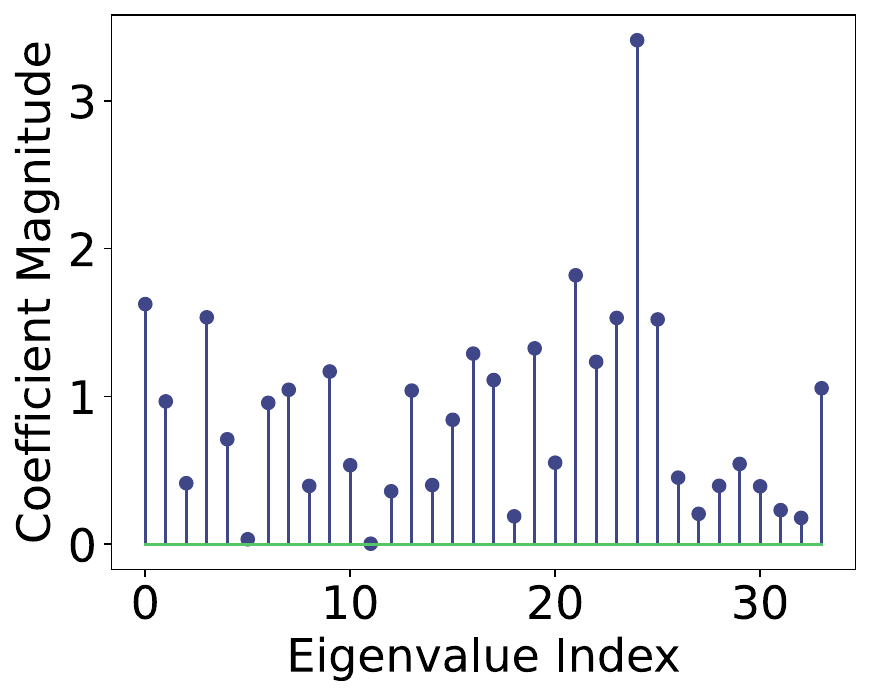
         \caption{Random filter.}
    \end{subfigure}         
    \begin{subfigure}[t]{0.32\textwidth}
         \centering
         \def\svgwidth{\textwidth}
         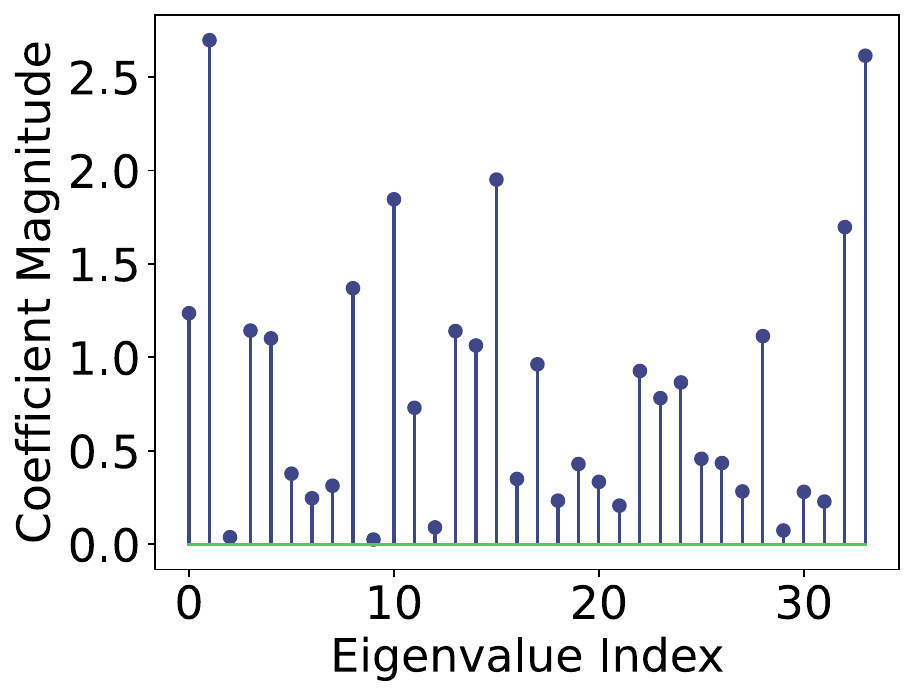
         \caption{Random filter.}
     \end{subfigure}
    \caption[Three random filters in graph Fourier space.]{Coefficient magnitudes $|F(\vtheta)|$ of three random filters $\vtheta$ in graph Fourier space as used for the spectral graph convolution. The KarateClub dataset was used to construct $\mL_\tsym$.}
    \label{fig:fourier_gc}
\end{figure*}

\begin{figure*}[tb]
     \centering
     \begin{subfigure}[t]{0.32\textwidth}
         \centering
        \def\svgwidth{\textwidth}
     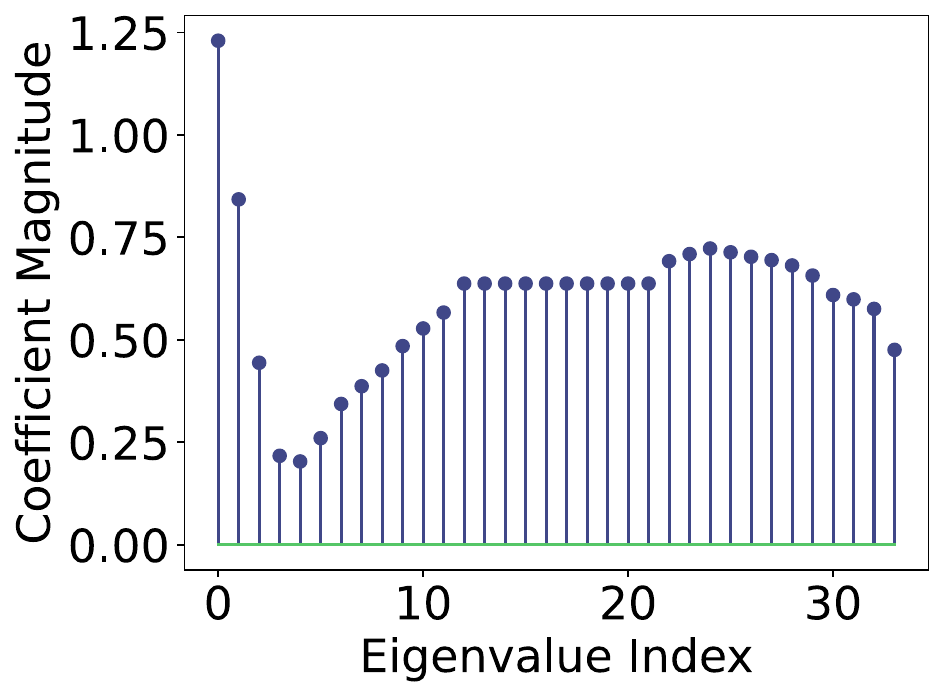
     \caption{$K=2$.}
     \end{subfigure}
     \hfill
     \begin{subfigure}[t]{0.32\textwidth}
         \centering
         \def\svgwidth{\textwidth}
         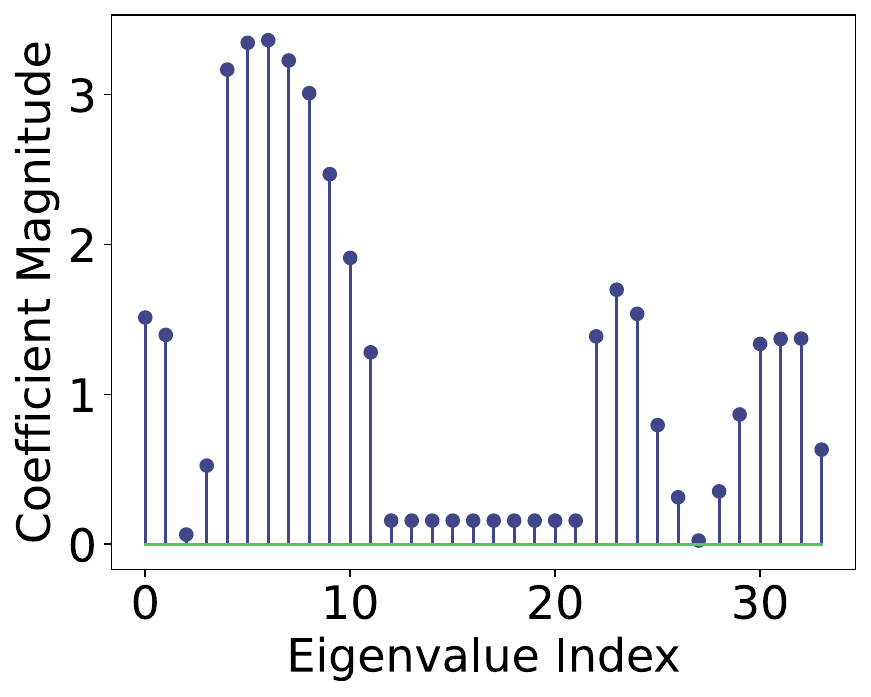
         \caption{$K=8$.}
    \end{subfigure}         
    \begin{subfigure}[t]{0.32\textwidth}
         \centering
         \def\svgwidth{\textwidth}
         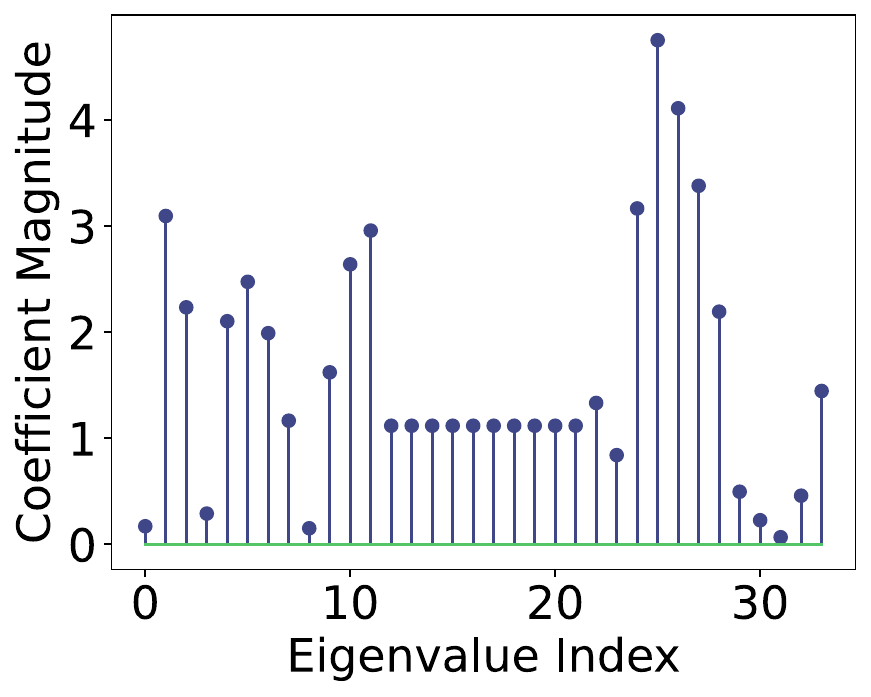
         \caption{$K=16$.}
     \end{subfigure}
    \caption[Chebyshev polynomials as filters in graph Fourier space.]{Coefficient magnitudes $|F(\vtheta)|$ of Chebyshev polynomials $F(\vtheta) = \sum_{k=0}^K w_kT_k(\vlambda)$ as filters in graph Fourier space with three different degrees $K$. The KarateClub dataset was used to construct $\mL_\tsym$.}
    \label{fig:fourier_poly}
\end{figure*}

\begin{figure*}[tb]
     \centering
     \begin{subfigure}[t]{0.32\textwidth}
         \centering
        \def\svgwidth{\textwidth}
     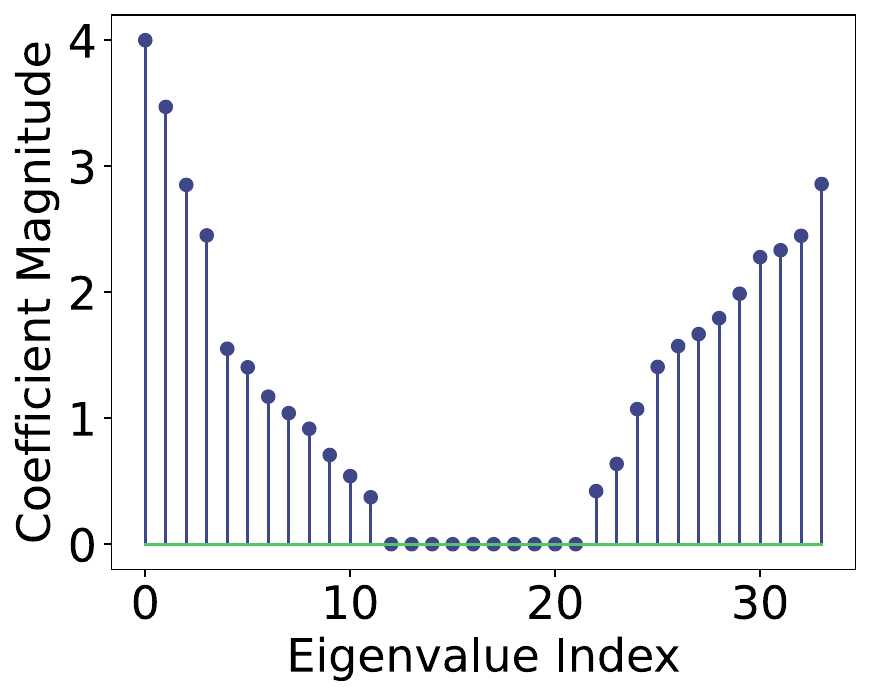
     \caption{$w=4$.}
     \end{subfigure}
     \hfill
     \begin{subfigure}[t]{0.32\textwidth}
         \centering
         \def\svgwidth{\textwidth}
         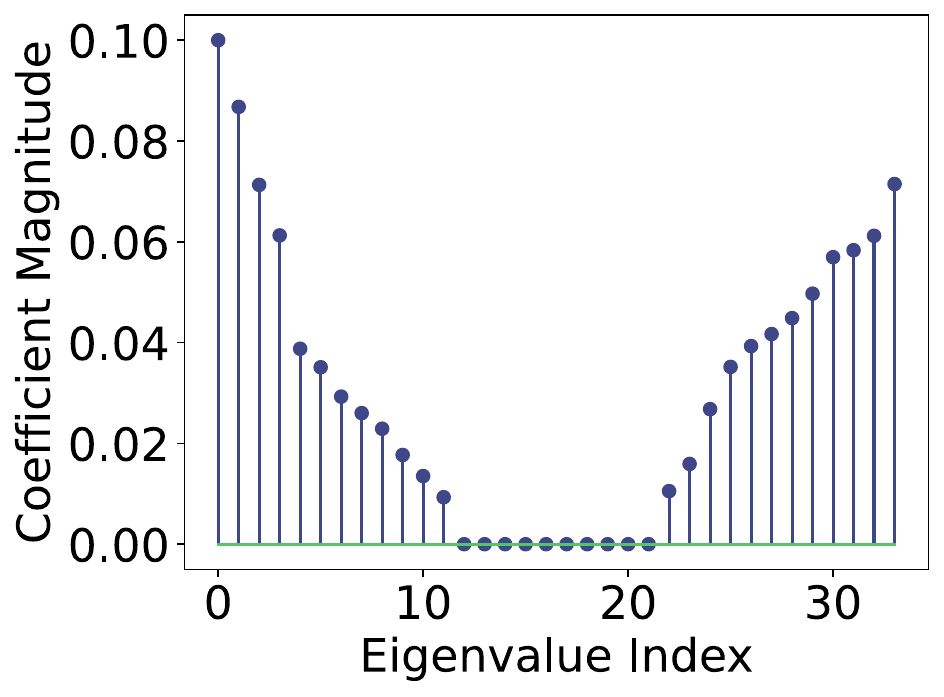
         \caption{$w=0.1$.}
    \end{subfigure}         
    \begin{subfigure}[t]{0.32\textwidth}
         \centering
         \def\svgwidth{\textwidth}
         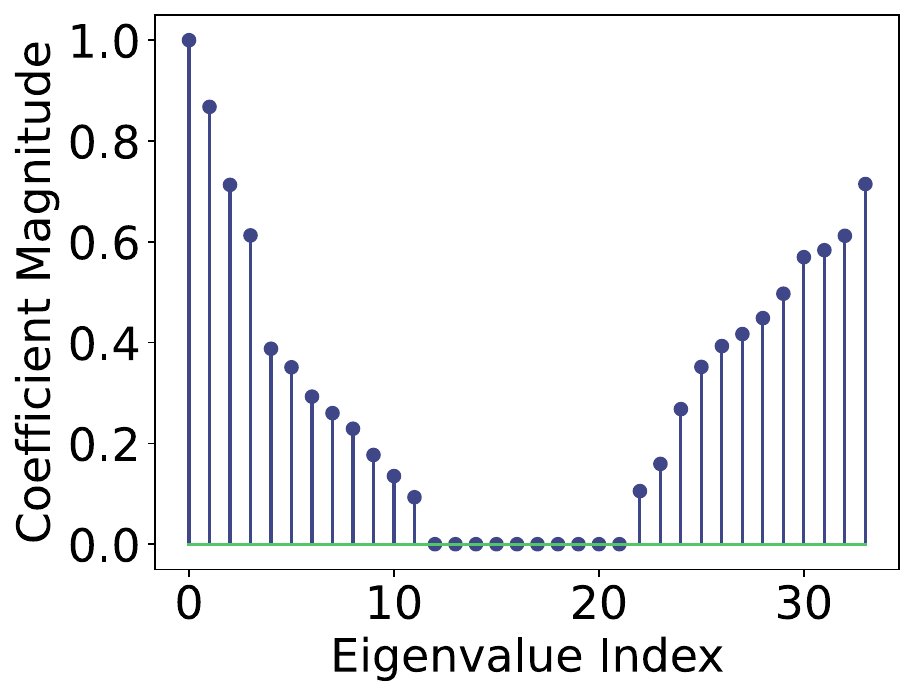
         \caption{$w=-1$.}
     \end{subfigure}
    \caption[Random GCN instantiations as filters in graph Fourier space.]{Coefficient magnitudes $|F(\vtheta)|$ of three filters $F(\vtheta) = w\vlambda$ as proposed for the GCN. The KarateClub dataset was used to construct $\mL_\tsym$.}
    \label{fig:fourier_gcn}
\end{figure*}

\begin{figure*}[tb]
     \centering
     \begin{subfigure}[t]{0.305\textwidth}
         \centering
        \def\svgwidth{\textwidth}
     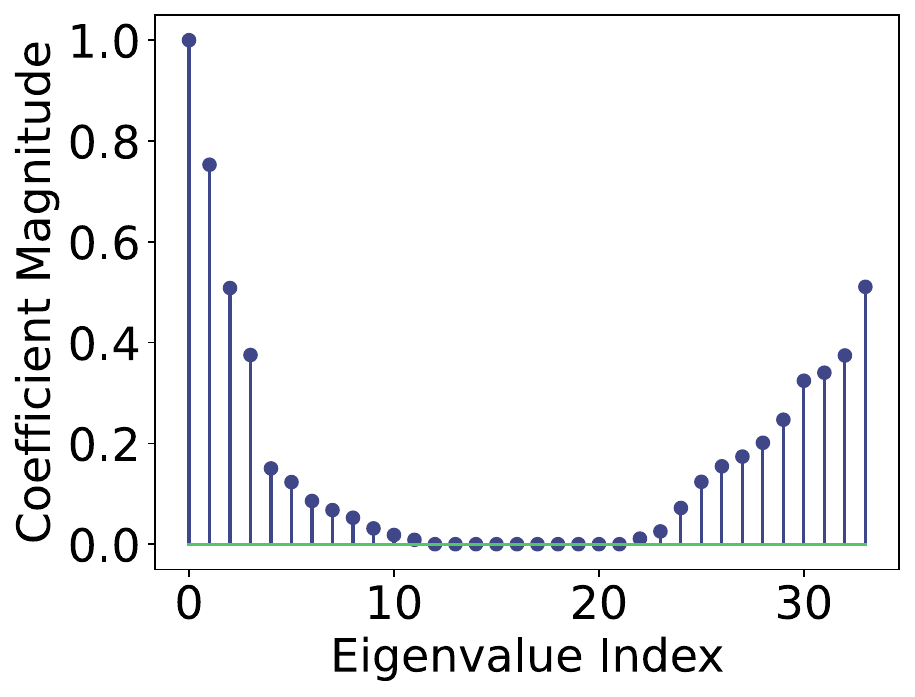
     \caption{Combining $k=2$ GCN filters.}
     \end{subfigure}
     \hfill
     \begin{subfigure}[t]{0.345\textwidth}
         \centering
         \def\svgwidth{\textwidth}
         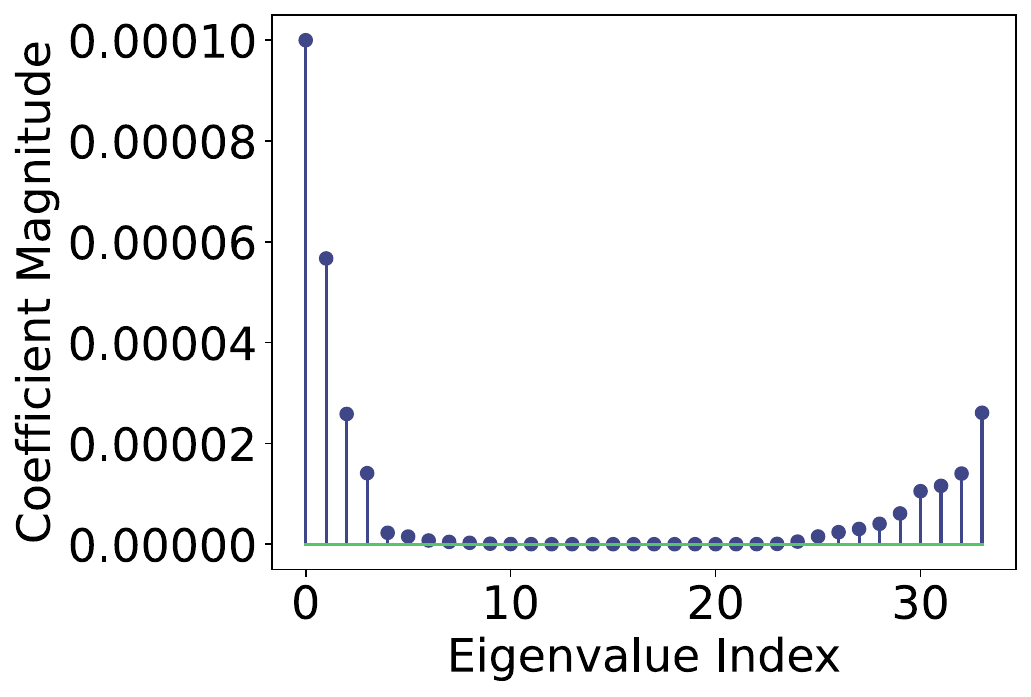
         \caption{Combining $k=4$ GCN filters.}
    \end{subfigure}         
    \begin{subfigure}[t]{0.33\textwidth}
         \centering
         \def\svgwidth{\textwidth}
         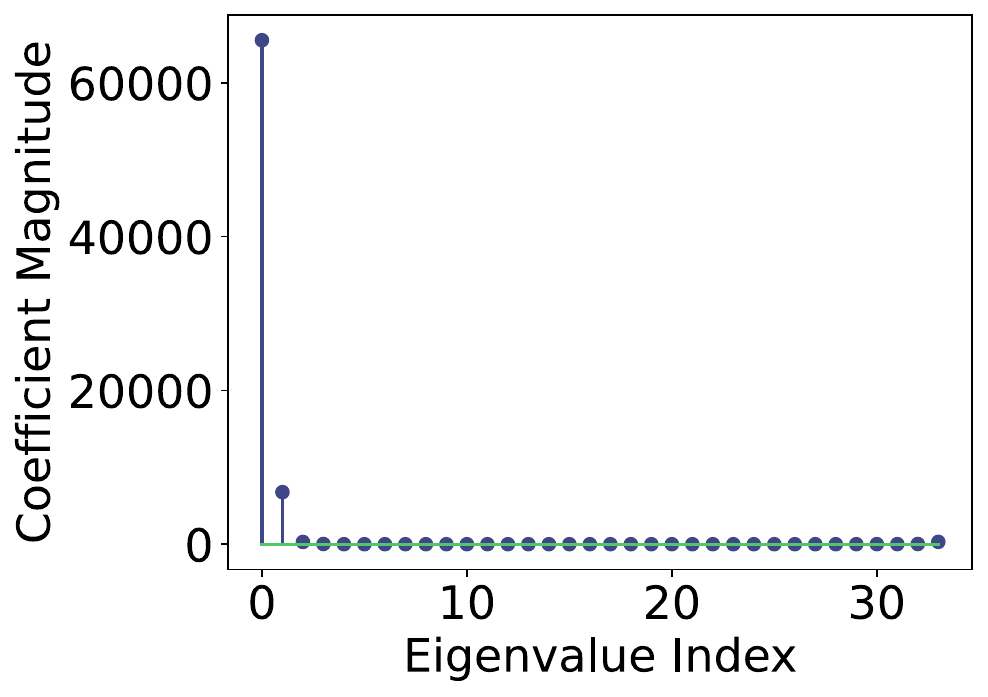
         \caption{Combining $k=16$ GCN filters.}
     \end{subfigure}
    \caption[Repeated GCN iterations in graph Fourier space.]{Coefficient magnitudes $|F(\vtheta)|$ of three filters $F(\vtheta) = w_k\vlambda\odot\dots\odot w_1\vlambda$ with different numbers of iterations as proposed for the GCN. The KarateClub dataset was used to construct $\mL_\tsym$.}
    \label{fig:fourier_gcn_depth}
\end{figure*}

\section{Over-Smoothing in Graph Fourier Space}
Our first analysis to improve our understanding of the over-smoothing phenomenon considers the filters of different graph convolutions. This section closely follows our previously published work presented in~\parencite{roth2025what}. 
As described in \Cref{sec:fundamentals:graph_conv}, the graph convolution is defined in the graph Fourier space 
\begin{equation}
F(\vtheta * \vx) = F(\vtheta) \odot F(\vx)
\end{equation}
where $\vx\in\R^n$ are the node attributes or the graph signal, $\vtheta\in\R^n$ is a filter, and $\odot$ is the element-wise product. The Fourier transform $F$ is defined as the graph Fourier transform given as $F(\vz) = \mU^\top\vz$ for any vector $\vz\in\R^n$, where $\mU$ is the matrix of eigenvectors of the graph Laplacian $\mL_\tsym$ (see \Cref{sec:fundamentals:spectral}).
Here, we visualize several exemplary filters to build intuition for the over-smoothing phenomenon and to support our theoretical analysis in the following sections.
Each coefficient $F(\vtheta)_i\in\R$ determines the extent to which the component $F(\vx)_i$, associated with the $i$-th Fourier basis vector, is amplified or dampened. A small magnitude of the coefficient indicates that the associated component gets dampened, while a large magnitude indicates an amplification of the component.
Ideally, a convolution can amplify and dampen arbitrary components, depending on the choice of the filter $\vtheta$. 

\subsection{Graph Convolutions as Spectral Filters}
The spectral graph convolution allows for any filter $\vtheta$, as detailed in \Cref{sec:fundamentals:graph_conv}. Thus, any filter $F(\vtheta)=\mU^\top\vtheta$ in the graph Fourier space can be represented. We visualize random choices for $\vtheta$ in \Cref{fig:fourier_gc}, visually confirming that this spectral graph convolution can amplify and dampen arbitrary components of the signal $\vx$.

\subsection{Polynomial Approximations as Spectral Filters}
For a more efficient runtime, polynomial approximations of $F(\vtheta)$ were proposed~\parencite{hammond2011wavelets,he2021bernnet}. These typically utilize the eigenvalues $\vlambda\in\R^n$ of the graph Laplacian as a basis. We consider approximations using Chebyshev polynomials $F(\vtheta) = \sum_{k=0}^K \evw_kT_k(\vlambda)$ where $T_k$ is the $k$-th order Chebyshev polynomial with $T_0(\vlambda)=\mathbf{1}$, $T_1(\vlambda)=\vlambda$, and $T_k(\vlambda)=2\cdot\vlambda\odot T_{k-1}(\vlambda) - T_{k-2}(\vlambda)$ for $k>1$~\parencite{hammond2011wavelets}. We visualize such polynomial filters of varying degrees in \Cref{fig:fourier_poly}. 
We observe a smooth polynomial structure of the filter coefficients, with similar eigenvalues being assigned similar values. 

\subsection{Message-Passing as Spectral Filters}
\label{sec:3:fourier:mp}
The GCN~\parencite{kipf2017semi} was derived as a first-order approximation of polynomial graph filters, so that the state of each node is only affected by its direct neighbors. It can be expressed as a graph convolution with a filter in the graph Fourier domain given by $F(\vtheta) = w\vlambda$, where $w\in\R$ is its single parameter. Note that the GCN adds self-loops for computing $\mL_\tsym$, which we ignore here for clarity.
We visualize filters $F(\vtheta) = w\vlambda$ in graph Fourier space for three values of $w\in\R$ in \Cref{fig:fourier_gcn}. As $w$ uniformly scales all filter coefficients by the same value, the relative magnitudes remain constant for any choice of $w\in\R$.
\Cref{fig:fourier_gcn} also visually highlights that the first coefficient $(w\vlambda)_1$ always has the largest magnitude among all filter coefficients, for any non-zero $w\in\R$. Thus, the corresponding smooth component gets amplified maximally for any parameter choice.

This already provides intuition for the over-smoothing phenomenon: under the first-order approximation used by the GCN, any expressible filter $F(\vtheta)$ amplifies the smoothest component the most, as any choice of the parameter $w$ scales all filter coefficients uniformly. When applying the GCN, it is not possible to amplify any component more than the component corresponding to $\mU_{:,1}$ for any choice of parameter $w$.
Thus, every such filter amplifies the smoothest frequency component the most, while relatively suppressing all other frequency components. 

\subsubsection{Shared Component Amplification}
The provided visualizations considered the single-input single-output (SISO) case, where $\vx\in\R^n$ and $\vtheta*\vx\in\R^n$ are vectors, i.e., each node has a single feature. The above methods are typically applied in the multi-input multi-output (MIMO) case, where the graph signal $\mX\in\R^{n\times d}$ has $d$ feature channels for each node. Based on \textcite{bruna2014spectral}, the graph convolution is applied in the MIMO case by applying a separate graph convolution with independent filters $\vtheta_{(p,q)}\in\R^n$ for every combination of input feature channel $i\in[d]$ and output feature channel $q\in[c]$, i.e., the resulting state is obtained as 
\begin{equation}
\label{eq:3:spectral_gcn}
    F\bracks{\mX^{\prime}_{:,q}} = \sum_{p\in[d]} F\bracks{\vtheta_{(p,q)}} \odot F\bracks{\mX_{:,p}}
\end{equation}
with $\mX^{\prime}\in\R^{n\times c}$. The GCN also utilizes this extension to the MIMO case~\parencite{kipf2017semi}, for which we have $F\bracks{\vtheta_{(p,q)}} = w_{(p,q)}\vlambda$ with $w_{(p,q)}\in\R$ for every pair of input channel $p$ and output channel $q$. The parameters $w_{(p,q)}$ can also be arranged into a matrix $\mW\in\R^{d\times c}$, where $\emW_{p,q} = e_{(p,q)}$. \Cref{eq:3:spectral_gcn} can be equivalently stated for the GCN in the typical notation as
\begin{equation}
    \mX^\prime = \mA_\tsym\mX\mW\, .
\end{equation}

We recall that for every choice of parameters $\emW_{p,q}$, each $F\bracks{\vtheta_{(p,q)}}$ amplifies the frequency components in the same way up to a scaling factor that is uniform across all components. Thus, the relative amplification of components is shared across all pairs of input and output feature channels. We refer to this as shared component amplification (SCA). The resulting representations are inherently constrained, and the gains of additional output channels are limited. We will study this phenomenon more deeply theoretically in the following sections and provide a formal definition in \Cref{def:sca}.

Importantly, SCA occurs regardless of the specific basis $\vlambda$ used. Even when a different basis $\hat{\vlambda}$ is chosen, the parameters $\emW_{p,q}$ lead to filters with the same relative amplification of components across all channel combinations. The component corresponding to the maximal absolute value of $\hat{\vlambda}$ is maximally amplified for every input and output combination of feature channels.
Instead, it is desirable to allow for different components in the input signal to be amplified across distinct output channels.

\subsubsection{Component Dominance}
SCA highlights a limitation in the approximation of a single graph convolution $\vtheta*\vx$ used in the GCN. For the GCN as a localized approximation, the resulting state for each node takes information only from its direct neighbors into account. The GCN is typically applied in an iterative fashion. Ignoring the non-linear activation functions, this has form
\begin{equation}
\begin{split}
F\bracks{\vtheta_k}\odot\dots\odot F\bracks{\vtheta_1}\odot F(\vx) &= w_k\vlambda \odot\dots\odot w_1\vlambda\odot F(\vx)\, ,
\end{split}   
\end{equation}
where $\vtheta_i\in\R^n$ is the filter for iteration $i\in[k]$, and $w_i\in\R$ is the corresponding parameter of the GCN.
We visualize random such filters $F(\vtheta) = \bracks{w_k\cdot\dots\cdot w_1}\vlambda^{(k)}$, where $\evlambda^{(k)}_i = \evlambda_i^k$ for different numbers of layers $k$ in \Cref{fig:fourier_gcn_depth}. We observe that with increased depth, all components except the first are increasingly suppressed, as their corresponding eigenvalues $\evlambda_i^k$ approach zero relative to $\evlambda_1^k$.

As in the single iteration case, the scalar $\bracks{w_k\cdot\ldots\cdot w_1}\in\R$ applies a uniform scaling of each coefficient $\evlambda^{k}_i$. As such, the maximal coefficient $\evlambda_1$ increasingly dominates the other coefficients with an increasing number of iterations $k$, independent of the chosen parameters $w_1,\dots,w_k$.
We refer to the phenomenon of a single component being dominantly amplified as component dominance, which we will formally define in \Cref{def:component_dominance}.
These insights provide an initial intuition for the properties of the GCN limiting their performance and the theoretical analysis presented in the following section.

\label{sec:understanding:zero}
\begin{figure}
     \centering
     \begin{subfigure}[b]{0.49\textwidth}
         \centering
        \def\svgwidth{\textwidth}
         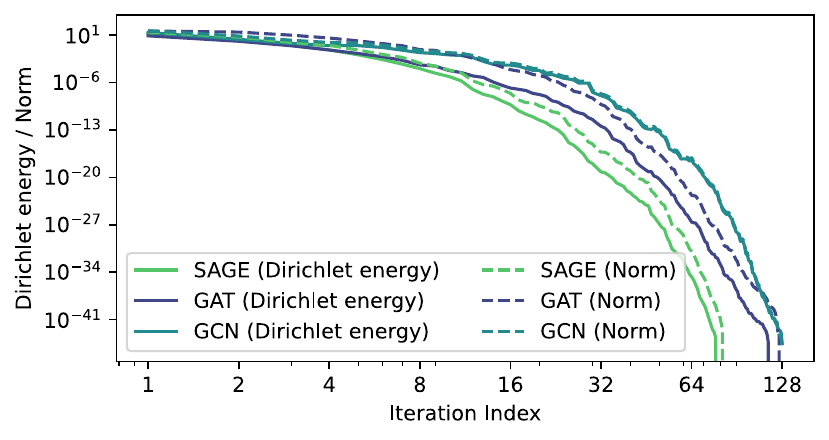
         \caption{Dirichlet energy converges to zero.}
     \end{subfigure}
     \hfill
     \begin{subfigure}[b]{0.49\textwidth}
         \centering
        \def\svgwidth{\textwidth}
         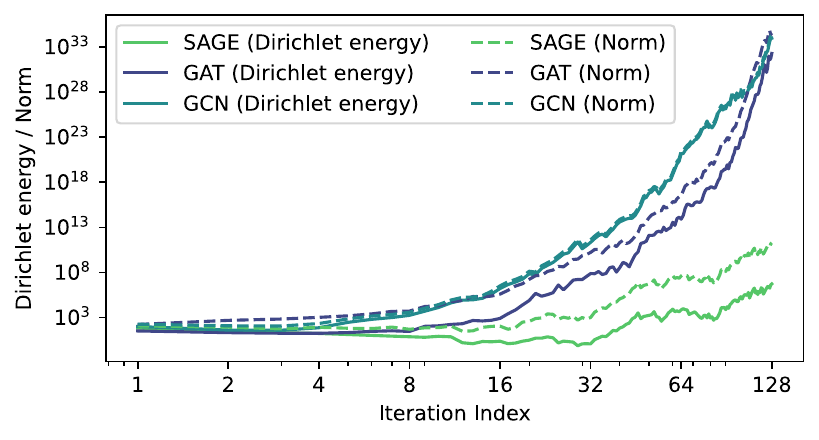
         \caption{Dirichlet energy does not converge.}
     \end{subfigure}
        \caption[Comparison of the Dirichlet energy and the norm of representations]{Comparison of the Dirichlet energy $E(\mX^{(k)})$ defined using the random walk Laplacian $\mL_\trw$ and the norm of the representations $\norm{\mX^{(k)}}_F^2$ on the Cora dataset~\parencite{mccallum2000automating}. The node representations $\mX^{(k)}$ are obtained for $128$ iterations of three MPNNs (graph convolutional network (GCN)~\parencite{kipf2017semi}, graph attention network (GAT)~\parencite{velickovic2017graph}, the SAGE convolution~\parencite{hamilton2017inductive}). Following \textcite{rusch2023survey}, random parameters are used for (a). We additionally present the same figure when all parameters are scaled by a factor of $2$ in (b). }
        \label{fig:constant}
\end{figure}

\section{Over-Smoothing and Vanishing Norms}
Several conflicting definitions and views for over-smoothing currently coexist, as detailed in \Cref{sec:understanding_sota}. Several studies prove that node representations converge to a multiple of the degree-proportional dominant eigenvector $\mD^{1/2}\mathbf{1}$ of $\mA_\tsym$ when repeatedly applying GCN iterations~\parencite{oono2020graph,cai2020anote,zhou2021dirichlet,giovanni2023understanding}. This can be shown using the Dirichlet energy $E(\mX) = \trt\bracks{\mX^\top\mL_\tsym\mX}$, as the degree-proportional state corresponds to the null space of $\mL_\tsym$, i.e., $\mL_\tsym\mD^{1/2}\mathbf{1} = \mathbf{0}$.
Others have observed an exponential convergence to a constant state~\parencite{rusch2022graph,rusch2023gradient,rusch2023survey,wu2023demystifying}, for example, using the Dirichlet energy $E_\mL = \trt\bracks{\mX^\top\mL\mX}$ defined using the unnormalized graph Laplacian $\mL$, as the constant state is in the null space of $\mL$, i.e., $\mL\mathbf{1} = \mathbf{0}$.

Mathematically, the matrix $\mX$ lies in the intersection of the null spaces of both $\mL_\tsym$ and $\mL$ if and only if $\mX = \mathbf{0}$, i.e., all entries of $\mX$ are zero. 
Thus, the norm of the representations needs to vanish in order for both properties to be satisfied.
To verify this claim, we theoretically study the effect of applying a GCN iteration on the norm of the representations. We find that the norm of $\mX$ is bounded by the largest singular value of $\mW$, leading to the same bound as shown for the Dirichlet energy (\parencite{oono2020graph,cai2020anote}). We formalize this in \Cref{prop:vanish}:

\begin{proposition}[Vanishing norm of node representations]
\label{prop:vanish}
    Let $\mA_\tsym\in\R^{n\times n}$ be a symmetrically normalized adjacency matrix and $\mX\in\R^{n\times d}$, $\mW\in\R^{d\times c}$ arbitrary matrices for some $n,d,c\in\mathbb{N}$. Further, let $\phi$ be a component-wise activation function with $|\phi(a)| \leq |a|$ for all $a\in\R$. Then,  
    \begin{equation}
        \norm{\phi\bracks{\mA_\tsym\mX\mW}}_F \leq \sigma_1 \norm{\mX}_F\, ,
    \end{equation}
    where $\sigma_1$ is the largest singular value of $\mW$.
\end{proposition}
\begin{proof}
    By definition of the Frobenius norm, we have
    \begin{equation}
        \norm{\phi\bracks{\mA_\tsym\mX\mW}}_F 
        = \sqrt{\sum_{i=1}^n\sum_{j=1}^{c}\phi\bracks{\mA_\tsym\mX\mW}_{i,j}^2}\, .
    \end{equation}
    Using the property $|\phi(a)|\leq |a|$ for all $a\in\R$ of the activation function $\phi$, we obtain
    \begin{equation}
    \begin{split}
    \norm{\phi\bracks{\mA_\tsym\mX\mW}}_F &\leq \sqrt{\sum_{i=1}^n\sum_{j=1}^{c}\bracks{\mA_\tsym\mX\mW}_{i,j}^2} \\
        &= \norm{\mA_\tsym\mX\mW}_F,
    \end{split}
    \end{equation}
    which is the Frobenius norm, ignoring the non-linear activation function.
    We now apply the sub-multiplicative property of the Frobenius norm with the spectral norm, yielding
    \begin{equation}
        \norm{\mA_\tsym\mX\mW}_F \leq \norm{\mA_\tsym}_2\cdot\norm{\mX}_F\cdot\norm{\mW}_2\, .
    \end{equation}
    We then use the fact that the spectral norm of a matrix is equal to its largest singular value, i.e., $\norm{\mA_\tsym}_2 = \lambda_1 = 1$ by properties of $\mA_\tsym$ and $\norm{\mW}_2 = \sigma_1$. Combining these equalities, we obtain
    \begin{equation}
        \norm{\phi\bracks{\mA_\tsym\mX\mW}}_F \leq \sigma_1 \norm{\mX}_F\, ,
    \end{equation}
    as claimed.
\end{proof}


\Cref{prop:vanish} corresponds to applying a single GCN iteration with 
\begin{equation}
\label{eq:3:constant_gcn}
    \mX^{(k)} = \phi\bracks{\mA_\tsym\mX^{(k-1)}\mW^{(k)}}\, .
\end{equation}
When repeatedly applying such GCN iterations, all entries of the state converge to zero when $\prod_{k=1}^\infty\sigma_1^{\mW^{(k)}} = 0$ with $\sigma_1^{\mW^{(k)}}$ being the largest singular value of $\mW^{(k)}$, i.e., we have $\lim_{k\to\infty}\mX^{(k)} = 0$. 
As the zero matrix is a special case of both the constant vector $\mathbf{1}$ and the degree-proportional vector $\mD^{1/2}\mathbf{1}$, the Dirichlet energy $E(\mX)$ defined using $\mL_\tsym$ and the Dirichlet energy $E_\mL(\mX)$ defined using $\mL$ both approach zero as $k\to\infty$. This explains the previous seemingly contradictory findings. 

We visualize both the Frobenius norm $\norm{\mX^{(k)}}_F$ and the Dirichlet energy $E_\mL\bracks{\mX^{(k)}}$ for node representations according to \Cref{eq:3:constant_gcn} in \Cref{fig:constant}. We observe a close similarity between both metrics. This is not only the case when the norm converges to zero, but also when the norm does not converge to zero. 
Thus, the norm of the node representations overshadows potential insights of the Dirichlet energy. As such, observing the Dirichlet energy to converge to zero seems to be a consequence of the norm converging to zero, and not of the representations converging to a degree-dependent or constant state. This calls into question the potential insights gained from the Dirichlet energy in this form. 

This vanishing norm is a critical concern when applying many message-passing iterations, which can lead to vanishing gradients~\parencite{bengio1994learning,hochreiter1997long,pascanu2013on} and hinder optimization.  
As previously shown, the over-smoothing phenomenon arises independently of the scale of node representations~\parencite{liu2020towards,chen2020measuring,giovanni2023understanding,maskey2023fractional}. Instead, we argue to separate the vanishing norm from over-smoothing and study both phenomena independently. This allows a more targeted analysis of the reasons and the effects of each property. Separating them also allows for better communication.
The vanishing norm always implies over-smoothing under this definition, since $E(\mathbf{0}) = 0$. This would imply over-smoothing even for trivial cases, such as $\mX^{(k)} = a\mX^{(k-1)}$ for $|a| < 1$. This suffers from the vanishing norm problem and over-smoothing in this definition, but we argue that it distracts from the underlying problem of over-smoothing.
Thus, when the representations converge to zero, it is a misleading special case. Several theoretical studies base their findings about over-smoothing on assumptions about the singular values of the feature transformations~\parencite{oono2020graph,cai2020anote,zhou2021dirichlet,wu2023demystifying}, which may oversimplify certain cases, as these primarily concern the magnitude of the features, rather than the smoothing property.

Other existing metrics, such as SMV~\parencite{liu2020towards} and MAD~\parencite{chen2020measuring} (see \Cref{sec:understanding:state:oversmoothing}), similarly argue to quantify the similarity of node representations independently of the feature magnitude. These incorporate feature normalization to consider over-smoothing independently of the norm. However, these metrics are not used for theoretical studies.
More recent theoretical studies on over-smoothing consider the Dirichlet energy $E\bracks{\frac{\mX^{(k)}}{\norm{\mX^{(k)}}_F}}$ of the normalized node representations, which can be considered as a study of over-smoothing while ignoring the effects of the norm~\parencite{giovanni2023understanding,maskey2023fractional}. We further argue that the Dirichlet energy alone might not provide sufficient insight into the underlying phenomenon. 

This observation aligns with our initial study on convolutional filters in the graph Fourier domain in \Cref{sec:3:fourier:mp}. We observed that all components are uniformly scaled, which may overshadow the amplification of the smooth component.
\section{A Theoretical Study on the Effect of Message-Passing on Representations}
\label{sec:understanding:rank}
Based on Section~\ref{sec:understanding:fourier}, we have developed some intuition about the effects of applying iterations of the GCN and provided some initial insights into the phenomena of shared component amplification (SCA) and component dominance (CD).
We have also observed that the norm of node representations has led to some conflicting findings and definitions regarding over-smoothing.
Here, we conduct an in-depth study on the theoretical properties of the GCN, which we will later generalize.
We consider linearized message-passing functions of the form
\begin{equation}
\label{eq:3:theory_gcn}
    \mX^{(k)} = \tilde{\mA}\mX^{(k-1)}\mW^{(k)}
\end{equation}
where $\tilde{\mA}\in\R^{n\times n}$, $\mX^{(k)}\in\R^{n\times d}$, and $\mW^{(k)}\in\R^{d\times d}$ for all $k\in\mathbb{N}$ and some $d\in\mathbb{N}$.
This corresponds to the equivalent node-wise update function
\begin{equation}
    \mX_{i,:}^{(k)} = \sum_{j\in \gN_i} \emA_{i,j}\mX^{(k-1)}_{j,:}\mW^{(k)}\, .
\end{equation}

\begin{figure*}
\definecolor{vibrantpink}{HTML}{fb9a99} 
\centering
\begin{tikzpicture}[x=\textwidth/4, y=\textwidth/4]
\definecolor{vibrantblue}{HTML}{a6cee3} 
\definecolor{vibrantgreen}{HTML}{b2df8a} 
\definecolor{vibrantpink}{HTML}{fb9a99} 
\definecolor{vibrantorange}{HTML}{fdbf6f} 
\definecolor{vibrantpurple}{HTML}{cab2d6} 
\definecolor{vibrantsand}{HTML}{ffff99} 
\usetikzlibrary{shapes}
\pgfdeclarelayer{background}
\pgfsetlayers{background,main}
\tikzset{
dot/.style = {circle, minimum size=#1,
              inner sep=0pt, outer sep=0pt, draw=black},
dot/.default =13pt  
}
    \tikzstyle{box} = [draw, draw=vibrantblue, thick,rounded corners,line width=1.8pt]

    \def\yconv{1.6}
    \def\ysiso{-0.2}
    \def\yover{0.7}
    \def\ymethods{-1.1}
    \def\xshift{0.5}
    \def\xout{1.4}
    \def\xpoly{0.0}

    \node[rectangle,draw=vibrantsand,minimum height=2cm,minimum width=3.8cm, rounded corners, line width=1mm] at (\xpoly-\xout,\ymethods) (chapter3) {};
    
    \node[scale=0.7,align=center] at (\xpoly-\xout, \ymethods+0.4) {\textbf{State-of-the-Art~\parencite{giovanni2023understanding}}};   
    \node[scale=0.7,align=center] at (\xpoly-\xout, \ymethods) {$\tilde{\mA} := \mA_\tsym$ \\ $\mW^{(1)} = \dots = \mW^{(k)}$ symmetric};   

    \node[rectangle,draw=vibrantblue,minimum height=2cm,minimum width=3.8cm, rounded corners, line width=1mm] at (\xpoly,\ymethods) (chapter5) {};
    \node[scale=0.7,align=center] at (\xpoly, \ymethods+0.4) {\textbf{\Cref{pr:asym_wany}}};   
    \node[scale=0.7,align=center] at (\xpoly, \ymethods) {$\tilde{\mA} := \mA_\tsym$ \\ $\mW^{(1)} \neq \dots \neq \mW^{(k)}$ arbitrary};   

    \node[rectangle,draw=vibrantblue,minimum height=2cm,minimum width=3.8cm, rounded corners, line width=1mm] at (\xpoly+\xout,\ymethods) (chapter6) {};

    \node[scale=0.7,align=center] at (\xpoly+\xout, \ymethods+0.4) {\textbf{\Cref{pr:anya_anyw}}};   
    \node[scale=0.7,align=center] at (\xpoly+\xout, \ymethods) {$\tilde{\mA}$ arbitrary \\ $\mW^{(1)} \neq \dots \neq \mW^{(k)}$ arbitrary};   

    \draw[->,line width=1.2pt] (chapter3) to node[sloped,above,color=black,scale=0.7,rotate=90] {} (chapter5);
    \draw[->,line width=1.2pt] (chapter5) to node[sloped,above,color=black,scale=0.7,rotate=90] {} (chapter6);    
    
\end{tikzpicture}
\caption[Connection between our theoretical study and the state-of-the-art.]{Limit behavior of $\mX^{(k)}$, where $\mX^{(k)} = \tilde{\mA}\mX^{(k-1)}\mW^{(k)}$, under various choices for $\tilde{\mA}$ and $\mW^{(1)},\dots,\mW^{(k)}$ from existing state-of-the-art, and our proposed extensions.}
\label{fig:3:theory_contributions}
\end{figure*}
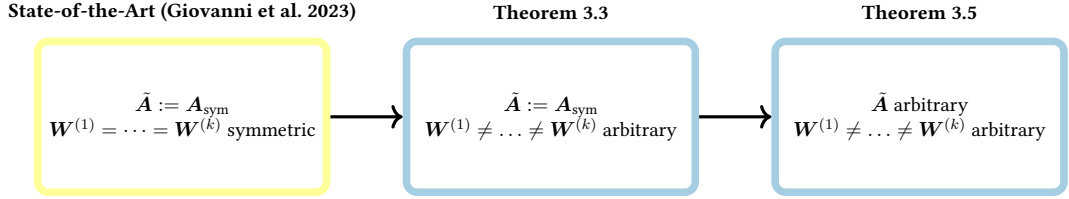

Theoretical insights are currently only available for the case when $\tilde{\mA}:=\mA_\tsym$ and the feature transformations $\mW^{(k)} = \mW$ are shared for every $k\in\mathbb{N}$ with the shared matrix $\mW$ also being symmetric~\parencite{giovanni2023understanding}. Their results only apply to the limit case $\mX^{(l)}$ with infinitely many iterations $l$.
We study the effect of such a message-passing step for any, not necessarily identical, matrices $\mW^{(1)},\dots,\mW^{(l)}$ of suitable shape, both for a single iteration and the repeated application. We initially consider the case $\tilde{\mA}:=\mA_\tsym$ and then further generalize our theoretical study to allow for any aggregation matrix $\tilde{\mA}$ instead of the symmetrically normalized adjacency matrix $\mA_\tsym$. 
We summarize this connection between our findings and the currently known case in \Cref{fig:3:theory_contributions}. We summarize our findings and provide comprehensive definitions in Section~\ref{sec:3:summary}. This analysis mostly follows one of our previous publications~\parencite{roth2023rank}.

\subsection{The Effect of GCN Iterations}
\label{sec:explaining}
We now assume the case $\tilde{\mA}:= \mA_\tsym$ in \Cref{eq:3:theory_gcn} for an undirected and non-bipartite graph, as used for the GCN~\parencite{kipf2017semi}.
Instead of focusing on the Dirichlet energy, we study the effect of the GCN on the node representations in general. To ease our theoretical study, we utilize the vectorized form of a single iteration of the GCN from \Cref{eq:3:theory_gcn}, resulting in
\begin{equation}
\label{eq:understanding:form}
    \tvec\bracks{\mA_\tsym\mX\mW} = \bracks{\mW^\top\otimes\mA_\tsym}\tvec(\mX) = \mT\tvec(\mX)
\end{equation}
with $\mT = \bracks{\mW^\top\otimes\mA_\tsym}\in\mathbb{R}^{nd\times nd}$ applied before any non-linear activation function. The Kronecker product $\otimes$ is a standard operation that is frequently used for theoretical studies of GNNs~\parencite{gu2020implicit,roth2022transforming,giovanni2023understanding,maskey2023fractional}. We will introduce its properties as required. The first important property of the Kronecker product used above is $\tvec(\mA\mB\mC) = \bracks{\mC^\top\otimes\mA}\tvec(\mB)$ for any matrices $\mA,\mB,\mC$ of suitable shape~\parencite{horn1991topics}.
As we allow for arbitrary matrices $\mW$, we will drop the transpose of $\mW$ in the notation for conciseness.
Next, we represent the node representations in the graph Fourier basis (see \Cref{sec:fundamentals:spectral}) $\mU$ using the property $\mU^\top\mU=\mI_n$ as follows:
\begin{equation}
\begin{split}
\label{eq:x_in_gf}
    \tvec(\mX) 
    &= \tvec(\mU\mB) \\
    &= \bracks{\mI_d\otimes \mU}\tvec(\mB)
\end{split}
\end{equation}
where $\mB = \mU^\top\mX\in\R^{n\times d}$ contains the graph Fourier transformed node representations or components. Each entry $\emB_{i,j}$ represents the coefficient or component of the $i$-th graph Fourier basis vector of feature channel $j$.
We express the graph Fourier decomposition of $\mX$ as
\begin{equation}
\begin{split}
    \tvec(\mX)
    &= \sum_{i=1}^n\bracks{\mI_d\otimes \mU_{:,i}}\tvec\bracks{\mB_{i,:}} \\
    &= \sum_{i=1}^n\mS_{(i)}\tvec\bracks{\mB_{i,:}}\, ,
\end{split}
\end{equation}
where $\mS_{(i)}=\bracks{\mI_d\otimes\mU_{:,i}}\in\R^{nd\times d}$ and $\tvec\bracks{\mB_{i,:}}\in\R^d$ is the vector containing the coefficients of the $i$-th graph Fourier basis vector across all feature channels. Combining this reformulation of the state $\mX$ with the vectorized GCN iteration (\Cref{eq:understanding:form}), we consider the effect of a GCN iteration
\begin{equation}
    \tvec\bracks{\mA_\tsym\mX\mW} = \sum_{i=1}^n\mT\mS_{(i)}\tvec\bracks{\mB_{i,:}}
\end{equation}
 on the different graph Fourier components. For each graph Fourier basis vector, we further construct disjoint subspaces 
 \begin{equation}
     \gQ_i = \tspan\left\{\,\mS_{(i)}\vv \mid \vv\in\R^d\,\right\}\, .
 \end{equation}
Every element $\tvec(\mZ)\in\gQ_i$ is the vectorized form of a rank-one matrix $\mZ = \mU_{:,i}\vc^\top\in\R^{n\times d}$ for some $\vc\in\R^d$. $\mZ$ only contains non-zero components corresponding to graph Fourier basis vector $\mU_{:,i}$. Thus, each column of $\mZ$ is a scalar multiple of the corresponding graph Fourier basis vector $\mU_{:,i}$. We find the first key property of GCN iterations to exploit this decomposition. Each subspace corresponding to a graph Fourier basis vector is invariant under applications of any GCN iteration $\mT$:

\begin{lemma}[Graph Fourier Components are invariant to GCN iterations]
\label{pr:symm_inv}
    Let $\mA_\tsym\in\R^{n\times n}$ be a symmetrically normalized adjacency matrix with an orthogonal matrix of eigenvectors $\mU$, and let $\mW\in\R^{d\times d}$ be any matrix. Consider the subspace $$\gQ_i = \tspan\left\{\,\bracks{\mI_d\otimes \mU_{:,i}}\vv \mid \vv\in\R^{d}\,\right\}\, .$$
    Then, $\gQ_i$ is invariant under $\mT = \mW\otimes\mA_\tsym$, i.e.,
    \begin{equation}
        \forall\vz\in\gQ_i\colon \mT\vz\in\gQ_i\, .
    \end{equation}
\end{lemma}

\begin{proof}
    Any $\vz\in\gQ_i$ can be expressed as a linear combination $\vz = \bracks{\mI_d\otimes\mU_{:,i}}\vb$ of basis vectors $\bracks{\mI_d\otimes\mU_{:,i}}$ of $\gQ_i$, where $\vb\in\R^{nd}$. By definition of $\mT = \mW\otimes\mA_\tsym$, we have
    \begin{equation}
        \mT\vz = \bracks{\mW\otimes\mA_\tsym}\bracks{\mI_d\otimes\mU_{:,i}}\vb\, .
    \end{equation}
    Following the mixed-product property $(\mA\otimes\mB)\cdot(\mC\otimes\mD) = (\mA\cdot\mC)\otimes(\mB\cdot\mD)$ of the Kronecker product and the matrix product for matrices of suitable shape and $\mA_\tsym\mU_{:,i} = \lambda_i\mU_{:,i}$, we have
    \begin{equation}
        \bracks{\mW\otimes\mA_\tsym}\bracks{\mI_d\otimes\mU_{:,i}}\vb = \bracks{\mI_d\otimes\mU_{:,i}}\bracks{\mW\otimes \lambda_i\mI_n}\vb\, .
    \end{equation}
    By defining $\vb^\prime = \bracks{\mW\otimes \lambda_i\mI_n}\vb$, this further simplifies to
    \begin{equation}
        \mT\vz = \bracks{\mI_d\otimes\mU_{:,i}}\vb^\prime\, .
    \end{equation}
    As $\mT\vz$ is a linear combination of the basis vectors from subspace $\gQ_i$, this concludes the proof.
\end{proof}

This proof exploits two central properties of the constructed subspaces. First, as the graph Fourier basis vectors are also eigenvectors of $\mA_\tsym$, applying $\mA_\tsym$ reduces to scaling each graph Fourier component by the corresponding eigenvalue. Second, applying the feature transformation $\mW$ cannot transform the given graph Fourier components into other graph Fourier components.
This discovery is central to our subsequent investigation, as it enables us to study the effect of a GCN iteration separately for each subspace and compare their effects.

\subsubsection{A Single GCN Iteration}
Given these invariant subspaces $\gQ_i$, we now study the effect of GCN iterations on each graph Fourier component independently.
Precisely, we study the effect of a vectorized GCN iteration $\mW\otimes\mA_\tsym$ on each set of graph Fourier basis vectors $\mI_d\otimes\mU_{:,i}$ of subspace $\gQ_i$, i.e., we consider $\bracks{\mW\otimes\mA_\tsym}\bracks{\mI_d\otimes\mU_{:,i}}$ for all $i\in[n]$. 
We find $\mA_\tsym$ to perform a scaling of the basis vectors $\mI_d\otimes\mU_{:,i}$ of each subspace $\gQ_i$ by the corresponding eigenvalue $\lambda_i$ of $\mA_\tsym$, i.e., 
\begin{equation}
    \bracks{\mI_d\otimes\mA_\tsym}\bracks{\mI_d\otimes\mU_{:,i}} = \lambda_i\bracks{\mI_d\otimes\mU_{:,i}}\, .
\end{equation}
The feature transformation $\mW$ performs an equivalent transformation to all sets of basis vectors $\mI_d\otimes\mU_{:,i}$. While this can change the norm $\norm{\bracks{\mW\otimes\mI_n}\bracks{\mI_d\otimes\mU_{:,i}}}_F$ arbitrarily, the norm of all subspaces equally changes, i.e., we have
\begin{equation}
    \frac{\norm{\bracks{\mW\otimes\mI_n}\bracks{\mI_d\otimes\mU_{:,i}}}_F}{\norm{\bracks{\mW\otimes\mI_n}\bracks{\mI_d\otimes\mU_{:,j}}}_F} = 1
\end{equation}
for all $i,j\in[n]$. Thus, the feature transformation cannot amplify components for one graph Fourier basis vector more than for another. Applying both $\mA_\tsym$ and $\mW$ then scales the basis vectors of each subspace $\gQ_i$ only relative to the corresponding eigenvalue $\lambda_i$. We formalize this finding in the following theorem:

\begin{theorem}[Shared Component Amplification for the GCN]
\label{pr:symm_iter}
Let $\mT = \mW\otimes\mA_\tsym\in\R^{nd\times nd}$, where $\mW\in\R^{d\times d}$ is any matrix and $\mA_\tsym\in\R^{n\times n}$ is the symmetrically normalized adjacency matrix with orthonormal eigenvectors $\mU\in\R^{n\times n}$. Let $\mS_{(i)} = \mI_d\otimes\mU_{:,i}$ with corresponding eigenvalue $\lambda_i$ of $\mA_\tsym$ for all $i\in[n]$. Then, for all $i,j\in [n]$,
    \begin{equation}
    \frac{\norm{\mT\mS_{(i)}}_F}{\norm{\mT\mS_{(j)}}_F} = \frac{|\lambda_i|}{|\lambda_j|}\, .    
    \end{equation}
\end{theorem}
\begin{proof}
    We start by expanding all matrices by their definitions:
    \begin{equation}
    \frac{\norm{\mT\mS_{(i)}}_F}{\norm{\mT\mS_{(j)}}_F} = \frac{\norm{\bracks{\mW\otimes\mA_\tsym}\bracks{\mI_d\otimes\mU_{:,i}}}_F}{\norm{\bracks{\mW\otimes\mA_\tsym}\bracks{\mI_d\otimes\mU_{:,j}}}_F}\, .    
    \end{equation}
    Using the mixed-product property of the Kronecker product and the matrix product, we have
    \begin{equation}
        \frac{\norm{\bracks{\mW\otimes\mA_\tsym}\bracks{\mI_d\otimes\mU_{:,i}}}_F}{\norm{\bracks{\mW\otimes\mA_\tsym}\bracks{\mI_d\otimes\mU_{:,j}}}_F} = \frac{\norm{\bracks{\mW\otimes\mA_\tsym\mU_{:,i}}}_F}{\norm{\bracks{\mW\otimes\mA_\tsym\mU_{:,j}}}_F}\, .
    \end{equation}
    As $\mU_{:,i}$ is an eigenvector of $\mA_\tsym$ with eigenvalue $\lambda_i$, we have
    \begin{equation}
        \frac{\norm{\bracks{\mW\otimes\mA_\tsym\mU_{:,i}}}_F}{\norm{\bracks{\mW\otimes\mA_\tsym\mU_{:,j}}}_F} = \frac{\norm{\bracks{\mW\otimes\lambda_i\mU_{:,i}}}_F}{\norm{\bracks{\mW\otimes\lambda_j\mU_{:,j}}}_F}\, .
    \end{equation}
    By sub-multiplicativity of the norm of the Kronecker product, this simplifies to
    \begin{equation}
        \frac{\norm{\mW\otimes\lambda_i\mU_{:,i}}_F}{\norm{\mW\otimes\lambda_j\mU_{:,j}}_F} = \frac{\norm{\mW}_F\cdot\norm{\lambda_i\mU_{:,i}}_F}{\norm{\mW}_F\cdot\norm{\lambda_j\mU_{:,j}}_F}\, .
    \end{equation}
    The norm $\norm{\mW}_F$ cancels, and with the absolute homogeneity property of the norm, we have
    \begin{equation}
        \frac{\norm{\mW}_F\cdot\norm{\lambda_i\mU_{:,i}}_F}{\norm{\mW}_F\cdot\norm{\lambda_j\mU_{:,j}}_F} = \frac{|\lambda_i|\cdot\norm{\mU_{:,i}}_F}{|\lambda_j|\cdot\norm{\mU_{:,j}}_F}\, .
    \end{equation}
    As each $\mU$ is chosen to be an orthonormal matrix, the Frobenius norm of $\mU_{:,i}$ is one, and we have
    \begin{equation}
        \frac{|\lambda_i|\cdot\norm{\mU_{:,i}}_F}{|\lambda_j|\cdot\norm{\mU_{:,j}}_F} = \frac{|\lambda_i|}{|\lambda_j|}\, .
    \end{equation}
    This concludes the proof.
\end{proof}

This property of the GCN can be interpreted in several ways.
As the eigenvalues are fixed and given by $\mA_\tsym$, the subspaces with larger eigenvalues will be amplified more strongly than those corresponding to eigenvalues with a smaller magnitude for any $\mW$. 
As the largest eigenvalue $\lambda_1 = 1$, with corresponding smooth eigenvector $\mU_{:,1} = \mD^{1/2}\mathbf{1}$ that is degree-proportional, each iteration relatively amplifies this smooth subspace by this fixed ratio compared to all other subspaces.

In addition to this relative amplification being fixed, it is also shared across all feature channels. As every $\vv\in\gQ_i$ is the vectorized form of a rank-one matrix, the same graph Fourier basis vector is relatively amplified by the same ratio of eigenvalue for every feature channel. Thus, a single GCN iteration relatively amplifies the smooth components across all feature channels by the ratio $\frac{\lambda_1}{\lambda_i}$ compared to all other components.
We will define a more general form of this phenomenon in \Cref{def:sca}, and refer to this phenomenon as shared component amplification (SCA). This finding also aligns with our spectral analysis of the filter used within the GCN in \Cref{sec:3:fourier:mp}.

As the ratio depends on the aggregation function $\mA_\tsym$ but not on the feature transformation $\mW$, we cannot construct or learn feature transformations that prevent the amplification of smooth signals for any of the feature channels. 

Compared to previous studies that identified bounds on the Dirichlet energy based on singular values of $\mW$~\parencite{oono2020graph,cai2020anote}, this equality provides deeper insights.
This relative amplification of the message-passing step is independent of any subsequent operations, such as non-linear activation functions.

\subsubsection{Repeated GCN Iterations}

We have identified that each GCN iteration amplifies the graph Fourier components relatively with a fixed ratio given by the corresponding eigenvalues of $\mA_\tsym$. We now consider the case when repeatedly applying GCN iterations
\begin{equation}
    \tvec\bracks{\mX^{(k)}} = \tvec\bracks{\mA_\tsym\mX^{(k-1)}\mW^{(k)}} = \bracks{\bracks{\mW^{(k)}}^\top\otimes\mA_\tsym}\tvec\bracks{\mX^{(k-1)}}
\end{equation}
with time-inhomogeneous feature transformations $\mW^{(k)}\in\R^{d\times d}$ for every $k\in\mathbb{N}$. As before, we will omit the transpose for notational simplicity, as we allow any matrices $\mW^{(k)}$. Although the following analysis only covers the linear case without activation functions, this linear case remains insufficiently explored. The extension to the non-linear case remains open for future work. The relative amplification of graph Fourier induced subspaces from \Cref{pr:symm_iter} can be directly extended to this iterated case:

\begin{proposition}[Component amplification for multiple GCN iterations]
\label{pr:symm_iterated}
    Let $\mT^{(k)} = \mW^{(k)}\otimes\mA_\tsym\in\R^{nd\times nd}$, where $\mW^{(k)}\in\R^{d\times d}$ is any matrix and $\mA_\tsym\in\R^{n\times n}$ is the symmetrically normalized adjacency matrix with orthonormal eigenvectors $\mU\in\R^{n\times n}$. Let $\mS_{(i)} = \mI_d\otimes\mU_{:,i}$ with corresponding eigenvalue $\lambda_i$ of $\mA_\tsym$ for all $i\in[n]$. Then, for all $i,j\in [n]$,
    \begin{equation}
        \frac{\norm{\mT^{(l)}\dots\mT^{(1)}\mS_{(i)}}_F}{\norm{\mT^{(l)}\dots\mT^{(1)}\mS_{(j)}}_F} = \frac{|\lambda_i^l|}{|\lambda_j^l|}
    \end{equation}
\end{proposition}

\begin{proof}
    This proof is mostly analogous to the proof of \Cref{pr:symm_iter}. We start by replacing each $\mT^{(k)}$ and $\mS_{(i)}$ by their definition:
    \begin{equation}
        \frac{\norm{\mT^{(l)}\dots\mT^{(1)}\mS_{(i)}}_F}{\norm{\mT^{(l)}\dots\mT^{(1)}\mS_{(j)}}_F} = \frac{\norm{\bracks{\mW^{(l)}\otimes\mA_\tsym}\dots\bracks{\mW^{(1)}\otimes\mA_\tsym}\bracks{\mI_d\otimes\mU_{:,i}}}_F}{\norm{\bracks{\mW^{(l)}\otimes\mA_\tsym}\dots\bracks{\mW^{(1)}\otimes\mA_\tsym}\bracks{\mI_d\otimes\mU_{:,j}}}_F}\, .
    \end{equation}
    By the mixed-product property of the Kronecker product and the matrix product, this simplifies to
    \begin{equation}
        \frac{\norm{\bracks{\mW^{(l)}\otimes\mA_\tsym}\dots\bracks{\mW^{(1)}\otimes\mA_\tsym}\bracks{\mI_d\otimes\mU_{:,i}}}_F}{\norm{\bracks{\mW^{(l)}\otimes\mA_\tsym}\dots\bracks{\mW^{(1)}\otimes\mA_\tsym}\bracks{\mI_d\otimes\mU_{:,j}}}_F} = \frac{\norm{\bracks{\mW^{(l)}\dots\mW^{(1)}}\otimes\bracks{\mA_\tsym^l\mU_{:,i}}}_F}{\norm{\bracks{\mW^{(l)}\dots\mW^{(1)}}\otimes\bracks{\mA_\tsym^l\mU_{:,j}}}_F}\, .
    \end{equation}
    As in the proof of \Cref{pr:symm_iter}, the sub-multiplicativity of the norm of the Kronecker product and $\mU_{:,i}$ being an eigenvector of $\mA_\tsym$ allow the simplification
    \begin{equation}
        \frac{\norm{\bracks{\mW^{(l)}\dots\mW^{(1)}}\otimes\bracks{\mA_\tsym^l\mU_{:,i}}}_F}{\norm{\bracks{\mW^{(l)}\dots\mW^{(1)}}\otimes\bracks{\mA_\tsym^l\mU_{:,j}}}_F} = \frac{\norm{\mW^{(l)}\dots\mW^{(1)}}_F\cdot\norm{\lambda_i^l\mU_{:,i}}_F}{\norm{\mW^{(l)}\dots\mW^{(1)}}_F\cdot\norm{\lambda_j^l\mU_{:,j}}_F}\, .
    \end{equation}
    The norm of the feature transformations cancels. By the absolute homogeneity property $\norm{\lambda_i\mU_{:,i}}_F=|\lambda_i|\cdot\norm{\mU_{:,i}}_F$ of the Frobenius norm, and $\norm{\mU_{:,i}}=\norm{\mU_{:,j}}$ due to orthonormality of $\mU$. We have  
    \begin{equation}
        \frac{\norm{\mW^{(l)}\dots\mW^{(1)}}_F\cdot\norm{\bracks{\lambda_i}^l\mU_{:,i}}_F}{\norm{\mW^{(l)}\dots\mW^{(1)}}_F\cdot\norm{\bracks{\lambda_j}^l\mU_{:,j}}_F} = \frac{|\bracks{\lambda_i}^l|}{|\bracks{\lambda_j}^l|}\, ,
    \end{equation}
    concluding the proof.
\end{proof}

This finding shows that repeatedly applying GCN iterations relatively amplifies the smooth component in the data by an exponential ratio in the number of iterations $l$. This holds for any feature transformations $\mW^{(k)}$.
As a consequence of this property, the subspace $\mS_{(1)}$ corresponding to the largest absolute eigenvalue $\lambda_1$ and smooth eigenvector $\mU_{:,1} = \mD^{1/2}\mathbf{1}$ of $\mA_\tsym$ dominates all other subspaces in the limit case:

\begin{corollary}[Component Dominance of $\mS_{(1)}$]
\label{pr:basis_limit}
    Let $\mT^{(k)} = \mW^{(k)}\otimes\mA_\tsym\in\R^{nd\times nd}$, where $\mW^{(k)}\in\R^{d\times d}$ is any matrix and $\mA_\tsym\in\R^{n\times n}$ is the symmetrically normalized adjacency matrix with orthonormal eigenvectors $\mU\in\R^{n\times n}$. Let $\mS_{(i)} = \mI_d\otimes\mU_{:,i}$ with corresponding eigenvalue $\lambda_i$ of $\mA_\tsym$ for all $i\in[n]$. Then, for all $i\in [n]$,
    \begin{equation}
        \lim_{l\to\infty} \frac{\norm{\mT^{(l)}\dots\mT^{(1)}\mS_{(i)}}_F}{\max_p \norm{\mT^{(l)}\dots\mT^{(1)}\mS_{(p)}}_F} = \begin{cases}0, & \text{if $i \neq 1$} \\ 1, & \text{otherwise}\, ,\end{cases}
    \end{equation}
    and the rate of convergence is given by $\frac{\lambda_i}{\lambda_1}$ for $i\neq 1$.
\end{corollary}

\begin{proof}
    Based on \Cref{pr:symm_iterated}, we have
    \begin{equation}
        \lim_{l\to\infty} \frac{\norm{\mT^{(l)}\dots\mT^{(1)}\mS_{(i)}}_F}{\max_p \norm{\mT^{(l)}\dots\mT^{(1)}\mS_{(p)}}_F} = \lim_{l\to\infty} \frac{|\lambda_i^l|}{\max_p |\lambda_p^l|}\, .
    \end{equation}
    As $\max_p \lambda_p = \lambda_1$ for $\mA_\tsym$, we have $\max_p |\lambda_p^l| = |\lambda_1^l| = 1$. As $|\lambda_i| < |\lambda_1|$ for all $i\in [n]$, we have in the limit
    \begin{equation}
        \lim_{l\to\infty}\frac{|\lambda_i^l|}{\max_p |\lambda_p^l|} = \begin{cases}
    1, & \text{if $i=1$}\\ 0, & \text{otherwise}\, .
    \end{cases}
    \end{equation}
    The convergence rate is given by $|\lambda_i| / |\lambda_1|$ for $i\neq 1$.
\end{proof}

We can already explain the observed over-smoothing phenomenon with \Cref{pr:symm_iterated} and \Cref{pr:basis_limit}. In each GCN iteration, the basis vectors of each subspace corresponding to a graph Fourier basis vector are relatively scaled by their respective eigenvalue. This is irrespective of the employed feature transformation $\mW^{(k)}$. When applying multiple GCN iterations, this scaling happens exponentially, and the graph Fourier components corresponding to $\lambda_1$ increasingly dominate all other graph Fourier components. We will define a more general form of this phenomenon in \Cref{def:component_dominance} and refer to it as component dominance.

Up to this point, we only considered the basis vectors. Node representations are linear combinations of these basis vectors, i.e., 
\begin{equation}
    \tvec(\mX^{(0)}) = \sum_{i=1}^n\bracks{\mI_d\otimes \mU_{:,i}} \tvec\bracks{\mB_{i,:}}
\end{equation}
with $\mB = \mU^\top\mX^{(0)}$. For each subspace $\gQ_i$ corresponding to a graph Fourier basis vector $\mU_{:,i}$, we have coefficients $\mB_{i,:}$.
As the differences in the norm of the basis vectors exponentially increase with the number of iterations $l$, the coefficients $\mB_{i,:}$ need to be increasingly large compared to the coefficients $\mB_{1,:}$ of the maximally amplified subspace $\gQ_1$ to contribute non-negligibly to $\mX^{(l)}$. 
As the feature transformation acts equivalently on each subspace $\gQ_i$, the signals across subspaces need to be very different. The signal in $\gQ_1$ needs to be exponentially small with increased depth in order not to dominate the representations. At the same time, the coefficients in other subspaces must be sufficiently large. 
This means that $\mX^{(0)}$, or equivalently $\mB_{1,:}$, needs to lie in an increasingly small subspace that converges to a measure-zero set under the Lebesgue measure for $l\to\infty$.
The probability that $\mB_{1,:}$ drawn from any bounded set lies inside this increasingly small subspace converges to zero. This corresponds to convergence in probability.
Due to time-inhomogeneous feature transformations $\mW^{(i)}\neq\mW^{(j)}$ for every $i\neq j$, $\mX^{(l)}$ may not converge for $l\to\infty$ but converges to the space $\gQ_1$. This is known as convergence in probability~\parencite{dudley2022real}, which we denote by $\plim$.
We formalize this in the following theorem:

\begin{theorem}[Representations converge to $\gQ_1$ in probability]
\label{pr:asym_wany}
Let $\mX^{(k)} = \mA_\tsym\mX^{(k-1)}\mW^{(k)}$, where $\mA_\tsym\in\R^{n\times n}$ is the symmetrically normalized adjacency matrix and $\mW^{(k)}\in\R^{d\times d}$ for every $k\in\mathbb{N}$. Then, 
    \begin{equation}
        \plim_{l\to\infty} \norm{\frac{\mX^{(l)}}{\norm{\mX^{(l)}}_F} - \mM^{(l)}}_F = 0
    \end{equation}
    for $\mX^{(0)}\in[-m,m]^{n\times d}$ for some $m\in\R$ with respect to the Lebesgue measure and some $\tvec\bracks{\mM^{(l)}} \in\gQ_1$ that depends on $l$ for every $l\in\mathbb{N}$.
\end{theorem}

\begin{proof}
    By defining $\mO^{(l)} = \mW^{(1)}\dots\mW^{(l)}$, we have
    \begin{equation}
        \mX^{(l)} = \mA_\tsym^l\mX^{(0)}\mW^{(1)}\dots \mW^{(l)} = \mA_\tsym^l\mX^{(0)}\mO^{(l)}\, .
    \end{equation}
    Using properties of the Kronecker product, we can equivalently write this in vectorized form as $\bracks{\bracks{\mO^{(l)}}^\top\otimes\mA_\tsym^l}\tvec(\mX^{(0)})$. We omit the transpose for clarity. Using the singular value decomposition $\mO^{(l)} = \mV^{(l)}\mSigma^{(l)}\bracks{\mN^{(l)}}^\top$ and eigendecomposition $\mA_\tsym^l = \mU\mLambda^l\mU^\top$, we obtain
    \begin{equation}
        \tvec\bracks{\mA_\tsym\mX^{(0)}\mO^{(l)}} = \bracks{\mO^{(l)}\otimes \mU}\bracks{\mI_d\otimes\mLambda^l}\vb\, ,
    \end{equation}
    where $\vb = \bracks{\mI_d\otimes\mU}^\top\tvec\bracks{\mX^{(0)}}$.
    We now choose $\vm^{(l)} = \frac{\bracks{\mO^{(l)}\otimes \mU}\bracks{\mI_d\otimes\mLambda^l}\vb_1}{\norm{\mX^{(l)}}_F}$ where $\vb_1$ is equal to $\vb$ for entries corresponding to $\gQ_1$ and zero for all other entries. 
    We thus have
    \begin{equation}
        \norm{\frac{\tvec\bracks{\mX^{(l)}}}{\norm{\tvec\bracks{\mX^{(l)}}}_F} - \vm^{(l)}}_F = \norm{\frac{\bracks{\mO^{(l)}\otimes \mU}}{\norm{\mX^{(l)}}_F} \bracks{\mI_d\otimes\mLambda^l}\bracks{\vb - \vb_1}}_F\, .
    \end{equation}
    To bound the numerator and denominator, we divide both by the largest singular value $\sigma_1^{(l)}$ of $\mO^{(l)}$, resulting in
    \begin{equation}
        \norm{\frac{\tvec\bracks{\mX^{(l)}}}{\norm{\tvec\bracks{\mX^{(l)}}}_F} - \vm^{(l)}}_F = \norm{\frac{\bracks{\frac{\mO^{(l)}}{\sigma_1^{(l)}}\otimes \mU}}{\norm{\frac{\mX^{(l)}}{\sigma_1^{(l)}}}_F} \bracks{\mI_d\otimes\mLambda^l}\bracks{\vb - \vb_1}}_F\, .
    \end{equation}
    Using the sub-multiplicativity of the matrix norm and the absolute homogeneity, we upper this norm by the following product:
    \begin{equation}
    \norm{\frac{\tvec\bracks{\mX^{(l)}}}{\norm{\tvec\bracks{\mX^{(l)}}}_F} - \vm^{(l)}}_F \leq \frac{\norm{\bracks{\frac{\mO^{(l)}}{\sigma_1^{(l)}}\otimes \mU}}_2}{\norm{\frac{\mX^{(l)}}{\sigma_1^{(l)}}}_F} \norm{\bracks{\mI_d\otimes\mLambda^l}\bracks{\vb - \vb_1}}_F
    \end{equation}
    We now proceed by bounding each of the three norms separately. First, by the definition of the Frobenius norm, we have:
    \begin{equation}
        \norm{ \bracks{\mI_d\otimes\mLambda^l}\bracks{\vb- \vb_1}}_F = \sqrt{\sum_{i=1}^{n}\sum_{r=1}^d \bracks{\lambda_i^l\bracks{\evb_{i,r} - \evb^\prime_{i,r}}}^2}\, .
    \end{equation}
    As $\evb_{1,r} = \evb^\prime_{1,r}$ for all $r\in[d]$ we can remove the terms where $i=1$:
    \begin{equation}
        \norm{ \bracks{\mI_d\otimes\mLambda^l}\bracks{\vb- \vb_1}}_F = \sqrt{\sum_{i=2,r=1}^{n,d} \bracks{\lambda_i^l\bracks{\evb_{i,r} - \evb^\prime_{i,r}}}^2}\, .
    \end{equation}
    This can be further bounded by the largest absolute eigenvalue $\lambda_2^l<1$:
    \begin{equation}
        \norm{ \bracks{\mI_d\otimes\mLambda^l}\bracks{\vb- \vb_1}}_F \leq \lambda_2^l\sqrt{\sum_{i=2,r=1}^{n,d} \bracks{\evb_{i,r} - \evb^\prime_{i,r}}^2}\, .
    \end{equation}
    The terms inside the square root can further be bounded by their maximal value $b_{\max} = \max_{i,r} |\evb_{i,r}|$:
    \begin{equation}
    \begin{split}
        \norm{\bracks{\mI_d\otimes\mLambda^l}\bracks{\vb- \vb_1}}_F 
        &\leq \lambda_2^l\cdot b_{\max}\cdot \sqrt{(n-1)\cdot d}\, .
    \end{split}
    \end{equation}
    As $l\to\infty$, this part converges to zero as $\lim_{l\to\infty} \lambda_2^l = 0$ due to $|\lambda_2|<1$.

    Next, we bound the norm in the numerator:
    \begin{equation}
         \norm{\frac{\mO^{(l)}}{\sigma_1^{(l)}}\otimes\mU}_2 = \norm{\frac{\mO^{(l)}}{\sigma_1^{(l)}}}_2\cdot\norm{\mU}_2 = 1
    \end{equation}
    using the property that $\mU$ and $\frac{\mO^{(l)}}{\sigma_1^{(l)}}$ have a maximal singular value of $1$.
    
    As the last part, we bound the denominator:
    \begin{equation}
        \frac{1}{\norm{\frac{\tvec(\mX^{(l)})}{\sigma_1^{(l)}}}_2} = \frac{1}{\sqrt{\frac{\tvec(\mX^{(l)})}{\sigma_1^{(l)}}^\top\frac{\tvec(\mX^{(l)})}{\sigma_1^{(l)}}}}\, ,
    \end{equation}
    which is given by the definition of the Frobenius norm.
    By substituting $\tvec\bracks{\mX^{(0)}} = \bracks{\mN^{(l)}\otimes\mU}\vb^{(l)}$ where $\vb^{(l)} = \bracks{\mN^{(l)}\otimes\mU}^\top\tvec\bracks{\mX^{(0)}}$, we get
    \begin{equation}
     \frac{1}{\norm{\frac{\tvec(\mX^{(l)})}{\sigma_1^{(l)}}}_2} = \frac{1}{\sqrt{\bracks{\vb^{(l)}}^\top\bracks{\frac{\bracks{\mN^{(l)}}^\top\bracks{\mO^{(l)}}^\top\mO^{(l)}\mN^{(l)}}{\bracks{\sigma_1^{(l)}}^2}\otimes\mU^\top\mA_\tsym^{2l}\mU}\vb^{(l)}}}\, .
    \end{equation}
    Due to orthonormality, we have $\mU\mU^\top = \mI_n$ and $\mN^{(l)}\bracks{\mN^{(l)}}^\top = \mI_d$. This allows the simplification
    \begin{equation}
        \frac{1}{\norm{\frac{\tvec\bracks{\mX^{(l)}}}{\sigma_1^{(l)}}}_2} = \frac{1}{\sqrt{\bracks{\vb^{(l)}}^\top\bracks{\bracks{\frac{\mSigma^{(l)}}{\bracks{\sigma_1^{(l)}}^2}}^2\otimes\mLambda^{2l}}\vb^{(l)}}}\, . 
    \end{equation}
    As all involved matrices are diagonal, we can further simplify this as a sum of element-wise multiplications:
    \begin{equation}
        \frac{1}{\norm{\frac{\tvec\bracks{\mX^{(l)}}}{\sigma_1^{(l)}}}_2} = \frac{1}{\sqrt{\sum_{i=1}^{n}\sum_{r=1}^d \bracks{b^{(l)}_{i,r}}^2 \bracks{\frac{(\sigma_r^{(l)})^2}{\bracks{\sigma_1^{(l)}}^2}}^2\lambda_{i}^{2l}}}\, .
    \end{equation}
    This can be bounded by the coefficient $b_{1,1}$ corresponding to the maximal singular value $\sigma_1^{(l)}$ and the maximal eigenvalue $\lambda_1$ of ${\mA_\tsym}$:
    \begin{equation}
        \frac{1}{\norm{\frac{\tvec\bracks{\mX^{(l)}}}{\sigma_1^{(l)}}}_2}  \leq \frac{1}{b^{(l)}_{1,1}} = \frac{1}{\bracks{\mN^{(l)}_{:,1} \otimes \mU_{:,1}}^\top\tvec\bracks{\mX^{(0)}}}\, .
    \end{equation}
    Combining these upper bounds on each term, we get
    \begin{equation}
        \norm{\frac{\tvec\bracks{\mX^{(l)}}}{\norm{\tvec\bracks{\mX^{(l)}}}_F} - \vm^{(l)}}_2 \leq \frac{1}{\bracks{\mN^{(l)}_{:,1} \otimes \mU_{:,1}}^\top\tvec\bracks{\mX^{(0)}}}\lambda_2^l\cdot\sqrt{n-1}\cdot\sqrt{d}\cdot b_{\max}\, .
    \end{equation}
    As the last step, we need to show that the probability of this term being larger than any fixed $\epsilon$ goes to zero for large $l$.
    \begin{equation}
    \begin{split}
        &\frac{1}{\bracks{\mN^{(l)}_{:,1} \otimes \mU_{:,1}}^\top\tvec\bracks{\mX^{(0)}}}\cdot\lambda_2^l\cdot\sqrt{n-1}\cdot\sqrt{d}\cdot b_{\max} > \epsilon \\
        \Leftrightarrow &\frac{1}{\eps}\cdot\lambda_2^l\cdot\sqrt{d}\cdot\sqrt{n-1}\cdot b_{\max} > \bracks{\mN^{(l)}_{:,1} \otimes \mU_{:,1}}^\top\tvec\bracks{\mX^{(0)}}
    \end{split}
    \end{equation}
    As the left-hand side exponentially converges to zero, the right-hand side also converges to zero. 
    Thus, the component $\bracks{\mN^{(l)}_{:,1} \otimes \mU_{:,1}}^\top\tvec\bracks{\mX^{(0)}}\in\R$ of the initial state needs to be increasingly close to zero.
    For a uniform distribution over any bounded interval, the probability of $\mX^{(0)}$ being in this increasingly small subspace converges to zero.
\end{proof}

\Cref{pr:asym_wany} confirms that the probability of having initial node representations $\mX^{(0)}$ for which $\mX^{(l)}$ is not getting dominated by the subspace $\gQ_1$ converges to zero. 
As $\gQ_1$ is constructed using the eigenvector $\mU_{:,1} = \mD^{1/2}\mathbf{1}$ of $\mA_\tsym$, $\mX^{(l)}$ gets dominated by this smooth vector in probability.
Based on the extensive proof, this phenomenon can only be prevented when $\bracks{\mN^{(l)}_{:,1} \otimes \mU_{:,1}}^\top\tvec\bracks{\mX^{(0)}}$, where $\mN^{(l)}_{:,1}$ is the first right singular vector of $\mW^{(1)}\dots\mW^{(l)}$, is increasingly close to zero for $l\to\infty$. 

Equivalently, this convergence can be stated in probability over the feature transformations $\mW^{(k)}$ instead of the initial node representations $\mX^{(0)}$. As the node representations need to align closely with the null space of the feature transformations, the probability can be taken over either one.
Due to this property describing a relative phenomenon between components, the normalization of node representations emerges naturally. The shared scalar multiple $\norm{\mX^{(l)}}_F$ does not affect this phenomenon.
When the feature transformations lead to a large norm $\norm{\mX^{(l)}}_F$, previous studies on over-smoothing claimed over-separation of node representations. \Cref{pr:asym_wany} confirms that the smooth component still dominates the resulting node features. Studying the unnormalized node representations does not provide sufficient insights into this underlying phenomenon.

For the special case of a fixed and shared feature transformation $\mW = \mW^{(k)}$ for every $k\in\mathbb{N}$, $\mX^{(0)}$ needs to align exactly with the null space of $\mW$. This is a more specific requirement and corresponds to $\mX^{(l)}$ being dominated by $\gQ_1$ for almost every initial $\mX^{(0)}$ with respect to the Lebesgue measure. We formalize this in the following Proposition:

\begin{proposition}[Representations converge to $\gQ_1$ for almost every $\mX^{(0)}$]
\label{pr:asym_wfixed}
Let $\mX^{(k)} = \mA_\tsym\mX^{(k-1)}\mW$, where $\mA_\tsym\in\R^{n\times n}$ is the symmetrically normalized adjacency matrix and $\mW\in\R^{d\times d}$ any matrix. Then, 
    \begin{equation}
        \lim_{l\to\infty} \norm{\frac{\mX^{(l)}}{\norm{\mX^{(l)}}_F} - \mM^{(l)}}_F = 0
    \end{equation}
    for almost every $\mX^{(0)}\in\R^{n\times d}$ with respect to the Lebesgue measure and some $\tvec(\mM^{(l)}) \in\gQ_1$ that depends on $l$ for every $l\in\mathbb{N}$.
\end{proposition}

\begin{proof}
    Based on \Cref{pr:asym_wany}, we have 
    \begin{equation}
        \norm{\frac{\tvec\bracks{\mX^{(l)}}}{\norm{\tvec\bracks{\mX^{(l)}}}_F} - \vm^{(l)}}_F \leq \frac{1}{\bracks{\mN_{:,1} \otimes \mU_{:,1}}^\top\tvec\bracks{\mX^{(0)}}}\cdot\lambda_2^l\cdot\sqrt{n-1}\cdot\sqrt{d}\cdot b_{\max}\, ,
    \end{equation}
    where $\mN$ are the right singular vectors of $\mW$ given by the singular value decomposition $\mW = \mV\mSigma\mN^\top$.
    Thus, the right-hand side is a constant instead of a layer-dependent value. As such, this inequality holds for sufficiently large $l$ when $\bracks{\mN_{:,1} \otimes \mU_{:,1}}^\top\tvec\bracks{\mX^{(0)}}\neq 0$, which is satisfied for almost every $\mX^{(0)}$.
\end{proof}

We summarize these two findings about $\mX^{(l)}$ in the limit $l\to\infty$ as follows. For time-inhomogeneous feature transformation, the probability that the GCN avoids domination by the smooth subspace converges to zero as $l\to\infty$, but always remains non-zero for any finite $l$. When the feature transformation is shared across GCN iterations, this dominance occurs for almost every initial matrix $\mX^{(0)}\in\R^{n\times d}$ with respect to the Lebesgue measure. Next, we show how this property can be equivalently observed as over-smoothing using the Dirichlet energy.

\subsubsection{Over-Smoothing as a Consequence}
\label{sec:3:over-smoothing_conseq}

We now connect our identified properties of the GCN with over-smoothing. Based on \Cref{pr:asym_wany}, we have identified that node representations $\mX^{(l)}$ converge in probability to the space of rank-one matrices induced by $\gQ_1$. Any $\tvec(\mM)\in\gQ_1$ is a rank-one matrix of the form $\mM = \mU_{:,1}\vb_{(\mM)}$ with the dominant eigenvector of $\mA_\tsym$ given by $\mU_{:,1} = \mD^{1/2}\mathbf{1}$ and column-wise scalars $\vb_{(\mM)}\in\R^d$ that depend on $\mM$. Thus, each column of $\mX^{(l)}$ increasingly aligns with the degree-proportional vector. 
In spectral graph theory (see \Cref{sec:fundamentals:spectral}), $\mU_{:,1}$ is the smoothest eigenvector of $\mL_\tsym$, corresponding to the lowest frequency of the graph Fourier transform.
As one common metric for observing over-smoothing~\parencite{cai2020anote,zhou2021dirichlet,giovanni2023understanding}, the Dirichlet energy $E(\mX) = \trt\bracks{\mX^\top\mL_\tsym\mX}$ has the property that the degree-proportional vector is in the null space of $\mL_\tsym$, i.e., $\mL_\tsym\mU_{:,1} = \mathbf{0}$. Thus, we have $E(\mM) = 0$ for any $\tvec(\mM)\in\gQ_1$. As $\frac{\mX^{(l)}}{\norm{\mX^{(l)}}_F}$ converges in probability to matrices $\tvec(\mM)\in\gQ_1$ as $l\to\infty$, we can equivalently show and observe that the Dirichlet energy $E\bracks{\frac{\mX^{(l)}}{\norm{\mX^{(l)}}_F}}$ of the normalized node representations converges to zero in probability. We formalize this in the following proposition:

\begin{proposition}[The linearized GCN exhibits Over-Smoothing]
\label{pr:over_smoothing_sym}
Let $\mX^{(k)} = \mA_\tsym\mX^{(k-1)}\mW^{(k)}$, where $\mW^{(k)}\in\R^{d\times d}$ is any matrix for all $k\in\mathbb{N}$ and $\mA_\tsym\in\R^{n\times n}$ is the symmetrically normalized adjacency matrix with $\lambda_2$ being its second largest eigenvalue in magnitude. Then,
    \begin{equation}
        \plim_{l\to\infty}E\bracks{\frac{\mX^{(l)}}{\norm{\mX^{(l)}}_F}} = 0
    \end{equation}
    with convergence rate $\lambda_2^2$ in probability for $\mX^{(0)}\in[-m,m]^{n\times d}$ for any $m\in\R$ with respect to the Lebesgue measure.
\end{proposition}

\begin{proof}
    Recall the definition of the Dirichlet energy and the Frobenius norm:
    \begin{equation}
        E\bracks{\frac{\mX^{(l)}}{\norm{\mX^{(l)}}_F}} = \frac{\trt\bracks{\bracks{\mX^{(l)}}^\top\mL_\tsym\mX^{(l)}}}{\trt\bracks{\bracks{\mX^{(l)}}^\top\mX^{(l)}}}\, .
    \end{equation}
    Using the identity $\trt\bracks{\mA^\top\mB} = \tvec(\mA)^\top\tvec(\mB)$ for any matrices $\mA,\mB$ of suitable shape yields
    \begin{equation}
        E\bracks{\frac{\mX^{(l)}}{\norm{\mX^{(l)}}_F}} = \frac{\tvec\bracks{\mX^{(l)}}^\top\tvec\bracks{\mL_\tsym\mX^{(l)}}}{\tvec\bracks{\mX^{(l)}}^\top\tvec\bracks{\mX^{(l)}}}\, .
    \end{equation}
    Let $\mO^{(l)} = \mW^{(1)}\dots\mW^{(l)}$. Based on the identity $\tvec\bracks{\mX^{(l)}} = \bracks{\mO^{(l)}\otimes\mA_\tsym^l}\tvec\bracks{\mX^{(0)}}$, we obtain
    \begin{equation}
        E\bracks{\frac{\mX^{(l)}}{\norm{\mX^{(l)}}_F}} = \frac{\tvec\bracks{\mX^{(0)}}^\top\bracks{\mO^{(l)}\otimes\mA_\tsym^l}^\top\bracks{\mI\otimes\mL_\tsym}\bracks{\mO^{(l)}\otimes\mA_\tsym^l}\tvec\bracks{\mX^{(0)}}}{\tvec\bracks{\mX^{(0)}}^\top\bracks{\mO^{(l)}\otimes\mA_\tsym^l}^\top\bracks{\mO^{(l)}\otimes\mA_\tsym^l}\tvec\bracks{\mX^{(0)}}}\, .
    \end{equation}
    Consider the singular value decomposition $\mO^{(l)} = \mV^{(l)}\mSigma^{(l)}\bracks{\mN^{(l)}}^\top$ and eigendecomposition $\mA_\tsym = \mU\mLambda\mU^\top$. Using the singular vectors and eigenvectors as basis vectors, we rewrite the vectorized node representations as $\tvec\bracks{\mX^{(0)}} = \bracks{\mN^{(l)}\otimes\mU}\tvec\bracks{\mB^{(l)}}$ for $\tvec\bracks{\mB^{(l)}} = \bracks{\mN^{(l)}\otimes\mU}^\top\tvec\bracks{\mX^{(0)}}$. We also utilize the mixed-product property $(\mA\otimes\mB)(\mC\otimes\mD) = (\mA\mC\otimes\mB\mD)$ for any matrices $\mA,\mB,\mC,\mD$ of suitable shape. This yields
    \begin{multline}
        E\bracks{\frac{\mX^{(l)}}{\norm{\mX^{(l)}}_F}} =  \\\frac{\bracks{\tvec\bracks{\mB^{(l)}}}^\top\bracks{\bracks{\mN^{(l)}}^\top\bracks{\mO^{(l)}}^\top\mO^{(l)}\mN^{(l)}}\otimes\bracks{\bracks{\mU}^\top\mA_\tsym^l\mL_\tsym\mA_\tsym^l\mU}\tvec\bracks{\mB^{(l)}}}{\bracks{\tvec\bracks{\mB^{(l)}}}^\top\bracks{\bracks{\mN^{(l)}}^\top\bracks{\mO^{(l)}}^\top\mO^{(l)}\mN^{(l)}}\otimes\bracks{\bracks{\mU}^\top\mA_\tsym^l\mA_\tsym^l\mU}\tvec\bracks{\mB^{(l)}}}\, .
    \end{multline}
    As $\mO^{(l)}\mN^{(l)} = \mV^{(l)}\mSigma^{(l)}$ and $\mA_\tsym^l\mU = \mU\mLambda^l$, this simplifies as
    \begin{equation}
        E\bracks{\frac{\mX^{(l)}}{\norm{\mX^{(l)}}_F}} = \frac{\tvec\bracks{\mB^{(l)}}^\top\bracks{\bracks{\mSigma^{(l)}}^2\otimes \mLambda^{2l}\bracks{\mI_n-\mLambda}}\tvec\bracks{\mB^{(l)}}}{\tvec\bracks{\mB^{(l)}}^\top\bracks{\bracks{\mSigma^{(l)}}^2\otimes \mLambda^{2l}}\tvec\bracks{\mB^{(l)}}}\, .
    \end{equation}
    As all involved matrices are diagonal, this can be simplified as a sum over element-wise products:
    \begin{equation}
        E\bracks{\frac{\mX^{(l)}}{\norm{\mX^{(l)}}_F}} = \frac{\sum_{i=1}^{n}\sum_{r=1}^d \bracks{\emB_{i,r}^{(l)}}^2\bracks{\sigma_r^{(l)}}^2\lambda_i^2\bracks{1-\lambda_i}}{\sum_{i=1}^{n}\sum_{r=1}^d \bracks{\emB_{i,r}^{(l)}}^2\bracks{\sigma_r^{(l)}}^2\lambda_i^2}\, ,
    \end{equation}
    where $\sigma_r^{(l)} = \emSigma_{r,r}^{(l)}$ and $\lambda_i = \emLambda_{i,i}$.
    We normalize the singular values using the largest singular value $\sigma_1^{(l)}$ in both the numerator and denominator. As the maximal eigenvalue $\lambda_1 = 1$ is one, normalization of eigenvalues is not required:
    \begin{equation}
        E\bracks{\frac{\mX^{(l)}}{\norm{\mX^{(l)}}_F}} = \frac{\sum_{i=1}^{n}\sum_{r=1}^d \bracks{\emB_{i,r}^{(l)}}^2\frac{\bracks{\sigma_r^{(l)}}^2}{\bracks{\sigma_1^{(l)}}^2}\lambda_i^2\bracks{1-\lambda_i}}{\sum_{i=1}^{n}\sum_{r=1}^d \bracks{\emB_{i,r}^{(l)}}^2\frac{\bracks{\sigma_r^{(l)}}^2}{\bracks{\sigma_1^{(l)}}^2}\lambda_i^2}\, .
    \end{equation}
    We bound both the numerator using the maximal absolute coefficient $b_{\max}^{(l)} = \max_{i,r} |\emB_{i,r}^{(l)}|$ and the denominator using coefficient $\emB_{1,1}^{(l)}$ corresponding to the first eigenvalue $\lambda_1=1$ and singular value $\sigma_1^{(l)}/\sigma_1^{(l)} = 1$:
    \begin{equation}
        E\bracks{\frac{\mX^{(l)}}{\norm{\mX^{(l)}}_F}} < \frac{(n-1)\cdot d\cdot \bracks{b_{\max}^{(l)}}^2\lambda_2^{2l}\bracks{1-\lambda_2}}{\bracks{\emB_{1,1}^{(l)}}^2}\, .
    \end{equation}

    For convergence in probability, we need to ensure that the probability that this upper bound is larger than any fixed epsilon converges to zero, i.e.,
    \begin{equation}
        \lim_{l\to\infty}\mathbb{P}\bracks{\frac{(n-1)\cdot d\cdot (b_{\max}^{(l)})^2\lambda_2^{2l}\bracks{1-\lambda_2}}{\bracks{\emB_{1,1}^{(l)}}^2} > \epsilon} = 0\, .
    \end{equation}
    As $\lambda_2^{2l} = 0$ for $l\to\infty$, it suffices to show that the ratio $\frac{\bracks{b_{\max}^{(l)}}^2}{\bracks{\emB_{1,1}^{(l)}}^2}$ is bounded in probability. Given that $\mX^{(0)}$ is sampled from a bounded domain, the probability that a randomly sampled $\mX^{(0)}$ violates this condition converges to zero. 
\end{proof}

Based on this finding, the probability that over-smoothing, as measured by the Dirichlet energy, occurs for $\mX^{(l)}$ converges to one for $l\to\infty$.
This finding further motivates applying the Dirichlet energy to normalized node representations $\frac{\mX^{(l)}}{\norm{\mX^{(l)}}_F}$ to empirically observe over-smoothing and its underlying theoretical property. Contrarily, applying the Dirichlet energy to the unnormalized node representations $\mX^{(l)}$ does not capture this property. For unnormalized node representations, we could not distinguish between avoiding over-smoothing and the smooth component dominating $\mX^{(l)}$ while all other components are negligible in comparison. This also aligns with our findings from \Cref{sec:understanding:zero}, in which we have shown that the norm of the node representations overshadows possible insights of the Dirichlet energy when applied to unnormalized node representations. It also aligns with our spectral study on the GCN in \Cref{sec:understanding:fourier}, in which we have shown that all filters of the GCN in the SISO case are equivalent up to a uniform scaling of all components. Previous work similarly uses the normalized state within the Dirichlet energy for improved insights into the behavior of MPNNs~\parencite{chen2020measuring,liu2020towards,giovanni2023understanding,maskey2023fractional}.
Next, we generalize our study to arbitrary aggregation matrices $\tilde{\mA}$ instead of the symmetrically normalized adjacency matrix $\mA_\tsym$.

\subsection{The Effect of Arbitrary Aggregations}
To this point, our theoretical study assumed the aggregation matrix to be the symmetrically normalized adjacency matrix $\mA_\tsym$. However, many other MPNNs adopt a similar update rule to \Cref{eq:understanding:form} that can be expressed 
\begin{equation}
\label{eq:update_anya}
    \mX^{(k)} = \tilde{\mA}\mX^{(k-1)}\mW^{(k)}\, ,
\end{equation}
where $\tilde{\mA}\in\R^{n\times n}$ is an arbitrary matrix containing any edge weights between any two nodes. Examples of such an $\tilde{\mA}$ include the sum aggregation, mean aggregation, attention-based scores, or negative edge weights. 
As described in \Cref{sec:understanding_sota}, most previous studies assumed a fixed aggregation function, typically $\mA_\tsym$ or $\mA_\trw$.
We find that the insights from \Cref{pr:symm_iter} (shared component amplification), \Cref{pr:basis_limit} (component dominance), and \Cref{pr:asym_wany} extend to this generalized case.

\subsubsection{A Single Iteration}
Specifically, we first extend \Cref{pr:symm_iter} to arbitrary aggregation functions $\tilde{\mA}$ and prove that feature transformations cannot influence the relative amplification of components whenever an update rule as in \Cref{eq:update_anya} is used:

\begin{proposition}[Component Amplification is independent of $\mW$]
\label{pr:jordan_dom}
Let $\mT = \mW\otimes\tilde{\mA}\in\R^{nd\times nd}$, where $\mW\in\R^{d\times d}$ and $\tilde{\mA}\in\R^{n\times n}$ are arbitrary matrices. Let $\mP_{(i)}\in\R^{n \times n_{(i)}}$ and $\mP_{(j)}\in\R^{n \times n_{(j)}}$ be two arbitrary matrices for some $n_{(i)},n_{(j)}\in\mathbb{N}$. Then,
    \begin{equation}
    \frac{\norm{\bracks{\mW\otimes\tilde{\mA}}\bracks{\mI_d\otimes\mP_{(i)}}}_F}{\norm{\bracks{\mW\otimes\tilde{\mA}}\bracks{\mI_d\otimes\mP_{(j)}}}_F} = \frac{\norm{\tilde{\mA}\mP_{(i)}}_F}{\norm{\tilde{\mA}\mP_{(j)}}_F}\, .
    \end{equation}
\end{proposition}
\begin{proof}
    The proof is analogous to the proof of \Cref{pr:symm_iter}.
    We begin by applying the mixed-product property of the Kronecker and matrix product:
    \begin{equation}
        \frac{\norm{\bracks{\mW\otimes\tilde{\mA}}\bracks{\mI_d\otimes\mP_{(i)}}}_F}{\norm{\bracks{\mW\otimes\tilde{\mA}}\bracks{\mI_d\otimes\mP_{(j)}}}_F} = \frac{\norm{\mW\otimes\tilde{\mA}\mP_{(i)}}_F}{\norm{\mW\otimes\tilde{\mA}\mP_{(j)}}_F}\, .
    \end{equation}
    Now, we use the sub-multiplicativity of the norm for Kronecker products:
    \begin{equation}
        \frac{\norm{\mW\otimes\tilde{\mA}\mP_{(i)}}_F}{\norm{\mW\otimes\tilde{\mA}\mP_{(j)}}_F} = \frac{\norm{\mW}_F\cdot\norm{\tilde{\mA}\mP_{(i)}}_F}{\norm{\mW}_F\cdot\norm{\tilde{\mA}\mP_{(j)}}_F}\, .
    \end{equation}
    Since $\norm{\mW}_F$ cancels out, we conclude:
    \begin{equation}
        \frac{\norm{\bracks{\mW\otimes\tilde{\mA}}\bracks{\mI_d\otimes\mP_{(i)}}}_F}{\norm{\bracks{\mW\otimes\tilde{\mA}}\bracks{\mI_d\otimes\mP_{(j)}}}_F} = \frac{\norm{\tilde{\mA}\mP_{(i)}}_F}{\norm{\tilde{\mA}\mP_{(j)}}_F}
    \end{equation}
\end{proof}

Matrices $\mP_{(i)}$ and $\mP_{(j)}$ correspond to bases of different components in the node representations.
The relative amplification of these components solely depends on the aggregation matrix $\tilde{\mA}$. It is thus fixed and independent of the learnable feature transformation $\mW$.
As with \Cref{pr:symm_iter}, it is also shared across feature channels, meaning that the same component gets amplified for all feature channels. We will define a generalized form of this phenomenon of shared component amplification (SCA) in \Cref{def:sca}.
For full-rank matrices $\tilde{\mA}$, $\mP_{(i)}$ can be constructed using their eigenvectors. 
In the general case for any $\tilde{\mA}$, $\mP_{(i)} = \mV_{:,i}$ can be selected as the singular vector $\mV_{:,i}$ utilizing the singular value decomposition $\tilde{\mA} = \mN\mSigma\mV^\top$. Each component then gets amplified by its corresponding singular value. We formalize this property in the following corollary:

\begin{theorem}[SCA in Message-Passing]
\label{pr:sca_svd}
Let $\mT = \mW\otimes\tilde{\mA}\in\R^{nd\times nd}$, where $\mW\in\R^{d\times d}$ and $\tilde{\mA}\in\R^{n\times n}$ are arbitrary matrices. Let the singular value decomposition of $\tilde{\mA}$ be $\tilde{\mA}=\mN\mSigma\mV^\top$ with $\mV_{:,i}\in\R^{n\times 1}$ and $\sigma_i = \emSigma_{i,i}$ for all $i\in[n]$. Then, for all $i,j\in [n]$,
    \begin{equation}
    \frac{\norm{\bracks{\mW\otimes\tilde{\mA}}\bracks{\mI_d\otimes\mV_{:,i}}}_F}{\norm{\bracks{\mW\otimes\tilde{\mA}}\bracks{\mI_d\otimes\mV_{:,j}}}_F} = \frac{\sigma_i}{\sigma_j}\, .
    \end{equation}
\end{theorem}

\begin{proof}
    Following \Cref{pr:jordan_dom}, we have 
    \begin{equation}
        \frac{\norm{\bracks{\mW\otimes\tilde{\mA}}\bracks{\mI_d\otimes\mV_{:,i}}}_F}{\norm{\bracks{\mW\otimes\tilde{\mA}}\bracks{\mI_d\otimes\mV_{:,j}}}_F} = \frac{\norm{\tilde{\mA}\mV_{:,i}}_F}{\norm{\tilde{\mA}\mV_{:,j}}_F}
    \end{equation}
    As each $\mV_{:,i}$ is a right singular vector of $\tilde{\mA}$, we have
    \begin{equation}
        \frac{\norm{\tilde{\mA}\mV_{:,i}}_F}{\norm{\tilde{\mA}\mV_{:,j}}_F} = \frac{\norm{\sigma_i\mN_{:,i}}_F}{\norm{\sigma_j\mN_{:,j}}_F}\, .    
    \end{equation}
    As $\mN$ is an orthonormal matrix, this reduces to
    \begin{equation}
        \frac{\norm{\sigma_i\mN_{:,i}}_F}{\norm{\sigma_j\mN_{:,j}}_F} = \frac{\sigma_i}{\sigma_j}\, .
    \end{equation}
\end{proof}

Based on the singular vectors $\mV$ of the aggregation matrix $\tilde{\mA}$, the singular vector $\mV_{:,1}$ with the maximal singular value $\sigma_1$ gets amplified maximally with a fixed ratio relative to all other singular vectors across all feature channels. As the subspaces induced by the singular vectors are not invariant to $\mW\otimes\tilde{\mA}$, \Cref{pr:sca_svd} cannot be extended to the iterated case.
However, for any matrix, $\mP_{(i)}$ can be selected to be generalized eigenvectors of $\tilde{\mA}$ based on the Jordan normal form. For all eigenvalues of the same magnitude, the corresponding generalized eigenvectors span subspaces that are invariant under $\tilde{\mA}$. 

\subsubsection{Repeated Iterations}

We consider the Jordan decomposition $\tilde{\mA} = \mP\mJ\mP^{-1}$ where $\mJ\in\R^{n\times n}$ is a block diagonal matrix that contains the Jordan blocks of $\tilde{\mA}$ and $\mP$ contains the generalized eigenvectors. 
Let $\mJ_{(i)}$ contain the Jordan blocks corresponding to all eigenvalues with the same magnitude.
We form the basis vectors $\mP_{(i)}$ that contain all columns of the generalized eigenvectors $\mP$ corresponding to all eigenvalues of $\tilde{\mA}$ with the same magnitude. Each $\mP_{(i)}$ spans a subspace that is invariant under $\tilde{\mA}$, i.e., we have $\tilde{\mA}\mP_{(i)} = \mP_{(i)}\mJ_{(i)}$. It is also known that the Jordan block $\mJ_{(i)}$ containing the eigenvalues with the largest magnitude dominates all other Jordan blocks $\mJ_{(j)}$ in the limit $l\to\infty$. We formalize this in the following proposition:

\begin{proposition}[Component Dominance for Arbitrary Aggregation Matrices]
    \label{pr:aany_cd}
Let $\mT^{(k)} = \mW^{(k)}\otimes\tilde{\mA}\in\R^{nd\times nd}$ for $k\in\mathbb{N}$, where $\mW^{(k)}\in\R^{d\times d}$ are arbitrary matrices and $\tilde{\mA}\in\R^{n\times n}$ has Jordan decomposition $\tilde{\mA} = \mP\mJ\mP^{-1}$. 
Let $r$ denote the number of distinct eigenvalue magnitudes of $\tilde{\mA}$, and for each $i\in[r]$, let $\mP_{(i)}\in\R^{n\times d_{(i)}}$ be the submatrix of $\mP$ consisting of the $d_{(i)}$ columns corresponding to eigenvalues of $\tilde{\mA}$ with magnitude $|\lambda_i|$. Then, for all $i\in [n]$,
    \begin{equation}
    \frac{\norm{\bracks{\mW^{(l)}\otimes\tilde{\mA}}\dots\bracks{\mW^{(1)}\otimes\tilde{\mA}}\bracks{\mI_d\otimes\mP_{(i)}}}_F}{\max_p\norm{\bracks{\mW^{(l)}\otimes\tilde{\mA}}\dots\bracks{\mW^{(1)}\otimes\tilde{\mA}}\bracks{\mI_d\otimes\mP_{(p)}}}_F} = \begin{cases}
    1, & \text{if $|\lambda_i|=\max_{j\in[n]}|\lambda_j|$}\\ 0, & \text{otherwise}\, .
    \end{cases}
    \end{equation}
\end{proposition}
\begin{proof}
Without loss of generality, we assume $\max_{j\in[n]}|\lambda_j| = |\lambda_1|$. We have $|\lambda_i|<|\lambda_1|$ for all $i\neq 1$. Based on \Cref{pr:jordan_dom}, we obtain
\begin{equation}
    \frac{\norm{\bracks{\mW^{(l)}\otimes\tilde{\mA}}\dots\bracks{\mW^{(1)}\otimes\tilde{\mA}}\bracks{\mI_d\otimes\mP_{(i)}}}_F}{\max_p\norm{\bracks{\mW^{(l)}\otimes\tilde{\mA}}\dots\bracks{\mW^{(1)}\otimes\tilde{\mA}}\bracks{\mI_d\otimes\mP_{(p)}}}_F} = \frac{\norm{\mP_{(i)}\mJ_{(i)}^l}_F}{\max_p\norm{\mP_{(p)}\mJ_{(p)}^l}_F}\, .
    \end{equation}
    We bound the numerator and denominator using the largest eigenvalue in magnitude $|\lambda_1|$ of $\tilde{\mA}$ to obtain
    \begin{equation}
        \frac{\norm{\mP_{(i)}\mJ_{(i)}^l}_F}{\max_p\norm{\mP_{(p)}\mJ_{(p)}^l}_F} = \frac{\norm{\mP_{(i)}\left(\frac{\mJ_{(i)}}{|\lambda_1|}\right)^l}_F}{\max_p\norm{\mP_{(p)}\bracks{\frac{\mJ_{(p)}}{|\lambda_1|}}^l}_F}\, .
    \end{equation}
    We now only consider the limit behavior of the numerator. For $i \neq 1$ all diagonal entries of $\mJ_{(i)}$ are less than one in magnitude, all entries of $\bracks{\frac{\mJ_{(i)}}{|\lambda_1|}}^l$ converge to zero. Thus, we have for $i\neq 1$, 
    \begin{equation}
        \lim_{l\to\infty}\norm{\mP_{(i)}\bracks{\frac{\mJ_{(i)}}{|\lambda_1|}}^l}_F = 0\, .
    \end{equation}
    We now consider the case $i=1$. Let $\vv^\top = \begin{bmatrix}1 &0 &\dots& 0\end{bmatrix}$. We have $\mJ_{(i)}\vv = \vv$ and $\mP_{(i)}v = (\mP_{(i)})_{:,1}$. By definition, we can set $\norm{\bracks{\mP_{(i)}}_{:,1}}_2 = 1$. We also use the norm inequality $\norm{\cdot}_F \geq \norm{\cdot}_2$. In total, we have
    \begin{equation}
        \norm{\mP_{(1)}\bracks{\frac{\mJ_{(1)}}{|\lambda_1|}}^l}_2 = \max_{\norm{u}_2\leq1}\norm{\mP_{(1)}\bracks{\frac{\mJ_{(1)}}{|\lambda_1|}}^l\vu}_2 \geq \norm{\mP_{(1)}\bracks{\frac{\mJ_{(1)}}{|\lambda_1|}}^l\vv}_2 = \norm{\bracks{\mP_{(1)}}_{:,1}}_2
    \end{equation}
    Thus, $i=1$ is the only index that does not converge to zero, which is therefore maximal. The numerator converges to zero for all other cases, while the denominator is lower bounded by 1. This concludes the proof.
\end{proof}

This confirms that the components of $\mX^{(0)}$ corresponding to the largest eigenvalue in magnitude get dominantly amplified compared to the components corresponding to all other eigenvalues.
Unless the initial components corresponding to $\mP_{(1)}$ have increasingly small magnitude, node representations $\mX^{(l)}$ will be dominated by the components from $\mP_{(1)}$. This is equivalent to \Cref{pr:basis_limit}. We can similarly show convergence in probability to subspace $\mP_{(1)}$.

\begin{theorem}[Representations converge to $\gQ_1$ in probability]
\label{pr:anya_anyw}
Let $\mX^{(k)} = \tilde{\mA}\mX^{(k-1)}\mW^{(k)}$, where $\tilde{\mA}\in\R^{n\times n}$ is a matrix with Jordan decomposition $\tilde{\mA} = \mP\mJ\mP^{-1}$, and each $\mW^{(k)}\in\R^{d\times d}$ for every $k\in\mathbb{N}$. Let $\mP_{(1)}\in\R^{n\times d_{(1)}}$ denote the submatrix of $\mP$ consisting of the $d_{(1)}$ columns corresponding to eigenvalues of $\tilde{\mA}$ whose magnitude is equal to $\max_{i\in[n]} |\lambda_i|$. Then, 
    \begin{equation}
        \plim_{l\to\infty} \norm{\frac{\mX^{(l)}}{\norm{\mX^{(l)}}_F} - \mP_{(1)}\mV^{(l)}}_F = 0
    \end{equation}
    for $\mX^{(0)}\in[-m,m]^{n\times d}$ sampled with respect to the Lebesgue measure for some $m\in\R$ and some $\mV^{(l)}\in\R^{d_{(1)}\times d}$ for each $l\in\mathbb{N}$.
\end{theorem}

\begin{proof}
Based on the proof of \Cref{pr:asym_wany}, we can upper bound the distance to a rank-one matrix using the inequality
    \begin{equation}
        \norm{\frac{\mX^{(l)}}{\norm{\mX^{(l)}}_F} - \mP_{(1)}\mV^{(l)}}_F \leq \frac{\norm{\frac{\mO^{(l)}}{\sigma_1^{(l)}}\otimes\mP}_2}{\norm{\frac{\mX^{(l)}}{\lambda_1\sigma_1^{(l)}}}_F}\norm{\bracks{\mI_d\otimes\frac{\mJ^l}{|\lambda_1|}}\bracks{\vb - \vb_1}}_2
    \end{equation}
    where $\sigma_1^{(l)}$ is the largest singular value of $\mO^{(l)}=\mW^{(l)}\dots\mW^{(1)}$,  $\vb = \bracks{\mI_d\otimes\mU}^\top\tvec\bracks{\mX^{(0)}}\in\R^{nd}$, and $\vb_1\in\R^{nd}$ is a copy of $\vb$ that contains zeros for all positions $\bracks{\evb_1}_i$ for which $|\emJ_{i,i}| \neq |\lambda_1|$.
    Equivalently to the proof of \Cref{pr:asym_wany}, we provide upper bounds for each of the three norms separately. First, we bound the Jordan block matrix $\mJ$ using the second-largest eigenvalue in magnitude $|\lambda_2|$:
    \begin{equation}
        \norm{\bracks{\mI_d\otimes\frac{\mJ^l}{|\lambda_1|}}\bracks{\vb - \vb_1}}_2 \leq n\cdot l^{n-1}\cdot \bracks{\frac{|\lambda_2|}{|\lambda_1|} }^l \cdot b_{\max}\cdot \sqrt{n\cdot d}\, .
    \end{equation}
    Next, we bound the nominator of the fraction using the largest entry of $\mP$:
    \begin{equation}
        \norm{\frac{\mO^{(l)}}{\sigma_1^{(l)}}\otimes\mP}_2 = n\cdot \max_{i,j} \emP_{i,j}\, .
    \end{equation}
    Equivalently to the proof of \Cref{pr:asym_wany}, we bound the denominator using a component of $\mX^{(l)}$ corresponding to an eigenvalue of $\tilde{\mA}$ with magnitude $|\lambda_1|$ and a maximal singular value $\sigma_1^{(l)}$ of $\mO^{(l)}$:
    \begin{equation}
        \frac{1}{\norm{\frac{\mX^{(l)}}{\lambda_1\sigma_1^{(l)}}}_F} \leq \frac{1}{\bracks{\mN_{:,1}^{(l)}\otimes\mP_{:,1}}^\top\tvec\bracks{\mX^{(0)}}}
    \end{equation}

    Combining these bounds, we have
    \begin{equation}
        \norm{\frac{\mX^{(l)}}{\norm{\mX^{(l)}}_F} - \mP_{(1)}\mV^{(l)}}_F \leq \frac{n\cdot \max_{i,j} \emP_{i,j}}{\bracks{\mN_{:,1}^{(l)}\otimes\mP_{:,1}}^\top\tvec\bracks{\mX^{(0)}}}n\cdot l^{n-1}\cdot \bracks{\frac{|\lambda_2|}{|\lambda_1|} }^l \cdot b_{\max}\cdot \sqrt{n\cdot d}
    \end{equation}
    Equivalently to the proof of \Cref{pr:asym_wany}, in order for this not to converge to zero, we need $\frac{1}{\bracks{\mN_{:,1}^{(l)}\otimes\mP_{:,1}}^\top\tvec\bracks{\mX^{(0)}}}$ to be unbounded. The probability of this occurring over a bounded domain converges to zero. Thus, we have convergence in probability.
\end{proof}

\Cref{pr:anya_anyw} confirms that for any aggregation matrix $\tilde{\mA}$ and any layer-dependent feature matrices $\mW^{(1)},\dots,\mW^{(l)}$, the node representations $\mX^{(l)}$ converge to a low-rank state in probability for initial states $\mX^{(0)}$. This rank is upper bounded by the number of eigenvalues of $\tilde{\mA}$ with maximal magnitude. Based on this property, feature channels contain increasingly similar and redundant information with increased depth.  

\subsubsection{Rank Collapse and Over-Correlation}
\sloppy Analogously to connecting these properties for the special case of $\tilde{\mA} = \mA_\tsym$ to over-smoothing, we now generalize the concept of over-smoothing and relate it to over-correlation.
Based on \Cref{pr:anya_anyw}, we have seen that node representations get closer to a matrix whose rank is bounded by the number of eigenvalues of $\tilde{\mA}$ with maximal magnitude.
Over-smoothing is a consequence of $\mA_\tsym$ having a unique eigenvalue with maximal magnitude, for which the corresponding eigenvector is $\mD^{1/2}\mathbf{1}$. 
For the general case, we now equivalently assume that $\tilde{\mA}$ has a unique eigenvalue with maximal magnitude, which holds for almost every $\tilde{\mA}$ with respect to the Lebesgue measure~\parencite{griffiths1997principles}.
For such an aggregation matrix $\tilde{\mA}$ with an arbitrary dominant eigenvector, this eigenvector will dominate every feature channel (i.e., column) of $\mX^{(l)}$. 
More generally than converging to a specific rank-one matrix, we refer to the phenomenon of converging to any rank-one matrix, as rank collapse of node representations $\mX^{(l)}$. We formalize this in the following proposition:

\begin{proposition}[Rank Collapse in MPNNs]
\label{pr:study:rank_collapse}
    Let $\mX^{(k)} = \tilde{\mA}\mX^{(k-1)}\mW^{(k)}$, where $\mW^{(k)}\in\R^{d\times d}$ is any matrix for all $k\in\mathbb{N}$ and $\tilde{\mA}\in\R^{n\times n}$ is any matrix with $|\lambda_1|>|\lambda_j|$ for all $j>1$ and corresponding eigenvector $\vu$. Then, 
    \begin{equation}
     \plim_{l\to\infty} \norm{\frac{\mX^{(l)}}{\norm{\mX^{(l)}}_F} - \vu\bracks{\vv^{(l)}}^\top}_F = 0
    \end{equation}
    in probability for $\mX^{(0)} \in [-m,m]^{n\times d}$ for any $m\in\R$ with respect to the Lebesgue measure and a layer-dependent $\vv^{(l)}$ for all $l\in\mathbb{N}$.
\end{proposition}

\begin{proof}
    Based on \Cref{pr:anya_anyw}, we have
    \begin{equation}
     \plim_{l\to\infty} \norm{\frac{\mX^{(l)}}{\norm{\mX^{(l)}}_F} - \mP_{(1)}\mV^{(l)}}_F = 0
    \end{equation}
    where $\mP_{(1)}$ contains all generalized eigenvectors of $\tilde{\mA}$ corresponding to eigenvalues with maximal magnitude. As we assume this eigenvalue $\lambda_1$ to be unique, $\mP_{(1)}$ reduces to a matrix with a single column, which we denote by $\vu$ and the corresponding matrix of coefficients $\mV^{(l)}$ reduces to a matrix with a single row, which we denote as the transposed vector $\bracks{\vv^{(l)}}^\top$.   
\end{proof}

As node representations approach a rank-one matrix, the feature channels become more similar and contain redundant information. The node representations of a rank-one matrix could be reduced to a single feature per node.
\Cref{pr:study:rank_collapse} also confirms that the probability of having feature transformations $\mW^{(k)}$ that prevent rank collapse converges to zero as the number of iterations goes to infinity. Thus, the effect of learnable parameters is limited for message-passing operations of this form. 

\Cref{pr:study:rank_collapse} also explains the observed phenomenon of over-correlation. It was found that feature channels tend to become increasingly correlated with an increasing number of iterations, even when over-smoothing is not observed~\parencite{jin2022feature}. Based on our findings above, when node representations are close to a rank-one matrix, the feature channels are also strongly correlated. Thus, over-correlation and over-smoothing describe similar properties of the underlying message-passing process.

As shown in \Cref{pr:anya_anyw}, when the aggregation function $\tilde{\mA}$ has multiple eigenvalues with maximal magnitude, the representation matrix still collapses to a low-rank subspace, though not necessarily of rank one. In this case, an over-correlation of feature channels may not be observed, but the underlying phenomenon still occurs.

\subsection{The Effect of Iteration-Dependent Attention Weights}
Various proposed message-passing methods utilize an iteration-dependent aggregation function $\tilde{\mA}^{(k)}$ instead of a fixed aggregation function. Most of them use learnable attention coefficients, as employed in the graph attention network (GAT)~\parencite{velickovic2017graph}, GATv2~\parencite{brody2022how}, or graph Transformer~\parencite{shi2021masked,rong2020self,rampasek2022recipe}. The corresponding update rule is given as
\begin{equation}
    \mX^{(k)} = \tilde{\mA}^{(k)}\mX^{(k-1)}\mW^{(k)}
\end{equation}
where the attention coefficients of $\tilde{\mA}^{(k)}\in\R^{n\times n}$ at iteration $k$ depend on the features $\mX^{(k-1)}$. We generally have $\tilde{\mA}^{(i)} \neq \tilde{\mA}^{(j)}$ for $i\neq j$. These are referred to as layer-dependent or time-inhomogeneous aggregation functions. 

While the attention coefficients are iteration-dependent, they share some key properties. Attention coefficients are typically softmax-activated, meaning that the sum over each row is one, and all edge weights are non-negative. This corresponds to a node being updated using a weighted mean over its neighbors. Matrices $\tilde{\mA}^{(k)}$ satisfying these properties are called row-stochastic.
Such matrices have been extensively studied due to their importance in various fields~\parencite{gallager1996discrete}. It is known that any row-stochastic matrix has the all-ones vector $\mathbf{1}$ as an eigenvector with corresponding eigenvalue $\lambda_1 = 1$. From the Perron-Frobenius theorem, it is also known that all other eigenvalues are smaller in magnitude, i.e., $|\lambda_i| < \lambda_1$ for all $i>1$~\parencite{gallager1996discrete}.

For a single iteration, \Cref{pr:jordan_dom} applies and components of $\mX^{(k-1)}$ are amplified across all columns depending on $\tilde{\mA}^{(k)}$. However, the generalized eigenvectors and singular vectors are typically distinct across different row-stochastic matrices $\tilde{\mA}^{(k)}$. Thus, our iterated findings cannot be directly applied. 

We build on a result regarding the product of time-inhomogeneous aggregation matrices defined over a common ergodic graph (i.e., strongly connected and aperiodic) from \textcite{wu2023demystifying}.
They assume all $\tilde{\mA}^{(k)}$ to have shared non-zero entries for all adjacent nodes, and each non-zero entry has a minimum value of $\epsilon>0$. They found that the product of any such time-inhomogeneous aggregation matrices converges to the column-wise constant state, i.e., $\lim_{l\to\infty} \prod_{k=1}^l \tilde{\mA}^{(k)} = \mathbf{1}\vu^\top$ for some vector $\vu\in\R^n$ indicating the constant value for each column.

This means that we can form two subspaces $\gQ_1$ and $\gQ_2$, with $\gQ_1$ corresponding to the constant vector and the other subspace corresponding to all other components. For $\mX^{(l)}$, the components corresponding to $\gQ_1$ get increasingly amplified, while the components corresponding to $\gQ_2$ become negligible.
As the constant vector corresponds to the null space of the unnormalized graph Laplacian $\mL = \mD - \mA$, we can show that the Dirichlet energy $E_\mL (\mX) = \trt\bracks{\mX^\top\mL\mX}$ converges to zero for the normalized state $\frac{\mX^{(l)}}{\norm{\mX^{(l)}}_F}$, independently of the chosen feature transformations:

\begin{proposition}[Iteration-dependent attention exhibits over-smoothing]
\label{pr:gat_smooth}
\sloppy Let $\mX^{(k)} = \tilde{\mA}^{(k)}\mX^{(k-1)}\mW^{(k)}$ where each $\mW^{(k)}\in\R^{d\times d}$ is any matrix and $\tilde{\mA}^{(k)}\in\mathbb{R}^{n\times n}$ is a row-stochastic matrix, representing the same ergodic graph with minimum non-zero value $\epsilon$ in $\tilde{\mA}$ for some $\epsilon>0$.
Then, 
\begin{equation}\plim_{l\to\infty} E_{\mL}\bracks{\frac{\mX^{(l)}}{\norm{\mX^{(l)}}_F}} = 0\,.
\end{equation}
    for $\mX^{(0)}\in[-m,m]^{n\times d}$ for any $m\in\R$ in probability with respect to the Lebesgue measure.
\end{proposition}

\begin{proof}
Based on \parencite{wu2023demystifying}, we use the fact that
    \begin{equation}
        \lim_{l\to\infty} \tilde{\mA}^{(l)}\dots\tilde{\mA}^{(1)} = 1 (\vy)^\top\, .
    \end{equation}
    We utilize the induced decomposition 
    \begin{equation}
        \tilde{\mA}^{(l)}\dots\tilde{\mA}^{(1)} = 1(\vy)^\top + \tilde{\mA}_{-}^{(l)}\, ,
    \end{equation}
    where $\tilde{\mA}_{-}^{(l)} = \tilde{\mA}^{(l)}\dots\tilde{\mA}^{(1)} - 1(\vy)^\top$ with $ \norm{\tilde{\mA}_{-}^{(l)}}_2 \leq Cq^l$ for some $C\in\R$ and $0<q<1$.
    As $\mathbf{1}$ is in the nullspace of $\mL$, i.e., $\mL\mathbf{1} = \mathbf{0}$, we have 
    \begin{equation}
        \bracks{\tilde{\mA}^{(l)}\dots\tilde{\mA}^{(1)}}^\top\mL\bracks{\tilde{\mA}^{(l)}\dots\tilde{\mA}^{(1)}} = \bracks{\tilde{\mA}_{-}^{(l)}}^\top\mL \tilde{\mA}_{-}^{(l)}\, .
    \end{equation}
    Based on the definition of $E_\mL$ and the property of the Kronecker product that $\tvec(\mA\mB\mC) = \bracks{\mC^\top\otimes\mA}\tvec(\mB)$ for matrices $\mA,\mB,\mC$ of compatible shape, we obtain
\begin{equation}
    E_\mL\bracks{\frac{\mX^{(l)}}{\norm{\mX^{(l)}}_F}} = \frac{\tvec\bracks{\mX^{(0)}}^\top\bracks{\bracks{\mO^{(l)}}^\top\mO^{(l)}\otimes\bracks{\tilde{\mA}_{-}^{(l)}}^\top\mL\tilde{\mA}_{-}^{(l)}}\tvec\bracks{\mX^{(0)}}}{\norm{\mX^{(l)}}_F^2}\, ,
\end{equation}
where $\mO^{(l)} = \mW^{(1)}\dots\mW^{(l)}$. By normalizing by the largest singular value $\sigma_1^{(l)}$ of $\mO^{(l)}$, we obtain
\begin{equation}
    E_\mL\bracks{\frac{\mX^{(l)}}{\norm{\mX^{(l)}}_F}} = \frac{\tvec\bracks{\mX^{(0)}}^\top\bracks{\frac{\bracks{\mO^{(l)}}^\top\mO^{(l)}}{\bracks{\sigma_1^{(l)}}^2}\otimes\bracks{\tilde{\mA}_{-}^{(l)}}^\top\bracks{\mL}\tilde{\mA}_{-}^{(l)}}\tvec\bracks{\mX^{(0)})}}{\frac{\norm{\mX^{(l)}}_F^2}{\bracks{\sigma_1^{(l)}}^2}}
\end{equation}
Using the sub-multiplicativity, we upper-bound each term individually and obtain 
\begin{equation}
\label{eq:bound_inhomogen}
   E_\mL\bracks{\frac{\mX^{(l)}}{\norm{\mX^{(l)}}_F}}
    \leq \frac{\norm{\tvec\bracks{\mX^{(0)}}}_2^2 \norm{\frac{\mO^{(l)}}{\sigma_1^{(l)}}}_2^2\norm{\mA^{(l)}_{-}}_2^2\norm{\mL}_2}{\frac{\norm{\mX^{(l)}}_F^2}{\bracks{\sigma_1^{(l)}}^2}}
\end{equation}
As $\lim_{l\to\infty} \norm{\tilde{\mA}_{-}^{(l)}}_2 \leq C q^l = 0$, $\norm{\mL}_2 \leq 2\cdot d_{\max}$ where $d_{\max}$ is the maximum degree of a node in the given graph, and $\norm{\frac{\mO^{(l)}}{\sigma_1^{(l)}}}_2 = 1$, the numerator can be upper bounded as
\begin{equation}
    E_\mL\bracks{\frac{\mX^{(l)}}{\norm{\mX^{(l)}}_F}}
    \leq \frac{\norm{\mX^{(0)}}_F^2\cdot2\cdot d_{\max}\cdot C^2\cdot q^{2l}}{\frac{\norm{\mX^{(l)}}_F^2}{\bracks{\sigma_1^{(l)}}^2}}\, .
\end{equation}
Next, we bound the denominator analogously to the proof of \Cref{pr:asym_wany} to obtain
\begin{equation}
    \frac{1}{\frac{\norm{\mX^{(l)}}_F^2}{\bracks{\sigma_1^{(l)}}^2}} \geq \frac{1}{\bracks{\bracks{\mN_{:,1}^{(l)}\otimes \mathbf{1}}^\top\tvec\bracks{\mX^{(0)}}}^2}\, ,
\end{equation}
where $\mO^{(l)} = \mV^{(l)}\mSigma^{(l)}\bracks{\mN^{(l)}}^\top$ is the singular value decomposition of $\mO^{(l)}$ and $\mN_{:,1}^{(l)}\in\R^{d\times 1}$ contains the right singular vector associated to $\sigma_1^{(l)}$ as its sole column.

Combining these upper bounds, we obtain
\begin{equation}
    E_\mL\bracks{\frac{\mX^{(l)}}{\norm{\mX^{(l)}}_F}} \leq \frac{\norm{\mX^{(0)}}_F^2\cdot 2\cdot d_{\max}\cdot C^2\cdot q^{2l}}{\bracks{\bracks{\mN_{:,1}^{(l)}\otimes \mathbf{1}}^\top\tvec\bracks{\mX^{(0)}}}^2}
\end{equation}
As the numerator converges to zero exponentially, $\mX^{(0)}$ needs to lie in an increasingly small subspace to avoid convergence to zero of the complete term. Since the measure of this subspace shrinks exponentially, the probability of randomly observing such an $\mX^{(0)}$ from a bounded set converges to zero.
\end{proof}

\Cref{pr:gat_smooth} confirms that our insights also hold for attention-based aggregation functions. Repeatedly applying such update steps leads to the dominance of the constant vector across all feature channels. As this constant vector is in the null space of the unnormalized graph Laplacian $\mL$, the Dirichlet energy converges to zero for initial representations $\mX^{(0)}$ with probability converging to one when $l\to\infty$. 
This finding can be similarly applied to any iteration-dependent aggregation functions $\tilde{\mA}^{(1)},\dots\tilde{\mA}^{(l)}$ whenever their product dominates certain components.
However, these theoretical insights do not extend to the multi-head case, where $h$ attention matrices $\mA^{(k)}_1,\dots,\mA^{(k)}_h$ are used for each iteration $k$.

\subsection{Graph Convolutions as Power Iteration}
\label{sec:understanding:rank:power}
To provide further intuition, we now show that the limit behavior of $\frac{\mX^{(l)}}{\norm{\mX^{(l)}}_F}$ for $l\to\infty$ is a special case of the classical power iteration algorithm~\parencite{kowalewski1909einfuehrung,mises1929praktische} when using a shared feature transformation, i.e., $\mW^{(1)} = \dots = \mW^{(l)}$. This connection to a well-established algorithm and behavior will deepen our understanding of the over-smoothing and rank collapse phenomena.
We will first recap the power iteration method and its general proof, before showing that many variations of graph convolutions can be seen as a special case of power iteration. Thus, these graph convolutions exhibit the same properties and standard textbook proofs can be applied. This subsection is based on~\parencite{roth2024simplifying}.

\subsubsection{Power Iteration}
The power method or power iteration is an algorithm to determine the dominant eigenvector and eigenvalue of a matrix $\mT\in\R^{p\times p}$. The key property used within power iteration is that repeatedly applying a matrix $\mT$ to a vector $\vx\in\R^p$ leads to the resulting vector $\mT^l\vx$ getting increasingly dominated by the eigenvector $\mU_{:,1}$ of $\mT$ corresponding to its eigenvalue with maximal magnitude. This holds for almost every matrix $\mT$, as it has a single eigenvalue with maximal magnitude, and almost every $\vx$, as its component corresponding to $\mU_{:,1}$ needs to be non-zero. This means that $\frac{\mT^k\vx}{\norm{\mT^k\vx}_2}$ becomes increasingly equal to plus or minus $\mU_{:,1}$. 
We will first present a commonly used proof for power iteration in general before relating this to graph convolutions and MPNNs: 

\begin{proposition}[Power Iteration~\parencite{knabner2017lineare}]\label{prop:power}
    Let $\mT\in\R^{p\times p}$ for some $p\in\mathbb{N}$ with an eigenvalue $|\lambda_1| > |\lambda_i|$ that is greater in magnitude than all other eigenvalues. Further, let $\mU_{:,1}\in\R^{p}$ be a normalized eigenvector of $\mT$ corresponding to $\lambda_1$. Then, for almost every $\vx\in\R^p$,
    \begin{equation}
        \frac{\mT^k\vx}{\norm{\mT^k\vx}_2} = \beta_k\mU_{:,1} + \vm^{(k)}
    \end{equation}
    for $\vm^{(k)}\in\R^p$ with $\lim_{k\to\infty}\norm{\vm^{(k)}}_2 = 0$ and $\beta_k\in\{-1,1\}$ for every $k\in\mathbb{N}$.
\end{proposition}

\begin{proof}
    The proof utilizes the Jordan normal form $\mT = \mP\mJ\mP^{-1}$. As $\mP$ contains the generalized eigenvectors of $\mT$, we have $\mU_{:,1} = \mP_{:,1}$ and $\emJ_{1,1} = \lambda_1$.
    This simplifies the $k$-th power of $\mT$ to $\mT^k = \mP\mJ^k\mP^{-1}$. We also decompose $\vx$ into a linear combination of $\mP$ by setting $\vc = \mP^{-1}\vx$. Based on these two properties, we have
    \begin{equation}
        \frac{\mT^k\vx}{\norm{\mT^k\vx}_2} = \frac{\mP\mJ^k\vc}{\norm{\mP\mJ^k\vc}_2}\, .
    \end{equation}
    By normalizing the numerator and denominator by $|\lambda_1|$ and the corresponding entry $\evc_1$ of $\vc$, we obtain
    \begin{equation}
        \frac{\mT^k\vx}{\norm{\mT^k\vx}_2} = \frac{\evc_1}{|\evc_1|}\bracks{\frac{|\lambda_1|}{|\lambda_1|}}^k\frac{\mP\bracks{\frac{\mJ}{|\lambda_1|}}^k\frac{\vc}{\evc_1}}{\norm{\mP\bracks{\frac{\mJ}{|\lambda_1|}}^k\frac{\vc}{|\evc_1|}}_2}\, .
    \end{equation}
    As $\mJ$ is a block diagonal matrix, which has $\lambda_1$ as its maximal entry $\emJ_{1,1}$ on its diagonal with a corresponding block size of $1$, normalizing by $|\lambda_1|$ results in all other entries converging to zero, i.e., we have
    \begin{equation}
        \lim_{k\to\infty}\bracks{\frac{\mJ}{|\lambda_1|}}^k = \begin{bmatrix}
            1 & 0 & \dots &  0 \\
            0 & 0 &\dots & 0 \\
            \vdots & \vdots & \ddots & 0 \\
            0 & 0 & \dots & 0
        \end{bmatrix}
    \end{equation}
    
    Thus, all other components $\evc_i$ for $i\neq1$ also converge to zero, and we obtain
    \begin{equation}
        \frac{\mT^k\vx}{\norm{\mT^k\vx}_2} = \frac{\evc_1}{|\evc_1|}\bracks{\frac{|\lambda_1|}{|\lambda_1|}}^k\frac{\mP_{:,1}}{\norm{\mP_{:,1}}_2} + \vm^{(k)} 
    \end{equation}
    for some $\vm^{(k)}$ with $\lim_{k\to\infty} \norm{\vm^{(k)}}_2=0$. We further have $\frac{\evc_1|\lambda_1^k|}{|\evc_1|\cdot|\lambda_1^k|}\in \{-1,1\}$. When the eigenvalue with maximal magnitude is positive, i.e., $\lambda_1 > 0$, the state $\frac{\mT^k\vx}{\norm{\mT^k\vx}_2}$ converges. Otherwise, it oscillates between the positive and negative normalized eigenvector but still converges in direction to $\mP_{:,1}$. 
     
    This proof requires $\evc_1\neq 0$, which is satisfied for almost every $\vx\in\R^p$ with respect to the Lebesgue measure on $\R^p$.
\end{proof}

We summarize this property as follows: When repeatedly applying a matrix $\mT$ to a vector, the eigenvector of $\mT$ corresponding to the eigenvalue of $\mT$ with maximal magnitude increasingly dominates the resulting vector. The components corresponding to all other directions becomes comparably negligible.
The condition that $\mT$ has a unique eigenvalue $|\lambda_1| > |\lambda_i|$ with maximal magnitude is satisfied for almost every matrix $\mT$ with respect to the Lebesgue measure~\parencite{tao2012topics}.

\subsubsection{Graph Convolutions as Power Iteration}
Many variations of graph convolutions or message-passing iterations can be similarly expressed as a matrix-vector product

\begin{equation}
\label{eq:linear_graph_conv}
    \tvec\bracks{\mX^{(k)}} = \tvec\bracks{\tilde{\mA}\mX^{(k-1)}\mW} = \bracks{\mW^\top\otimes \tilde{\mA}}^k\tvec\bracks{\mX^{(0)}} = \mS^k\vx^{(0)}
\end{equation}
where $\mS = \bracks{\mW^\top\otimes\tilde{\mA}}\in\R^{nd\times nd}$ and $\vx^{(0)} = \tvec\bracks{\mX^{(k)}}\in\R^{nd}$. \Cref{prop:power} directly applies to \Cref{eq:linear_graph_conv} under the assumption that $\mS = \mW^\top\otimes\tilde{\mA}$ has a unique eigenvalue with maximal magnitude. Based on power iteration, $\tvec\bracks{\mX^{(k)}}$ results in a state that is dominated by the corresponding eigenvector of $\mS$.

While this applies to any linear operator $\mS$, the special form of $\mS$ in \Cref{eq:linear_graph_conv} allows us to define the property more precisely.
As $\mS = \mW^\top\otimes\tilde{\mA}$ is a Kronecker product of two matrices, its eigenvalues and eigenvectors have a specific form. 
In the following, we denote $\lambda_i^{(\mM)}$ as the $i$-th largest eigenvalue in magnitude of $\mM$ and $\mU^{(\mM)}_{:,i}$ as the corresponding eigenvector of matrix $\mM$.
Based on the mixed-product property of the Kronecker product and the matrix product, each eigenvector $\mU_{:,m}$ of $\mS$ is a Kronecker products of the eigenvectors of $\mW^\top$ and $\tilde{\mA}$, i.e., the eigenvectors are given by $\mU_{:,(i-1)n+j}^{(\mS)}=\mU_{:,i}^{(\mW^\top)}\otimes\mU_{:,j}^{(\tilde{\mA})}$ for $i\in[d]$ and $j\in[n]$~\parencite{horn1991topics}. The corresponding eigenvalue is given by the product of eigenvalues $\lambda_i^{(\mW^\top)}$ of $\mW^\top$ and $\lambda_j^{(\tilde{\mA})}$ of $\tilde{\mA}$ as $\lambda_{:,(i-1)n+j}^{(\mS)} = \lambda_i^{(\mW^\top)}\lambda_j^{\tilde{\mA}}$ for $i\in[d]$ and $j\in[n]$~\parencite{horn1991topics}.
Based on these additional properties of the considered $\mS$, power iteration shows that repeatedly applying such graph convolutions results in $\tvec(\mX^{(k)})$ becoming increasingly dominated by $\mU_{:,1}^{(\mS)} = \mU_{:,1}^{(\mW^\top)}\otimes\mU_{:,1}^{(\tilde{\mA})}$. While the insight by this Proposition is similar to Theorem~\ref{pr:asym_wfixed} and Theorem~\ref{pr:anya_anyw}, the relation to the well-known power iteration method gives it additional understanding and intuition. We formalize this property in the following corollary:

\begin{corollary}[Power Iteration with a Kronecker Product]
    \label{prop:power_kron}
    Let $\mS = \mW^\top \otimes\tilde{\mA}\in\R^{nd\times nd}$ for any $\mW\in\R^{d\times d}$ and $\tilde{\mA}\in\R^{n\times n}$ with an eigenvalue $|\lambda_1^{(\mS)}| > |\lambda_i^{(\mS)}|$ that is greater in magnitude than all other eigenvalues. Then, for almost every $\vx^{(0)}$,
    \begin{equation}
        \frac{\mS^k\vx^{(0)}}{\norm{\mS^k\vx^{(0)}}_2} = \beta_k\bracks{\mU_{:,1}^{(\mW^\top)}\otimes\mU_{:,1}^{(\tilde{\mA)}}} + \vm^{(k)}
    \end{equation}
    for $\vm^{(k)}$ with $\lim_{k\to\infty}\norm{\vm^{(k)}}_2 = 0$ and $\beta_k\in\{-1,1\}$ for every $k\in\mathbb{N}$.
\end{corollary}

 \begin{proof}
    Based on Theorem~\ref{prop:power}, we know that 
    \begin{equation}
        \frac{\mS^k\vx^{(0)}}{\norm{\mS^k\vx^{(0)}}_2} = \beta_k\mU_{:,1}^{(\mS)} + \vm^{(k)}
    \end{equation}
    for $\beta_k\in\{-1,1\}$ and some $\vm^{(k)}$ with $\lim_{k\to\infty}\norm{\vm^{(k)}}_2 = 0$.
    Based on the mixed-product property of the Kronecker product and the matrix product, the eigenvector $\mU_{:,1}^{(\mS)}$ is given by the Kronecker product $\mU_{:,1}^{(\mS)} = \mU_{:,1}^{(\mW^\top)} \otimes \mU_{:,1}^{(\tilde{\mA})}$ of eigenvectors of $\tilde{\mA}$ and $\mW^\top$.
    Substituting the definition of eigenvector $\mU_{:,1}^{(\mS)}$ concludes this corollary.
 \end{proof}

While \Cref{prop:power_kron} already establishes the connection to our previous findings, we additionally state \Cref{prop:power_kron} in matrix notation to further highlight the equivalence:

\begin{corollary}[Power Iteration with a Kronecker Product]
    \label{prop:power_kron_matrix}
    Let $\mX^{(k)} = \tilde{\mA}\mX^{(k-1)}\mW$ for $\mW\in\R^{d\times d}$ and $\tilde{\mA}\in\R^{n\times n}$ matrices with eigenvalues $|\lambda_1^{(\mW)}| > |\lambda_i^{(\mW)}|$ and $|\lambda_1^{(\tilde{\mA})}| > |\lambda_i^{(\tilde{\mA})}|$ that are greater in magnitude than all other eigenvalues of $\mW$ and $\tilde{\mA}$, respectively.  Then,
    \begin{equation}
        \frac{\mX^{(k)}}{\norm{\mX^{(k)}}_F} = \beta_k\mU_{:,1}^{(\tilde{\mA)}}\bracks{\mU_{:,1}^{(\mW)}}^\top + \mM^{(k)}
    \end{equation}
    for almost every $\mX^{(0)}\in\R^{n\times d}$ and $\beta_k\in\{-1,1\}$ and $\mM^{(k)}\in\mathbb{R}^{n\times d}$ for every $k\in\mathbb{N}$ with $\lim_{l\to\infty}\norm{\mM^{(l)}}_2 = 0$.
\end{corollary}

Despite \Cref{prop:power_kron} and \Cref{prop:power_kron_matrix} considering a simplified case of graph convolutions and message-passing methods, they highlight the underlying property behind over-smoothing and more general phenomena. As the aggregation function $\tilde{\mA}$ is fixed, its dominant eigenvector $\mU_{:,1}^{(\tilde{\mA}}$ is also predefined. The dominance of this vector across each feature channel cannot be prevented by the feature transformation $\mW$.
Thus, the learnable parameters given by $\mW$ have a limited effect on the resulting representations.
As feature channels become increasingly similar and contain more redundant information, additional feature channels and parameters similarly have a limited effect.
For specific choices of the aggregation function $\tilde{\mA}$, we know in advance to which vector each feature channel becomes more similar with increased depth. For $\mA_\tsym$, we have $\mU_{:,1} = \mD^{1/2}\mathbf{1}$, resulting in each feature channel becoming increasingly similar to this degree-proportional state. For $\mA_\trw$ or other row-stochastic matrices, we have $\mU^{(\mA_\trw)} = \mathbf{1}$, resulting in each feature channel becoming increasingly similar to the constant vector. As such, over-smoothing can be observed as a special case of this phenomenon.  

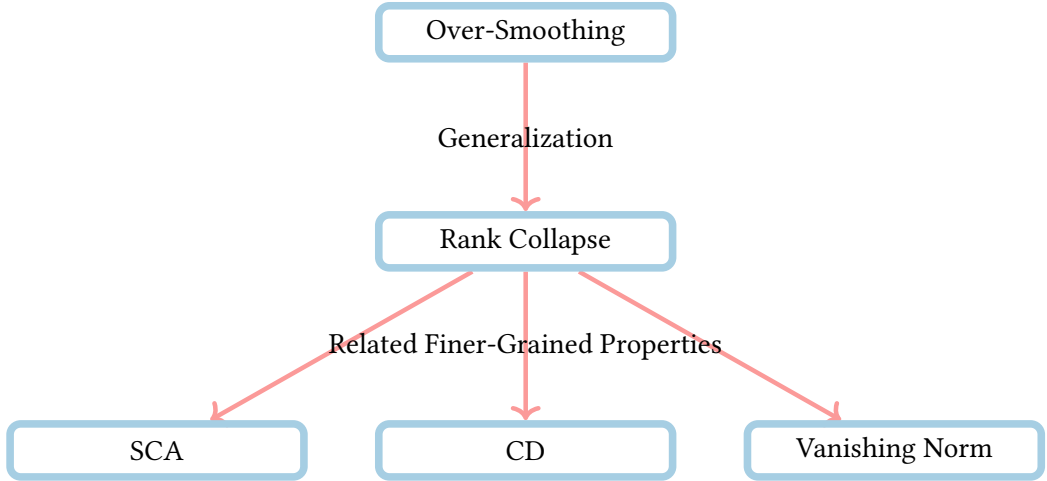
\begin{figure*}
\definecolor{vibrantpink}{HTML}{fb9a99} 
\centering
\begin{tikzpicture}[x=\textwidth/4, y=\textwidth/4]
\definecolor{vibrantblue}{HTML}{a6cee3} 
\definecolor{vibrantgreen}{HTML}{b2df8a} 
\definecolor{vibrantpink}{HTML}{fb9a99} 
\definecolor{vibrantorange}{HTML}{fdbf6f} 
\definecolor{vibrantpurple}{HTML}{cab2d6} 
\definecolor{vibrantsand}{HTML}{ffff99} 
\usetikzlibrary{shapes}
\pgfdeclarelayer{background}
\pgfsetlayers{background,main}
\tikzset{
dot/.style = {circle, minimum size=#1,
              inner sep=0pt, outer sep=0pt, draw=black},
dot/.default =13pt  
}
    \tikzstyle{box} = [draw, draw=vibrantblue, thick,rounded corners,line width=1.8pt]

    \def\yconv{1.6}
    \def\ysiso{-0.2}
    \def\yover{0.7}
    \def\ymethods{-1.1}
    \def\xshift{0.5}
    \def\xout{1.4}
    \def\xpoly{0.0}
    \def\yshift{0.8}
    \def\boxheight{20}
    \def\boxwidth{110}

    \node[rectangle,draw=vibrantblue,minimum height=\boxheight,minimum width=\boxwidth, rounded corners, line width=1mm] at (\ymethods,\xpoly+\yshift) (os) {};
    \node[scale=1.0,align=center] at (\ymethods,\xpoly+\yshift) {Over-Smoothing};  

    \node[scale=1.0,align=center] at (\ymethods,\xpoly+0.4) {Generalization};

    \node[rectangle,draw=vibrantblue,minimum height=\boxheight,minimum width=\boxwidth, rounded corners, line width=1mm] at (\ymethods,\xpoly) (rc) {};  
    \node[scale=1.0,align=center] at (\ymethods,\xpoly) {Rank Collapse}; 

    \node[scale=1.0,align=center] at (\ymethods,\xpoly-0.4) {Related Finer-Grained Properties}; 

    \node[rectangle,draw=vibrantblue,minimum height=\boxheight,minimum width=\boxwidth, rounded corners, line width=1mm] at (\ymethods-\xout,\xpoly-\yshift) (sca) {};
    \node[scale=1.0,align=center] at (\ymethods-\xout,\xpoly-\yshift) {SCA};   

    \node[rectangle,draw=vibrantblue,minimum height=\boxheight,minimum width=\boxwidth, rounded corners, line width=1mm] at (\ymethods,\xpoly-\yshift) (cd) {};
    \node[scale=1.0,align=center] at (\ymethods,\xpoly-\yshift) {CD}; 

    \node[rectangle,draw=vibrantblue,minimum height=\boxheight,minimum width=\boxwidth, rounded corners, line width=1mm] at (\ymethods+\xout,\xpoly-\yshift) (vn) {};
    \node[scale=1.0,align=center] at (\ymethods+\xout,\xpoly-\yshift) {Vanishing Norm}; 

    \begin{pgfonlayer}{background} 
    \draw[vibrantpink,->,line width=1.7pt] (os) to node[sloped,above,color=black,scale=1.0,rotate=90] {} (rc);
    \draw[vibrantpink,->,line width=1.7pt] (rc) to node[sloped,above,color=black,scale=1.0,rotate=90] {} (sca); 
    \draw[vibrantpink,->,line width=1.7pt] (rc) to node[sloped,above,color=black,scale=1.0,rotate=90] {} (cd); 
    \draw[vibrantpink,->,line width=1.7pt] (rc) to node[sloped,above,color=black,scale=1.0,rotate=90] {} (vn); 
    \end{pgfonlayer}
    
\end{tikzpicture}
\caption[Connections between our identified phenomena and properties.]{The connections between our identified phenomena and properties.}
\label{fig:3:summary}
\end{figure*}

\section{Summary}
\label{sec:3:summary}
We now summarize our extensive findings from this section, introduce formal definitions, and relate our findings to a broader context. 
While many variations of MPNNs exist, we consider iterative update functions, for which each iteration $k$ constructs an updated representation for each node $v_i$ by computing 
\begin{equation}
\label{eq:3:summ:node_wise}
    \mX_{i,:}^{(k)} = \sum_{v_j\in N_i} \tilde{\emA}_{i,j}\mX_{j,:}^{(k-1)}\mW^{(k)}\, ,
\end{equation}
where $\tilde{\mA}\in\R^{n\times n}$ encodes the graph structure and the utilized aggregation function as edge weights, $\mW^{(k)}\in\R^{d\times d}$ is an iteration-dependent feature transformation, and $\mX^{(k)}\in\R^{n\times d}$ is the matrix of node representations after iteration $k$. This form can represent various established methods, e.g., when the edge weights represent the mean or the sum aggregation. The edge weights may also be obtained by a more complex procedure, e.g., as a function of the node representations. We consider linearized iterations, in which we omit the activation function.

Based on our theoretical findings, we first define two properties that we have shown to hold for all message-passing methods of the form given in \Cref{eq:3:summ:node_wise}. These are shared component amplification (SCA), which occurs for each individual message-passing iteration, and component dominance, a phenomenon that occurs with increased depth. We then relate these properties to the observable phenomenon of rank collapse of node representations, with over-smoothing as a special case of rank collapse. We also propose to consider our identified vanishing norm phenomenon as a separate challenge. These connections are visualized in \Cref{fig:3:summary}.
Based on our findings, we also introduce the rank-one distance as a novel metric to quantify the rank collapse of node representations. While we show that all message-passing methods of the form given by \Cref{eq:3:summ:node_wise} exhibit SCA and CD, we also introduce a message-passing framework that does not have these properties.

For notational clarity, we will summarize our findings using an equivalent matrix notation of \Cref{eq:3:summ:node_wise} as
\begin{equation}
    \mX^{(k)} = \tilde{\mA}\mX^{(k-1)}\mW^{(k)}\, .
\end{equation}
For further notational simplicity, we use the equivalent vectorized form
\begin{equation}
\label{eq:summary:vect}
    \tvec\bracks{\mX^{(k)}} = \mT^{(k)}\tvec\bracks{\mX^{(k-1)}}\, ,
\end{equation}
where $\mT^{(k)} = \bracks{\mW^{(k)}}^\top\otimes\tilde{\mA}\in\R^{nd\times nd}$ is the Kronecker product $\otimes$ of matrices $\bracks{\mW^{(k)}}^\top$ and $\tilde{\mA}$. We start by defining SCA.

\subsection{Shared Component Amplification}

Following our findings from \Cref{pr:symm_iter}, \Cref{pr:jordan_dom}, and \Cref{pr:sca_svd}, we define SCA as follows:

\begin{definition}[Shared Component Amplification (SCA)]
\label{def:sca}
Let $\gM$ be a set of matrices with $\mM\in\gM$ satisfying $\mM\in\R^{nd\times nd}$.
    We say that $\gM$ exhibits \textbf{shared component amplification (SCA)} if there exist vectors $\vp_{(i)}\in\R^{n}$ for $i\in[n]$ where the set $\{\vp_{(1)},\dots,\vp_{(n)}\}$ spans $\R^n$, and there exist constants $a_1,\dots,a_n\in\R$ satisfying
    \begin{equation}
        \frac{\norm{\mM\bracks{\mI_d\otimes\vp_{(i)}}}_F}{\norm{\mM\bracks{\mI_d\otimes\vp_{(j)}}}_F} = \frac{a_i}{a_j}
    \end{equation}
    for all $i,j\in[n]$ and all matrices $\mM\in\gM$.
\end{definition}

To provide an informal explanation for this definition, we first note that by construction of matrices $\mI_d\otimes\vp_{(i)}\in\R^{nd\times d}$, any linear combination $\tvec(\mX) = \bracks{\mI_d\otimes\vp_{(i)}}\vc$ for $\vc\in\R^d$ induces a rank-one matrix $\mX$.
Thus, when the set of matrices $\gM$ satisfies the SCA property, components corresponding to $\vp_{(i)}$ are amplified by a fixed relative amount across all feature channels, independently of the matrix $\mM\in\gM$.
Thus, it is not possible to amplify one component maximally for one of the feature channels while amplifying another component maximally for a different feature channel. Based on \Cref{pr:jordan_dom}, any iteration of the form given by \Cref{eq:3:summ:node_wise} with an arbitrary aggregation matrix $\tilde{\mA}$ and all linear feature transformation $\mW$ exhibits SCA, which we state here to align with \cref{def:sca}:

\begin{theorem}[Shared Component Amplification]
    \label{pr:sca}
    Let $\tilde{\mA}\in\R^{n\times n}$ be any matrix and $\gM = \left\{\,\mW\otimes\tilde{\mA}\mid \mW \in \R^{d\times d}\,\right\}$ be a set of matrices. Then, $\gM$ exhibits shared component amplification.
\end{theorem}

As \Cref{pr:sca} is equivalent to \Cref{pr:jordan_dom}, we refer to the proof provided there. Central to our proof is the property of a matrix $\mM = \mW\otimes\tilde{\mA}$ that each eigenvector $\vu^{(\mM)}$ of $\mM$ is a Kronecker product $\vu^{(\mM)} = \vu^{(\mW)}\otimes\vu^{(\tilde{\mA})}$ of eigenvectors $\vu^{(\mW)}$ of $\mW$ and $\vu^{(\tilde{\mA})}$ of $\tilde{\mA}$. Such eigenvectors are vectorized rank-one matrices $\vu^{(\mM)} = \tvec\bracks{\vu^{(\tilde{\mA})}\bracks{\vu^{(\mW)}}^\top}$. For singular vectors of $\mM$, this is equivalent. As $\mM$ amplifies components corresponding to eigenvectors and singular vectors, $\mM$ amplifies these components sharedly across all feature channels. 

This confirms that whenever a message-passing iteration of the form given by \Cref{eq:3:summ:node_wise} is applied, components get amplified with a fixed ratio for all feature channels and all feature transformation $\mW$. This severely limits the potential of such MPNNs, as more than one feature channel leads to redundancy.
With \Cref{pr:symm_iter}, we considered the special case of setting the aggregation matrix $\tilde{\mA} = \mA_\tsym$ as the symmetrically normalized adjacency matrix $\mA_\tsym$. In this case, the eigenvectors of $\mA_\tsym$ can be selected as vectors $\vp_{(i)}$, and the components corresponding to each eigenvector get relatively amplified by the ratio of their respective eigenvalues. Consequently, the eigenvector $\mU_{:,1} = \mD^{1/2}\mathbf{1}$ gets maximally amplified for every feature channel, resulting in a smoothing of representations. 
Based on \Cref{pr:sca_svd}, we found that for arbitrary aggregation matrices $\tilde{\mA}$, the right singular vectors can be used as vectors $\vp_{(i)}$. Thus, the component of the representation matrix corresponding to the maximal singular value is maximally amplified across all feature channels and for all learnable parameters.

Thus, MPNNs of the form given in \Cref{eq:3:summ:node_wise} are inherently limited. A single message-passing iteration cannot amplify distinct components across feature channels, which holds true for any choice of learnable parameters $\mW$. This also limits the effect of the optimization process and of additional parameters or feature channels.

\subsection{Component Dominance}

In addition to SCA occurring for each individual message-passing iteration of the form given in \Cref{eq:3:summ:node_wise}, we also identified a related property for repeatedly applying such iterations. Based on our findings from \Cref{pr:symm_iterated}, \Cref{pr:basis_limit}, and \Cref{pr:aany_cd}, we define CD as follows:

\begin{definition}[Component Dominance (CD)]
\label{def:component_dominance}
    A sequence of matrices $\bracks{\mM^{(i)}}_{i\in\mathbb{N}}$, satisfying $\mM^{(k)}\in\R^{m\times m}$ for any $m\in\mathbb{N}$, exhibits \textbf{component dominance (CD)} if there exist matrices $\mP_{(1)}\in\R^{m\times m_1},\dots,\mP_{(p)}\in\R^{m\times m_p}$ with $p>1$ and $m_1+\dots+m_p = m$ such that the columns of $\mP_{(1)},\dots,\mP_{(p)}$ together span $\R^m$ and
    \begin{equation}
        \lim_{l\to\infty}\frac{\norm{\mM^{(l)}\dots\mM^{(1)}\mP_{(i)}}_F}{\norm{\mM^{(l)}\dots\mM^{(1)}\mP_{(1)}}_F} = 
        \begin{cases}
            1 & \text{if } i = 1, \\
            0 & \text{if } i \ne 1
        \end{cases}
    \end{equation}
    for all $i\in[p]$.
\end{definition}

When a sequence of transformations satisfies this property, the component corresponding to $\mP_{(1)}$ gets increasingly amplified compared to the amplification of all other components. We found that any MPNN of the form given in \Cref{eq:3:summ:node_wise} exhibits CD:

\begin{theorem}[Component Dominance]
\label{pr:cd}
    Let $\bracks{\mM^{(i)}}_{i\in\mathbb{N}}$ be a sequence of matrices of the form 
    \[\mM^{(i)} = \bracks{\mW^{(i)}}^\top\otimes\tilde{\mA}\,,\] where each $\mW^{(i)}\in\R^{d\times d}$ and $\tilde{\mA}\in\R^{n\times n}$ is a fixed matrix whose eigenvalues do not all have the same magnitude. Then, the sequence  $\bracks{\mM^{(i)}}_{i\in\mathbb{N}}$ exhibits component dominance.
\end{theorem}

\Cref{pr:cd} is equivalent to \Cref{pr:aany_cd}, whose proof we refer to for further details.
Our finding confirms that MPNNs of the form given in \Cref{eq:3:summ:node_wise} dominantly amplify a single component while relatively filtering out all other components.
When using $\mA_\tsym$ as the aggregation matrix, we have shown in \Cref{pr:symm_iterated} and \Cref{pr:basis_limit} that the component corresponding to the dominant eigenvector $\mD^{1/2}\mathbf{1}$ of $\mA_\tsym$ gets dominantly amplified across all feature channels.
As shown in \Cref{pr:aany_cd}, for an arbitrary aggregation matrix $\tilde{\mA}$, 
$\mP_{(1)}$ is of the form $\mP_{(1)} = \mI_d\otimes\mU_{(1)}$ where $\mU_{(1)}$ contains all generalized eigenvectors of $\tilde{\mA}$ with an eigenvalue of maximal magnitude. Due to the Kronecker product, this dominant amplification of eigenvectors $\mU_{(1)}$ is shared across all feature channels, similarly to SCA. 
In \Cref{sec:understanding:rank:power}, we have related this phenomenon to the classical power iteration method~\parencite{kowalewski1909einfuehrung,mises1929praktische}. From power iteration, it is similarly known that repeatedly applying a matrix $\mM$ to a vector results in the dominance of the component corresponding to the dominant eigenvector of $\mM$. In our case, all eigenvectors of $\mM$ are given as a Kronecker product of eigenvectors of $\mW$ and $\tilde{\mA}$.

\subsection{Rank Collapse}
With SCA and CD, we describe properties of any message-passing method of the form given in \Cref{eq:3:summ:node_wise} utilizing any aggregation matrix $\tilde{\mA}$, which may represent any graph structure. We do not directly observe the identified amplification of components, rather we observe node representations $\mX^{(k)}$ that get iteratively updated by these message-passing functions. 
However, SCA and CD have direct consequences on these node representations. 
When the message-passing steps satisfy both the SCA and CD properties, the dominantly amplified component identified by CD is constrained by the maximally amplified subspace induced by SCA. 
As a consequence, node representations $\mX^{(k)}$ converge to a low-rank subspace, following our findings from \Cref{pr:asym_wfixed}, \Cref{pr:anya_anyw}, \Cref{pr:study:rank_collapse}. We define rank collapse as the case when node representations $\mX^{(k)}$ become increasingly close to a rank-one state as the number of iterations $k$ increases:

\begin{definition}[Rank Collapse]
\label{def:rank_collapse}
    A sequence of matrices $\bracks{\mX^{(k)}}_{k\in\mathbb{N}}$ exhibits \textbf{rank collapse} if there exists a sequence of rank-one matrices $\bracks{\mY^{(k)}}_{k\in\mathbb{N}}$ such that
    \begin{equation}
        \lim_{k\to\infty} \norm{\frac{\mX^{(k)}}{\norm{\mX^{(k)}}_F} - \mY^{(k)}}_F = 0\, .
    \end{equation}
\end{definition}

With \Cref{pr:study:rank_collapse}, we have shown that rank collapse occurs for any message-passing method given by \Cref{eq:3:summ:node_wise} under slight assumptions:

\begin{theorem}[Rank collapse in probability for MPNNs.]
\label{pr:summ:rank_collapse}
Let $\mX^{(k)} = \tilde{\mA}\mX^{(k-1)}\mW^{(k)}$ for any matrix $\tilde{\mA}\in\R^{n\times n}$ with a dominant eigenvalue $|\lambda_1| > |\lambda_i|$ for all $i>1$, and for a sequence of matrices $\bracks{\mW^{(k)}}_{k\in\mathbb{N}}$ with $\mW^{(k)}\in\R^{d\times d}$. Then, for any initial $\mX^{(0)}\in[-m,m]^{n\times d}$, where $m\in\R$, the sequence $\bracks{\mX^{(k)}}_{k\in\mathbb{N}}$ exhibits rank collapse in probability with respect to the Lebesgue measure over $\mX^{(0)}$. Moreover, the rank-one matrices $\mY^{(k)}$ are of the form $\mY^{(k)} = \mU_{:,1}\vn_{(k)}^\top$, where $\mU_{:,1}$ is the eigenvector of $\tilde{\mA}$ corresponding to eigenvalue $\lambda_1$, and $\vn_{(k)}\in\R^d$ is an arbitrary vector for $k\in\mathbb{N}$.
\end{theorem}

\Cref{pr:summ:rank_collapse} is equivalent to \Cref{pr:study:rank_collapse} and we refer to the provided proof there. The condition $|\lambda_1|>|\lambda_i|$ for all $i>1$ is satisfied for almost every aggregation matrix $\tilde{\mA}$. As such, whenever an MPNN of the form in \Cref{eq:3:summ:node_wise} is utilized, the collapse of node representations $\mX^{(k)}$ to a rank-one matrix typically occurs. Our insights also highlight the importance of considering normalized node representations, as the increasing dominance of a single component across all feature channels may otherwise go undetected.
Specifically, based on the proof, all columns of these rank-one matrices are scaled versions of the same dominant eigenvector of $\tilde{\mA}$, induced by the dominant amplification of this eigenvector (CD). 
We have further related this phenomenon to a special case of power iteration~\parencite{kowalewski1909einfuehrung,mises1929praktische} in \Cref{sec:understanding:rank:power}, providing additional understanding. As \Cref{pr:summ:rank_collapse} holds for iteration-dependent feature transformations $\mW^{(k)}$, it is slightly more general.

\subsection{Over-Smoothing}
Based on \Cref{sec:explaining}, we have found that over-smoothing occurs as a special case of rank collapse, when the dominant eigenvector $\mU_{:,1}$ of $\tilde{\mA}$ has a particular form. Thus, it can also be seen as the message-passing steps exhibiting SCA and CD, with additional constraints on the specific form of the subspaces.
Over-smoothing is typically considered a convergence to a smooth subspace, where either all feature channels become constant vectors or degree-proportional vectors.
The dominant eigenvector of $\tilde{\mA}$ is equal to one of these vectors for established aggregation matrices, e.g., for the symmetrically normalized adjacency matrix $\mA_{\tsym}$, it is $\mD^{1/2}\mathbf{1}\in \R^n$, and for row-stochastic matrices it is $\mathbf{1}\in\R^n$. As this behavior was related to Laplacian smoothing in the work of \textcite{li2018deeper}, it is commonly referred to as over-smoothing. Based on our findings, we define over-smoothing as a special case of rank collapse:

\begin{definition}[Over-Smoothing]
\sloppy    A sequence of matrices $\bracks{\mX^{(k)}}_{k\in\mathbb{N}}$ exhibits \textbf{over-smoothing} if it exhibits rank collapse and the sequence of rank-one matrices $(\mY^{(k)})_{k\in\mathbb{N}}$ is of the form $\mY^{(k)} = \mathbf{1}\bracks{\vu^{(k)}}^\top$ or $\mY^{(k)} = \mD^{1/2}\mathbf{1}\bracks{\vu^{(k)}}^\top$ for any sequence of vectors $\bracks{\vu^{(k)}}_{k\in\mathbb{N}}$.
\end{definition}

For example, in \Cref{pr:asym_wany}, we have provided the novel insight that over-smoothing occurs for the GCN without non-linear activation functions and arbitrary feature transformations:

\begin{theorem}
\label{pr:3:summ:over}
Let $\mX^{(k)} = \mA_\tsym\mX^{(k-1)}\mW^{(k)}$ for a symmetrically normalized adjacency matrix $\mA_\tsym\in\R^{n\times n}$ and for a sequence of matrices $\bracks{\mW^{(k)}}_{k\in\mathbb{N}}$ with $\mW^{(k)}\in\R^{d\times d}$. Then, for any initial $\mX^{(0)}\in[-m,m]^{n\times d}$, where $m\in\R$, the sequence $\bracks{\mX^{(k)}}_{k\in\mathbb{N}}$ exhibits over-smoothing in probability with respect to the Lebesgue measure over $\mX^{(0)}$.
\end{theorem}

As \Cref{pr:3:summ:over} is equivalent to \Cref{pr:asym_wany}, we refer to the proof provided there.
In \Cref{sec:3:over-smoothing_conseq}, we also relate this property to the Dirichlet energy as a commonly employed metric to analyse over-smoothing empirically and theoretically. As $\mL_\tsym\mD^{1/2}\mathbf{1} = \mathbf{0}\in\R^n$, when the Dirichlet energy $E\bracks{\frac{\mX^{(k)}}{\norm{\mX^{(k)}}_F}}$ converges to zero as $k\to\infty$, the sequence of node representations exhibits over-smoothing. Similarly, we have $\mL\mathbf{1}=\mathbf{0}\in\R^n$, so when the Dirichlet energy $E_\mL$ computed with the unnormalized graph Laplacian $\mL$ converges to zero, over-smoothing also occurs. 
With \Cref{pr:gat_smooth}, we have also shown that attention scores with iteration-dependent aggregation matrices $\tilde{\mA}^{(k)}$ similarly exhibit over-smoothing.
Some previous works defined over-smoothing solely as convergence to a constant state. We have shown in \Cref{sec:understanding:zero} that unnormalized node representations $\mX^{(k)}$ converging to the zero matrix as $k\to\infty$ have been interpreted as convergence to a constant state. In contrast, the normalized node representations converge to a non-constant valued state. In the following, we further summarize our arguments for separating over-smoothing and rank collapse from the norm of the representations.

\subsection{Vanishing Norm}
In \Cref{sec:understanding:zero}, we found that the norm of node representations may converge to zero when repeatedly applying message-passing steps, depending on their parameters. In contrast, SCA and CD are properties that describe the relative behavior of different components and their amplification. Consequently, we also defined rank collapse and over-smoothing independently of the norm of the node representations.
However, the vanishing norm phenomenon may still significantly affect the performance of MPNNs.
Therefore, we argue that the vanishing norm phenomenon warrants further study, albeit separately from SCA, CD, rank collapse, and over-smoothing. By studying these phenomena separately, research efforts can communicate their considered challenge more precisely, develop targeted solutions, and investigate the importance of each fine-grained property more clearly.
We formally define vanishing norm as the following property:

\begin{definition}[Vanishing Norm]
    A sequence of matrices $\bracks{\mX^{(k)}}_{k\in\mathbb{N}}$ exhibits \textbf{vanishing norm} if 
    \begin{equation}
        \lim_{k\to\infty} \norm{\mX^{(k)}}_F = 0\, .
    \end{equation}
\end{definition}

Based on \Cref{prop:vanish}, we identified vanishing norm to occur in MPNNs when the learnable parameters across all iterations are sufficiently close to zero. The vanishing norm property is closely related to vanishing gradients, which have been a key challenge in MLPs and RNNs~\parencite{bengio1994learning,hochreiter1997long,pascanu2013on}. 

\subsection{Quantifying Rank Collapse and Over-Smoothing}
Based on the insights and definitions we have provided, we now introduce a novel metric for quantifying rank collapse, which generalizes over-smoothing. This will enable us to empirically analyze the effectiveness of further methods without the need to study them theoretically.
When using metrics to quantify over-smoothing, many instances of the more general phenomenon of rank collapse would remain undetected. 
Unifying them into a common metric to identify all methods leading to rank collapse, including all forms of over-smoothing, would be ideal. Based on our theoretical insights and \Cref{def:rank_collapse} on rank collapse, we propose to quantify the similarity of a sequence of states to rank-one matrices. The left and right singular vectors of the maximal singular value give the best rank-one approximation.
While we recommend computing this distance, we also propose a more efficient method when obtaining the left and right singular vectors is not computationally feasible. When $\mX$ is a rank-one matrix, any row and any column allow us to determine all values of $\mX$. Consequently, choosing any row and column of $\mX$ and forming a normalized rank-one matrix using their outer product and computing the distance to $\frac{\mX}{\norm{\mX}_2}$ approximates the similarity to a rank-one matrix. We refer to this metric as the rank-one distance (ROD) and define it as follows:

\begin{definition}[Rank-One Distance (ROD)]
\label{def:rod}
Let $\mX\in\R^{n\times d}$ be a matrix, $\vu = \mX_{:,i}$ a column of $\mX$ with $i = \argmax_{k} \norm{\mX_{:,k}}$ and $j = \argmax_{k} \norm{\mX_{k,:}}$ the row of $\mX$ with maximal norm. To account for the correct sign, we set $\vv = \mX_{j,:}$ if $\emX_{j,i} > 0$ and $\vv = - \mX_{j,:}$ otherwise. The rank-one distance (ROD) is defined as
    \begin{equation}
        \textrm{ROD}(\mX) = \norm{\frac{\mX}{\norm{\mX}_2} - \frac{\vu\vv^\top}{\norm{\vu\vv^\top}_2}}_*\, ,
    \end{equation}
    where $\norm{\cdot}_*$ is the nuclear norm.
\end{definition}

Our definition of ROD aligns well with \Cref{def:rank_collapse} for rank collapse. With this metric, we can identify methods that exhibit rank collapse without requiring a theoretical analysis. Similar to over-smoothing being a particular case of rank collapse, the Dirichlet energy of the normalized state converging to zero is a special case of the ROD converging to zero.

\subsection{Avoiding Rank Collapse and Over-Smoothing}
\label{sec:skp}
Based on our findings, rank collapse can be prevented by dealing with either SCA or CD. To recall the notation, we restate the vectorized form of our considered message-passing iterations given by \Cref{eq:3:summ:node_wise} as
\begin{equation}
\label{eq:skp_restate}
    \tvec\bracks{\mX^{(k)}} = \mT^{(k)}\tvec\bracks{\mX^{(k-1)}}
\end{equation}
where $\mT^{(k)} = \bracks{\mW^{(k)}}^\top\otimes\tilde{\mA}\in\R^{nd\times nd}$ for some $\mW^{(k)}\in\R^{d\times d}$ and $\tilde{\mA}\in\R^{n\times n}$. A central aspect of SCA in these MPNNs is that all eigenvectors and singular vectors of $\mT^{(k)}$ are given as Kronecker products of eigenvectors and singular vectors of $\mW^{(k)}$ and $\tilde{\mA}$. 
This property always holds when $\mT^{(k)}$ can be decomposed as a Kronecker product. To avoid SCA, it is necessary to use a transformation $\mT^{(k)}$ that is not decomposable into a single Kronecker product.
However, applying a dense matrix $\mT^{(k)}$ of shape $(nd \times nd)$ is computationally infeasible. We aim to remain within the framework of applying feature transformations and aggregation matrices. This can be achieved by noting that any matrix 
\begin{equation}
    \mT^{(k)} = \sum_{i=1}^{r}\mW_{(i)}^{(k)}\otimes\tilde{\mA}_{(i)}
\end{equation}
can be decomposed into a sum of Kronecker products (SKP)~\parencite{cao2021sum} for some $\mW_{(i)}^{(k)}\in\R^{d\times d}$ and $\tilde{\mA}_{(i)}\in\R^{n\times n}$ for all $i\in[r]$. For any matrix $\mT^{(k)}$, there exists a trivial decomposition into a sum of $r=d^2$ Kronecker products, where each $\mW_{(i)}$ has a single non-zero entry and $\tilde{\mA}_{(i)}$ contains the corresponding part of $\mT^{(k)}$.
Applying this choice to the node representations as in \Cref{eq:skp_restate}, i.e., $\mT^{(k)}\tvec\bracks{\mX^{(k-1)}}$, in the usual matrix form results in the iterative form
\begin{equation}
\label{eq:3:multi}
    \mX^{(k)} = \sum_{i=1}^{r} \tilde{\mA}_{(i)}\mX^{(k-1)}\mW_{(i)}^{(k)}\, .
\end{equation}
This SKP-based formulation allows us to avoid storing or applying $\mT^{(k)}\in\R^{nd\times nd}$ directly. Instead, the matrices $\tilde{\mA}_{(1)},\dots,\tilde{\mA}_{(r)}$ and $\mW_{(1)},\dots,\mW_{(r)}$ are stored and applied separately.
This formulation is capable of maximally amplifying distinct components in the node representations across different feature channels, avoiding SCA. 
When each aggregation matrix $\tilde{\mA}_{(i)}$ has different eigenvectors and singular vectors, the corresponding feature transformation $\mW_{(i)}^{(k)}$ controls the mixing of the amplified components for each feature channel. As such, for $r$ pairs of aggregation and feature transformation matrices, we can amplify at least $r$ different components across $r$ different feature channels. We formalize this in the following proposition:

\begin{proposition}[An SKP can avoid SCA]
\label{prop:kronecker_sum}
    \sloppy Let $\mV\in\R^{n\times d}$ with $\mathrm{rank}(\mV) = r$. Then, there exist matrices $\tilde{\mA}_{(1)},\dots,\tilde{\mA}_{(r)}\in\R^{n\times n}$ and $\mW_{(1)},\dots,\mW_{(r)}\in\R^{d\times d}$ such that for any matrix $\mP\in\R^{n\times d}$ with $\mP\neq p\cdot\mV$ for any $p\in\R$, and $\norm{\mP}_F = \norm{\mV}_F$, the following inequality holds:
    \begin{equation}
        \norm{\sum_{i=1}^r\tilde{\mA}_{(i)}\mV\mW_{(i)}}_F > \norm{\sum_{i=1}^r\tilde{\mA}_{(i)}\mP\mW_{(i)}}_F\, .
    \end{equation}
\end{proposition}

\begin{proof}
    Without loss of generality, assume that the first $r$ columns of $\mV$ are linearly independent.
    Define the matrices $\tilde{\mA}_{(i)} = \frac{\mV_{:,i}\mV_{:,i}^\top}{\norm{\mV_{:,i}\mV_{:,i}^\top}_F}$, which are symmetric and rank-one. Let $\mW_{(i)}$ be the matrix with a one at position $(i,i)$ and zeros at all other positions, i.e., $\bracks{\mW_{(i)}}_{i,i} = 1$.
    The matrix $\mS = \sum_{i=1}^r \bracks{\mW_{(i)}}^\top\otimes\tilde{\mA}_{(i)}$ has an eigenvector $\tvec(\mV)$ with corresponding eigenvalue $\lambda_1 = 1$, and all other eigenvalues of $\mS$ are zero.
    Using the equivalent vector norm, we thus have $\norm{\mS\tvec(\mV)}_2 = \lambda_1\norm{\mV}_F$ and $\norm{\mS\tvec(\mP)}_2 < \lambda_1\norm{\mP}_F$ for all vectors $\tvec(\mP)\neq p\tvec(\mV)$ for any $p\in\R$. The assumption $\norm{\mV}_F = \norm{\mP}_F$ concludes the proof.
\end{proof}

\Cref{prop:kronecker_sum} confirms that a sum of $r$ Kronecker products can maximally amplify any desired features of rank $r$.
This finding confirms that multiple terms effectively prevent SCA and, consequently, rank collapse and over-smoothing.  
Message-passing methods utilizing a single edge weight between any two nodes can also be seen as using a single computational graph $\tilde{\mA}$. In contrast, our proposed SKP corresponds to the use of multiple computational graphs $\tilde{\mA}_{(1)},\dots,\tilde{\mA}_{(r)}$. While each $\tilde{\mA}_{(i)}$ can be interpreted as a distinct graph with independent edges and weights, they can also represent multiple edge relation types over a shared node set. Such a combination of $\tilde{\mA}_{(1)},\dots,\tilde{\mA}_{(r)}$ is also referred to as a multigraph.
While each computational graph may utilize a distinct structure, a shared structure with different edge weights already suffices for amplifying distinct components.

With this formulation, each aggregation matrix $\tilde{\mA}_{(i)}$ can amplify distinct components, and the learnable parameters $\mW_{(i)}$ control how strongly these components are amplified for each feature channel. This enables more effective optimization, as the parameters have a greater impact on the resulting representations. 
Even with $r=2$ computational graphs, an SKP can amplify distinct components across feature channels, thereby mitigating SCA and rank collapse.

This finding provides a general framework for MPNNs that utilize an SKP, or equivalently, multiple computational graphs. This SKP-based framework is not tied to specific choices of aggregation functions, but rather establishes a general desirable property for MPNNs when the goal is to amplify distinct components across feature channels. Many choices for aggregation functions and feature transformations are possible within the SKP framework.
The SKP form is similar to multi-head attention, where each $\tilde{\mA}_{(i)}$ corresponds to an attention head. However, as these are softmax-activated, each $\tilde{\mA}_{(i)}$ is a row-stochastic matrix, which all have the all-ones vector $\mathbf{1}$ as their dominant eigenvector. Our proof requires distinct dominant eigenvectors so that each parameter matrix $\mW_{(i)}$ controls which components are maximally amplified in each feature channel.

This finding also reveals a connection between convolutional neural networks (CNNs)\linebreak[1] \parencite{fukushima1980neocognitron,lecun1989backpropagation} and the avoidance of SCA in message-passing. Convolution on a grid can be interpreted as a special case of an SKP with $r$ terms, where $r$ corresponds to the number of elements in the filter or kernel. Each relative connection in the convolution can be represented by a connectivity matrix $\tilde{\mA}_{(i)}$ over positions on the grid, with a corresponding distinct linear transformation $\mW_{(i)}$ applied for each position. This perspective helps explain why SCA — and, consequently, over-smoothing — is a phenomenon that arises specifically in the context of graph neural networks.

\begin{figure*}[tb]
         \centering
        \def\svgwidth{0.90\textwidth}
     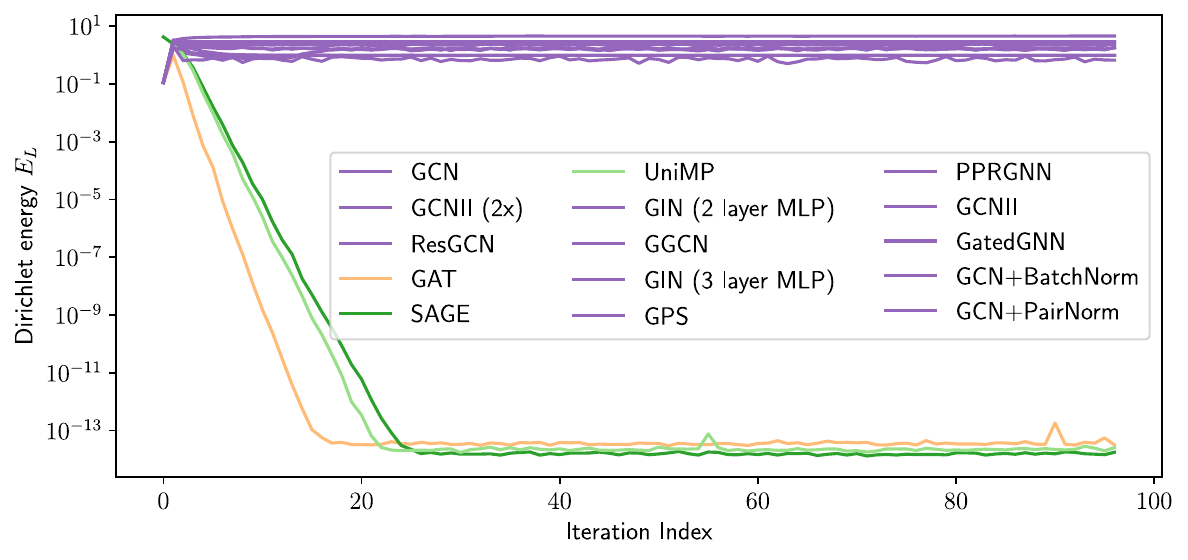
     \caption[Dirichlet energy using the unnormalized graph Laplacian.]{Dirichlet energy $E_{\mL}\bracks{\mX^{(k)}} = \trt\bracks{\bracks{\mX^{(k)}}^\top\mL\mX^{(k)}}$ using the unnormalized graph Laplacian $\mL$ across $96$ iterations of message passing for the KarateClub dataset~\parencite{zachary1977information} and $15$ methods. All parameters are randomly initialized, and visualized values are averaged over $50$ runs.}
      \label{fig:dirichlet_unnorm}     
\end{figure*}
\begin{figure*}[tb]
         \centering
        \def\svgwidth{0.95\textwidth}
     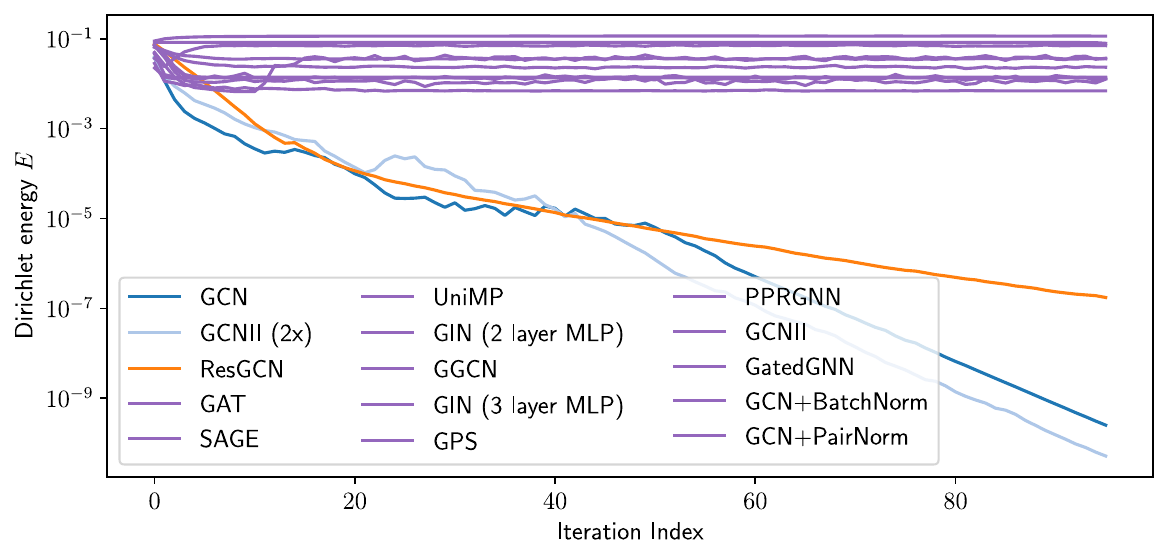
     \caption[Dirichlet energy using the symmetrically normalized graph Laplacian.]{Dirichlet energy $E\bracks{\mX^{(k)}} = \trt\bracks{\bracks{\mX^{(k)}}^\top\mL_\tsym\mX^{(k)}}$ using the symmetrically normalized graph Laplacian $\mL_\tsym$ across $96$ iterations of message passing for the KarateClub dataset~\parencite{zachary1977information} and $15$ methods. All parameters are randomly initialized, and visualized values are averaged over $50$ runs.}
      \label{fig:dirichlet_symm}     
\end{figure*}
\begin{figure*}[tb]
         \centering
        \def\svgwidth{0.95\textwidth}
         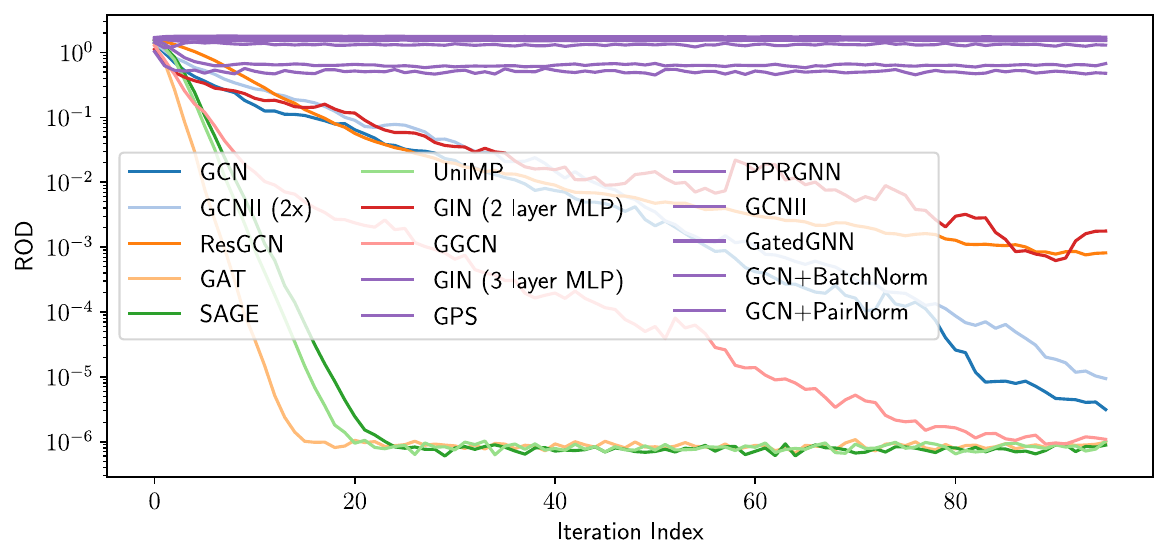
     \caption[Rank-one distance (ROD).]{Rank-one distance (ROD) across $96$ iterations for the KarateClub dataset~\parencite{zachary1977information} and $15$ methods. All parameters are randomly initialized, and visualized values are averaged over $50$ runs.}
      \label{fig:rank_one}     
\end{figure*}

\section{Evaluation}
\label{sec:3:eval}
We now empirically validate our theoretical insights and extend them to additional MPNNs and non-linear activation functions. 
Based on the definition of rank collapse in \Cref{def:rod} and the corresponding rank-one distance (ROD) metric, we first evaluate several existing methods in their ability to avoid rank collapse and compare this metric to two versions of the Dirichlet energy.
As the second part, we evaluate a sum of Kronecker products (SKP) within MPNNs, rather than a single Kronecker product, to evaluate its ability to approximate target functions for various benchmark tasks. 

\subsection{Quantifying Rank Collapse Using ROD}
\label{sec:3:exp:rod}
Based on our theoretical insights and definitions for rank collapse and over-smoothing, we introduced the rank-one distance (ROD) as a metric capturing a more general phenomenon than over-smoothing. 
Here, we compare the Dirichlet energy as a metric to quantify over-smoothing with the ROD. Instead of only indicating whether the node representations are similar to a rank-one matrix formed by specific columns, ROD determines the similarity to any rank-one matrix.
We evaluate these metrics for several established methods, allowing us to gain insights into the ability of those methods to avoid rank collapse or over-smoothing. We provide a reproducible implementation of these experiments~\footnote{\href{https://github.com/roth-andreas/simplifying-over-smoothing}{https://github.com/roth-andreas/simplifying-over-smoothing}}. These experiments are based on \parencite{roth2024simplifying}.

\paragraph{Experimental Setup}
We consider the update functions of various MPNNs, interleaved with a nonlinear activation function. In general, these are of the form
\begin{equation}
\label{eq:rod_eval}
    \mX^{(k)} = \phi\bracks{f\bracks{\mA,\mX^{(k-1)}}}
\end{equation}
where $\phi$ is set to be the ReLU activation function, and $f$ is a message-passing method, which we will detail in the following.
For the graph structure $\mA$, we use the small KarateClub dataset~\parencite{zachary1977information}. It consists of a single undirected graph with $n=34$ nodes and $156$ edges. Initial node features $\mX^{(0)}\in\R^{n\times d}$ are randomly initialized from a normal distribution with $\emX_{i,j}\sim \mathcal{N}(0,1)$.
We consider three metrics based on the normalized node representations $\frac{\mX^{(k)}}{\norm{\mX^{(k)}}_F}$: The Dirichlet energy 
\begin{equation}
    E_\mL\bracks{\frac{\mX^{(k)}}{\norm{\mX^{(k)}}_F}} = \frac{\trt\bracks{\bracks{\mX^{(k)}}^\top\mL\mX^{(k)}}}{\norm{\mX^{(k)}}_F^2}
\end{equation} 
utilizing the unnormalized graph Laplacian $\mL\in\R^{n\times n}$, which is zero when $\mX^{(k)} = \mathbf{1}\vu^\top$ for any $\vu\in\R^d$. As the second metric, we evaluate the Dirichlet energy 
\begin{equation}
    E\bracks{\frac{\mX^{(k)}}{\norm{\mX^{(k)}}_F}} = \frac{\trt\bracks{\bracks{\mX^{(k)}}^\top\mL_\tsym\mX^{(k)}}}{\norm{\mX^{(k)}}_F^2}
\end{equation} 
utilizing the symmetrically normalized graph Laplacian $\mL_\tsym$, which is zero when $\mX^{(k)} = \mD^{1/2}\mathbf{1}\vu^\top$ for any $\vu\in\R^d$. Lastly, the rank-one distance (ROD), which is zero for any rank-one matrix $\mX^{(k)} = \vv\vu^\top$ for any $\vv\in\R^n$ and $\vu\in\R^d$, as defined in \Cref{def:rod}.
We compute all metrics for $\mX^{(k)}$ for each $k\in[0,96]$. With more iterations, numerical instabilities prevented us from computing meaningful values.

\paragraph{Evaluated Models}
We consider several established MPNNs as function $f$ in \Cref{eq:rod_eval}.
The first set of methods utilizes the symmetrically normalized adjacency matrix $\mA_\tsym$ for aggregation. The dominant eigenvector of $\mA_\tsym$ is the degree-proportional vector. These are the graph convolutional network (GCN)~\parencite{kipf2017semi} and the GCN with the previous state added to the output using a residual connection~\parencite{he2016deep} as in the ResGCN~\parencite{kipf2017semi}. We also evaluate GCNII~\parencite{chen2020simple} that further adds initial residual connections that combine the initial state with the output of each iteration. As we observed differences depending on the magnitude of feature transformation, we further evaluate a variant in which all feature transformations are scaled by a factor of 2 (GCNII (2$\times$)). Additionally, we consider the personalized PageRank graph neural network (PPRGNN)~\parencite{roth2022transforming}, which proposes a weighting scheme on the initial residual connections that ensures convergence of node representations to a fixed point for an infinite number of iterations while maintaining a portion of the initial representations.

The second set of methods utilizes an averaging function as the aggregation function, potentially weighted using a learnable function. These are the graph attention network (GAT)~\parencite{velickovic2017graph}, and the SAGE convolution~\parencite{hamilton2017inductive}. The dominant eigenvector of matrices representing such an averaging aggregation is the constant vector.
As graph transformers similarly average over all nodes in a graph, we evaluate the unified message-passing model (UniMP)~\parencite{shi2021masked} and the general, powerful, and scalable graph transformer (GPS)~\parencite{rampasek2022recipe}, which combines the graph transformer framework with local message-passing and feature normalization.

We also evaluate several other methods that were specifically proposed to mitigate over-smoothing, regarding their ability to avoid the more general rank collapse. The generalized GCN (GGCN)~\parencite{yan2022two} allows for negative edge weights, which they have shown to reduce the smoothing effect between adjacent nodes.

Additionally, we consider adding normalization layers. We evaluate batch normalization~\parencite{ioffe2015batch} as a well-established normalization procedure and PairNorm~\parencite{zhao2020pairnorm} as a normalization that was specifically proposed to prevent over-smoothing. We combine both normalization techniques with the GCN to evaluate their ability to prevent rank collapse for a method known to suffer from over-smoothing.

As another established mechanism within MPNNs, we evaluate the ability of gating to prevent rank collapse. These use gating coefficients to control the mixing of aggregated features with the previous state. Here, we evaluate the gated graph neural network (GatedGNN)~\parencite{li2016gated}.

While the above methods apply linear feature transformation, the graph isomorphism network (GIN)~\parencite{xu2019how} employs a multi-layer perceptron (MLP), which is known to be a universal approximator~\parencite{hornik1989multilayer}. We evaluate two such instantiations: one that applies a two-layer MLP (GIN 2-layer MLP), and one that uses a three-layer MLP (GIN 3-layer MLP) as a feature transformation.

We use the standard implementation from Pytorch-Geometric~\parencite{fey2019fast} when available, or use the implementation provided with the proposed methods.
All learnable parameters are randomly initialized based on their provided implementations. We repeat this process for $50$ random seeds.

\paragraph{Results}
We present mean values for the Dirichlet energy $E_\mL\left(\frac{\mX^{(k)}}{\|\mX^{(k)}\|_F}\right)$ using the unnormalized graph Laplacian in \Cref{fig:dirichlet_unnorm}. We observe that this Dirichlet energy quickly decreases to small values for GAT, SAGE, and UniMP. After around $20$ iterations, the Dirichlet energy of these models stabilizes at around $10^{-13}$, which appears to be the machine precision. These three models utilize a (weighted) mean aggregation, which is known to have the constant vector as the dominant eigenvector. All other methods remain roughly constant throughout all iterations.

For comparison, we also present mean values for the Dirichlet energy $E\left(\frac{\mX^{(k)}}{\|\mX^{(k)}\|_F}\right)$ using the symmetrically normalized graph Laplacian in \Cref{fig:dirichlet_symm}. We observe this Dirichlet energy to decrease towards zero with an increased number of iterations for the GCN, ResGCN, and GCNII (2$\times$). These methods utilize the symmetrically normalized adjacency matrix as their aggregation function. While convergence to zero can be proven for GCN and ResGCN in the linear case, for GCNII (2$\times$), the influence of the constant initial features seemingly diminishes as the number of iterations increases. All other methods remain at a roughly constant Dirichlet energy.

As the third metric, we evaluate the ROD in \Cref{fig:rank_one}. The ROD decreases to zero when $\mX^{(k)}$ becomes closer to any rank-one matrix. We observe this to be the case for all methods, for which either one of $E_\mL\bracks{\frac{\mX^{(k)}}{\norm{\mX^{(k)}}_F}}$ or $E\bracks{\frac{\mX^{(k)}}{\norm{\mX^{(k)}}_F}}$ decreases to zero, as these are special cases. Additionally, we observe the ROD for the GIN (2-layer MLP) to approach zero. Since the sum aggregation has neither a constant nor a degree-proportional vector as its dominant eigenvector, the representations appear to converge to a different vector that is shared across all feature channels. The GIN utilizing a 3-layer MLP as a feature transformation successfully avoids this rank-one state.
Additionally, we observe that the node representations for the GGCN, which allows for negative edge weights, also converge to a rank-one matrix. While negative edge weight can cause the dominant eigenvector to also be significantly different, GGCN still utilizes a single computational graph, exhibits SCA, and amplifies the same signal component across all feature channels.

Methods that avoid rank collapse include both normalization methods, GCN+BatchNorm and GCN+PairNorm, as well as the methods GatedGNN, PPRGNN, GCNII, GPS, and GIN (3-layer MLP). These appear to be promising directions for addressing the underlying problem of rank collapse.

\subsection{The Effectiveness of a Sum of Kronecker Products}
\label{sec:validation}
\definecolor{color1}{HTML}{404788}
\definecolor{color2}{HTML}{238A8D}
\definecolor{color3}{HTML}{55C667}
\definecolor{color4}{HTML}{FDE725}

\begin{figure}[tb]
     \centering
     \begin{subfigure}[b]{0.32\textwidth}
         \centering
        \def\svgwidth{\textwidth}
         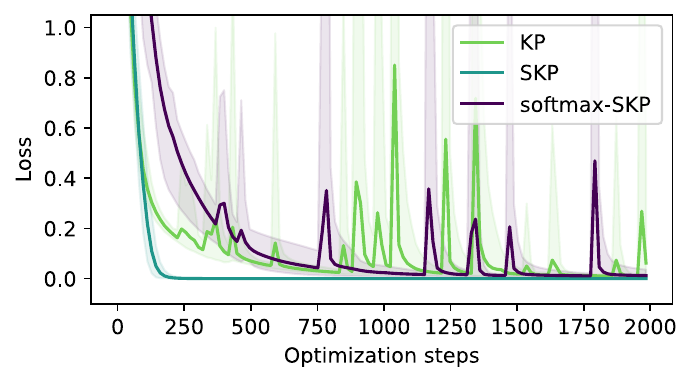
         \caption{Cora}
         \label{fig:constant_zero}
     \end{subfigure}
     \hfill
     \begin{subfigure}[b]{0.32\textwidth}
         \centering
        \def\svgwidth{\textwidth}
         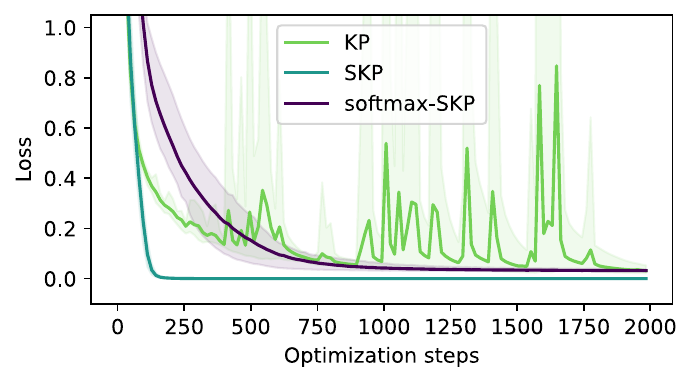
         \caption{Citeseer}
     \end{subfigure}
     \hfill
     \begin{subfigure}[b]{0.32\textwidth}
         \centering
        \def\svgwidth{\textwidth}
         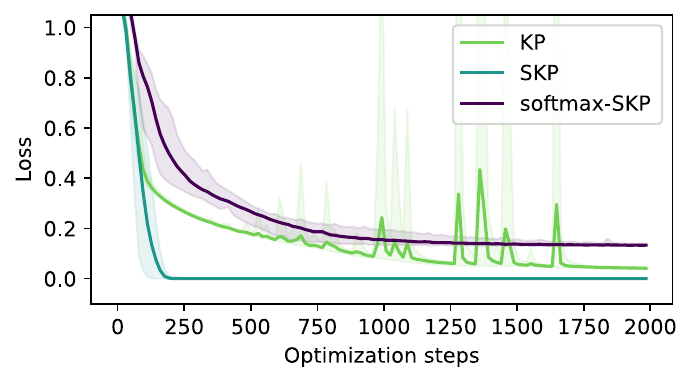
         \caption{Pubmed}
     \end{subfigure}
          \begin{subfigure}[b]{0.32\textwidth}
         \centering
        \def\svgwidth{\textwidth}
         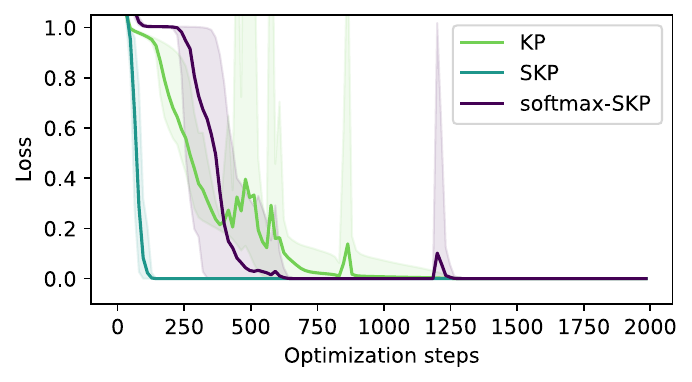
         \caption{Texas}
     \end{subfigure}
     \hfill
     \begin{subfigure}[b]{0.32\textwidth}
         \centering
        \def\svgwidth{\textwidth}
         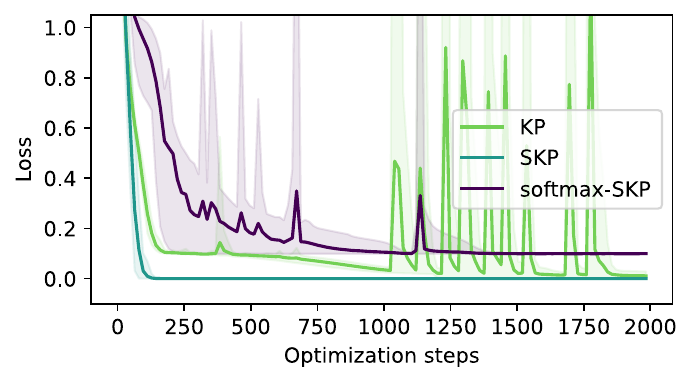
         \caption{Wisconsin}
     \end{subfigure}
     \hfill
     \begin{subfigure}[b]{0.32\textwidth}
         \centering
        \def\svgwidth{\textwidth}
         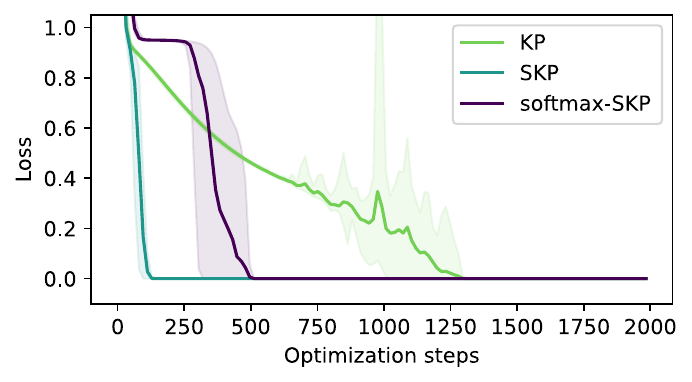
         \caption{Cornell}
     \end{subfigure}
          \begin{subfigure}[b]{0.32\textwidth}
         \centering
        \def\svgwidth{\textwidth}
         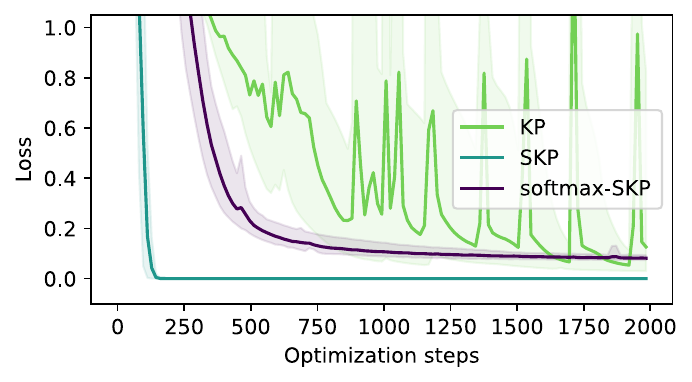
         \caption{Film}
     \end{subfigure}
     \hfill
     \begin{subfigure}[b]{0.32\textwidth}
         \centering
        \def\svgwidth{\textwidth}
         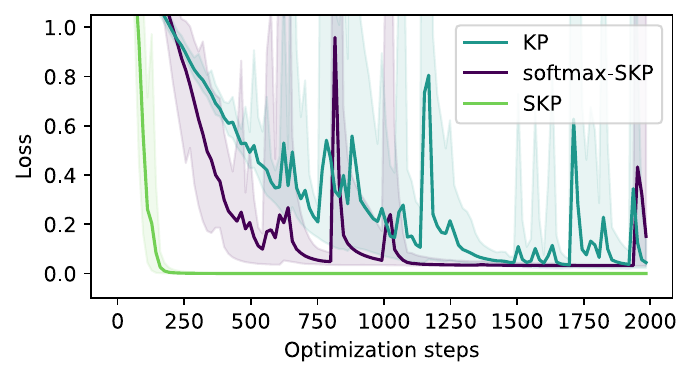
         \caption{Chameleon}
     \end{subfigure}
     \hfill
     \begin{subfigure}[b]{0.32\textwidth}
         \centering
        \def\svgwidth{\textwidth}
         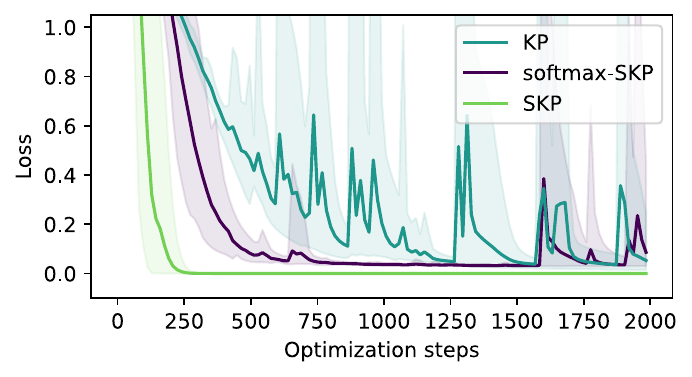
         \caption{Squirrel}
     \end{subfigure}
        \caption[Comparison of the training error during optimization.]{Training error during optimization for nine benchmark tasks for node classification when using all labels for optimization. The evaluated methods are a Kronecker product (KP), a sum of Kronecker products (SKP), and a sum of Kronecker products where each aggregation function is node-wise softmax-activated (softmax-SKP). All edge weights and feature transformations are directly learnable. Each method uses $l=8$ iterations and $d=32$ feature channels. Average values across $10$ runs are in bold, with the minimum and maximum values presented as a shaded area.}
        \label{fig:3:skp_results}
\end{figure}

As our theoretical findings indicate a limited effectiveness of learnable parameters for common forms of message passing, we now empirically investigate the optimization properties of several message-passing frameworks. For this, we consider nine common benchmark datasets for node classification and track the ability of different message-passing frameworks to fit the training data. Ideally, employed methods are sufficiently expressive in their ability to closely approximate desired functions. We provide a reproducible implementation of these experiments~\onlineresource{https://github.com/roth-andreas/rank_collapse}{https://github.com/roth-andreas/rank_collapse}. This set of experiments is based on \textcite{roth2023rank}.

\paragraph{Datasets}
We consider nine graphs for node classification. These are Cora, Citeseer, and Pubmed~\parencite{yang2016revisiting}, which are citation networks for which each node corresponds to a document, and a reference between documents is represented as an edge. The task is to classify each document based on a bag-of-words representation and the references. These are considered to be homophilic datasets, as referenced documents mostly share the same class~\parencite{yan2022two}.
The other six considered datasets are referred to as heterophilic datasets, as these contain fewer shared classes between neighboring nodes~\parencite{yan2022two}.
These consist of three web page graphs: Texas, Wisconsin, Cornell~\parencite{pei2020geom}, each of which represents web pages from one university and their links, as nodes and edges, respectively. The task is to classify each web page into one of the categories student, project, course, staff, and faculty~\parencite{pei2020geom}.  
The next two datasets are the Wikipedia graphs Chameleon~\parencite{rozemberczki2021multi} and Squirrel~\parencite{rozemberczki2021multi}. Similarly, nodes correspond to web pages from Wikipedia related to the corresponding topic, and edges correspond to their mutual links. The task is to categorize each web page based on its monthly traffic. 
Lastly, the actor co-occurrence dataset Film~\parencite{tang2009social} represents actors as nodes and their co-occurrence on Wikipedia web pages as edges. The task is to categorize the number of words on the Wikipedia page of the actor. For each dataset, we utilize the largest connected component.
For this experiment, we aim to approximate the target function and assume all node labels are available for optimization using the cross-entropy loss.

\paragraph{Experimental Setup}
Given one of these graphs with $n$ nodes and $c$ labels, we use the provided initial node representations $\mX\in\R^{n\times h}$, where $h$ is the input feature dimension of the corresponding task, and discrete node labels $\mY\in\{0,1\}^{n\times c}$ encoded as a one-hot vector for each node.
We first apply a linear feature transformation to obtain node representations $\mX^{(0)}\in\R^{n\times d}$ with the number of feature channels $d\in\mathbb{N}$ being a tunable parameter. We then update the node representations
\begin{equation}
\label{eq:sec3:eval}
    \tvec\bracks{\mX^{(k)}} = \phi\bracks{\mT^{(k)}\tvec\bracks{\mX^{(k-1)}}}
\end{equation}
using an iteration-dependent linear message-passing step $\mT^{(k)}$ and extend our theoretical insights by incorporating a non-linear activation function $\phi$. We introduce our different choices for $\mT^{(k)}$ in the following paragraph. After $l$ such iterations, each node representation $\mX^{(l)}_{i,:}$ is mapped to class predictions $\hat{\mY}\in[0,1]^{n\times c}$ using an affine transformation and a softmax activation. The cross-entropy loss between $\mY$ and $\hat{\mY}$ is computed and optimized for $\num{2000}$ steps using the Adam optimizer~\parencite{kingma2015adam} and a learning rate of $0.001$. For each framework chosen for $\mT^{(k)}$, this process is repeated for ten random initializations. 

\paragraph{Methods}
Based on our theoretical findings, we evaluate three different frameworks as the message-passing function $\mT^{(k)}$ in \Cref{eq:sec3:eval}: a single Kronecker product, a sum of Kronecker products, and a sum of Kronecker products for which each aggregation matrix is softmax activated. We extend these methods slightly from our theory to provide further insights by allowing for iteration-dependent aggregation matrices and a non-linear activation function after each iteration.

The single Kronecker product corresponds to the established framework that utilizes a single feature transformation $\mW^{(k)}\in\R^{d\times d}$ and a single aggregation matrix $\tilde{\mA}^{(k)}\in\R^{n\times n}$ for each iteration $k$, i.e., we have
\begin{equation}
    \mT^{(k)}_{\mathrm{KP}} = \mW^{(k)} \otimes\tilde{\mA}^{(k)}\, .
\end{equation}
Most of our theoretical insights have assumed that the message-passing is of this form, as it is used within many established approaches, such as GCN~\parencite{kipf2017semi} and GraphSAGE~\parencite{hamilton2017inductive}. It can also represent the sum aggregation~\parencite{xu2019how}, negative edge weights~\parencite{bo2021beyond}, and many more~\parencite{chien2021adaptive,yan2022two}. To cover all of these approaches in our experiments, we allow for all entries of the feature transformation $\mW^{(k)}$ and the weights of all given edges to be directly learnable for $\tilde{\mA}^{(k)}$ for each $k\in\mathbb{N}$. Based on our theoretical insights, a single Kronecker product exhibits shared component amplification, limiting the effectiveness of such approaches (see \Cref{sec:3:summary}). 

Next, we evaluate the effectiveness of our identified solution to shared component amplification using a sum of Kronecker products (see \Cref{sec:skp}). For this, we utilize $h=2$ terms and set
\begin{equation}
    \mT^{(k)}_{\mathrm{SKP}} = \sum_{i=1}^2\mW^{(k)}_{(i)}\otimes\tilde{\mA}_{(i)}^{(k)}
\end{equation}
where $\mW_{(i)}^{(k)}\in\R^{d\times d}$ and $\tilde{\mA}_{(i)}^{(k)}\in\R^{n\times n}$ for $i\in\{1,2\}$. As with $\mT_{\mathrm{KP}}^{(k)}$, the entries of $\tilde{\mA}_{(1)}^{(k)}$ and $\tilde{\mA}_{(2)}^{(k)}$ corresponding to existing edges are directly learnable. Based on \Cref{prop:kronecker_sum}, this form enables more effective feature transformations, as it allows for controlled mixing of amplified components across different feature channels.

The third considered framework is inspired by attention-based methods, such as the graph attention network (GAT)~\parencite{velickovic2017graph}, the TransformerConv~\parencite{shi2021masked}~\parencite{brody2022how}. The message-passing step is defined as
\begin{equation}
    \mT^{(k)}_{\mathrm{softmax-SKP}} = \sum_{i=1}^2\mW^{(k)}_{(i)}\otimes\tau\bracks{\tilde{\mA}_{(i)}^{(k)}}
\end{equation}
where $\tau$ applies a row-wise softmax activation function, as is used for attention-based aggregation functions. As with $\mT^{(k)}_{\mathrm{KP}}$ and $\mT^{(k)}_{\mathrm{SKP}}$, all edge weights in $\tilde{\mA}_{(1)}^{(k)}$ and $\tilde{\mA}_{(2)}^{(k)}$ are directly learnable.

\begin{figure}[tb]
     \centering
     \begin{subfigure}[b]{0.32\textwidth}
         \centering
        \def\svgwidth{\textwidth}
         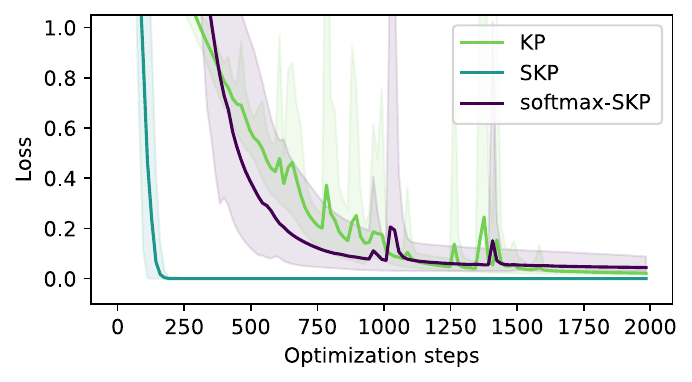
         \caption{$d=16$}
     \end{subfigure}
     \hfill
     \begin{subfigure}[b]{0.32\textwidth}
         \centering
        \def\svgwidth{\textwidth}
         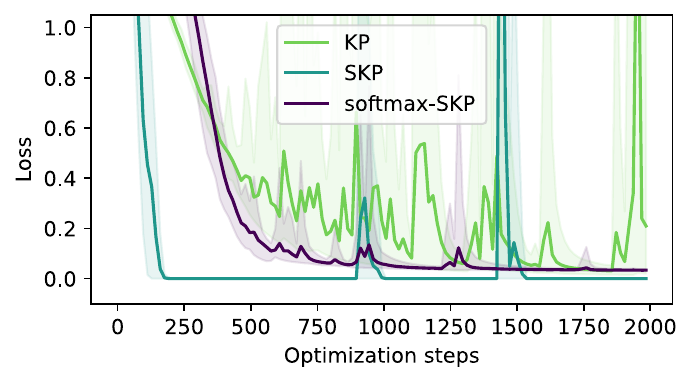
         \caption{$d=64$}
     \end{subfigure}
     \hfill
     \begin{subfigure}[b]{0.32\textwidth}
         \centering
        \def\svgwidth{\textwidth}
         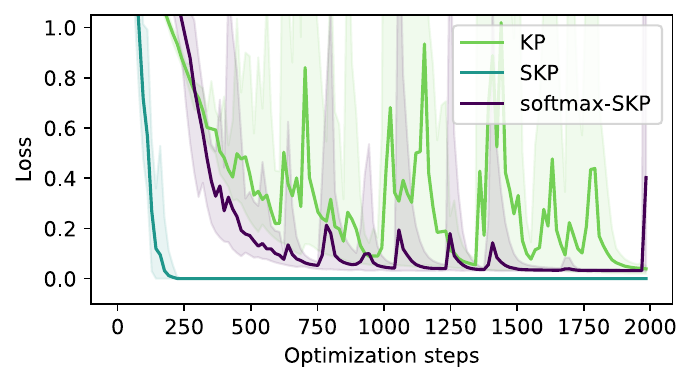
         \caption{$d=128$}
     \end{subfigure}
        \caption[Comparison of the training error during optimization for different numbers of parameters.]{Training error during optimization for the squirrel benchmark tasks when using all labels for optimization, as in \Cref{fig:3:skp_results}. The number of feature channels $d$ is varied in $d\in\{16,64,128\}$ and the number of iterations is set as $l=8$. Average values across $10$ runs are in bold, with the minimum and maximum values presented as a shaded area.}
        \label{fig:3:skp:parameters}
\end{figure}

\begin{figure}[tb]
     \centering
     \begin{subfigure}[b]{0.32\textwidth}
         \centering
        \def\svgwidth{\textwidth}
         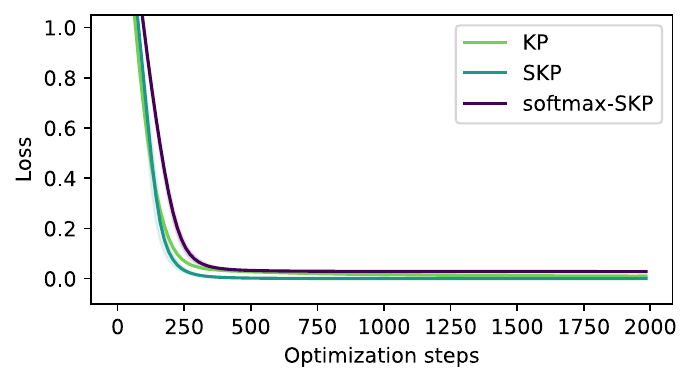
         \caption{$l=1$}
     \end{subfigure}
     \hfill
     \begin{subfigure}[b]{0.32\textwidth}
         \centering
        \def\svgwidth{\textwidth}
         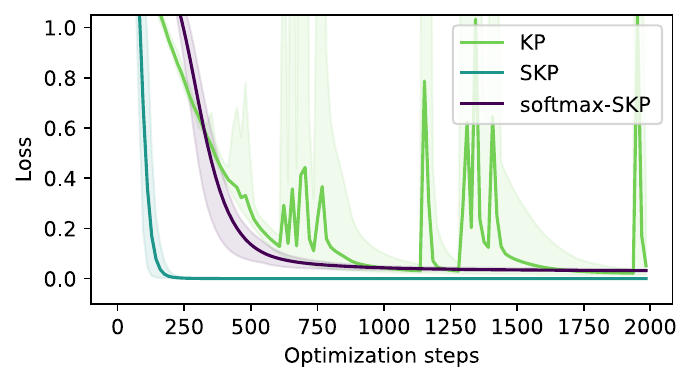
         \caption{$l=4$}
     \end{subfigure}
     \hfill
     \begin{subfigure}[b]{0.32\textwidth}
         \centering
        \def\svgwidth{\textwidth}
         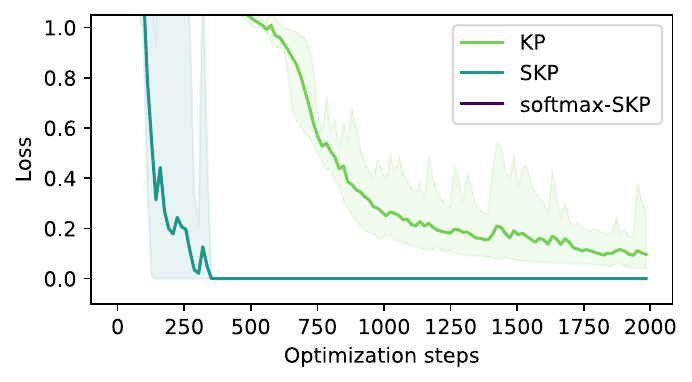
         \caption{$l=32$}
     \end{subfigure}
        \caption[Comparison of the training error during optimization for different numbers of iterations.]{Training error during optimization for the squirrel benchmark tasks when using all labels for optimization, as in \Cref{fig:3:skp_results}. The number of iterations $l$ is varied in $l\in\{1,4,32\}$ and the number of feature channels is set as $d=32$. Average values across $10$ runs are in bold, with the minimum and maximum values presented as a shaded area.}
        \label{fig:3:skp:layers}
\end{figure}

\begin{figure}[tb]
     \centering
     \begin{minipage}[t]{0.49\textwidth}
         \centering
        \def\svgwidth{\textwidth}
        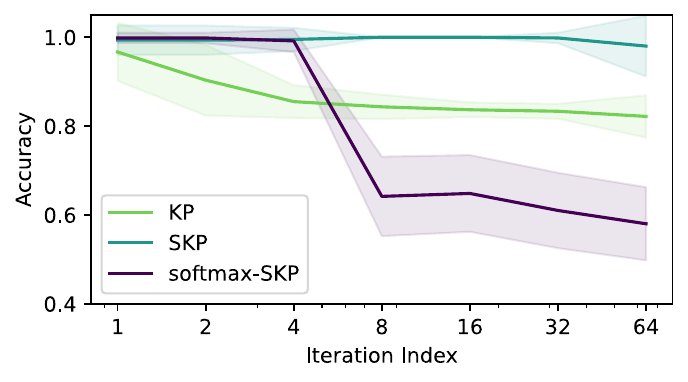
         \caption[Training accuracy for methods with a different number of iterations for the synthetic $3$-class classification task.]{Obtained training accuracy for methods with a number of iterations $l\in\{1,2,4,8,16,32,64\}$ for the synthetic $3$-class classification task with four nodes. Average accuracy over $10$ runs in bold, standard deviation as shaded area.}
         \label{fig:3:synt:acc}
     \end{minipage}
     \hfill
     \begin{minipage}[t]{0.49\textwidth}
         \centering
        \def\svgwidth{\textwidth}
        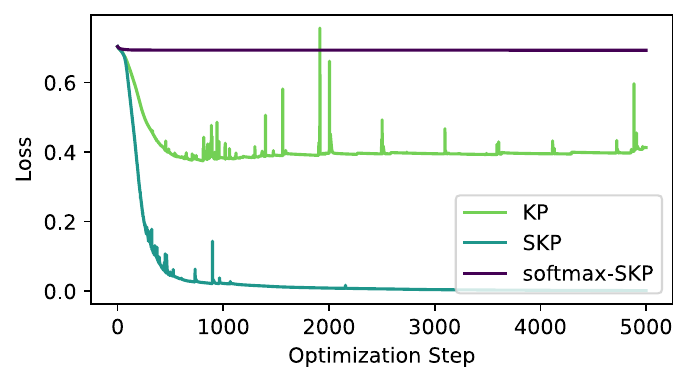
         \caption[Optimization error for the synthetic $3$-class classification task.]{Optimization error for $l=8$ iterations and the synthetic $3$-class classification task with four nodes. Averaged over 50 random seeds.}
         \label{fig:3:synt:loss}
     \end{minipage}
\end{figure}

\paragraph{Results}
In \Cref{fig:3:skp_results}, we visualize the training errors during optimization for all nine datasets, using $l=8$ message-passing iterations and $d=32$ feature channels. We observe that the SKP consistently achieves smooth error minimization, resulting in an error close to zero after fewer than 250 optimization steps for all datasets. For KP and softmax-SKP, the error minimization is slower in all cases and exhibits unstable behavior. For many datasets, the error after $2000$ optimization steps also remains higher than the obtained loss for SKP. The smoothness of the optimization and the convergence speed both indicate that optimizing the parameters within the SKP framework is more effective than for the KP framework or the softmax-SKP framework.

We additionally present an ablation study varying the number of feature channels $d\in\{16,64,128\}$ for the Squirrel dataset in \Cref{fig:3:skp:parameters}. While the optimization remains fast and smooth for the SKP in all cases, it remains slower and unstable for both the KP and softmax-SKP. Even when the SKP uses $d=16$ feature channels, the optimization is significantly better than the KP and softmax-SKP, even with $d=256$ feature channels. This aligns with our theoretical findings, as we identified that the KP amplifies components shared across all feature channels, making additional feature channels redundant.

In \Cref{fig:3:skp:layers}, we present an additional ablation study that varies the number of iterations. For a single iteration ($l=1$), the optimization remains smooth for all methods, but the SKP minimizes the error slightly faster and achieves a slightly lower error. However, the difference is rather small.
Aligning with our theoretical insights, the more iterations $l$ we employ, the more significant the difference becomes. While for $l=32$ iterations, SKP remains mostly smooth and achieves an error close to zero, the error for KP and softmax-SKP is increasingly unstable, and convergence is slower.

\subsubsection{Synthetic Dataset}
\label{sec:synthetic}
We additionally extend this experiment to a more challenging synthetic task.
When node representations collapse into a lower-dimensional subspace, it becomes more challenging to assign them to different classes. We begin with an observation about three nodes of a graph.
When their node representations collapse into a one-dimensional subspace, the representations are positioned at a specific point on a line. One node is located in the center, with each of the two other nodes positioned on opposite sides. A linear classifier cannot classify the two outer nodes in the same class while classifying the center node in a different class. Representations need to be in a higher-dimensional subspace for solving this task. Our synthetic task follows this observation. We consider a slightly more challenging version with four nodes and the multi-label classification task, where the labels are given as
\begin{equation}\mathbf{Y} = \begin{bmatrix}
    $1$ & $1$ & $1$ \\
    $1$ & $0$ & $0$ \\
    $0$ & $1$ & $0$ \\
    $0$ & $0$ & $1$
\end{bmatrix}.
\end{equation}
Each pair of two nodes shares the same label for one of the tasks. When node representations lie in a subspace of two or fewer dimensions, a linear decision boundary cannot correctly classify all nodes. Thus, this task is challenging for methods that yield node representations in a lower-dimensional subspace.

\paragraph{Experimental Setup}
Initial node representations $\mathbf{X}^{(0)}\in\R^{4\times 6}$ are sampled from a normal distribution with mean zero and standard deviation set to one. Aggregation matrices and feature transformation are randomly initialized from a normal distribution. As we consider a multi-label classification task, the predictions are obtained using a sigmoid activation, rather than the softmax activation used for node classification tasks. We consider methods with $l\in\{1,2,4,8,16,32,64\}$ iterations for message-passing. Optimization for each method and number of iterations is separately performed for $\num{5000}$ optimization steps using the Adam optimizer\parencite{kingma2015adam} with a learning rate of \num{0.001}. 
Each experiment was repeated for $50$ random seeds. The employed methods are the same as for the nine benchmark tasks, i.e., KP, SKP, and softmax-SKP.

\paragraph{Results}
We present the average accuracy for all three methods across the evaluated numbers of iterations in \Cref{fig:3:synt:acc}. While the SKP retains an accuracy of almost $100\%$ across all iterations, the KP degrades visibly already from the second iteration. The softmax-SKP retains an average accuracy of $100\%$ up to four iterations, and drops to around $70\%$ for more iterations. In \Cref{fig:3:synt:loss}, we also present the change in the optimization loss for $\num{5000}$ optimization steps when using eight iterations. All three methods exhibit significantly different behavior, with the loss of the SKP converging to a value close to zero. For the KP, the loss converges to a value around $0.3$. For the softmax-SKP, the loss remains constant at its initial value throughout optimization. These insights further confirm that optimization is ineffective for KP and softmax-SKP methods but significantly improved with SKP.
    \cleardoublepage
    \clearpage

\chapter{Preventing Shared Component Amplification With Multiple Computational Graphs}
\label{chap:4}
\textit{
Many established message-passing neural networks (MPNNs) exhibit shared component amplification (SCA), a phenomenon where applying a single message-passing operation amplifies the same component across all feature channels. 
This inherently limits the expressiveness of such message-passing steps and the utility of the learnable parameters.
Since any message-passing operation using a single computational graph exhibits the SCA property, we propose using multiple computational graphs within each message-passing step and present two methods for deriving such approaches. First, starting with existing MPNNs, we propose splitting the computational graph into multiple edge relations. Second, we define the spectral graph convolution for multiple feature channels and introduce a novel approximation for message-passing that naturally incorporates multiple computational graphs.
We prove that both approaches avoid SCA and demonstrate their effectiveness on downstream tasks.
}

\begin{figure*}[t]
\newcommand{\rectangle}[1]{%
  \begin{tikzpicture}[baseline=-0.5ex]
    \node[minimum size=0.6em, draw=#1, fill=#1, inner sep=0pt] {};
  \end{tikzpicture}%
}
\definecolor{vibrantpink}{HTML}{fb9a99} 
\definecolor{vibrantsand}{HTML}{ffff99} 
\definecolor{vibrantblue}{HTML}{a6cee3}

\centering
\begin{tikzpicture}[x=\textwidth/4, y=\textwidth/4]
\definecolor{vibrantblue}{HTML}{a6cee3} 
\definecolor{vibrantgreen}{HTML}{b2df8a} 
\definecolor{vibrantpink}{HTML}{fb9a99} 
\definecolor{vibrantorange}{HTML}{fdbf6f} 
\definecolor{vibrantpurple}{HTML}{cab2d6} 
\definecolor{vibrantsand}{HTML}{ffff99} 
\usetikzlibrary{shapes}
\pgfdeclarelayer{background}
\pgfsetlayers{background,main}
\tikzset{
dot/.style = {circle, minimum size=#1,
              inner sep=0pt, outer sep=0pt, draw=black},
dot/.default =13pt  
}
    \tikzstyle{box} = [draw, draw=vibrantgreen, thick,rounded corners,line width=1.8pt]

    \def\ysiso{0.7}
    \def\ymimo{-0.2}
    \def\xgc{-1.1}
    \def\xmp{1.00}
    
    \draw[box,draw=vibrantorange] (-2, 0.37) rectangle (1.53, 1.03); 
    \node at (-1.80, \ysiso) {SISO}; 
        
    \draw[box,draw=vibrantorange] (-2, -1.32) rectangle (1.53, 0.15); 
    \node at (-1.80, \ymimo-0.4) {MIMO}; 
    \node[scale=1.0] at (\xgc,\ymimo -0.4) {$=$};

    \draw[box] (\xgc -0.45, -1.4) rectangle (\xgc + 0.45, 1.25); 
    \node at (\xgc, 1.1) {Graph Convolution};
    \node[fill=vibrantsand,draw=vibrantsand,minimum width=3.5cm,minimum height=2cm,draw,ellipse] (scgc) at (\xgc,\ysiso) {};
    \node[scale=0.7] at (\xgc, \ysiso) {$\vtheta*\vx=\mU \mathrm{diag}(\vw)\mU^\top\vx$};
    \node[color=black] at (\xgc, \ysiso-0.12) {\scriptsize\parencite{bruna2014spectral}};

    \node[fill=vibrantsand,draw=vibrantsand,minimum width=3.5cm,minimum height=2cm,draw,ellipse] (mcgc) at (\xgc,\ymimo) {};
    \node[scale=0.7] at (\xgc, \ymimo) {$\displaystyle\mX^\prime_{:,q} = \sum_{p=1}^d\vtheta_{(p,q)}*\mX_{:,p}$};
    \node[fill=vibrantpink,draw=vibrantpink,minimum width=3.5cm,minimum height=2cm,draw,ellipse] (mimogc) at (\xgc,\ymimo-0.8) {};
    \node[scale=0.7] at (\xgc, \ymimo - 0.8) {$\displaystyle\tTheta*\mX = \sum_{k=1}^n \mA_{(k)}\mX\mW_{(k)}$};
    \node[color=black] at (\xgc, \ymimo - 0.96) {\scriptsize (\Cref{sec:uniformity:mimo-gc})};

    \draw[box] (\xmp - 0.45, -1.4) rectangle (\xmp + 0.45, 1.25);   
    \node at (\xmp, 1.1) {Message-Passing};
    \node[fill=vibrantsand,draw=vibrantsand,minimum width=3.5cm,minimum height=2cm,draw,ellipse] (scmpnn) at (\xmp,\ysiso) {};    
    \node[scale=0.7] at (\xmp, \ysiso) {$\vtheta * \vx \approx w \tilde{\mA}\vx$}; 

    \coordinate (x) at (\xmp,-0.15);
        \node[fill=vibrantsand,draw=vibrantsand,minimum width=3.5cm,minimum height=2cm,draw,ellipse] (a) at (\xmp,\ymimo) {}; 
    \node[scale=0.7] at (\xmp,\ymimo) {$\displaystyle\mX^\prime_{:,q} = \sum_{p=1}^dw_{(p,q)}\tilde{\mA}\mX_{:,p}$};

    \node[fill=vibrantpink,draw=vibrantpink,minimum width=3.5cm,draw,ellipse,minimum height=2cm] (lmcgc) at (\xmp,\ymimo-0.8) {};
    \node[scale=0.7] at (\xmp,\ymimo -0.8) {$\displaystyle\tTheta * \mX \approx \sum_{k=1}^K \tilde{\mA}_{(k)}\mX\mW_{(k)}$}; 

    \node [box, draw=vibrantblue, text width=3.8cm, scale=0.7, fill=vibrantblue] (sca) at (\xmp - 0.9, \ymimo) {Shared Component Amplification (SCA) for a single iteration.};


    \draw[->,line width=1.2pt] (scgc) to (scmpnn);
    \node[color=black] at ($(\xmp,\ysiso+0.05)!0.5!(\xgc,\ysiso+0.05)$){\scriptsize\parencite{kipf2017semi}};
    \draw[->,line width=1.2pt] (scmpnn) to (a);
    \node[color=black] at ($(\xmp,\ymimo)!0.5!(\xmp,\ysiso)$){\scriptsize\parencite{kipf2017semi}};

    \draw[vibrantpink,->,line width=1.8pt] (sca) to (lmcgc);
    \draw[vibrantpink,->,line width=1.8pt] (a) to (lmcgc);    
    \node[color=black] at ($(\xmp, \ymimo)!0.5!(\xmp,\ymimo -0.75)$){\scriptsize \Cref{sec:4:splitting}};


    \draw[black,->,line width=1.2pt] (scgc) to (mcgc);
    \node[color=black] at ($(\xgc,\ymimo)!0.5!(\xgc,\ysiso)$){\scriptsize\parencite{bruna2014spectral}};
    
    \draw[vibrantpink,draw=vibrantpink,->,line width=1.8pt] (mimogc) to node[above, color=black,scale=0.7] {}  (lmcgc);
    \node[color=black] at ($(\xgc,\ymimo -0.74)!0.5!(\xmp,\ymimo -0.74)$){\scriptsize \Cref{sec:uniformity:mimo-gc}};

\end{tikzpicture}
\caption[Connection between the contributions in this chapter and previous work.]{Connection between the contributions in this chapter, their connection to \Cref{sec:understanding}, and related work. Yellow parts (\protect\rectangle{vibrantsand}) indicate previous work, blue parts (\protect\rectangle{vibrantblue}) findings from \Cref{sec:understanding}, and pink parts (\protect\rectangle{vibrantpink}) indicate the contributions presented in this chapter.}
\label{fig:4:overview}
\end{figure*}
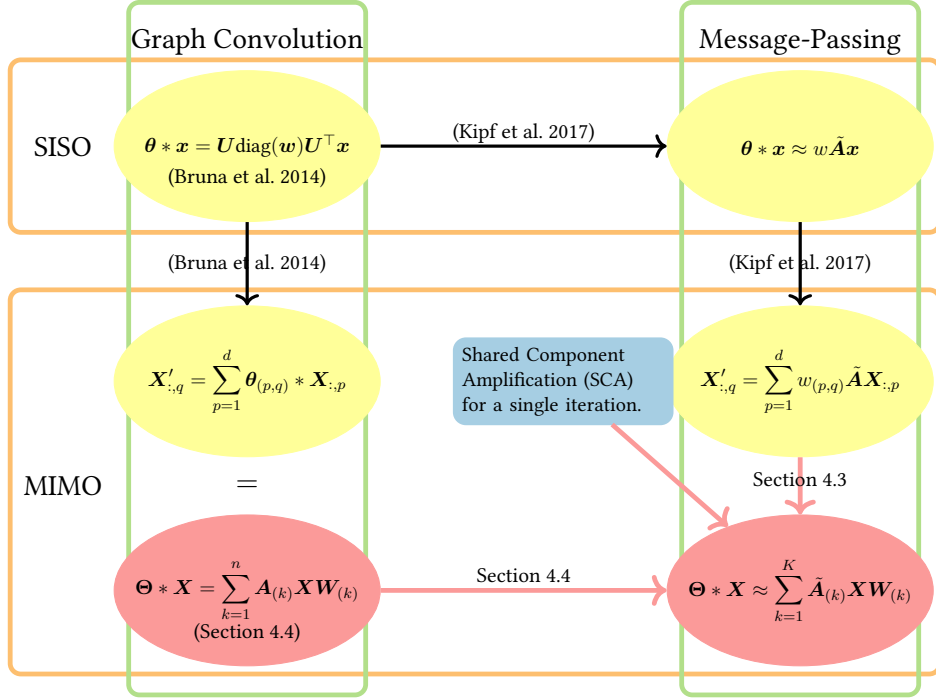

\section{Introduction}
While message-passing neural networks (MPNNs) have achieved promising results, their impact on real-world applications has been limited~\parencite{bechler2025position}. 
Various phenomena have been identified in the literature that are related to performance limitations~\parencite{arnaiz2025oversmoothing}. These include over-squashing~\parencite{alon2021on}, under-reaching~\parencite{alon2021on,sun2022position}, heterophily~\parencite{yan2022two}, over-smoothing~\parencite{li2018deeper,oono2020graph}, over-correlation~\parencite{jin2022feature}, and rank collapse~\parencite{roth2023rank}. However, due to an incomplete theoretical and empirical understanding of these phenomena, modifications lack theoretical grounding, and the relative importance of each phenomenon remains unclear.

We identified shared component amplification (SCA) as one theoretically derived property underlying the observed over-smoothing and rank collapse phenomena, as discussed in \Cref{sec:understanding}. SCA refers to the property that a single message-passing step amplifies the same component for each feature channel and any set of parameters. As a result, such message-passing steps cannot make some feature channels more similar, while simultaneously making others more dissimilar.
Unlike over-smoothing, which is typically considered in the iterated or limit case, we identified SCA as a property of a single message-passing operation in \Cref{def:sca}.
The property of SCA occurs whenever a single computational graph is used for message-passing, as shown in \Cref{pr:sca}. 
While repeatedly applying message-passing steps that exhibit SCA can lead to over-smoothing and rank collapse of the node representations, we focus on a single iteration in this chapter. 
As shown in \Cref{pr:sca}, a single computational graph always exhibits SCA, and using multiple computational graphs can prevent it, as shown in \Cref{prop:kronecker_sum}. We will present two frameworks for deriving and utilizing multiple computational graphs. These follow two of our previous publications (\parencite{roth2024preventing} and \parencite{roth2025what}). We provide a visual overview in \Cref{fig:4:overview}.

In Section~\ref{sec:4:splitting}, we introduce the multi-relational split (MRS) framework, with which we propose to decompose the computational graph of existing MPNNs into multiple edge relations. We propose to do this by assigning each edge to one of multiple relation types. This is motivated by the fact that messages between two nodes may serve a different goal. We prove that the MRS framework can effectively prevent SCA.

In Section~\ref{sec:mc-mpnns}, we present a novel derivation of message-passing methods. Instead of approximating MPNNs from the graph convolution in the single-input single-output (SISO) case, as was done with the GCN, we approximate MPNNs directly in the multi-input multi-output (MIMO) case. For that, we first derive the MIMO graph convolution (MIMO-GC) using the convolution theorem and graph Fourier transform. We find that in this multi-channel case, the MIMO-GC naturally utilizes multiple computational graphs, each amplifying a single distinct component. We then approximate the MIMO-GC as localized MIMO-GCs (LMGCs) by localizing the aggregation step. This serves as a general framework with strong theoretical foundations. This framework enables us to study the properties of message-passing methods in a fixed yet flexible form. We prove conditions under which LMGCs prevent SCA and further demonstrate that they allow for injectivity of each message-passing step while applying a linear function.

We empirically evaluate both approaches in \Cref{sec:4:eval}, presenting results for the MRS framework in \Cref{sec:4:eval:mrs} and for the LMGC framework in \Cref{sec:4:eval:lmgc}.

\section{Related Work}
We now provide an overview of previous work aimed at avoiding over-smoothing and rank collapse. Specifically, we present approaches related to the SCA property, which we are addressing in this chapter. Based on our proposed solution in \Cref{sec:skp} of utilizing multiple computational graphs, we also provide an overview of existing methods that change the computational graph for message-passing.

\paragraph{Solutions for Rank Collapse and Over-Smoothing}
Many different strategies for mitigating rank collapse and over-smoothing have been proposed. To retain information, various methods have been proposed to use residual connections either from the previous iteration~\parencite{bresson2017residual,li2020deepergcn,chen2020simple,li2021training,scholkemper2025residual} or the initial state~\parencite{klicpera2019predict,chen2020simple,gu2020implicit,roth2022transforming}. Other works have proposed combining the intermediate representations after each iteration into a joint final state~\parencite{xu2018representation,fey2019just}. Further methods propose to limit the effect of update steps using a gating mechanism~\parencite{bresson2017residual,rusch2023gradient,dwivedi2023benchmarking,finkelshtein2024cooperative}. Other approaches have shown the benefits of applying normalization techniques to node representations~\parencite{zhao2020pairnorm,li2020deepergcn,scholkemper2025residual}. \textcite{zhou2021dirichlet} propose to regularize the Dirichlet energy of the node representations to encourage obtaining less similar representations. 

Other lines of work have proposed methods that amplify higher-frequency components rather than low-frequency components using a residual connection~\parencite{eliasof2023improving,giovanni2023understanding} using negative edge weights~\parencite{bo2021beyond,yan2022two}, or combining multiple aggregation functions~\parencite{corso2020principal,tailor2022do,rosenbluth2023some}. \textcite{jin2022feature} minimize the correlation between different feature channels in the obtained node representations using regularization.

While these methods can effectively mitigate over-smoothing, they are designed for the multi-iteration message-passing setting. As SCA is a property of a single message-passing iteration, these approaches are not designed to deal with SCA. With our introduction of SCA (\Cref{def:sca}), we can develop targeted strategies to mitigate this property.

\paragraph{Changing the Computational Graph}
Various approaches have been proposed to modify the provided graph into a more effective computational graph for message passing. 
\textcite{alon2021on} identified that sparsely connected regions lead to the over-squashing of an exponential amount of information into a fixed-size vector. 
Several works have proposed changing the topology of a graph to reduce sparse connections between regions~\parencite{topping2022understanding,abboud2022shortest,barbero2024locality}. 
Other works have found that node representations become similar faster for densely connected regions or graphs~\parencite{rong2020dropedge,yan2022two}. They have proposed using a sparsified computational graph~\parencite{rong2020dropedge,yan2022two,nguyen2023revisiting,jamadandi2024spectral,rubio2025gnns}.
With Co-GNNs, \textcite{finkelshtein2024cooperative} propose to discard some of the edges during message passing dynamically. 

Several other approaches can be seen as using multiple computational graphs.
Some methods were proposed for multi-relational graphs~\parencite{schlichtkrull2018modeling,vashishth2020composition}. These works assume a given dataset for which multiple relation types are provided. \textcite{butler2023convolutional} propose multigraph neural networks that extend polynomial filters on graphs to given multigraphs. \textcite{suresh2021breaking} propose to add a second edge relation in addition to proximity that is based on the structural similarity of nodes and uses both edge relations for message-passing. \textcite{yang2020factorizable} propose to disentangle a graph into multiple factorized graphs, each capturing a distinct latent relation to obtain disentangled node representations. \textcite{giunchiglia2022towards} propose to utilize an explanability method and discard all representations and edges seen as task-irrelevant. Relatedly, \textcite{guo2024esgnn} propose the edge splitting GNN that separates available edges of a graph into task-relevant and task-irrelevant ones, producing two computational graphs. \textcite{luan2022revisiting} propose an adaptive channel mixing that use the adjacency matrix, the graph Laplacian, and a residual connection as three separate computational graphs to adaptively amplify high- and low-frequency components. \textcite{rossi2023edge} consider directed graphs and propose to use the graph with reverse edges as a second computational graph. \textcite{eliasof2024feature} propose to utilize distinct edge weights for each feature channel. While all of these works make valuable contributions to MPNNs, the general effect of multiple computational graphs and how they should be designed have not been studied extensively.

\section{Splitting The Computational Graph of MPNNs}
\label{sec:4:splitting}
As the first approach to utilizing multiple computational graphs within MPNNs, we now propose a framework that modifies existing MPNNs. The framework starts with any existing message-passing operation and assigns each edge to one of multiple relational types. Depending on the selected relation for an edge, messages are exchanged and transformed differently. This results in a multi-relational graph. This section is based on \textcite{roth2024preventing}.

\subsection{Multi-Relational Split MPNNs}
\label{sec:4:split:theory}
We consider an attributed graph $\gG=(\mA,\mX)$ in matrix form with adjacency matrix $\mA\in\R^{n\times n}$ and node representations $\mX\in\R^{n\times d}$. The corresponding set of nodes $\gV=\{v_1,\dots,v_n\}$, where $v_i$ corresponds to row $i$ of $\mX$, and the $i$-th row and column of $\mA$. We recall our definition of a message-passing step from \Cref{eq:2:message-passing}, which obtains an updated state for node $v_i$ as
\begin{equation}
\label{eq:4:split_mpnn}
    \bracks{\textrm{MPNN}(\mA,\mX)}_{i,:} = \phi\bracks{\mX_{i,:},\bigoplus\left\{\left\{\,\alpha_{(i,j)}\cdot\psi\bracks{\mX_{i,:},\mX_{j,:}}\mid v_j\in \gN_i\,\right\}\right\}}\, ,
\end{equation}
where $\psi,\phi\colon \R^{d} \times \R^d \to \R^d$ are two functions on tuples, $\oplus$ is a permutation invariant set aggregation function, and $\alpha_{(i,j)}\in\R$ is an edge-specific weighting scalar.

For splitting any such MPNN into a version that utilizes multiple computational graphs, we introduce an edge relation assignment function $f(v_i,v_j)\in [l]$ that assigns each edge $(v_i,v_j)$ of the graph to one of $l$ relation types. Depending on the relation type $k\in[l]$, a different initialization $\psi_k$ of the feature transformation $\psi$ is applied. In general, we define such multi-relational split MPNNs (MRS-MPNNs) as 

\begin{equation}
    \bracks{\textrm{MRS-MPNN}(\mA,\mX)}_{i,:} = \phi\bracks{\mX_{i,:},\bigoplus\left\{\left\{\,\alpha_{(i,j)}\cdot\psi_{f\bracks{v_i,v_j}}\bracks{\mX_{i,:},\mX_{j,:}}\mid v_j\in \gN_i\,\right\}\right\}}\, .
\end{equation}

Thus, any MPNN of the form given \Cref{eq:4:split_mpnn} can be directly converted into an MRS-MPNN by defining an edge relation assignment function $f$ and instantiating $l$ transformations $\psi_1,\dots,\psi_l$. Before analyzing some of the beneficial properties of MRS-MPNNs, we now provide several examples of established MPNNs and their corresponding MRS-MPNNs. We do not include non-linear activation functions in the following examples, as these are typically applied after applying a message-passing step.

\paragraph{GCN}
As one of the most commonly employed MPNNs, the graph convolutional network (GCN) applies a linear feature transformation $\mW$ and aggregates node representations using a symmetric degree normalization. Stated in our MRS framework, the MRS-MPNN can be expressed as
\begin{equation}
    \bracks{\textrm{MRS-GCN}(\mA,\mX)}_{i,:} = \sum_{j\in \gN_i} \frac{1}{\sqrt{d_i d_j}}\mW_{f\bracks{v_i,v_j}}\mX_{j,:}
\end{equation}
where $\mA_\tsym$ is the symmetrically normalized adjacency matrix and $\mW_{k}\in\R^{d\times d^\prime}$ are independent linear feature transformations for all $k\in[l]$. Apart from the selection of the feature transformation, all other parts remain unchanged from the GCN. Thus, setting all feature transformations $\mW_{k}$ to the same matrix would recover the original GCN.
This can equivalently be expressed in matrix notation as
\begin{equation}
\label{eq:4:mrs-gcn}
    \textrm{MRS-GCN}(\mA,\mX) = \sum_{k\in[l]}\tilde{\mA}_{(k)}\mX\mW_{k}\, ,
\end{equation}
where $\bracks{\tilde{\mA}_{(k)}}_{i,j} = \bracks{\mA_{\tsym}}_{i,j}$ when $f\bracks{v_i,v_j} = k$ and zero otherwise. \Cref{eq:4:mrs-gcn} highlights the form that utilizes multiple computational graphs, which can avoid SCA, as outlined in \Cref{sec:skp}.

\paragraph{SAGE} 
Similarly, the SAGE convolution~\parencite{hamilton2017inductive} can be expressed in the MRS framework as
\begin{equation}
    \textrm{MRS-SAGE}(\mA,\mX) = \mX\mV + \sum_{k\in[l]}\tilde{\mA}_{(k)} \mX\mW_{(k)}\, ,
\end{equation}
where $\bracks{\tilde{\mA}_{(k)}}_{i,j} = (\mA_\trw)_{i,j}$ when $f\bracks{v_i,v_j} = k$ and zero otherwise, similar to the MRS-GCN.

\paragraph{GAT} 
The MRS framework can also be applied to adaptively obtained edge weights and multiple attention heads, as used in the graph attention network (GAT)~\parencite{velickovic2017graph}. For $H\in\mathbb{N}$ attention heads, the MRS-GAT is defined as
\begin{equation}
    \textrm{MRS-GAT}(\mA,\mX) = \sum_{h\in[H]}\sum_{k\in[l]}\tilde{\mA}^{(h)}_{(k)}\mX\mW^{(h)}_{(k)}\, ,
\end{equation}
where $\tilde{\mA}^{(h)}\in\R^{n\times n}$ is the matrix containing the attention coefficients of head $h\in[H]$, and $\bracks{\tilde{\mA}^{(h)}_{(k)}}_{i,j} = \bracks{\tilde{\mA}^{(h)}}_{i,j}$ when $f\bracks{v_i,v_j} = k$ and zero otherwise.

\paragraph{GIN} 
The graph isomorphism network (GIN)~\parencite{xu2019how} is an established method that enables the injectivity of a message-passing step by utilizing a non-linear multi-layer perceptron (MLP) $\omega\colon \R^d\to\R^d$ as its feature transformation. Converted with the MRS framework, this leads to the update step
\begin{equation}
    \textrm{MRS-GIN}(\mA,\mX) = \sum_{k\in[l]}\tilde{\mA}_{(k)} \omega_{(k)}(\mX)\, ,
\end{equation}
where $\omega_{(k)}\colon\R^d\to\R^d$ is a node-wise (i.e., row-wise) MLP for $k\in[l]$, and $\bracks{\tilde{\mA}_{(k)}}_{i,j} = (\mA)_{i,j}$ when $f\bracks{v_i,v_j} = k$ and zero otherwise.

\paragraph{GatedGCN} 
The GatedGCN~\parencite{dwivedi2023benchmarking} extends the GCN by a gating mechanism that allows for incoming messages to be discarded. The corresponding MRS adaptations are given by
\begin{equation}
    \bracks{\textrm{MRS-GatedGCN}(\mA,\mX)}_{i,:} = \bracks{\mX\mV}_{i,:} + \sum_{j\in \gN_i}\frac{e_{(i,j)}\odot\mX_{j,:}\mW_{f\bracks{v_i,v_j}}}{\epsilon + \sum_{k\in \gN_i}e_{(i,k)}}
\end{equation}
where $e_{(i,j)}\in\R^d$ is an obtained vector of edge attributes, and $\mV,\mW_{(1)},\dots,\mW_{(l)}\in\R^{d\times d}$.

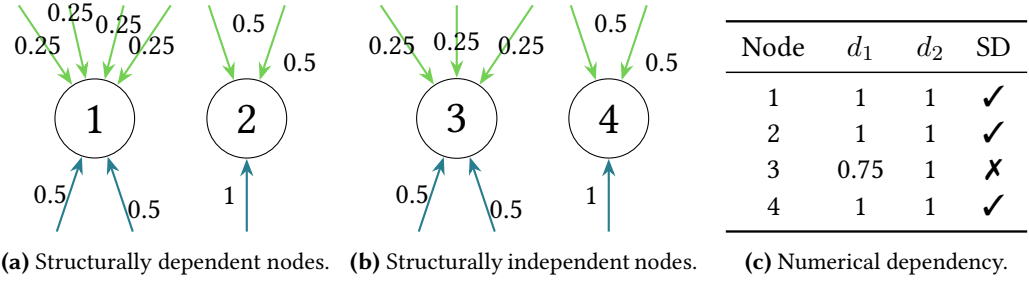
\begin{figure*}
\definecolor{viridis1}{RGB}{72,21,103}
\definecolor{viridis2}{RGB}{40,125,142}
\definecolor{viridis3}{RGB}{115,208,85}
     \begin{subfigure}[b]{0.33\textwidth}
         \centering
         
         \begin{tikzpicture}
\begin{scope}[every node/.style={circle,draw,scale=1.6}]
    \node (D) at (1.5,1.5) {1} ;
    \node (E) at (3.5,1.5) {2} ;
\end{scope}

\begin{scope}[>={Stealth[viridis3]},
              every edge/.style={draw=viridis3,thick}]
    \path [->] (0.5,3.0) edge node[xshift=-1mm,pos=0.5] {0.25} (D); 
    \path [->] (1.16,3.0) edge node[xshift=0mm,pos=0.1] {0.25} (D); 
    \path [->] (1.88,3.0) edge node[xshift=0mm,pos=0.25] {0.25} (D); 
    \path [->] (2.5,3.0) edge node[xshift=1mm,pos=0.5] {0.25} (D); 
    \path [->] (3.0,3.0) edge["0.5"] (E); 
    \path [->] (4.0,3.0) edge["0.5"] (E); 
\end{scope}
\begin{scope}[>={Stealth[viridis2]},
              every edge/.style={draw=viridis2,thick}]
    \path [->] (1.0,0.0) edge node[xshift=-2.5mm,pos=0.5] {0.5} (D); 
    \path [->] (2.0,0.0) edge node[xshift=2.5mm,pos=0.3] {0.5} (D);
    \path [->] (3.5,0.0) edge["1"] (E); 
\end{scope}
\end{tikzpicture}
        
         \caption{Structurally dependent nodes.}
     \end{subfigure}
     \hfill
\begin{subfigure}[b]{0.33\textwidth}
         \centering
         
         \begin{tikzpicture}
\begin{scope}[every node/.style={circle,draw,scale=1.6}]
    \node (D) at (1.5,1.5) {3} ;
    \node (E) at (3.5,1.5) {4} ;
\end{scope}

\begin{scope}[>={Stealth[viridis3]},
              every edge/.style={draw=viridis3,thick}]
    \path [->] (0.5,3.0) edge node[xshift=-2mm] {0.25} (D); 
    \path [->] (1.5,3.0) edge node[xshift=0mm] {0.25} (D); 
    \path [->] (2.5,3.0) edge node[xshift=2mm] {0.25} (D); 
    \path [->] (3.0,3.0) edge["0.5"] (E); 
    \path [->] (4.0,3.0) edge["0.5"] (E); 
\end{scope}
\begin{scope}[>={Stealth[viridis2]},
              every edge/.style={draw=viridis2,thick}]
    \path [->] (1.0,0.0) edge node[xshift=-2.5mm,pos=0.5] {0.5} (D); 
    \path [->] (2.0,0.0) edge node[xshift=2.5mm,pos=0.3] {0.5} (D);
    \path [->] (3.5,0.0) edge["1"] (E); 
\end{scope}
\end{tikzpicture}
        
         \caption{Structurally independent nodes.}
     \end{subfigure}
     \hfill
\begin{subfigure}[b]{0.32\textwidth}
         \centering
         
         \begin{tabular}{cccc} \toprule
  Node & $d_1$ & $d_2$ & SD \\ \midrule
  1 & 1 & 1 & \ding{51} \\
  2 & 1 & 1 & \ding{51} \\
  3 & 0.75 & 1 & \ding{55} \\
  4 & 1 & 1 & \ding{51} \\
  \bottomrule
  \end{tabular}
        
         \caption{Numerical dependency.}
     \end{subfigure}
        \caption[Structurally dependent and independent nodes.]{Exemplary visualization of structurally dependent (a) and structurally independent (b) nodes for two computational graphs. Arrows in each color represent the incoming edges for each computational graph. The numbers beside the arrows indicate the edge weights. The numerical dependency (c) indicates the weighted in-degrees of the four considered nodes, and SD indicates whether each node is structurally dependent to node $1$.}
        \label{fig:struct_independ}
\end{figure*}

\paragraph{Theoretical Properties of MRS-MPNNs}

We now analyze the beneficial properties of MRS-MPNNs. For this analysis, we consider MRS-MPNNs of the form 
\begin{equation}
    \mX^\prime = \sum_{k\in[l]} \tilde{\mA}_{(k)}\mX\mW_{(k)}
\end{equation}
where $\tilde{\mA}_{(k)}\in\R^{n\times n}$ can represent any graph structure and edge weights, and $\mW_{(k)}\in\R^{d\times d}$ is a linear feature transformation for all $k\in[l]$. This corresponds to multiple computational graphs, as proposed in \Cref{sec:skp}. While we have shown in \Cref{prop:kronecker_sum} that this form can generally avoid SCA, we now study the conditions in more detail. Based on the definition of SCA in \Cref{def:sca}, we need to ensure that different components of $\mX$ can be amplified across distinct feature channels.

To ease our notation, we introduce a vector $\vd^{(i)}$ that captures the weighted in-degrees for node $i$ and all computational graphs: 

\begin{definition}[Weighted In-Degrees]
    Let $\tilde{\mA}_{(k)}\in\R^{n\times n}$ for $k\in[l]$. For each node $i\in[n]$, the vector $\vd^{(i)}\in\R^l$ contains the weighted in-degrees $\evd^{(i)}_k = \sum_{m\in[n]} \bracks{\tilde{\mA}_{(k)}}_{i,m}$ for all edge relations $k\in[l]$.
\end{definition}

This allows us to define the notion of structural dependence between two nodes:

\begin{definition}[Structural Dependence and Independence]
    Two nodes $i,j\in[n]$ are structurally dependent when $\vd^{(i)} = p\cdot\vd^{(j)}$ for some $p\in\R$, otherwise they are called structurally independent.
\end{definition}

We visualize an example in \Cref{fig:struct_independ}. We can now show that the structural independence of nodes serves as a sufficient condition to avoid SCA:

\begin{theorem}
\label{pr:4:split_independence}
    Let $\mX^\prime = \sum_{k\in[l]} \tilde{\mA}_{(k)}\mX\mW_{(k)}$ for $\tilde{\mA}_{(k)}\in\R^{n\times n}$, $\mW_{(k)}\in\R^{d\times d^\prime}$ for all $k\in[l]$, and $\mX\in\R^{n\times d}$ be a rank-one matrix. Then, $\mX_{i,:}^\prime$ and $\mX_{j,:}^\prime$ are linearly independent for almost every (a.e.) matrices $\mW_{(1)},\dots,\mW_{(l)}$ and a.e. rank-one matrix $\mX$ if nodes $i\in[n]$ and $j\in[n]$ are structurally independent.
\end{theorem}

\begin{proof}
    We require
    \begin{equation}
    \bracks{\sum_{k=1}^l \tilde{\mA}_{(k)}\mX\mW_{(k)}}_{i,:} - c_j\cdot\bracks{\sum_{k=1}^l \tilde{\mA}_{(k)}\mX\mW_{(k)}}_{j,:} \neq 0
    \end{equation}
    for all choices of $c_j\in\R$. This condition is satisfied for a.e. choice of matrices $\mW_{(k)}$ for $k\in[l]$ if
    \begin{equation}
            \bracks{\tilde{\mA}_{(k)}\mX}_{i,:} - c_j\cdot\bracks{\tilde{\mA}_{(k)}\mX}_{j,:} \neq 0\, .
    \end{equation}
    for at least one $k\in[l]$. For a.e. rank-one matrix $\mX$ this inequality further holds whenever
    \begin{equation}
            \bracks{\tilde{\mA}_{(k)}}_{i,:} - c_j\cdot\bracks{\tilde{\mA}_{(k)}}_{j,:} \neq 0\, .
    \end{equation}
    for at least one $k\in[l]$. Based on the definition of structural independence, this cannot be satisfied for any $c_j\in\R$, and the resulting node representations are linearly independent.
\end{proof}

We note the limitation that \Cref{pr:4:split_independence} requires an a.e. condition on the state $\mX$. Thus, there exist specific constellations of matrices $\tilde{\mA}_{(k)}$ so that the resulting representations for nodes $i$ and $j$ remain linearly dependent. However, such matrices $\mX$ have measure zero, and we will further lift this condition in \Cref{prop:4:lin_indepdence}.
In all other cases, a single iteration with an MRS-MPNN can amplify distinct components for each feature channel for nodes $i$ and $j$, preventing SCA. Their node representations can remain similar for one feature channel, while becoming dissimilar for another. The similarity depends on the assigned edge weights, as specific choices can ensure that $\tilde{\mA}_{(k)}\mX$ is similar for nodes $i$ and $j$, while specific other choices can lead to $\tilde{\mA}_{(k)}\mX$ being dissimilar for nodes $i$ and $j$. For which feature channels the representations will be similar or dissimilar then depends on the corresponding feature transformations $\mW_{(k)}$. As these are typically learnable parameters, the optimization process can effectively control which computational graph should have larger weights for a specific downstream task. This shows that an MRS-MPNN can obtain more informative node representations from a given state for any two structurally independent nodes.
As rank-one matrices are not fixed points, such cases also prevent the convergence to a rank-one state. 
It also means that such functions are injective for structurally independent nodes and a.e. $\mX$ and $\mW_{(k)}$, which improves the expressivity of MPNNs. This already highlights a close similarity between avoiding SCA, and as a consequence, rank collapse and over-smoothing, and improving the expressivity to distinguish between structural differences. Two structurally independent nodes have distinct representations after applying one or more iterations of an MRS-MPNN for a.e. $\mX$.  
The two node representations remain linearly independent, even after applying a non-linear activation $\sigma$, assuming $\sigma$ is injective and applied component-wise.

Having more structurally independent nodes further allows these nodes to have pairwise linearly independent representations.
For a set of structurally independent nodes, each resulting node representation is linearly independent from the set of representations from all other nodes.
We formalize this result in the following theorem: 

\begin{theorem}
\label{pr:4:split:rank}
    Let $p$ be the number of structurally independent nodes, i.e., for all $i\in[p]$, it holds that $\vd^{(i)} \neq \sum_{j\in[p],j\neq i}p_j\vd^{(j)}$ for any $p_j\in\R$. Let $\sigma$ be a component-wise and injective activation function. Then for $\tilde{\mA}_{(k)}\in\R^{n\times n}$, a.e. $\mW_{(k)}\in\R^{d\times d^\prime}$, and a.e. rank-one matrix $\mX\in\R^{n\times d}$, it holds that
    \begin{equation}
    \mathrm{rank}\bracks{\sigma\bracks{\sum_{k=1}^l \tilde{\mA}_{(k)}\mX\mW_{(k)}}} \geq p\, .
    \end{equation}
\end{theorem}

\begin{proof}
    For notational simplicity, we assume the first nodes $[p]$ to be structurally independent. We then require
    \begin{equation}
    \bracks{\sigma\bracks{\sum_{k=1}^l \tilde{\mA}_{(k)}\mX\mW_{(k)}}}_{i,:} - \sum_{j\in[p],j\neq i} c_j\bracks{\sigma\bracks{\sum_{k=1}^l \tilde{\mA}_{(k)}\mX\mW_{(k)}}}_{j,:} \neq 0
    \end{equation}
    for all choices of $c_j\in\R,j\in[p],j\neq i$. This condition is satisfied for a.e. choice of matrices $\mW_{(k)}$ for $k\in[l]$ if
    \begin{equation}
            \bracks{\sum_{k=1}^l \tilde{\mA}_{(k)}\mX\mW_{(k)}}_{i,:} - \sum_{j\in[p],j\neq i} c_j\bracks{\sum_{k=1}^l \tilde{\mA}_{(k)}\mX\mW_{(k)}}_{j,:} \neq 0\, .
    \end{equation}

    Furthermore, for a.e. choice of matrices $\mW_{(k)}$ for $k\in[l]$ this inequality holds whenever 
    \begin{equation}
            \bracks{\tilde{\mA}_{(k)}\mX}_{i,:} - \sum_{j\in[p],j\neq i} c_j\bracks{\tilde{\mA}_{(k)}\mX}_{j,:} \neq 0\, .
    \end{equation}
    for at least one $k\in[l]$. For a.e. rank-one matrix $\mX$ this inequality further holds whenever
    \begin{equation}
            \bracks{\tilde{\mA}_{(k)}}_{i,:} - \sum_{j\in[p],j\neq i} c_j\bracks{\tilde{\mA}_{(k)}}_{j,:} \neq 0\, .
    \end{equation}
    for at least one $k\in[l]$. Based on the definition of structural independence, this cannot be satisfied for any $c_j\in\R$, and the resulting node representations are linearly independent.

    As this holds equivalently for any of the $p$ structurally independent nodes, this results in $p$ linearly independent representations, implying a minimal rank of $p$.
\end{proof}

This aligns with the goal of constructing a message-passing step that does not exhibit SCA. When we have $p$ structurally independent nodes, we can amplify distinct components for at least $p$ feature channels, and the learnable parameters given by $\mW_{(k)}$ control which components get amplified for which feature channel. This allows the learnable parameters to be more effective and the optimization to be more impactful. Adding feature channels can also capture new combinations of amplified components. When linearly independent representations are desired, maximizing the number of structurally independent nodes helps achieve this goal.
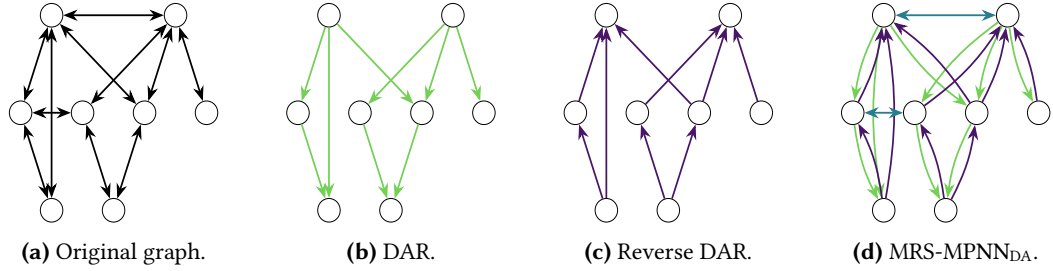
\begin{figure*}
\definecolor{viridis1}{RGB}{72,21,103}
\definecolor{viridis2}{RGB}{40,125,142}
\definecolor{viridis3}{RGB}{115,208,85}
     \centering
     \begin{subfigure}[b]{0.21\textwidth}
         \centering
        \resizebox{\textwidth}{\textwidth}{
         \begin{tikzpicture}
\begin{scope}[every node/.style={circle,draw,scale=1.0}]
    \node (A) at (0.5,3) {};
    \node (B) at (2.5,3) {};
    \node (C) at (0,1.5) {};
    \node (D) at (1,1.5) {} ;
    \node (E) at (2,1.5) {} ;
    \node (F) at (3,1.5) {} ;
    \node (G) at (0.5,0) {} ;
    \node (H) at (1.5,0) {} ;
\end{scope}

\begin{scope}[>={Stealth[black]},
              every edge/.style={draw=black,thick}]
    \path [<->] (A) edge (B);
    \path [<->] (A) edge (C);
    \path [<->] (A) edge (E);
    \path [<->] (A) edge (G);
    \path [<->] (B) edge (D);
    \path [<->] (B) edge (E); 
    \path [<->] (B) edge (F); 
    \path [<->] (C) edge (G); 
    \path [<->] (D) edge (H); 
    \path [<->] (E) edge (H); 
    \path [<->] (C) edge (D); 
\end{scope}
\end{tikzpicture}
        }
         \caption{Original graph.}
     \end{subfigure}
     \hfill
     \begin{subfigure}[b]{0.21\textwidth}
         \centering
         \resizebox{\textwidth}{\textwidth}{
         \begin{tikzpicture}
\begin{scope}[every node/.style={circle,draw,scale=1.0}]
    \node (A) at (0.5,3) {};
    \node (B) at (2.5,3) {};
    \node (C) at (0,1.5) {};
    \node (D) at (1,1.5) {} ;
    \node (E) at (2,1.5) {} ;
    \node (F) at (3,1.5) {} ;
    \node (G) at (0.5,0) {} ;
    \node (H) at (1.5,0) {} ;
\end{scope}

\begin{scope}[>={Stealth[viridis3]},
              every edge/.style={draw=viridis3,thick}]
    \path [->] (A) edge (C);
    \path [->] (A) edge (E);
    \path [->] (A) edge (G);
    \path [->] (B) edge (D);
    \path [->] (B) edge (E); 
    \path [->] (B) edge (F); 
    \path [->] (C) edge (G); 
    \path [->] (D) edge (H); 
    \path [->] (E) edge (H); 
\end{scope}
\end{tikzpicture}
        }
         \caption{DAR.}
     \end{subfigure}
     \hfill
     \begin{subfigure}[b]{0.21\textwidth}
         \centering
         \resizebox{\textwidth}{\textwidth}{
         \begin{tikzpicture}
\begin{scope}[every node/.style={circle,draw}]
    \node (A) at (0.5,3) {};
    \node (B) at (2.5,3) {};
    \node (C) at (0,1.5) {};
    \node (D) at (1,1.5) {} ;
    \node (E) at (2,1.5) {} ;
    \node (F) at (3,1.5) {} ;
    \node (G) at (0.5,0) {} ;
    \node (H) at (1.5,0) {} ;
\end{scope}

\begin{scope}[>={Stealth[viridis1]},
              every edge/.style={draw=viridis1,thick}]
    \path [<-] (A) edge (C);
    \path [<-] (A) edge (E);
    \path [<-] (A) edge (G);
    \path [<-] (B) edge (D);
    \path [<-] (B) edge (E); 
    \path [<-] (B) edge (F); 
    \path [<-] (C) edge (G); 
    \path [<-] (D) edge (H); 
    \path [<-] (E) edge (H); 
\end{scope}
\end{tikzpicture}
        }
         \caption{Reverse DAR.}
     \end{subfigure}
     \hfill
     \begin{subfigure}[b]{0.21\textwidth}
         \centering
         \resizebox{\textwidth}{\textwidth}{
         \begin{tikzpicture}
\begin{scope}[every node/.style={circle,draw,scale=1.0}]
    \node (A) at (0.5,3) {};
    \node (B) at (2.5,3) {};
    \node (C) at (0,1.5) {};
    \node (D) at (1,1.5) {} ;
    \node (E) at (2,1.5) {} ;
    \node (F) at (3,1.5) {} ;
    \node (G) at (0.5,0) {} ;
    \node (H) at (1.5,0) {} ;
\end{scope}

\begin{scope}[>={Stealth[viridis3]},
              every edge/.style={draw=viridis3,thick,bend right=10}]
    \path [->] (A) edge (C);
    \path [->] (A) edge (E);
    \path [->] (A) edge (G);
    \path [->] (B) edge (D);
    \path [->] (B) edge (E); 
    \path [->] (B) edge (F); 
    \path [->] (C) edge (G); 
    \path [->] (D) edge (H); 
    \path [->] (E) edge (H); 
\end{scope}
\begin{scope}[>={Stealth[viridis1]},
              every edge/.style={draw=viridis1,thick,bend left=10}]
    \path [<-] (A) edge (C);
    \path [<-] (A) edge (E);
    \path [<-] (A) edge (G);
    \path [<-] (B) edge (D);
    \path [<-] (B) edge (E); 
    \path [<-] (B) edge (F); 
    \path [<-] (C) edge (G); 
    \path [<-] (D) edge (H); 
    \path [<-] (E) edge (H); 
\end{scope}
\begin{scope}[>={Stealth[viridis2]},
              every edge/.style={draw=viridis2,thick}]
    \path [<->] (A) edge (B); 
    \path [<->] (C) edge (D); 
\end{scope}
\end{tikzpicture}
        }
         \caption{$\textrm{MRS-MPNN}_{\textrm{DA}}$.}
     \end{subfigure}
        \caption[Different computational graphs described for MRS-MPNNs.]{Different computational graphs described for $\textrm{MRS-MPNN}_{\textrm{DA}}$. The original graph is represented in (a), the obtained directed acyclic relation (DAR) using the node-degree as a partial ordering in (b), and the reverse DAR using the reverse partial ordering in (c). In (d), these two computational graphs are combined with the relation covering all other edges of the original graph. }
        \label{fig:comp_graphs}
\end{figure*}

\subsection{Obtaining Multiple Relations Using a Partial Ordering}
\label{sec:method}

Based on the general MRS framework and the theoretical properties we presented in \Cref{pr:4:split_independence} and \Cref{pr:4:split:rank}, an MRS-MPNN can improve the obtained representations and enable learnable parameters to be more effective. It remains open how the edge relation assignment function $f(v_i,v_j)$ can be defined. Ideally, such a function $f$ ensures the existence of structurally independent nodes. Finding the optimal function $f$ is highly task-dependent, and prior knowledge can be leveraged in many cases. 
For many applications, different edge relations can be naturally constructed, as nodes and edges correspond to distinct objects. For example, in traffic networks, roads leading to highways could be assigned to a different relation than nodes within a residential area. For chemical data, the type of bond could determine the relation type. We envision considerable potential for future work to identify specific edge splits for tasks of interest.

Here, we propose a generic option that utilizes graph-theoretical properties. 
Consider the following motivational example. A multi-head attention-based MPNN can also be viewed as a graph with multiple relations between all connected nodes. Attention normalizes edge weights to sum to one for each node within one computational graph, and all edge weights are positive. Graphs are typically ergodic, which means that they are strongly connected, i.e., a path exists between all pairs of nodes, and aperiodic, i.e., the length of their cycles does not have a common divisor $>1$. Such graphs are unable to produce structurally independent nodes:
\begin{proposition}
    Let $\tilde{\mA}_{(k)}\in\R^{n\times n}$ represent an ergodic graph for all $k\in[l]$ such that $\sum_{j\in[n]} \bracks{\tilde{\mA}_{(k)}}_{i,j} = 1$ for all $i\in[n]$ and $\bracks{\tilde{\mA}_{(k)}}_{i,j} \geq 0$ for all $i,j\in[n]$. Then, any pair of nodes is structurally dependent.
\end{proposition}
\begin{proof}
    For every relation $k\in[l]$ and every node $i\in[n]$, we have $\sum_{j\in[n]} \bracks{\tilde{\mA}_{(k)}}_{i,j} = 1$.
    This is a contradiction to the definition of structural independence, which requires $\sum_{j\in[n]} \bracks{\mA_{(k)}}_{p,j} \neq \sum_{j\in[n]} \bracks{\mA_{(k)}}_{q,j}$ for some $p\neq q$. 
\end{proof}

We note the importance of the ergodic assumption. For a graph that is not ergodic, the same cannot be stated. A graph that does not contain any ergodic subgraphs is a directed acyclic graph (DAG), which we define for a relation as follows:

\begin{definition}[Directed Acyclic Relation (DAR)]
\label{def:dag}
    A relation $\tilde{\mA}$ is called a directed acyclic relation (DAR) if there exists a strict partial ordering $\prec$ on the nodes $i,j\in[n]$ such that all edges $\tilde{\emA}_{i,j}\neq 0$, satisfy $i\prec j$.
\end{definition}

We further refer to a node with no incoming edges, i.e., $\tilde{\mA}_{i,:} = \mathbf{0}\in\R^n$, as a root node in $\tilde{\mA}$. We find that for two DARs with distinct root nodes, these nodes are always structurally independent:

\begin{proposition}
    Let $\tilde{\mA}_{(1)},\tilde{\mA}_{(2)}\in\R^{n\times n}$ be two DARs with distinct root nodes. Then, $\tilde{\mA}_{(1)}$ and $\tilde{\mA}_{(2)}$ have structurally independent nodes.
\end{proposition}
\begin{proof}
    Let $i$ be a root node of $\tilde{\mA}_{(1)}$ that is not a root node of $\tilde{\mA}_{(2)}$. Then $\sum_{j\in[n]}\bracks{\tilde{\emA}_{(1)}}_{i,j} = 0$ and $\sum_{j\in[n]}\bracks{\tilde{\emA}_{(2)}}_{i,j} \neq 0$. For any other node $k\in[n]$, we have $\sum_{j\in[n]}\bracks{\tilde{\emA}_{(1)}}_{k,j} \neq 0$, which implies structural independence.
\end{proof}

Constructing two DARs with distinct root nodes is one way to ensure structurally independent nodes, which then allows linearly independent node representations. Transforming a given graph $G=(\mA,\mX)$ into a directed acyclic graph is a well-known algorithmic challenge, which arises in topological sorting~\parencite{cormen1989introduction} and finding feedback arc sets~\parencite{garey1979computers}.
Instead, based on our \Cref{def:dag}, we start by defining a strict partial ordering $\prec$ on the nodes and only utilize the edges that satisfy the relation. Starting with the original aggregation function $\tilde{\mA}$, this results in the first DAR with entries $\bracks{\tilde{\mA}_{(1)}}_{i,j} = \tilde{\mA}_{i,j}$ when $i\prec j$ and zero otherwise. We obtain the second DAR, as the reverse DAR using the reverse strict partial ordering, i.e., the entries are $\bracks{\tilde{\mA}_{(2)}}_{i,j} = \tilde{\mA}_{i,j}$ for $j\prec i$ and zero otherwise. As we employ a partial ordering, some nodes may not be related by $\prec$. For these remaining edges, we propose a third edge relation $\tilde{\mA}_{(3)}$. Their combination results in the original adjacency matrix, i.e., $\tilde{\mA} = \tilde{\mA}_{(1)} + \tilde{\mA}_{(2)} + \tilde{\mA}_{(3)}$. We define the corresponding edge relation assignment function as
\begin{equation}
    f(i,j) = \begin{cases}
        1, & \text{if } i \prec j \\
        2, & \text{if } j \prec i \\
        3, & \text{otherwise}\, .
    \end{cases}
\end{equation}

Figure~\ref{fig:comp_graphs} visualizes various computational graphs and edge relations. We refer to this instantiation of the MRS framework applied to an MPNN as $\textrm{MRS-MPNN}_{DA}$ due to utilizing multiple directed acyclic relations. While we cannot guarantee that the third graph leads to more structurally independent nodes, additional graphs cannot hurt the number of structurally independent nodes. 

To apply the $\textrm{MRS-MPNN}_{DA}$, we additionally need to define the strict partial ordering $\prec$. 
The partial ordering needs to align with the considered task. While any method ensures structurally independent nodes, the edges within one relation type should be similar for the optimization process to obtain suitable parameters more effectively. When the messages sent across edges from different edge relations benefit from similar feature transformation, the gain of the MRS framework will also be less significant.
Graph traversal algorithms or centrality measures can provide such an ordering. We propose to use the node degree as an efficient and general ordering. Messages sent from higher-degree nodes to lower-degree nodes differ from those sent from lower-degree nodes to higher-degree nodes for specific tasks. For molecular data, the degree of a node corresponds to other atom types than those of lower degree~\parencite{wells2012structural}. Sending messages based on relations between atom types may be promising. As the choice of edge relation assignment function has a significant impact on node representations and the optimization process, this allows future methods to investigate which split specifically makes sense for a given task and which prior knowledge may be beneficial to incorporate for that task.

\section{What Can We Learn From MIMO Graph Convolutions?}
\label{sec:uniformity:mimo-gc}
The MRS framework converts the computational graph of a given MPNN into multiple computational graphs, allowing MPNNs to avoid SCA while maintaining other proposed benefits.
We now propose a novel derivation of message-passing methods from the spectral graph convolution. This will not only demonstrate the close connection between utilizing multiple computational graphs in MPNNs but also provide a more general framework for designing future message-passing methods. This section is based on \textcite{roth2025what}.

\subsection{SISO Graph Convolution}
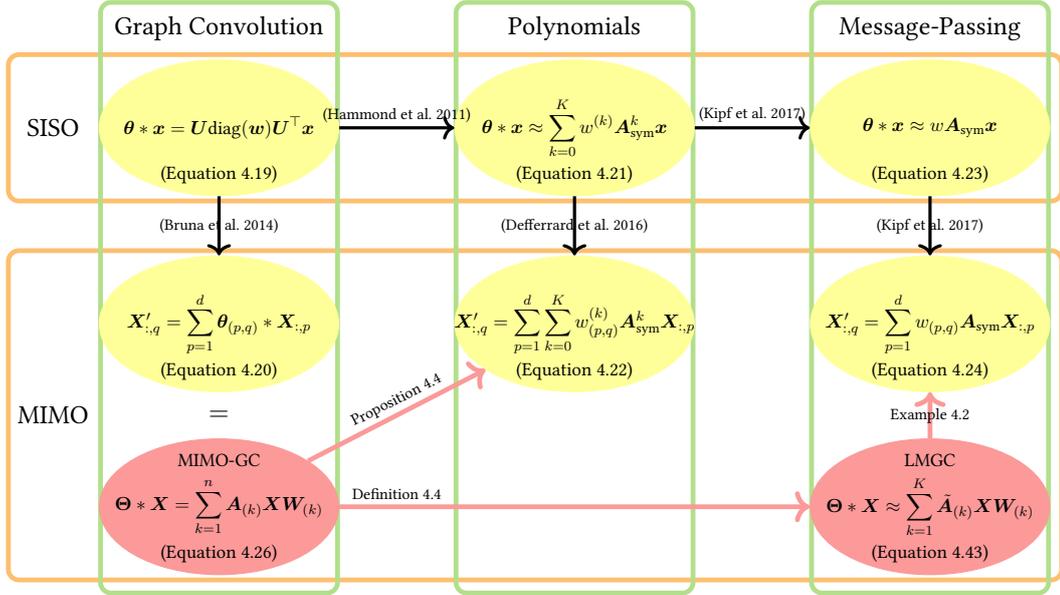
\begin{figure*}[t]
\newcommand{\rectangle}[1]{%
  \begin{tikzpicture}[baseline=-0.5ex]
    \node[minimum size=0.6em, draw=#1, fill=#1, inner sep=0pt] {};
  \end{tikzpicture}%
}
\definecolor{vibrantpink}{HTML}{fb9a99} 
\definecolor{vibrantsand}{HTML}{ffff99} 

\centering
\begin{tikzpicture}[x=\textwidth/4, y=\textwidth/4]
\definecolor{vibrantblue}{HTML}{a6cee3} 
\definecolor{vibrantgreen}{HTML}{b2df8a} 
\definecolor{vibrantpink}{HTML}{fb9a99} 
\definecolor{vibrantorange}{HTML}{fdbf6f} 
\definecolor{vibrantpurple}{HTML}{cab2d6} 
\definecolor{vibrantsand}{HTML}{ffff99} 
\usetikzlibrary{shapes}
\pgfdeclarelayer{background}
\pgfsetlayers{background,main}
\tikzset{
dot/.style = {circle, minimum size=#1,
              inner sep=0pt, outer sep=0pt, draw=black},
dot/.default =13pt  
}
    \tikzstyle{box} = [draw, draw=vibrantgreen, thick,rounded corners,line width=1.8pt]

    \def\ysiso{0.7}
    \def\ymimo{-0.05}
    \def\yours{-0.75}
    \def\xgc{-1.2}
    \def\xmp{1.50}
    \def\xpoly{0.15}
    \def\ymargin{0.28}
    \def\eqmargin{0.18}
    \def\captionscale{0.9}
    \def\eqscale{0.65}
    
    \draw[box,draw=vibrantorange] (\xgc-0.8, \ysiso-\ymargin) rectangle (\xmp+0.5, \ysiso+\ymargin); 
    \node[scale=\captionscale] at (-1.83, \ysiso) {SISO}; 
        
    \draw[box,draw=vibrantorange] (\xgc-0.8, \yours-\ymargin) rectangle (\xmp+0.5, \ymimo + \ymargin);
    \node[scale=\captionscale] at (-1.83, \ymimo-0.35) {MIMO};
    \node[scale=\captionscale] at (\xgc,\ymimo -0.35) {$=$}; 

    \draw[box] (\xgc -0.45, \yours-\ymargin-0.05) rectangle (\xgc + 0.45, \ysiso+\ymargin+0.2);
    \node[scale=\captionscale] at (\xgc, \ysiso+\ymargin+0.1) {Graph Convolution};
    \draw[box] (\xmp - 0.45, \yours-\ymargin-0.05) rectangle (\xmp + 0.45,\ysiso+\ymargin+0.2); 
    \node[scale=\captionscale] at (\xmp, \ysiso+\ymargin+0.1) {Message-Passing};
    \draw[box] (\xpoly -0.45, \yours-\ymargin-0.05) rectangle (\xpoly + 0.45,\ysiso+\ymargin+0.2);
    \node[scale=\captionscale] at (\xpoly, \ysiso+\ymargin+0.1) {Polynomials};
    
    \node[fill=vibrantsand,draw=vibrantsand,minimum width=3.15cm,minimum height=1.8cm,draw,ellipse] (scgc) at (\xgc,\ysiso) {};
    \node[scale=\eqscale] at (\xgc, \ysiso - \eqmargin) {(\autoref{eq:4:siso_gc})}; 
    \node[scale=\eqscale] at (\xgc, \ysiso) {$\displaystyle\vtheta*\vx=\mU \mathrm{diag}(\vw)\mU^\top\vx$};

    \node[fill=vibrantsand,draw=vibrantsand,minimum width=3.15cm,minimum height=1.8cm,draw,ellipse] (mcgc) at (\xgc,\ymimo) {};
    \node[scale=\eqscale] at (\xgc, \ymimo) {$\displaystyle\mX^\prime_{:,q} = \sum_{p=1}^d\vtheta_{(p,q)}*\mX_{:,p}$};
    \node[scale=\eqscale] at (\xgc, \ymimo - \eqmargin) {(\autoref{eq:4:siso_gc_mimo})}; 

    \node[fill=vibrantpink,draw=vibrantpink,minimum width=3.15cm,minimum height=1.8cm,draw,ellipse] (mimogc) at (\xgc,\yours) {};
    \node[scale=\eqscale] at (\xgc, \yours + \eqmargin) {MIMO-GC}; 
    \node[scale=\eqscale] at (\xgc, \yours - \eqmargin) {(\autoref{eq:4:mimo_gc})}; 
    \node[scale=\eqscale] at (\xgc, \yours) {$\displaystyle\tTheta*\mX = \sum_{k=1}^n \mA_{(k)}\mX\mW_{(k)}$};

    \node[fill=vibrantsand,draw=vibrantsand,minimum width=3.15cm,minimum height=1.8cm,draw,ellipse] (scmpnn) at (\xmp,\ysiso) {};    
    \node[scale=\eqscale] at (\xmp, \ysiso) {$\displaystyle\vtheta * \vx \approx w \mA_\tsym\vx$}; 
    \node[scale=\eqscale] at (\xmp, \ysiso - \eqmargin) {(\autoref{eq:4:siso_gcn})}; 
    \coordinate (x) at (\xmp,\ymimo);
        \node[fill=vibrantsand,draw=vibrantsand,minimum width=3.15cm,minimum height=1.8cm,draw,ellipse] (a) at (\xmp,\ymimo) {}; 
    \node[scale=\eqscale] at ($ (x) + (0:0.0) $) {$\displaystyle\mX^\prime_{:,q} = \sum_{p=1}^d w_{(p,q)}\mA_\tsym\mX_{:,p}$};
        \node[scale=\eqscale] at (\xmp, \ymimo - \eqmargin) {(\autoref{eq:4:mimo_gcn})}; 
    \node[fill=vibrantpink,draw=vibrantpink,minimum width=3.15cm,draw,ellipse,minimum height=1.8cm] (lmcgc) at (\xmp,\yours) {};
    \node[scale=\eqscale] at (\xmp,\yours) {$\displaystyle\tTheta * \mX \approx \sum_{k=1}^K \tilde{\mA}_{(k)}\mX\mW_{(k)}$}; 
    \node[scale=\eqscale] at (\xmp,\yours + \eqmargin) {LMGC}; 
    \node[scale=\eqscale] at (\xmp,\yours - \eqmargin) {(\autoref{eq:4:lmgc})}; 


    \node[fill=vibrantsand,draw=vibrantsand,minimum width=3.15cm,minimum height=1.8cm,draw,ellipse] (scpoly) at (\xpoly,\ysiso) {};
    \node[scale=\eqscale]  at (\xpoly, \ysiso) {$\displaystyle\vtheta * \vx \approx \sum_{k=0}^K w^{(k)} \mA_\tsym^k\vx$}; 
    \node[scale=\eqscale] at (\xpoly, \ysiso - \eqmargin) {(\autoref{eq:4:siso_poly})}; 

    \node[fill=vibrantsand,draw=vibrantsand,minimum width=3.15cm,minimum height=1.8cm,draw,ellipse] (mcpoly) at (\xpoly,\ymimo) {};
    \node[scale=\eqscale] at (\xpoly, \ymimo) {$\displaystyle\mX^\prime_{:,q} = \sum_{p=1}^d\sum_{k=0}^K w_{(p,q)}^{(k)}\mA_\tsym^k\mX_{:,p}$};   
    \node[scale=\eqscale] at (\xpoly, \ymimo - \eqmargin) {(\autoref{eq:4:mimo_poly})}; 

    \def\citescale{0.55}

    \draw[->,line width=1.2pt] (scpoly) to (scmpnn);
    \node[color=black] at (\xpoly+0.675,\ysiso+0.05) {\tiny\parencite{kipf2017semi}};
    \draw[->,line width=1.2pt] (scmpnn) to (a);
    \node[color=black] at (\xmp,0.325){\tiny\parencite{kipf2017semi}};

    \draw[black,->,line width=1.2pt] (scgc) to (mcgc);
    \node[color=black] at (\xgc,0.325){\tiny\parencite{bruna2014spectral}};

    \draw[->,line width=1.2pt] (scgc) to (scpoly);
    \node[color=black] at (\xgc+0.675,\ysiso+0.05) {\tiny\parencite{hammond2011wavelets}};

    \draw[->,line width=1.2pt] (scpoly) to (mcpoly);
    \node[color=black] at (\xpoly,0.325){\tiny\parencite{defferrard2016convolutional}};

    \draw[vibrantpink,->,line width=1.8pt] (mimogc) to (lmcgc);
    \node[color=black] at ($(\xgc,\yours+0.05)!0.25!(\xmp,\yours+0.05)$){\tiny\Cref{def:lmgc}};
    \draw[vibrantpink,->,line width=1.8pt] (mimogc)  to (mcpoly);
    \node[color=black,transform shape,anchor=base,text depth=0pt] at ($(\xgc,\yours-0.04)!0.5!(\xpoly,\ymimo-0.04)$) {\rotatebox{27}{\tiny\Cref{prop:4:mimo_gc_poly}}};
    
    \draw[vibrantpink,->,line width=1.8pt] (lmcgc) to (a);
    \node[color=black] at ($(\xmp,\yours)!0.5!(\xmp,\ymimo)$){\tiny\Cref{ex:4:lmgc_gcn}};

\end{tikzpicture}
\caption[Connection between the MIMO-GC, the LMGC and related work.]{Connection between the MIMO-GC, the LMGC and previous work on defining the graph convolution, polynomial, and localized approximations in the SISO and the MIMO case. Yellow parts (\protect\rectangle{vibrantsand}) indicate previous work, while pink parts (\protect\rectangle{vibrantpink}) indicate our contributions.}
\label{fig:4:mimo_overview}
\end{figure*}

We first provide an overview of related work on the definition of graph convolution and its approximations. We visualize the key elements of this overview in \Cref{fig:4:mimo_overview}. 
Initially, the single-input single-output case (SISO) was considered where the node representations $\vx\in\R^n$ consist of a single feature for each node. This is mapped to updated node representations $\vx^\prime\in\R^n$ that also consist of a single feature for each node.
\textcite{bruna2014spectral} defined the SISO graph convolution as
\begin{equation}
\label{eq:4:siso_gc}
    \vtheta * \vx = \mU\mathrm{diag}(\vw)\mU^\top\vx
\end{equation}
where $\vw = \mU^\top\vtheta\in\R^n$ using the graph Fourier transform $\mU^\top$ as the Fourier transform $F$ within the convolution theorem. We refer to our introduction of the graph Fourier transform in \Cref{sec:fundamentals:graph_fourier} and the graph convolution in \Cref{sec:fundamentals:graph_conv} for more details.
Most node representations $\mX\in\R^{n\times d}$ have some number $d$ of features for each node. \textcite{bruna2014spectral} propose to extend the SISO graph convolution to this multi-input multi-output (MIMO) case by applying a separate SISO graph convolution to every combination of input and output feature channels, i.e., the $q$-th feature channel of the resulting node representations $\mX^\prime\in\R^{n\times c}$ is updated as
\begin{equation}
\label{eq:4:siso_gc_mimo}
    \mX^\prime_{:,q} = \sum_{p=1}^d \vtheta_{(p,q)} * \mX_{:,p}\, ,
\end{equation}
where each $\vtheta_{(p,q)}\in\R^n$ is a vector for every $p\in[d]$ and $q\in[c]$.
\textcite{hammond2011wavelets} propose Chebyshev polynomials that approximate the SISO graph convolution. \textcite{sandryhaila2013discrete} propose polynomial filters for graphs in the SISO case more generally. For polynomial approximations, the updated state $\vx^\prime\in\R^n$ is set as
\begin{equation}
\label{eq:4:siso_poly}
    \vx^\prime = \sum_{k=0}^K \evw^{(k)}\mA_\tsym^k\vx
\end{equation}
where $\evw^{(k)}\in\R$ are the polynomial coefficients and the symmetrically normalized adjacency matrix $\mA_\tsym$ is the variable of the polynomial.
\textcite{defferrard2016convolutional} propose to extend polynomial filters to the MIMO case equivalently to the extension of SISO graph convolutions by applying a polynomial filter to each combination of input and output feature channel, i.e.,
\begin{equation}
\label{eq:4:mimo_poly}
    \mX^\prime_{:,q} = \sum_{p=1}^d\sum_{k=0}^K \evw^{(k)}_{(p,q)}\mA_\tsym^k\mX_{:,p}
\end{equation}
where each $\evw_{(p,q)}^{(k)}\in\R$ is a coefficient for each $p,q,k$. \textcite{gama2018mimo} similarly extend polynomial filters to the MIMO case. \textcite{kipf2017semi} propose the GCN as a localized first-order approximation of polynomial filters on graphs. 
\begin{equation}
\label{eq:4:siso_gcn}
    \vx^\prime = w\mA_\tsym\vx\, ,
\end{equation}
where $w\in\R$ is the single parameter of this approach. The GCN adds self-loops to the adjacency matrix before normalization, which we omit here for notational simplicity.
Equivalently to \textcite{bruna2014spectral}, \textcite{kipf2017semi} also extend the GCN to the MIMO case by applying the approximation to each combination of input and output channel, i.e.,
\begin{equation}
\label{eq:4:mimo_gcn}
    \mX^\prime_{:,q} = \sum_{p=1}^d w_{(p,q)}\mA_\tsym\mX_{:,p}
\end{equation}
for each $q\in[c]$ and $w_{(p,q)}\in\R$ for all $p,q$. 
Subsequently, many other MPNNs have been proposed as variations of the GCN~\parencite{velickovic2017graph,xu2019how,roth2022transforming}. 
This highlights that methods such as GCN and its modifications approximate the graph convolution in the SISO case, but are applied in the MIMO case. We will propose an approximation that avoids this indirect route.

\subsection{MIMO Graph Convolution}
\label{sec:mcgc}
Instead of approximating the SISO graph convolution, we propose to perform the approximation directly in the MIMO case. For that, we first need to define the graph convolution in the MIMO case. Given a node representations matrix $\mX\in\R^{n\times d}$, we want to perform the graph convolution in the Fourier domain
\begin{equation}
    \bracks{\tTheta * \mX}_{i,:} = F^{-1}(F(\tTheta) \boxdot F(\mX))
\end{equation}
based on the convolution theorem~\parencite{neil1963convolution}, where $\boxdot$ is the node-wise multiplication $\bracks{F(\tTheta)\boxdot F(\mX)}_{i,:} = F(\tTheta)_{i,:,:}\cdot F(\mX)_{i,:}$. Following \textcite{bruna2014spectral}, we define the Fourier transform $F$ as the graph Fourier transform given by the eigenvectors of $\mA_\tsym$.
The convoluted node representation matrix $\mX^\prime = \tTheta * \mX\in\R^{n\times c}$ can have $c\in\mathbb{N}$ feature channels for every node. The filter $\tTheta\in\R^{n\times c\times d}$ needs to contain the element-wise mapping from $d$ to $c$ feature channels $\tTheta_{i,:,:}\in\R^{c\times d}$. Based on these definitions, we obtain the MIMO graph convolution as follows:

\begin{theorem}[MIMO Spectral Graph Convolution (MIMO-GC)]
\label{pr:4:mimo_gc}
    Let $\mX\in\R^{n\times d}$ and $\tTheta\in\R^{n\times c\times d}$ and the Fourier transform $F(\mM) := \mU^\top\times_1\tM$ be defined using mode-1 multiplication $\times_1$ and the eigenvector matrix $\mU\in\R^{n\times n}$ of $\mA_\tsym\in\R^{n\times n}$ for any tensor $\tM\in\R^{n\times d_1 \times \dots\times d_k}$ for any $d_1,\dots,d_k,k\in\mathbb{N}$.
    Then,
    
    \begin{align}
        \bracks{\tTheta * \mX}_{i,:} &= \bracks{\sum_{k=1}^n \mA_{(k)}\mX\mW_{(k)}}_{i,:}  \label{eq:4:mimo_gc}\\
        &= \sum_{j=1}^n \mX_{j,:}\mW_{(i,j)} \label{eq:4:mimo_gc_node}
    \end{align}
    where $\mA_{(k)} = \mU_{:,k}\mU_{:,k}^\top\in\R^{n\times n}$, $\mW_{(k)} = \bracks{\mU^\top\times_1\tTheta}_{k,:,:}^\top\in\R^{d\times c}$ and $\mW_{(i,j)} = \bracks{\sum_{k=1}^n \emU_{i,k}\emU_{j,k}\mW_{(k)}}^\top\in\R^{c\times d}$.
\end{theorem}

\begin{proof}
    We first apply the graph Fourier transform $F$ to $\tTheta$ and $\mX$ using the mode-1 multiplication and $\mU^\top$. For $\tTheta$, we get the graph Fourier transformed filter
    \begin{equation}
        \hat{\tW}_{:,p,q} = \bracks{\mU^\top \times_1 \tTheta}_{:,p,q} = \mU^\top \tTheta_{:,p,q}
    \end{equation}
    with $\hat{\tW}\in\R^{n\times c\times d}$, and for $\mX$, we have
    \begin{equation}
        \hat{\mX} = F(\mX) = \mU^\top \mX\, .
    \end{equation}
    Based on this, we get
    \begin{equation}
        \tTheta * \mX = \mU \bracks{\hat{\tW} \boxdot \mU^\top\mX}\, ,
    \end{equation}
    using the inverse graph Fourier transform $F^{-1}(\mM) = \mU\mM$ for any $\mM$ of suitable shape.
    Similarly to the SISO graph convolution, we replace the node-wise product $\boxdot$ with a matrix multiplication by diagonalizing $\hat{\tW}$ as the block matrix of diagonal blocks in the matrix 
    \begin{equation}
       \mE = \begin{bmatrix}
           \hat{\etW}_{1,1,1} & 0 & 0 & & \hat{\etW}_{1,1,d} & 0 & 0 \\
           0 & \ddots & 0 & \dots & 0 & \ddots & 0 \\
           0 & 0 & \hat{\etW}_{n,1,1} & & 0 & 0 & \hat{\etW}_{n,1,d} \\
           & \vdots & & & & \vdots & \\
           \hat{\etW}_{1,c,1} & 0 & 0 & & \hat{\etW}_{1,c,d} & 0 & 0 \\
           0 & \ddots & 0 & \dots & 0 & \ddots & 0 \\
           0 & 0 & \hat{\etW}_{n,c,1} & & 0 & 0 & \hat{\etW}_{n,c,d} \\
       \end{bmatrix} \in\R^{nc\times nd}\, .
    \end{equation}
    This form allows us to simplify the vectorized MIMO graph convolution as
    \begin{equation}
    \label{eq:4:mimo_gc:vec}
        \tvec(\tTheta * \mX) = \bracks{\mI_c\otimes\mU}\mE\tvec\bracks{\mU^\top\mX}\, .
    \end{equation}
    We can further decompose $\mE$ as a sum of Kronecker products as
    \begin{equation}
        \mE = \sum_{k=1}^n \hat{\mW}_{k,:,:} \otimes \mI_n^{(k)}\, ,
    \end{equation}
    where $\mI_n^{(k)}\in\R^{n\times n}$ contains a single one at position $(k,k)$ and zeros at all other positions. Substituting this decomposition in \Cref{eq:4:mimo_gc:vec}, we obtain
    \begin{equation}
        \tvec(\tTheta * \mX) = \bracks{\sum_{k=1}^n \hat{\mW}_{k,:,:}\otimes \mU_{:,k}\mU_{:,k}^\top}\tvec(\mX)\, .
    \end{equation}
    Based on the property $(\mA\otimes \mB)\tvec(\mC) = \mB\mC\mA^\top$ of the Kronecker product for matrices $\mA,\mB,\mC$ of suitable shape, we state the MIMO graph convolution in the original matrix form as
    \begin{equation}
        \tTheta * \mX = \sum_{k=1}^n\mU_{:,k}\mU_{:,k}^\top\mX\hat{\mW}_{k,:,:}^\top\, .
    \end{equation}
    Defining $\mA_{(k)} = \mU_{:,k}\mU_{:,k}^\top$ and $\mW_{(k)} = \hat{\mW}_{k,:,:}$ concludes the proof.
\end{proof}

Due to the generality of the MIMO-GC, approximations of the graph convolution applied in the MIMO case should closely mimic the MIMO-GC.
The MIMO-GC can be equivalently expressed as applying the SISO graph convolution for every combination of input and output feature channel, as given in \Cref{eq:4:siso_gc_mimo}. Knowing the general form of the graph convolution for multiple feature channels allows us to study its properties. These properties can then be used to improve approximations.

The first observation is that the MIMO-GC utilizes $n$ multiple computational graphs $\mA_{(k)}\in\R^{n\times n}$, each being a rank-one matrix $\mA_{(k)}=\mU_{:,k}\mU_{:,k}^\top$. Thus, each such computational graph can only amplify exactly one component in the graph that corresponds to the eigenvector $\mU_{:,k}$ of $\mA_\tsym$. The corresponding feature transformation $\mW_{(k)}\in\R^{d\times c}$ encodes how much this component is amplified from each of the $d$ input feature channels for each of the $c$ output feature channels. This aligns well with our findings from \Cref{sec:understanding}, particularly our definition of shared component amplification (SCA). In \Cref{def:sca}, we have shown that a single computational graph can only amplify the same component maximally for all feature channels jointly. Each computational graph $\mA_{(k)}$ is orthogonal, meaning that it also amplifies independent components in the data. By utilizing multiple computational graphs, the MIMO-GC allows for the amplification of arbitrary components for each feature channel independently.  We formally show the universality of the MIMO-GC to map almost any given node representations $\mX\in\R^{n\times d}$ to any desirable node representation $\mX^\prime\in\R^{n\times c}$ in the following proposition:
\begin{proposition}
\label{pr:4:mimo_gc:universality}
    Let $\mX\in\R^{n\times d}$ and $\mX^\prime\in\R^{n\times c}$. Then, there exists a $\tTheta\in\R^{n\times c\times d}$ such that
    \begin{equation}
        \tTheta * \mX = \mX^\prime
    \end{equation}
    for almost every $\mX$.
\end{proposition}

\begin{proof}
    We decompose $\mX$ and $\mX^\prime$ as a sum of rank-one matrices, each with all columns equal to a scaled version of an eigenvector $\mU_{:,k}\in\R^{n\times 1}$ for all $k\in[n]$. For $\mX^\prime$, we obtain the decomposition
    \begin{equation}
        \mX^\prime = \sum_{k=1}^n\mU_{:,k}\vc_{(k)}^\top
    \end{equation}
    for some $\vc_{(k)}\in\R^c$ for all $k\in[n]$. Similarly, we use the decomposition
    \begin{equation}
        \mX = \sum_{k=1}^n \mU_{:,k}\vd_{(k)}^\top
    \end{equation}
    for some $\vd_{(k)}\in\R^d$ for all $k\in[n]$. Inserting this for the MIMO-GC, we obtain
    \begin{equation}
        \tTheta * \mX = \sum_{k=1}^n\mU_{:,k}\vd_{(k)}^\top\mW_{(k)}\, .
    \end{equation}
    By defining $\mW_{(k)}$ so that $\vc_{(k)} = \vd_{(k)}\mW_{(k)}$ which is possible for all $\vd_{(k)} \neq \mathbf{0}$, we conclude the proof.
\end{proof}

This confirms the universal power of the MIMO-GC to represent arbitrary such linear functions. The second interpretation of the MIMO-GC uses the form given by \Cref{eq:4:mimo_gc_node}. For every combination of nodes $i$ and $j$, a distinct feature transformation $\mW_{(i,j)}$ is applied. While these feature transformations are intertwined through $\mW_{(k)}$, this form highlights the benefits of applying distinct feature transformations $\mW_{(i,j)}$ for every message between pairs of nodes. As \Cref{pr:4:mimo_gc:universality} holds for this equivalent form, applying distinct feature transformations improves the ability of the MIMO-GC to represent arbitrary functions and amplify distinct components between every combination of input and output channel. For example, this form helps with understanding the benefits of recent MPNNs that have proposed the use of edge-wise distinct feature transformations, e.g., in neural sheaf diffusion~\parencite{hansen2020sheaf,bodnar2022neural}.

While obtaining the MIMO-GC is valuable for gaining a better understanding and intuition for applying MPNNs in the MIMO case, the MIMO-GC will typically not be applied directly for the same reasons that the SISO-GC is not applied. First, the graph Fourier transform is graph-dependent, so applying a filter $\tTheta$ to a different graph is typically not reasonable. In addition, the computational complexity of obtaining the graph Fourier transform and applying it is infeasible for many graphs. The graph Fourier transform requires obtaining all eigenvectors of $\mA_\tsym$ and applying $n$ dense matrices, in addition to applying $n$ feature transformations $\mW_{(k)}$, which has a total runtime complexity of $\mathcal{O}(n^2\cdot c\cdot d)$.

\paragraph{How Can We Benefit From the MIMO-GC?}
Instead of directly applying the MIMO-GC, we aim to use it as a starting point to derive approximations and other methods, similar to how the SISO-GC is utilized. Understanding existing approaches for MIMO computation helps design novel approaches. For example, we show in the following that any polynomial MIMO filter can be represented as a MIMO-GC with constraints on the feature transformation matrices:
\begin{proposition}
\label{prop:4:mimo_gc_poly}
    Let $\mX\in\R^{n\times d}$, $P\in\mathbb{N}$, and $\mV_{(p)}\in\R^{d\times c}$ for $0\leq p\leq P$ and any $d,c\in\mathbb{N}$. Then, there exists a filter $\tTheta\in\R^{n\times c \times d}$ such that
    \begin{equation}
        \sum_{p=0}^P \mA_\tsym^p\mX\mV_{(p)} = \tTheta_{\mathrm{poly}} * \mX\, .
    \end{equation}
\end{proposition}

\begin{proof}
    Based on the eigendecomposition $\mA_\tsym^p = \sum_{k=1}^n \mU_{:,k}\mU_{:,k}^\top\lambda^p_k$ and the definition of the MIMO-GC, we obtain
    \begin{equation}
        \sum_{p=0}^P \mA_\tsym^p\mX\mV_{(p)} = \sum_{k=1}^n\sum_{p=0}^P \mU_{:,k}\mU_{:,k}^\top\lambda^p_k\mX\mV_{(p)}
    \end{equation}
    By defining $\mW_{(k)} = \sum_{p=0}^P \lambda_k^p\mV_{(p)}$, we obtain the form of the MIMO-GC.
\end{proof}

This further confirms the expressive power of the MIMO-GC. As an example, consider the update step from the GCN as a first-order polynomial that can be expressed as MIMO-GC in the following form:
\begin{example}
\label{ex:4:mimo_gc_gcn}
    Let $\mX^\prime = \mA_\tsym\mX\mW$ for some $\mX\in\R^{n\times d}$, $\mW\in\R^{d\times c}$, $\mW\in\R^{d\times c}$, and $\mA_\tsym\in\R^{n\times n}$ for $n,d,c\in\mathbb{N}$. Then, 
    \begin{equation}
        \mX^\prime = \sum_{k=1}^n \mU_{:,k}\mU_{:,k}^\top\mX\mW_{(k)}
    \end{equation}
    where $\mW_{(k)} = \lambda_k\mW$ and $\lambda_k$ is the eigenvalue of $\mA_\tsym$ corresponding to eigenvector $\mU_{:,k}$.
\end{example}

As such, the GCN is equal to a MIMO-GC, which applies the same feature transformation to all components, with the only difference being that all components are then scaled by their respective eigenvalue of $\mA_\tsym$ jointly across all feature channels. This also serves as an additional intuitive explanation for the SCA phenomenon, as the components are scaled according to the aggregation matrix $\mA_\tsym$. The feature transformation cannot scale one component more than another. This similarly holds if any other single computational graph $\tilde{\mA}$ was used, as the feature transformation is always jointly applied to all components with the fixed scaling of all feature channels only depending on the spectrum of $\tilde{\mA}$. For $\mA_\tsym$, we have $\mU_{:,1} = \mD^{1/2}\mathbf{1}$ which will be amplified maximally as it corresponds to the maximal eigenvalue $\lambda_1 = 1$ of $\mA_\tsym$. This leads to the well-known phenomenon of over-smoothing~\parencite{oono2020graph}.

Apart from gaining a better understanding of existing approaches by relating them to the MIMO-GC, we can also design novel methods that share similar beneficial properties to the MIMO-GC.
To amplify multiple components, such as approximations of the MIMO-GC, multiple computational graphs should be utilized. The learnable feature transformations can then control which components get amplified for each feature channel. Equivalently, we have shown that methods can employ distinct feature transformations between all pairs of nodes. Such a formulation enables the learnable parameters to be more effective, which in turn facilitates optimization and yields more informative node representations.

\subsection{Localized MIMO Graph Convolutions}
\label{sec:mc-mpnns}

Following \Cref{eq:4:mimo_gc_node}, the MIMO-GC applies a distinct feature transformation between any pair of nodes $i$ and $j$. While the MIMO-GC utilizes a fully connected graph, we propose a localized approximation that applies a distinct feature transformation between every connected pair of nodes. This is equivalent to utilizing multiple computational graphs, as shown for the MIMO-GC (\Cref{pr:4:mimo_gc}). Formally, we define the localized MIMO-GC (LMGC) as follows:

\begin{definition}[Localized MIMO Graph Convolution (LMGC)]
\label{def:lmgc}
Let $\gG=(\mA,\mX)$ be a graph in matrix form with $\mA\in\R^{n\times n}$ and $\mX\in\R^{n\times d}$. A localized MIMO graph convolution (LMGC) is a permutation equivariant function $f(\mA,\mX)\in\R^{n\times c}$ of the form
\begin{equation}
\label{eq:4:lmgc}
\begin{split}
    \bracks{f(\mA,\mX)}_{i,:} &= \sum_{j\in N_i} \mX_{j,:}\tilde{\mW}_{(i,j)}  \\
    &= \bracks{\sum_{k\in[K]} \tilde{\mA}_{(k)}\mX\mW_{(k)}}_{i,:}
    \end{split}
\end{equation}
for some $K\in\mathbb{N}$, where each $\mW_{(k)}\in\R^{d\times c}$ is a feature transformation, and the matrices $\tilde{\mA}_{(k)}\in\R^{n\times n}$ contain scalar edge weights satisfying $\bracks{\tilde{\emA}_{(k)}}_{i,j} = 0$ whenever $\mA_{i,j} = 0$. The edge-specific feature transformations are defined as $\tilde{\mW}_{(i,j)} = \sum_{k=1}^K \bracks{\tilde{\emA}_{(k)}}_{i,j}\mW_{(k)}$ for all $i,j\in[n]$.
\end{definition}

The LMGC provides a general framework for message-passing operations that are based on the MIMO-GC. Similar to the MIMO-GC, the LMGC is defined using multiple computational graphs $\tilde{\mA}_{(1)},\dots,\tilde{\mA}_{(K)}$, or equivalently, distinct feature transformations $\tilde{\mW}_{(i,j)}$ for every edge from node $j$ to node $i$. These distinct feature transformations are linear combinations of the feature transformations of each computational graph. The coefficients $\bracks{\tilde{\emA}_{(k)}}_{i,j}\in\R$ can also be interpreted as the $K$ edge weights for the connection from node $j$ to node $i$. The definition of the LMGC framework leaves the definition of these edge weights and the number of computational graphs open. The feature transformations $\mW_{(k)}$ are typically learnable parameters. 
To provide a better understanding of this framework and its generality, we now demonstrate how several existing MPNNs are specific instantiations of the LMGC framework. We start with the GCN~\parencite{kipf2017semi}, as one of the most commonly employed message-passing methods:

\begin{example}[GCN~\parencite{kipf2017semi}]
\label{ex:4:lmgc_gcn}
    Let $\mW\in\R^{d\times c}$ be a feature transformation. The function 
    \begin{equation}
        f_{\mathrm{GCN}}(\mA,\mX) = \mA_\tsym\mX\mW
    \end{equation}
    is an LMGC with $K=1$, $\tilde{\mA}_{(1)}=\mA_\tsym$, and $\mW_{(1)} = \mW$.
\end{example}

As $K=1$ for the GCN, this method exhibits SCA as shown in \Cref{pr:sca} and does not benefit from multiple computational graphs. 
For other message-passing methods, it is more straightforward to see the equivalence in the node-wise aggregation form. The SAGE convolution used within GraphSAGE~\parencite{hamilton2017inductive} is such an example:

\begin{example}[SAGE~\parencite{hamilton2017inductive}]
    Let $\mV_{(1)},\mV_{(2)}\in\R^{d\times c}$ be feature transformations. The function 
    \begin{equation}
        \bracks{f_{\mathrm{SAGE}}(\mA,\mX)}_{i,:} = \bracks{\mX\mV_{(1)}}_{i,:} + \sum_{j\in N_i} \frac{1}{d_i} \bracks{\mX\mV_{(2)}}_{j,:}
    \end{equation}
    is an LMGC with $K=2$, $\mW_{(1)} = \mV_{(1)}$, $\mW_{(2)} = \mV_{(2)}$, and $\bracks{\tilde{\emA}_{(1)}}_{i,i} = 1$ for every $i\in[n]$ and zero otherwise, and $\bracks{\tilde{\mA}_{(2)}}_{(i,j)} = \frac{1}{d_i}$ for every $i\in[n]$ and $j\in N_i$ and zero otherwise. 
\end{example}

The LMGC is also similar to multi-head attention, as used for the graph attention network (GAT)~\parencite{velickovic2017graph}. The softmax used to compute the attention coefficients constrains them and prevents the resulting matrices $\tilde{\mA}_{(k)}$ from having independent eigenvectors that can be selectively amplified:

\begin{example}[GAT~\parencite{velickovic2017graph}]
    Let $H\in\mathbb{N}$ be the number of attention heads, $\mV_{(h)}\in\R^{d\times c}$ be feature transformations, and $\alpha_{(h)}^{(i,j)}\in\R$ be attention coefficients for every $h\in[H]$ and edge $i,j\in[n]$ with $\mA_{i,j}\neq 0$. The function
    \begin{equation}
    \bracks{f_{\mathrm{GAT}}(\mA,\mX)}_{i,:} = \sum_{h\in[H]}\sum_{j\in N_i} \alpha_{(h)}^{(i,j)} \mX_{j,:}\mV_{(h)}
    \end{equation}
    is an LMGC with $K=H$, $\bracks{\tilde{\emA}_{(k)}}_{i,j} = \alpha_{(k)}^{(i,j)}$, and $\mW_{(k)}=\mV_{(k)}$ for all $k\in[K]$.
\end{example}

Any other linear message-passing step can be expressed as an equivalent LMGC, including more complex ones like the GatedGCN~\parencite{dwivedi2023benchmarking} or neural sheaf diffusion~\parencite{hansen2020sheaf}. 
The LMGC framework allows a more streamlined development of novel MPNNs and the analysis of their theoretical properties, as we only need to consider the effect of different choices used for $\bracks{\tilde{\emA}_{(k)}}_{i,j}$. 
We now study properties of LMGCs induced by specific choices of edge weights $\bracks{\tilde{\emA}_{(k)}}_{i,j}$.

\subsubsection{Theoretical Properties of LMGCs}
Several of our previous results already apply to specific instantiations of the LMGC.
Based on \Cref{pr:sca}, we have shown that any LMGC with $K=1$ exhibits the SCA property. Following \Cref{pr:4:split:rank} and \Cref{pr:4:split_independence}, we have shown that structurally independent nodes can avoid SCA. We will now further study the properties of LMGCs with specific choices for edge weights $\bracks{\tilde{\emA}_{(k)}}_{i,j}$.

While a linear message-passing step can be expressed as an LMGC, non-linear feature transformations cannot be represented by the framework. However, such transformations are utilized within the graph isomorphism network (GIN) to allow for the injectivity of each message-passing step. This is a key property of message-passing to distinguish the same set of non-isomorphic graphs as the Weisfeiler-Leman (WL) test~\parencite{weisfeiler1968reduction,xu2019how,morris2019weisfeiler}. While the GIN cannot be represented as an LMGC, we now show that the LMGC is sufficient for allowing injectivity in message-passing: 

\begin{proposition}[Injectivity]
\label{prop:4:injectivity}
    Let $\gX\subset\R^d$ be a countable set and $K \in\mathbb{N}_{>0}$. Consider the LMGC function
    \begin{equation}
    f\bracks{\vx_{(i)},\gX_i} = \sum_{\vx_{(j)}\in \gX_i} \tilde{\mW}_{(i,j)}\vx_{(j)}\, ,
    \end{equation} 
    where $\gX_i\subset\gX$ is a finite multiset and $\tilde{\mW}_{(i,j)} = \sum_{k=1}^K a^{(k)}_{(i,j)}\mW_{(k)}$ with scalar coefficients $a^{(k)}_{(i,j)}\in\R$ and matrices $\mW_{(k)}\in\R^{d\times c}$. Then, $f\bracks{\vx_{(i)},\gX_i}$ is injective with respect to the pair $\bracks{\vx_{(i)}, \gX_i}$ for a.e. choice of coefficients $a^{(k)}_{(i,j)}$ and $\mW_{(k)}$.
\end{proposition}

\begin{proof}
    To prove injectivity, we need to show that
    \begin{equation}
        f\bracks{\vx_{(i)},\gX_i} - f\bracks{\vx_{(p)},\gX_p} \neq 0
    \end{equation}
    for any $\vx_{(i)},\vx_{(p)}\in\gX$ and $\gX_i,\gX_p\subset\gX$ with $\vx_{(i)}\neq\vx_{(p)}$ or $\gX_i\neq\gX_p$.
    
    Substituting the definition of $f$, we get
    \begin{equation}
        \sum_{\vx_{(j)}\in \gX_i} \sum_{k=1}^K a^{(k)}_{(i,j)}\mW_{(k)}\vx_{(j)} - \sum_{\vx_{(q)}\in \gX_p} \sum_{k=1}^K a^{(k)}_{(p,q)}\mW_{(k)}\vx_{(q)} \neq 0\, .
    \end{equation}

    This can be rewritten as a linear combination of the feature transformations $\mW_{(k)}$: 
    \begin{equation}
\sum_{k=1}^K \mW_{(k)} \bracks{ \sum_{\vx_{(j)} \in \gX_i} a^{(k)}_{(i,j)} \vx_{(j)} - \sum_{\vx_{(q)} \in \gX_p} a^{(k)}_{(p,q)} \vx_{(q)} } \neq 0\, .
    \end{equation}
    Based on the a.e. constraint on the feature transformations $\mW_{(k)}$, this is zero if and only if the weighted sums are zero simultaneously. As such, we require
    \begin{equation}
        \sum_{\vx_{(j)}\in \gX_i} a^{(k)}_{(i,j)}\vx_{(j)} - \sum_{\vx_{(q)}\in \gX_p} a^{(k)}_{(p,q)}\vx_{(q)} \neq 0
    \end{equation}
    for at least one $k\in[K]$. This is zero only on a set of measure zero. Hence, for a.e. choice of coefficients $\{a^{(k)}_{(i,j)}\}$, we have $f(\vx_{(i)},\gX_i) - f(\vx_{(p)},\gX_p) \neq 0$.
\end{proof}
    
This injectivity result is very similar to the injectivity of the GIN~\parencite{xu2019how}. \Cref{prop:4:injectivity} confirms that linear MPNNs are sufficient to allow for injectivity in message-passing. Thus, such an LMGC achieves the same expressive power for distinguishing non-isomorphic graphs as the WL test~\parencite{weisfeiler1968reduction}, which is maximal for any MPNN~\parencite{xu2019how,morris2019weisfeiler}. This is already achievable with a single computational graph. Based on the proof, the representation resulting from each computational graph is injective. With multiple computational graphs, we show that their combination yields linearly independent representations. This allows an LMGC to avoid SCA and amplify distinct components across feature channels:

\begin{theorem}[Linear Independence]
\label{prop:4:lin_indepdence}
    Let $\gX\subset\R^d$ be a countable set and $K \in\mathbb{N}_{>0}$. Consider the LMGC function
    \begin{equation}
    f\bracks{\vx_{(i)},\gX_i} = \sum_{\vx_{(j)}\in \gX_i} \tilde{\mW}_{(i,j)}\vx_{(j)}\, ,
    \end{equation} 
    where $\gX_i\subset\gX$ is a finite multiset and $\tilde{\mW}_{(i,j)} = \sum_{k=1}^K a^{(k)}_{(i,j)}\mW_{(k)}$ with scalar coefficients $a^{(k)}_{(i,j)}\in\R$ and matrices $\mW_{(k)}\in\R^{d\times c}$. 
    Suppose $(\vx_{(i)}, \gX_i)$ and $(\vx_{(p)},\gX_p)$ satisfy $\vx_{(i)}\neq\vx_{(p)}$ or $\gX_i \neq m\cdot\gX_p$ for any $m\in\mathbb{N}$
    Then, $f\bracks{\vx_{(i)},\gX_i}$ and $f\bracks{\vx_{(p)},\gX_p}$ are linearly independent for a.e. choice of coefficients $a^{(k)}_{(i,j)}$ and $\mW^{(k)}$.
\end{theorem}

\begin{proof}
To prove linear independence, we need to show that
    \begin{equation}
        f\bracks{\vx_{(i)},\gX_i} - b\cdot f\bracks{\vx_{(p)},\gX_p} \neq 0
    \end{equation}
    for any $b\in\R$.
    Following the proof of \Cref{prop:4:injectivity}, this is equal to 
    \begin{equation}
\sum_{k=1}^K \mW_{(k)} \bracks{ \sum_{\vx_{(j)} \in \gX_i} a^{(k)}_{(i,j)} \vx_{(j)} - b\cdot \sum_{\vx_{(q)} \in \gX_p} a^{(k)}_{(p,q)} \vx_{(q)} } \neq 0\, .
    \end{equation}
    This is satisfied for a.e. $\mW_{(k)}$ if and only if
    \begin{equation}
        \sum_{\vx_{(j)} \in \gX_i} a^{(k)}_{(i,j)} \vx_{(j)} - b\cdot \sum_{\vx_{(q)} \in \gX_p} a^{(k)}_{(p,q)} \vx_{(q)} \neq 0\, .
    \end{equation}
    for at least one $k\in[K]$. For the case that not all $\vx_{(j)}\in\gX_i$ and $\vx_{(q)}\in\gX_p$ are linearly dependent, this holds for all $k\in[K]$ for a.e. coefficients $a^{(k)}_{(i,j)}$. 
    For the case that all $\vx_{(j)}\in\gX_i$ and $\vx_{(q)}\gX_p$ are linearly dependent and $K=1$, $(\vx_{(i)},\gX_i)$ and $f\bracks{\vx_{(p)},\gX_p}$ will be linearly dependent, and $b$ can be chosen accordingly. However, each computational graph requires a different $b$ for a.e. choice of coefficients. Hence, for $K>1$ we have $f\bracks{\vx_{(i)},\gX_i} - b\cdot f\bracks{\vx_{(p)},\gX_p} \neq 0$ for any $b\in\R$.  
\end{proof}

This finding shows that LMGCs avoid SCA for a.e. choice of feature transformations and edge weights for $K>1$. This enables the LMGC to yield more similar adjacent node representations across some of the feature channels, while simultaneously allowing for less similar features in other channels. This connection also highlights the close connections between improving the expressive power of an MPNN and solving the underlying phenomena behind over-smoothing and rank collapse. 
It also highlights the connection between SCA and the structural expressivity of the WL test. As linear independence subsumes injectivity, avoiding SCA and over-smoothing with such an LMGC also results in a method that is maximally expressive for message-passing methods in distinguishing non-isomorphic graphs.
While the linear independence requires $\gX_i\neq m\cdot \gX_p$ for any $m$, injectivity is still ensured for this case. Thus, in the second iteration, we will have $\vx_{(i)}\neq \vx_{(p)}$, resulting in linearly independent node representations. Linear independence further improves the optimization process, as it allows the feature transformations to control the amplification of components for each feature channel individually.

Each $a_{(k)}^{(i,j)}$ in \Cref{prop:4:injectivity} and \Cref{prop:4:lin_indepdence} can be independently obtained, e.g., by a function $a_{(k)}^{(i,j)} = \phi_k\bracks{\vx_{(i)},\vx_{(j}}\in\R$ of the node representations $\vx_{(i)}$ and $\vx_{(j)}$. Given the measure-zero set for coefficients $a_{(k)}^{(i,j)}$ for which injectivity and linear independence are not given, many functions $\phi_k$ can be defined that avoid this measure-zero set for the edge weights as their output. Neural networks and MLPs specifically can then be designed to represent or approximate such functions.
However, the methods presented as examples above do not satisfy this condition. Due to the softmax applied within attention-based methods like GAT~\parencite{velickovic2017graph} and GATv2\parencite{brody2022how}, the coefficients form a measure-zero set. The averaging prevents such methods from distinguishing multisets with different multiplicities~\parencite{xu2019how}, e.g., $\gX_{(p)} = \{\{\vx\}\}$ and $\gX_{(q)} = \{\{\vx,\vx\}\}$ for any $\vx\in\gX$. 
Other methods were proposed that utilize the tanh activation function for coefficients, which generally allows such methods to avoid measure-zero subspaces. 

Further properties of LMGCs can be similarly studied by considering the effect of coefficients $\bracks{\tilde{\emA}_{(k)}}_{(i,j)}$. To provide an example that follows the definition of the LMGC and can satisfy \Cref{prop:4:injectivity} and \Cref{prop:4:lin_indepdence}, we introduce the following instantiation of the LMGC framework.

\subsubsection{Defining an LMGC}
\label{sec:4:lmgc:instance}
Based on the definition of LMGCs in \Cref{def:lmgc}, we only need to select the number of computational graphs $K$ and the method to obtain edge weights $\bracks{\tilde{\emA}_{(k)}}_{(i,j)}$ for all $k\in[K]$ and all edges. Following \Cref{prop:4:injectivity} and \Cref{prop:4:lin_indepdence}, we apply a neural network based on the node representations of the adjacent nodes for each edge. We define the coefficients similar to GATv2 while applying a tanh activation function as the last operation, as in FAGCN~\parencite{bo2021beyond} and GGCN~\parencite{yan2022two}:
\begin{equation}
    \bracks{\tilde{\emA}_{(k)}}_{(i,j)} := \sigma_2\bracks{\sigma_1\bracks{\mX_{i,:}\mW_{(1)} || \dots ||\mX_{i,:}\mW_{(K)} ||\mX_{j,:}\mW_{(1)}||\dots ||\mX_{j,:}\mW_{(K)}}\vv_{(k)}}
\end{equation}
where $||$ is the concatention operator, $\sigma_1$ is the LeakyReLU activation function, $\sigma_2$ is the tanh activation function, $\mW_{(1)},\dots,\mW_{(K)}\in\R^{d\times c}$ are the feature transformations of the LMGC, and $\mX_{i,:},\mX_{j,:}\in\R^{1\times d}$ are the $i$-th and $j$-th row of $\mX$, respectively, and $\vv_{(k)}\in\R^{2\cdot K\cdot c}$ is a vector of learnable parameters that is distinct per computational graph. While this construction of the edge weights is designed as a combination of previous methods, future work can further study which functions $\phi_k$ are best suited to satisfy the properties required in \Cref{prop:4:injectivity} and \Cref{prop:4:lin_indepdence}.

\begin{table*}[tb]
\caption[Results of the MRS framework for different splits.]{Results of the MRS framework applied to the GCN and the ZINC12k dataset. Training and testing mean absolute error (MAE) are independently obtained using separate hyperparameter optimizations for the learning rate and number of iterations. Step time refers to the runtime for one optimization step in milliseconds (ms). Average results over three random seeds are reported.}
\label{tab:ordering}
\begin{center}
\begin{tabular}{lccc}
\toprule
\multirow{2}{*}{Method} & Step Time & \multicolumn{2}{c}{ZINC12k (MAE)} \\
& (ms) & Training & Testing \\
\midrule
GCN & $\mathbf{4.3}\pm0.1$ & $0.051\pm0.002$ & $0.404\pm0.011$\\
\midrule
$\textrm{MRS-GCN}_{\textrm{DA}}$ (random) & $5.7\pm0.1$ & $0.006\pm0.001$ & $0.623\pm0.003$\\
$\textrm{MRS-GCN}_{\textrm{DA}}$ (Features) & $5.8\pm0.4$ & $0.011\pm0.003$ & $0.390\pm0.004$\\
$\textrm{MRS-GCN}_{\textrm{DA}}$ (PPR) & $6.8\pm0.4$ & $0.010\pm0.002$ & $0.358\pm0.006$\\
$\textrm{MRS-GCN}_{\textrm{DA}}$ (Degree) & $5.8\pm0.2$ & $\mathbf{0.003}\pm0.001$ & $\mathbf{0.318}\pm0.031$\\
\bottomrule
\end{tabular}
\end{center}
\end{table*}

\begin{figure*}[tb]

     \centering
     \begin{minipage}[t]{0.49\textwidth}
         \centering
        \def\svgwidth{\textwidth}
         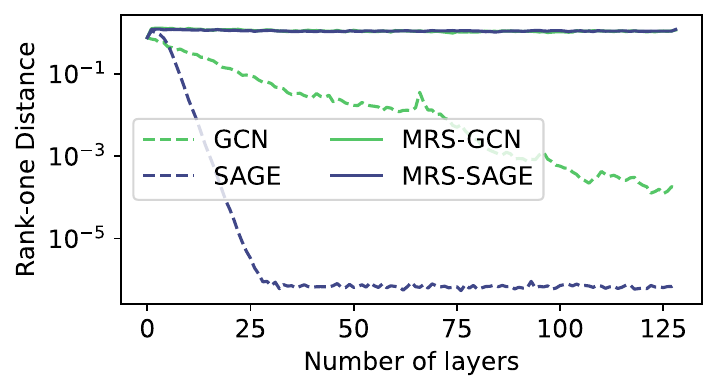

     \caption[Rank-one distance when applying the MRS framework.]{Rank-one distance (ROD) on Cora~\parencite{mccallum2000automating} after up to $128$ iterations of the GCN, SAGE, and their respective MRS versions. Average values over $50$ random parameter initializations are displayed.}
      \label{fig:dirichlet}
     \end{minipage}
     \hfill
     \begin{minipage}[t]{0.49\textwidth}
         \centering
         \def\svgwidth{\textwidth}
         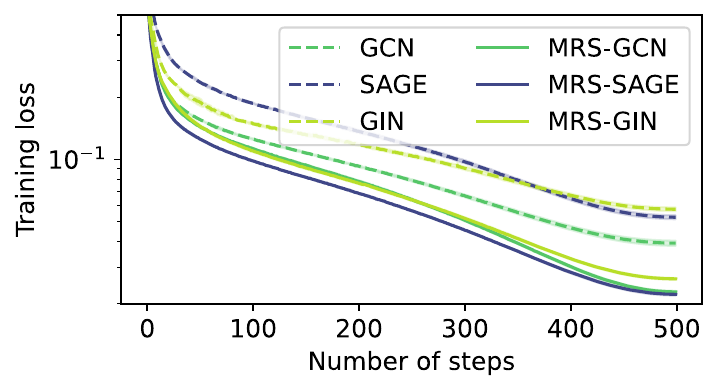
         \caption[Training error during optimization when applying the MRS framework.]{Training error during optimization when applying the MRS framework for the ZINC dataset using the methods GCN, SAGE, GIN, and their respective MRS versions. The learning rate and the number of iterations are tuned for each method. Mean values over three random initializations.}
         \label{fig:training_loss}
    \end{minipage}         
\end{figure*}

\begin{table*}[tb]
\caption[Results on ZINC for applying the MRS framework.]{Results on ZINC for applying the MRS framework to the five methods GCN~\parencite{kipf2017semi}, SAGE~\parencite{hamilton2017inductive}, GAT~\parencite{velickovic2017graph}, GIN~\parencite{xu2019how}, GatedGCN~\parencite{dwivedi2023benchmarking}. Training and testing are obtained using independent hyperparameter optimization of the number of iterations and the learning rate. Results are averaged over three random initializations. Pairwise best results in \textbf{bold}.}
\label{tab:zinc_full}
\begin{center}
\begin{tabular}{lcc}
\toprule
\multirow{2}{*}{Numer of iterations} & \multicolumn{2}{c}{ZINC (MAE)} \\ & Training & Testing\\
\midrule
GCN& $0.053\pm0.001$ & $0.155\pm0.003$ \\
$\textrm{MRS-GCN}_{\textrm{DA}}$ & $\mathbf{0.023}\pm0.000$ & $\mathbf{0.134}\pm0.001$ \\
\midrule
SAGE & $0.039\pm0.001$ & $0.123\pm0.002$ \\
$\textrm{MRS-SAGE}_{\textrm{DA}}$ & $\mathbf{0.022}\pm0.000$ & $\mathbf{0.106}\pm0.002$ \\
\midrule
GAT & $0.049\pm0.002$ & $0.149\pm0.001$\\
$\textrm{MRS-GAT}_{\textrm{DA}}$ & $\mathbf{0.026}\pm0.001$ & $\mathbf{0.128}\pm0.000$\\
\midrule
GIN & $0.058\pm0.001$ & $0.123\pm0.004$ \\
$\textrm{MRS-GIN}_{\textrm{DA}}$ & $\mathbf{0.026}\pm0.000$ & $\mathbf{0.106}\pm0.004$ \\
\midrule
GatedGCN & $0.051\pm0.002$ & $0.099\pm0.011$\\
$\textrm{MRS-GatedGCN}_{\textrm{DA}}$ & $\mathbf{0.011}\pm0.003$ & $\mathbf{0.088}\pm0.004$\\
\bottomrule
\end{tabular}
\end{center}
\end{table*}

\begin{table*}[tb]
\caption[Ablation study on ZINC12k for applying the MRS framework.]{Ablation study on ZINC12k for combining the MRS framework with residual connections (+ Res), jumping knowledge~\parencite{xu2018representation} (+ JK), and Laplacian positional encoding~\parencite{dwivedi2023benchmarking} (+ LapPE). The number of iterations indicates the number of message-passing iterations applied. The learning rate is tuned for each entry independently. The mean average error (MAE) over three random initializations is reported. Pairwise best results in \textbf{bold}.}

\label{tab:layer}
\begin{center}
\begin{tabular}{lcccccc}
\toprule
Method & 1 & 2 & 4 & 8 & 16 & 32 \\
\midrule
GCN & $0.591$ & $0.493$ & $0.421$ & $0.404$ & $0.417$ & $0.440$ \\
$\textrm{MRS-GCN}_{\textrm{DA}}$ & $\mathbf{0.525}$ & $\mathbf{0.445}$ & $\mathbf{0.343}$ & $\mathbf{0.318}$ & $\mathbf{0.318}$ & $\mathbf{0.338}$ \\
\midrule
GCN + Res & $0.567$ & $0.471$ & $0.427$ & $0.403$ & $0.370$ & $0.336$ \\
$\textrm{MRS-GCN}_{\textrm{DA}}$ + Res & $\mathbf{0.508}$ & $\mathbf{0.402}$ & $\mathbf{0.295}$ & $\mathbf{0.257}$ & $\mathbf{0.252}$ & $\mathbf{0.250}$\\
\midrule
GCN + JK & $0.588$ & $0.496$ & $0.424$ & $0.409$ & $0.413$ & $0.435$\\
$\textrm{MRS-GCN}_{\textrm{DA}}$ + JK & $\mathbf{0.526}$ & $\mathbf{0.442}$ & $\mathbf{0.324}$ & $\mathbf{0.303}$ & $\mathbf{0.311}$ & $\mathbf{0.304}$ \\
\midrule
GCN + LapPE & $0.498$ & $0.437$ & $0.392$ & $0.367$ & $0.383$ & $0.444$ \\
$\textrm{MRS-GCN}_{\textrm{DA}}$ + LapPE & $\mathbf{0.441}$ & $\mathbf{0.363}$ & $\mathbf{0.292}$ & $\mathbf{0.272}$ & $\mathbf{0.297}$ & $\mathbf{0.317}$ \\
\bottomrule
\end{tabular}
\end{center}
\end{table*}

\begin{table*}[tb]
\caption[Results on applying the MRS framework for five directed heterophilic benchmark graphs.]{Results on applying the MRS framework to the Dir-GNN~\parencite{rossi2023edge} for five directed heterophilic benchmark graphs. Equivalent to the Dir-GNN, SAGE is used for Roman-Empire, while the GCN is used for all other tasks. Reported values are the mean and standard deviation over five runs. Best results in \textbf{bold}, second-best \underline{underlined}.}
\label{tab:mrs:hetero}
\begin{center}
\footnotesize
\begin{tabular}{lccccc}
\toprule
Method & Squirrel & Chameleon & Arxiv-year & Snap-patents & Roman-Empire \\
\midrule
MLP    & $28.77\pm1.56$ & $46.21\pm2.99$ & $36.70\pm0.21$ & $31.34\pm0.05$ & $64.94\pm0.62$ \\
GCN & $53.43\pm2.01$ & $64.82\pm2.24$ & $46.02\pm0.26$ & $51.02\pm0.06$ & $73.69\pm0.74$ \\
H\_2GCN & $37.90\pm2.02$ & $59.39\pm1.98$ & $49.09\pm0.10$ & OOM & $60.11\pm0.52$ \\
GPR-GNN & $54.35\pm0.87$ & $62.85\pm2.90$ & $45.07\pm0.21$ & $40.19\pm0.03$ & $64.85\pm0.27$ \\
LINKX & $61.81\pm1.80$ & $68.42\pm1.38$ & $56.00\pm0.17$ & $61.95\pm0.12$ & $37.55\pm0.36$ \\
FSGNN & $74.10\pm1.89$ & $78.27\pm1.28$ & $50.47\pm0.21$ & $65.07\pm0.03$ & $79.92\pm0.56$ \\
ACM-GCN & $67.40\pm2.21$ & $74.76\pm2.20$ & $47.37\pm0.59$ & $55.14\pm0.16$ & $69.66\pm0.62$ \\
GloGNN & $57.88\pm1.76$ & $71.21\pm1.84$ & $54.79\pm0.25$ & $62.09\pm0.27$ & $59.63\pm0.69$ \\
Grad. Gating & $64.26\pm2.38$ & $71.40\pm2.38$ & $63.30\pm1.84$ & $69.50\pm0.39$ & $82.16\pm0.78$ \\
\midrule
DiGCN & $37.74\pm1.54$ & $52.24\pm3.65$ & OOM & OOM & $52.71\pm0.32$ \\
MagNet & $39.01\pm1.93$ & $58.22\pm2.87$ & $60.29\pm0.27$ & OOM & $88.07\pm0.27$ \\
Dir-GNN & $\underline{75.31}\pm1.92$ & $\underline{79.71}\pm1.26$ & $\underline{64.08}\pm0.30$ & $\underline{73.95}\pm0.05$ & $\underline{91.23}\pm0.32$ \\
\midrule
$\textrm{MRS-Dir-GNN}_{\textrm{DA}}$ & $\mathbf{76.01}\pm1.90$ & $\mathbf{80.17}\pm1.88$ & $\mathbf{66.03}\pm0.20$ & $\mathbf{74.72}\pm0.05$ & $\mathbf{91.87}\pm0.42$ \\
\bottomrule
\end{tabular}
\end{center}
\end{table*}

\section{Evaluation}
\label{sec:4:eval}
We now evaluate the MRS framework and the LMGC framework, both of which propose multiple computational graphs but follow different derivations. In \Cref{sec:4:eval:mrs}, we apply the MRS framework to established MPNNs and compare their empirical performance. In \Cref{sec:4:eval:lmgc}, we compare the performance of the LMGC derived from the MIMO-GC with several existing MPNNs.

\subsection{The MRS Framework}
\label{sec:4:eval:mrs}
Starting with the evaluation of the MRS framework, we empirically study the effect of applying this method to several MPNNs across various tasks. Specifically, we study the benefits of MRS on optimization and then use it on a set of benchmark tasks. This set of experiments is based on \textcite{roth2024preventing}. Our reproducible code is publicly available~\onlineresource{https://github.com/roth-andreas/splitting-computational-graphs}{https://github.com/roth-andreas/splitting-computational-graphs}.

\subsubsection{More Effective Parameters}
Based on our insights from \Cref{sec:4:split:theory}, the MRS framework enables more effective feature transformations and learnable parameters, resulting in improved optimization. We conduct several experiments and ablations using the ZINC dataset~\parencite{sterling2015zinc}. It consists of around $\num{250000}$ graphs, for which each node corresponds to an atom and the edges describe bonds between them. 
ZINC is a graph-level regression task, where the task is to predict the constrained solubility of each molecule represented as a graph.
In addition, we use ZINC12k, a subset consisting of $\num{12000}$ graphs as a computationally less consuming alternative, as proposed by \textcite{dwivedi2023benchmarking}. We use the experimental setup and implementation given by \textcite{dwivedi2022long} and the improvements proposed by \textcite{tonshoff2024where}. Following their experimental setup, each method utilizes at most $\num{500000}$ learnable parameters, and AdamW~\parencite{loshchilov2019decoupled} is employed as the optimizer. All experiments are repeated three times using random seeds, and the average and standard deviation of the results are presented.

\paragraph{The Effect of Different Partial Orderings}
Following \Cref{sec:method}, we convert an MPNN into an MRS-MPNN by defining a strict partial ordering, leading to three directed acyclic relations. Here, we consider four different strict partial orderings and study their effectiveness. 
To show that any ordering improves the optimization, we first employ a random ordering. However, messages exchanged for each relation do not exhibit any similarity, which can hurt generalization. Second, we evaluate a feature-based approach where all initial features are summed. Third, we use personalized PageRank~\parencite{page1999pagerank} as a node centrality measure. Lastly, we use the node degree as a simple and efficient ordering. We apply and compare all splits to the GCN, referring to our methods as $\textrm{MRS-GCN}_\textrm{DA}$. While these are relatively simple orderings, we aim to show the benefit of splitting the edges in general.
For each method, the learning rate and the number of layers are tuned.

Results are presented in \Cref{tab:ordering}. We observe that all MRS versions achieve a significantly reduced training loss, by up to $94\%$ for the node degree-based splitting. Even the random split achieves a similarly low training loss. This provides further evidence that any split leads to improvements in optimization. The generalization is also improved for all methods, except the random split. As there are no similarities between the edges in a random relation, the generalization performance is not improved. The node degree also achieves the lowest testing loss, reducing the error by $21\%$ compared to the GCN. The runtime is slightly higher for all MRS methods, with the degree MRS resulting in $5.8$ milliseconds, compared to $4.3$ milliseconds for each optimization step of the GCN. Based on these initial results, we will use the node degree as a strict partial ordering for all other experiments.

\paragraph{Preventing Rank Collapse}
As the goal of employing multiple computational graphs was to prevent rank collapse by dealing with SCA as one of its underlying phenomena, we now evaluate the ability of the MRS framework to prevent rank collapse. Following \Cref{def:rod} of the rank-one distance (ROD) as a metric to quantify the distance of representations to any rank-one matrix, we now measure ROD for the GCN and SAGE as two methods known to exhibit rank collapse and their respective MRS versions. Using a random graph from the ZINC dataset, we apply the respective message-passing methods interleaved with a ReLU activation function for $128$ iterations. We compute ROD after each iteration. 
We present average results across $50$ graphs in \Cref{fig:dirichlet}. While the ROD decreases quickly for the GCN and SAGE, it remains constant for MRS-GCN and MRS-SAGE across all iterations. This provides further evidence that preventing SCA also deals with rank collapse and over-smoothing.

\paragraph{MRS Applied to Various MPNNs}
To gain a better understanding of the impact of the MRS framework on message-passing, we now present results where we applied MRS to various established MPNNs. We consider the GCN~\parencite{kipf2017semi}, SAGE~\parencite{hamilton2017inductive}, GAT~\parencite{velickovic2017graph}, GIN~\parencite{xu2019how}, and GatedGCN~\parencite{dwivedi2023benchmarking}. For all methods, we optimize the learning rate and number of message-passing iterations. We present results for these methods and their respective MRS versions in \Cref{tab:zinc_full}. The lowest achieved training loss is significantly reduced for all MRS versions, confirming the benefits of optimizing our MRS-MPNNs. The testing loss is also improved in all cases, though not as significantly. This indicates that the identified optimization benefits can also lead to improvements in generalization. In \Cref{fig:training_loss}, we also visualize the training loss during optimization for the GCN, SAGE, GIN, and their respective MRS versions. It confirms that optimization is improved, as the loss is reduced more quickly, leading it to converge to a lower final value.

In addition to these message-passing methods, MPNNs typically combine these methods with other approaches like residual connections~\parencite{he2016deep,chen2020simple}, jumping knowledge~\parencite{xu2018representation}, or Laplacian positional encoding~\parencite{dwivedi2023benchmarking}. We present results for combining these methods with the GCN and the MRS-GCN in \Cref{tab:layer}, which includes detailed results for various numbers of iterations in $\{1,2,4,8,16,32\}$. In all cases, the MRS-GCN achieves better results. Crucially, as studied with SCA, improvements are already significant after a single message-passing iteration. The best result for the MRS-GCN is achieved in combination with residual connections and $32$ message-passing iterations. While for the other methods, $32$ iterations do not achieve the lowest testing error, this suggests potential for future work to investigate vanishing gradient issues. 

\subsubsection{The MRS framework and State-of-the-art methods}
With these promising results, we now also apply the MRS framework to state-of-the-art methods for complex tasks. We consider the set of large-scale directed heterophilic benchmark graphs based on the implementation of \textcite{rossi2023edge}. These datasets are the Squirrel and Chameleon graphs~\parencite{pei2020geom}, the arxiv-year and snap-patents graphs~\parencite{lim2021large}, and the roman-empire graph~\parencite{platonov2023a}. With Dir-GNN~\parencite{rossi2023edge} having achieved state-of-the-art results for these datasets by using the graph with reverse edges as a second computational graph, we now apply the MRS framework to this method to study its effectiveness for more complex methods. Dir-GNN applies Dir-GCN to squirrel, chameleon, arxiv-year, and snap-patents, and Dir-SAGE to roman-empire. We use the equivalent MRS version for each of these datasets. Following their experimental setup, we use ten fixed splits for squirrel, chameleon, and roman-empire, and five fixed splits for arxiv-year and snap-patents. Following their hyperparameters, we also optimize the learning rate, number of iterations, and dropout ratio using a grid search. We also provide baseline comparisons for ten state-of-the-art methods. 

We present average results in \Cref{tab:mrs:hetero}. The MRS-Dir-GNN consistently improves results for all tasks. We observe that the differences in results are more significant for the two larger graphs (arxiv-year and snap-patents), aligning with our previous findings that more complex functions can be represented. For squirrel, chameleon, and roman-empire, the other methods already achieve a low training error, leaving less potential for improved capabilities.

\begin{table}
\centering
\begin{tabular}{lr}
\toprule
Method  & MSE \\
\midrule
GATv2 & $0.12\pm0.04$ \\
FAGCN & $0.68\pm0.02$ \\
ACM & $0.49\pm0.02$ \\
GIN & $\underline{0.08}\pm0.03$\\
LMGC & $\mathbf{25\cdot10^{-9}}\pm86\cdot10^{-11}$ \\
\bottomrule
\end{tabular}
\caption[Minimal training error for the universality task.]{Minimal training mean squared error (MSE) for the universality task. Given random representations $\mX$ and $\mX^\prime$ for a random ER-graph, a single iteration of each of the considered methods is applied to $\mX$. The MSE between the output and $\mX^\prime$ is minimized for $\num{4000}$ optimization steps. The average and standard deviation are shown over three random seeds. The best result is in \textbf{bold}, the second-best \underline{underlined}.}
\label{tab:approx}
\end{table}

\begin{table*}[t]
\centering
\begin{tabular}{lrrr}
\toprule
& \multicolumn{3}{c}{ZINC}  \\
 Method  & Train & Test & Time per Epoch (s)\\
\midrule
GATv2 & $0.077\pm0.001$ & $0.114\pm0.004$ & $52.7\pm0.0$\\
FAGCN & $0.093\pm0.008$ & $0.130\pm0.003$ & $52.9\pm1.1$\\
ACM & $0.109\pm0.003$& $0.128\pm0.001$ & $73.1\pm0.9$\\
GIN & $\underline{0.068}\pm0.001$ & $\underline{0.088}\pm0.002$ & $\mathbf{35.5}\pm1.0$\\
LMGC & $\mathbf{0.054}\pm0.001$ & $\mathbf{0.080}\pm0.001$ & $\underline{47.5}\pm0.5$\\
\bottomrule
\end{tabular}
\caption[Results for evaluating the LMGC on ZINC]{Mean absolute error (MAE) results on the ZINC dataset. Each method uses at most $\num{100000}$ learnable parameters. Training and testing MAE are obtained using independent hyperparameter optimizations. Time per epoch is in seconds (s). Best results in \textbf{bold}, second-best \underline{underlined}.}
\label{tab:zinc}
\end{table*}

\begin{table*}[t]
\centering
\begin{tabular}{lrrr}
\toprule
& \multicolumn{3}{c}{ZINC12k}  \\
 Method  & Train & Test & Time per Epoch (s)\\
\midrule
GATv2 & $\underline{0.005}\pm0.003$ & $0.377\pm0.024$ & $2.5\pm0.1$\\
FAGCN & $0.016\pm0.006$ & $0.365\pm0.018$ & $2.5\pm0.0$\\
ACM & $0.019\pm0.005$ & $0.278\pm0.006$ & $3.4\pm0.0$\\
GIN & $0.018\pm0.009$ & $\underline{0.272}\pm0.009$ & $\mathbf{1.7}\pm0.0$ \\
LMGC & $\mathbf{0.002}\pm0.001$ & $\mathbf{0.241}\pm0.018$ & $\underline{2.4}\pm0.3$ \\
\bottomrule
\end{tabular}
\caption[Results for evaluating the LMGC on ZINC12k]{Mean absolute error (MAE) results on the ZINC12k dataset. Each method uses at most $\num{100000}$ learnable parameters. Training and testing MAE are obtained using independent hyperparameter optimizations. Time per epoch is in seconds (s). Best results in \textbf{bold}, second-best \underline{underlined}.}
\label{tab:zinc12k}
\end{table*}

\begin{table*}[t]
\footnotesize
\centering
\begin{tabular}{lrrrrr}
\toprule
Method & Basic & + LapPE & + Jumping Knowledge & + Residual & + All three \\
\midrule
GATv2 & $0.377\pm0.024$ & $0.341\pm0.040$ & $0.388\pm0.017$ & $0.311\pm0.016$ & $0.294\pm0.019$ \\
FAGCN & $0.365\pm0.018$ & $0.349\pm0.038$ & $0.352\pm0.042$ & $0.289\pm0.019$ & $0.232\pm0.012$ \\
ACM & $0.278\pm0.006$ & $0.281\pm0.019$ & $0.288\pm0.008$ & $0.266\pm0.017$ & $0.238\pm0.006$ \\
GIN & $\underline{0.272}\pm0.009$ & $\underline{0.259}\pm0.012$ & $\underline{0.267}\pm0.020$ & $\underline{0.240}\pm0.005$ & $\underline{0.228}\pm0.014$\\
LMGC & $\mathbf{0.241}\pm0.018$ & $\mathbf{0.234}\pm0.009$ & $\mathbf{0.233}\pm0.019$ & $\mathbf{0.215}\pm0.006$ & $\mathbf{0.203}\pm0.004$\\
\bottomrule
\end{tabular}
\caption[Results for an ablation study on ZINC12k]{Mean absolute error (MAE) results on the ZINC12k dataset. Each method is combined with Laplacian positional encoding (+ LapPE)~\parencite{dwivedi2023benchmarking}, jumping knowledge~\parencite{xu2018representation}, residual connections~\parencite{he2016deep}, and a combination of these three modifications. At most $\num{100000}$ learnable parameters are used. Each result is obtained using an independent hyperparameter optimization. Best results in \textbf{bold}, second-best \underline{underlined}.}
\label{tab:zinc12k_ablation}
\end{table*}

\begin{table*}[t]
\small
\centering
\begin{tabular}{lrrrrrr}
\toprule
Method & Texas & Cornell & Wisconsin & Film & Chameleon & Squirrel \\
\midrule
GATv2 & $71.6\pm1.0$ & $66.1\pm0.6$ & $79.1\pm2.0$ & $35.1\pm0.2$ & $\underline{47.1}\pm0.3$ & $\underline{35.1}\pm0.2$ \\
FAGCN & $\underline{73.5}\pm1.8$ &  $\underline{68.1}\pm1.9$ & $\underline{80.2}\pm1.8$ & $\underline{36.0}\pm0.3$ & $46.9\pm0.5$ & $34.6\pm0.3$ \\
ACM & $72.3\pm0.4$ & $65.1\pm0.7$ & $74.2\pm0.9$ & $35.8\pm0.3$ & $45.5\pm0.9$ & $34.5\pm0.1$\\
GIN & $70.5\pm1.1$ & $66.1\pm1.0$ & $79.0\pm0.6$ & $34.1\pm0.3$ & $46.1\pm0.4$ & $34.6\pm0.5$ \\
LMGC & $\mathbf{74.2}\pm2.2$ & $\mathbf{68.9}\pm2.2$ & $\mathbf{81.4}\pm1.1$ & $\mathbf{36.3}\pm0.4$ & $\mathbf{49.8}\pm0.8$ & $\mathbf{35.9}\pm0.5$ \\
\bottomrule
\end{tabular}
\caption[Results for the LMGC for six heterophilic node classification tasks.]{Obtained testing accuracies on six heterophilic node classification tasks. An independent hyperparameter optimization is performed for each entry. Each method uses at most $\num{100000}$ learnable parameters. Best results in \textbf{bold}, second-best \underline{underlined}.}
\label{tab:lmgc:hetero}
\end{table*}

\subsection{The LMGC Framework}
\label{sec:4:eval:lmgc}
As the second approach to obtain multiple computational graphs, we now evaluate the benefits of the LMGC framework. As described in \Cref{prop:4:injectivity}, the LMGC can be constructed to be injective on multisets, as GIN~\parencite{xu2019how}, while also being able to adaptively filter incoming messages similar to the attention-based message-passing methods~\parencite{velickovic2017graph,brody2022how}. At the same time, the LMGC also allows for negative edge weights as in FAGCN~\parencite{bo2021beyond}, which can act as a high-pass filter. The LMGC also prevents SCA as shown in \Cref{prop:4:lin_indepdence}, leading to more informative node representations and improved optimization.
We consider graph-level and node-level tasks to confirm these properties. These experiments are based on \textcite{roth2025what}. Our reproducible implementation is publicly available~\onlineresource{https://github.com/roth-andreas/mimo-graph-convolutions}{https://github.com/roth-andreas/mimo-graph-convolutions}.

\subsubsection{Evaluated Methods}
Based on our identified benefits of the LMGC, we compare it with the following four established and related methods. 

\paragraph{GIN} GIN~\parencite{xu2019how} ensures injectivity on multisets by applying a multi-layer perceptron on the multisets of node representations. GIN is maximally expressive for MPNNs and equivalent to the expressive power of the WL test. While this property is similar to \Cref{prop:4:injectivity}, GIN cannot be represented by an LMGC due to its non-linear feature transformation. We evaluate a standard version of the GIN that utilizes a two-layer MLP with ReLU activations after aggregation.

\paragraph{GATv2} As one effective attention-based message-passing operation, GATv2~\parencite{brody2022how} uses learnable attention coefficients as edge weights. With these, GATv2 can weight the importance of incoming messages adaptively. It utilizes $H$ attention heads. It is equivalent to an LMGC with $K=H$ computational graphs and coefficients
\begin{equation}
    \bracks{\tilde{\emA}_{(k)}}_{i,j} = \sigma_2\bracks{\sigma_1\bracks{\mX_{i,:}\mW_{(k)} + \mX_{j,:}\mW_{(k)}}\vv_{(k)}}
\end{equation}
where $\mW_{(k)}\in\R^{d\times c}$ are the feature transformations, $\vv_{(k)}\in\R^c$ is a vector, $\sigma_1$ is the LeakyReLU activation function and $\sigma_2$ is a node and head-wise softmax activation. To ensure fairness, we employ $K=4$ computational graphs for GATv2 and our LMGC instantiation.

\paragraph{ACM} As one method that has been proposed to utilize multiple computational graphs, we evaluate the adaptive-channel mixing (ACM) method~\parencite{luan2022revisiting}. They propose to use a normalized adjacency matrix to amplify low-frequency components, a normalized graph Laplacian to amplify high-frequency components, and a residual connection as three computational graphs. Expressed in the LMGC framework this corresponds to setting $K=3$ and selecting the matrices of coefficients as $\mA^{(1)} = \mA_\tsym$, $\mA^{(2)} = \mL_\tsym$ and $\mA^{(3)} = \mI_n$. To ensure fairness, we only use the third computational graph when we use residual connections for all methods.

\paragraph{FAGCN} The frequency adaptation GCN (FAGCN)~\parencite{bo2021beyond} allows for negative edge weights, which can act as a high-pass filter. Nodes with a negative edge weight can result in more distant representations after aggregation. This method was specifically designed for heterophilic graphs, where high-frequency components are beneficial. Stated in the LMGC framework, this corresponds to setting $K=1$ and defining the coefficients as
\begin{equation}
    \tilde{\emA}^{(1)}_{(i,j)}=\frac{\sigma\bracks{[\mX_{i,:}||\mX_{j,:}]\vv}}{\sqrt{d_i}\sqrt{d_j}}
\end{equation}
where $\vv\in\R^{2d}$ is a vector of learnable parameters, $\sigma$ is the tanh activation function and $d_i,d_j$ are the degrees of nodes $i$ and $j$, respectively.

\subsubsection{Universality}
Based on \Cref{pr:4:mimo_gc:universality}, the MIMO-GC can represent any function that maps almost any node representation matrix $\mX$ to any other node representation matrix $\mX^\prime$. To evaluate the ability of our considered MPNNs to represent such a mapping, we empirically evaluate their capacity to fit such a function. Given a random Erdős–Rényi graph~\parencite{erdos1959on} with $n$ nodes, we sample $\mX\in\R^{n\times d}$ and $\mX^\prime\in\R^{n\times d}$ element-wise from a standard normal distribution. We apply a single message-passing iteration, and minimize the mean-squared error (MSE) between the output and $\mX^\prime$ for $\num{4000}$ optimization steps using the Adam optimizer. We consider $n=16$ nodes, an edge probability of $p=10\%$, a feature dimension of $d=16$, and tune the learning rate in $\{0.03,0.01,0.003\}$.

We present average results over three random seeds in \Cref{tab:approx}. While GIN achieves the lowest optimization error of the four baselines with $0.08$, the LMGC achieves a significantly reduced error of $25\cdot10^{-9}$. This provides further evidence of the optimization benefits of multiple computational graphs.

\subsubsection{Graph Regression}
Now, we evaluate the considered methods for graph-level tasks. As in \Cref{sec:4:eval:mrs}, we consider the ZINC~\parencite{sterling2015zinc} and ZINC12k dataset as a subset of $\num{12000}$ graphs~\parencite{dwivedi2023benchmarking}. For these tasks, the maximal expressivity of GIN allows it to achieve strong results. Based on \Cref{prop:4:injectivity}, the LMGC is theoretically able to match this expressive power. As in \Cref{sec:4:eval:mrs}, we use the implementation of \textcite{tonshoff2024where} based on the Long Range Graph Benchmark~\parencite{dwivedi2022long}. For all five methods, we perform the same hyperparameter optimization on the number of iterations in $\{6,8,10\}$ and the learning rate in $\{0.001,0.0003,0.0001\}$ as proposed by \textcite{tonshoff2024where}. Each combination is repeated for four random seeds. Each model uses at most $\num{100000}$ parameters.

Results for ZINC and ZINC12k are presented in \Cref{tab:zinc} and \Cref{tab:zinc12k}, respectively. The LMGC achieves the best results for training and testing sets, with the GIN achieving roughly $10\%$ lower scores, and all other methods significantly worse due to their lower expressivity. An epoch for ZINC takes GIN $35.5$ seconds while the LMGC takes $47.5$ seconds, which is slightly faster compared to GATv2.

We present an additional ablation study for ZINC12k in \Cref{tab:zinc12k_ablation}. We combine all methods with Laplacian positional encoding~\parencite{kreuzer2021rethinking}, jumping knowledge~\parencite{xu2018representation}, and residual connections~\parencite{he2016deep}, and a combination of all three of these methods. The LMGC also achieves the lowest testing errors for all these combinations. These results provide further evidence that the LMGC can match the expressivity of GIN.

\subsubsection{Node Classification}
For node-level tasks, the ability of attention-based methods to filter information and the ability of the FAGCN and ACM to amplify high-frequency components are essential. We consider the six heterophilic graph benchmark datasets Texas, Cornell, Wisconsin, Film, Chameleon, and Squirrel, with ten data splits, as proposed by \textcite{pei2020geom}. We integrate our considered methods into the implementation provided by \textcite{rusch2023gradient}. Each model uses at most $\num{100000}$ parameters. The learning rate is tuned in $\{0.01,0.003,0.001\}$ and dropout ratio in $\{0.0,0.25,0.5\}$ using a grid search. For the selected hyperparameters for each method, all ten splits are rerun five times, and the average results are presented.

These average results are presented in \Cref{tab:lmgc:hetero}. While the LMGC achieves the best results for all considered tasks, the differences here are less significant. These datasets are challenging due to their heterophilic nature but are relatively small-scale. Due to the beneficial optimization properties of the LMGC, results for larger datasets are expected to improve even further.
    \cleardoublepage
    \clearpage
\chapter{Preventing Component Dominance Based on Personalized PageRank}
\label{chap:5}
\textit{
Many message-passing neural networks (MPNNs) tend to achieve worse performance across various tasks when the number of message-passing iterations is increased. 
We identified component dominance (CD) as one of multiple related phenomena, referring to a single component of the initial representations getting dominantly amplified. This leads to a loss of all other information, and node representations become increasingly independent of the initial features. 
In this chapter, we establish the connection between CD and the PageRank algorithm. As personalized PageRank was proposed as a solution for a similar shortcoming of PageRank, we adapt these modifications for MPNNs. This results in the personalized PageRank graph neural network (PPRGNN), a method that allows for infinitely many message-passing iterations, while maintaining information from the initial representations. We propose an efficient method for performing a fixed-point iteration with an absolute stopping criterion to obtain the solution. This approach further improves our understanding of the CD phenomenon in MPNNs and provides an effective solution for it.
}
\section{Introduction}
When processing graph-structured data, relevant information often needs to be combined from distant, non-adjacent nodes~\parencite{stachenfeld2021graph,rampasek2021hierarchical,alon2021on,freitas2021a,dwivedi2022long}. Message-passing neural networks (MPNNs), a widely used framework for such tasks, iteratively aggregate the representations of the neighbors of each node. Due to their localized nature, $l$ message-passing iterations capture information only from the $l$-hop neighborhood of each node. As such, multiple iterations need to be performed to combine distant information. Additionally, non-linear activation functions are typically applied only after each message-passing iteration. Therefore, more iterations allow the extraction of more complex node representations. At the same time, the resulting representations should preserve information about the local neighborhood of each node and its initial features.

However, the performance of many MPNNs has been found to degrade when more message-passing iterations are applied. In the literature, over-smoothing is frequently mentioned as a phenomenon related to this performance degradation~\parencite{li2018deeper,oono2020graph}. Our theoretical study in \Cref{sec:understanding} revealed that many MPNNs exhibit shared component amplification (SCA) (\Cref{def:sca}) and component dominance (CD) (\Cref{def:component_dominance}), two properties that underlie over-smoothing and rank collapse.
In \Cref{chap:4}, we extensively studied the design of MPNNs that avoid SCA and demonstrated their effectiveness in representing more complex functions.

In this chapter, we focus on designing methods that mitigate CD in MPNNs.
Specifically, we relate CD to the PageRank algorithm~\parencite{page1999pagerank}, where a similar phenomenon has been identified. 
This connection to a well-established algorithm with rich theoretical foundations provides a better understanding and intuition for CD in MPNNs. It allows the design of novel MPNNs inspired by modifications proposed for PageRank.
To address this issue in PageRank, the personalized PageRank (PPR)~\parencite{page1999pagerank} modification was proposed, which avoids this property. 
We adapt this principle to MPNNs by combining the result of message-passing iterations with the initial node features. We guarantee convergence of node representations to a steady state by using the geometric series of the exponential function. This allows the use of infinitely many message-passing iterations, while ensuring convergence and that the initial node representations affect the final node representations. We introduce an efficient fixed-point iteration scheme with an absolute stopping criterion for computing the final node representations. This chapter is based on \textcite{roth2022transforming}.

\section{Related Work}
We now review relevant work on dealing with CD and the connection between MPNNs and PageRank. In \Cref{sec:related:pagerank}, we introduce the PageRank algorithm and personalized PageRank (PPR) as a modification, which forms the basis of our approach. We then provide an overview of related graph neural network methods.

\subsection{PageRank and Personalized PageRank}
\label{sec:related:pagerank}
PageRank~\parencite{page1999pagerank} was initially proposed as an algorithm to quantify the importance of web pages. The key idea of PageRank is to assign higher scores to web pages linked to by many other relevant web pages, and a lower score otherwise.
In PageRank, the set of $n$ web pages is modeled as a directed graph, where each node represents a web page and each edge corresponds to a link from one web page to another. PageRank utilizes the column-stochastic normalized adjacency matrix $\mA_{\mathrm{pr}} = \mA\mD^{-1}\in\R^{n\times n}$, where $\mA\in\{0,1\}$ is the adjacency matrix indicating links, and $\mD\in\R^{n\times n}$ is the diagonal out-degree matrix. For a simplified version of PageRank, the relevance scores are given by the fixed point of
\begin{equation}
    \vr = \mA_{\mathrm{pr}}\vr\, ,
\end{equation}
where $\vr\in\R^n$ is the vector containing the relevance scores.
This solution exists as $\mA_{\mathrm{pr}}$ has a leading eigenvalue $\lambda_1 = 1$ with strictly greater magnitude than all other eigenvalues, and $\vr$ is a corresponding eigenvector of $\mA_{\mathrm{pr}}$~\parencite{page1999pagerank}. Although $\vr$ can be computed in closed form, it is commonly obtained efficiently via power iteration using the iterative update step
\begin{equation}
\label{eq:pr_lim}
    \vr^{(k)} = \mA_{\mathrm{pr}}\vr^{(k-1)}\, ,
\end{equation}
initialized with some $\vr^{(0)}\in\R^n$. As $k\to\infty$, $\vr^{(k)}$ converges to a scaled version of $\vr$, i.e., $\lim_{k\to\infty} \vr^{(k)} = c\cdot\vr$ for some scalar $c\in\R$ depending on $\vr^{(0)}$.
As a well-established algorithm, PageRank has a rich theoretical foundation, many adaptations, and intuitive interpretations. A common view is the probabilistic interpretation, also called the random surfer model~\parencite{page1999pagerank}. $\mA_{\mathrm{pr}}$ represents a stochastic transition matrix on the graph with uniform transition probabilities over outgoing edges from each node. In this interpretation, a surfer randomly clicks links on each webpage.
When repeatedly clicking on random links, the PageRank scores correspond to the probability of being at a given web page. With an increased number of clicks or iterations, these probabilities become increasingly independent of the initial web page. The matrix $\mA_{\mathrm{pr}}$ can also be viewed as a random walk matrix, where $k$ denotes the length of the random walk. For almost every initial distribution $\vr^{(0)}$, PageRank results in the same stationary distribution $\vr$. Thus, this version of PageRank does not allow a prior distribution $\vr^{(0)}$ to influence the resulting scores. To address this limitation, \textcite{page1999pagerank} also introduced the PPR algorithm, which introduces a restart probability $\alpha\in(0,1)$. The relevance scores are equal to the fixed point of the equation
\begin{equation}
\label{eq:ppr}
    \vs = (1-\alpha)\cdot\mA_{\mathrm{pr}}\vs + \alpha\cdot\vr^{(0)}\, ,
\end{equation}
where $\vs\in\R^n$ is the vector of resulting scores. The parameter $\alpha$ is interpreted as the probability of restarting the random walk, often referred to as a teleportation probability.
As a result, $\vs$ is a combination of random walks of different lengths, each starting with the initial distribution $\vr^{(0)}$. This becomes more intuitive when reformulating Equation~\ref{eq:ppr} as a truncated geometric series applied to vector $\vr^{(0)}$, i.e., 
\begin{equation}
\label{eq:ppr_geometric}
    \vs^{(k)} = \sum_{i=0}^k \alpha\cdot(1-\alpha)^i\cdot\mA_{\mathrm{pr}}^i\vr^{(0)}\, .
\end{equation}
The $i$-th term corresponds to a random walk of length $i$. 
As with PageRank, PPR converges when taking the limit of infinitely many iterations, i.e., $\lim_{k\to\infty}\vs^{(k)} = \vs$.
Since $0<1-\alpha<1$, the contribution of longer random walks diminishes, ensuring convergence.

\subsection{Personalized PageRank for Graph Neural Networks}
\label{sec:related}
We identified two primary approaches at the intersection of PPR and message passing in graph neural networks (GNNs). We will now provide details about these proposed methods.

\subsubsection{APPNP.}
\textcite{klicpera2019predict} initially identified the connection between PageRank and graph neural networks (GNNs), noting that the property of losing locality in PageRank occurs similarly in GNNs. Due to this close similarity, they proposed replacing message-passing iterations with the PPR algorithm (Equation~\ref{eq:ppr}). Given initial node features $\mX\in\R^{n\times d}$, they define the teleportation state as a node-wise transformed state $\mH^{(0)} = g(\mX)$, where $g$ is instantiated as a row-wise multi-layer perceptron (MLP)~\parencite{amari1967a}. 
Using $\mH^{(0)}$ as the initial distribution, they apply PPR using iterations of the form 
\begin{equation}
\label{eq:appnp}
    \mH^{(k)} = (1-\alpha)\cdot\mA_\tsym\mH^{(k-1)} + \alpha\cdot\mH^{(0)}\, ,
\end{equation}
where the symmetrically normalized adjacency matrix $\mA_\tsym$ is chosen instead of $\mA_{\mathrm{pr}}$ in PPR to better align with the graph convolutional network (GCN)~\parencite{kipf2017semi}.
Similar to PPR, $\mH^{(k)}$ converges to a fixed point as $k\to\infty$.
They approximate this limit by applying a fixed number of iterations, which they refer to as approximate personalized propagation of neural predictions (APPNP). 
As with PPR, the parameter $\alpha\in(0,1)$ serves as the restart probability. 
Notably, APPNP does not contain any learnable parameters in the propagation steps. Consequently, given a fixed $\mH^{(0)}$, the resulting state is unaffected by the optimization process, limiting the effect of learnable parameters.

\subsubsection{Implicit Graph Neural Networks.}
The implicit framework~\parencite{ghaoui2021implicit} obtains representations via fixed-point equations. While proposed independently of the connection to PPR, the implicit graph neural network (IGNN)~\parencite{gu2020implicit} defines such a fixed-point equation for graph-structured data. Specifically, they propose the non-linear fixed-point equation
\begin{equation}
\label{eq:ignn}
    \mH = \phi\bracks{\mA_\tsym\mH\mW + \mH^{(0)}}
\end{equation}
where $\phi$ is a component-wise non-expansive function, $\mA_\tsym$ is the symmetrically normalized adjacency matrix, $\mW\in\R^{d\times d}$ is a feature transformation matrix, and $\mH^{(0)} = g(\mX)$ is a node-wise transformation of the initial node features $\mX$. For better intuition, when $\phi$ is the identity function, \Cref{eq:ignn} can be expressed iteratively as 
\begin{equation}
    \mH^{(k)} = \sum_{i=0}^k \mA_\tsym^i\mH^{(0)}\mW^i
\end{equation}
which reveals a geometric series similar to Equation~\ref{eq:ppr_geometric}. As $k\to\infty$, $\mH^{(k)}$ converges to the fixed-point $\mH = \lim_{k\to\infty}\mH^{(k)}$ under conditions we describe in the following.

\textcite{gu2020implicit} prove that a fixed-point solution exists if the largest eigenvalue $\lambda_1$ of the matrix $\mW^\top\otimes\mA_\tsym\in\R^{nd\times nd}$ defined by the Kronecker product $\otimes$ has magnitude less than one.
Since $\mA_\tsym$ is a fixed matrix with a unique eigenvalue of maximum magnitude by the Perron-Frobenius Theorem~\parencite{berman1994nonnegative}, \textcite{gu2020implicit} propose to constrain $\mW$ such that the maximum eigenvalue in magnitude $\lambda_1$ satisfies $|\lambda_1^{(\mW)}|<1$. Because optimization of $\mW$ with gradient descent does not guarantee $|\lambda_1|<1$, they use projected gradient descent~\parencite{goldstein1964convex}. After each gradient descent step, the updated parameter matrix $\mW$ is projected onto the closest matrix in the allowed set of feature transformations, i.e., the set of matrices $\mW$ for which $|\lambda_1|<1$. Since this set forms an $\ell_1$-ball, efficient projection algorithms exist~\parencite{duchi2008efficient}. They propose to obtain $\mH$ by fixed-point iteration with a stopping criterion.

However, since the space of parameter matrices is restricted to a subspace of $\R^{d\times d}$, the fixed-point solutions are similarly constrained. Projected gradient descent further limits the effectiveness of each optimization step because it may alter the direction away from the steepest descent and reduce the step size. Additionally, the projection increases the complexity of the method and complicates the implementation. It is also challenging to combine this method with other approaches, as choosing different aggregation matrices $\tilde{\mA}$ instead of $\mA_\tsym$ changes the eigenvalue constraints imposed on $\mW$. 

\subsection{Approaches Related to Component Dominance}
Several other modifications to MPNNs have been proposed to prevent information loss in models with many message-passing iterations. Although the connection to PPR and CD is less direct, these approaches target similar limitations of MPNNs.
Many methods use residual connections to maintain information from previous states~\parencite{bresson2017residual,klicpera2019predict,li2020deepergcn,chen2020simple,li2021training,scholkemper2025residual}. Other approaches combine node representations from all iterations into joint representations~\parencite{xu2018representation,fey2019just}. Various gating methods enable models to selectively discard information from neighboring representations~\parencite{li2016gated,bresson2017residual,rusch2023gradient,dwivedi2023benchmarking,finkelshtein2024cooperative}. Other approaches found success in rescaling the feature transformation matrix $\mW$~\parencite{zhao2020pairnorm,oono2020graph}. \textcite{jin2022feature} propose regularizing correlations between feature channels. 
Following our definition of CD in \Cref{def:component_dominance}, these methods can similarly be interpreted as targeting CD specifically, rather than over-smoothing more broadly.
\section{Transforming PageRank into an Infinite-Depth Graph Neural Network}
We first detail the connection between PageRank and the component dominance (CD) property exhibited by many MPNNs. We then show how PPR can be adapted for MPNNs to ensure locality of node representations and prevent CD. We also provide details on convergence guarantees, how to efficiently obtain node representations, and how to perform optimization efficiently. 

\subsection{MPNNs and PageRank}
Many established message-passing operations can be formulated as
\begin{equation}
\label{eq:5:mp}
    \mX^{(k)} = \tilde{\mA}\mX^{(k-1)}\mW
\end{equation}
where $\mX^{(k-1)}\in\R^{n\times d}$ describes the current state of each node, $\tilde{\mA}\in\R^{n\times n}$ represents the structure of the graph, possibly with arbitrary edge weights for every connected pair of nodes. The matrix $\mW\in\R^{d\times d}$ is a linear feature transformation that we assume to be shared across all iterations.
Such message-passing operations can be equivalently expressed as the matrix-vector product 
\begin{equation}
\label{eq:mp_pr}
    \tvec\left(\tilde{\mA}\mX^{(k-1)}\mW\right) = \mT\vx^{(k-1)}
\end{equation}
where $\vx^{(k-1)} = \tvec\left(\mX^{(k-1)}\right)$ is the vectorized form of the node representations and $\mT = \mW^\top\otimes\tilde{\mA}\in\R^{nd\times nd}$ combines the graph structure and the feature transformation as a single matrix using general properties of the Kronecker product $\otimes$. 

In \Cref{pr:cd}, we have proven that all such message-passing steps $\mT$ exhibit component dominance, as defined in \Cref{def:component_dominance}. CD refers to the phenomenon that a single component of $\mX^{(0)}$ gets increasingly amplified with more iterations. Specifically, the components of $\mX^{(0)}$ corresponding to the eigenvalues with maximum magnitude of $\mT$ get dominantly amplified. We have also provided the connection to power iteration in \Cref{sec:understanding:rank:power}. As \Cref{eq:mp_pr} can be equivalently written as $\vx^{(k)} = \mT^k\vx^{(0)}$ using the matrix power $\mT^k$, power iteration also describes the property that the state $\vx^{(k)}$ gets increasingly dominated by the eigenvector of $\mT$ corresponding to the largest eigenvalue in magnitude.

These phenomena are also closely related to PageRank as given in \Cref{eq:pr_lim}. \Cref{eq:mp_pr} can be seen as applying a transition function $\mT$ to the scores $\vx^{(k-1)}$, though the entries of $\mT$ may be negative and may not sum to one for any column. As the eigenvalue with maximal magnitude of $\mT$ may also be negative and have magnitude greater than one, convergence of $\mX^{(k)}$ for $k\to\infty$ is typically not given. As such, applying PPR with transition matrix $\mT$ would also not ensure convergence. The various interpretations of PageRank still help our understanding of the phenomenon. Repeatedly applying a transition function $\mT$ results in a state given by a random walk of corresponding length, although the transition scores are not probabilistic. These random walks increasingly lose locality of their starting point, and $\mX^{(k)}$ becomes increasingly independent of the initial representations $\mX^{(0)}$. Combining this with our theoretical study in \Cref{sec:understanding} and connection to power iteration in \Cref{sec:understanding:rank:power}, the node representations $\mX^{(k)}$ increasingly align with the dominant eigenvector of $\mT$. This leads to a loss of information, as the other components have a diminishing effect on $\mX^{(k)}$.

\subsection{Personalized PageRank Graph Neural Network}
\label{sec:methods}
Given the close connection between PageRank~\parencite{page1999pagerank} and applying message-passing iterations, we want to investigate further how we can utilize the benefits of PPR to similarly allow for infinitely many message-passing iterations while retaining local information for every node state. Based on PPR, we want to add a chance to teleport to the initial state, which is equivalent to restarting the message-passing process from the initial state. When ensuring convergence of the message-passing process with depth, we ensure that the initial state always contributes to the final state, which thus ensures locality in the representation of each node. As the personalization matrix, we use the transformed initial state $\mH^{(0)} = f_\theta(\mX)$, as proposed by \textcite{klicpera2019predict}.

Following \Cref{eq:mp_pr}, here we consider the message-passing step $\mT = \mW^\top\otimes\tilde{\mA}$ instead of $\mA_{\mathrm{pr}}$ as used for PPR as given in \Cref{eq:ppr_geometric}. Equivalently to the PPR reformulation of PageRank, for message-passing operations of the form in \Cref{eq:5:mp}, this would lead to
\begin{equation}
\label{eq:ppr_gcn}
    \mH^{(k)} = (1-\alpha)\cdot\tilde{\mA}\mH^{(k-1)}\mW + \alpha\cdot\mH^{(0)}\, .
\end{equation}
However, for any fixed $\alpha$, convergence could not be guaranteed as the largest eigenvalue of $(1-\alpha)(\mW^\top\otimes\tilde{\mA})$ may be larger than one. This can cause the representations from longer message-passing iterations to dominate the final representation. The constant initial state $\mH^{(0)}$ would then have an increasingly small effect on that resulting representation, and we would still have a loss of information. IGNN resolves this by constraining the eigenvalues of $\tilde{\mA}$ and $\mW$. We want to allow any unconstrained $\mW$ and $\tilde{\mA}$, representing any aggregation function with arbitrary edge weights. While the geometric series given in Equation~\ref{eq:ppr_geometric} only converges when $\mT = \mW^\top\otimes\tilde{\mA}$ has maximum eigenvalue less than one for any $\alpha\in(0,1)$, there may not exist a fixed $\alpha\in(0,1)$ that guarantees the spectral radius of $(1-\alpha)\mT$ to be less than one. However, we propose to ensure convergence of $\mT^k$ using the exponential function formula
\begin{equation}
\label{eq:exp_series}
    \tvec\bracks{\mH} = \lim_{l\to\infty}\sum_{k=0}^l\frac{\mT^k}{k!}\tvec\bracks{\mH^{(0)}}
\end{equation}
applied to the transformed initial state $\mH^{(0)} = f_\theta(\mX)$. Note that this convergence holds for any $\mT$, including any graph structure or aggregation function, any learnable parameters, and even any number of computational graphs.

However, restating \Cref{eq:ppr_gcn} based on this geometric series cannot be done, as any fixed $\alpha\in\R$ can only lead to exponential decay but not factorial decay.
Instead, to achieve the factorial decay $k!$ in Equation~\ref{eq:exp_series}, the chance of teleporting back to the initial state needs to increase with $k$. Equivalently, the chance of extending the random walk needs to be reduced. We observe that $\frac{1}{k!} = \frac{1}{k \cdot (k-1)\dots 1}$ can be decomposed into such a product of increasing values. An iteration-dependent coefficient $\alpha^{(k)}$ can achieve this. 

Since the longest random walk starts from the first iteration $k=0$, the corresponding coefficient $\alpha^{(0)}$ should be the smallest to ensure proper weighting. For a fixed number of total iterations, this can be directly done. Let $l$ be the total number of iterations. We define the representations iteratively as
\begin{equation}
\label{def:pprgnn}
    \mH^{(k,l)} = \phi\bracks{\alpha^{(k,l)}\tilde{\mA}\mH^{(k-1,l)}\mW+\mH^{(0)}}
\end{equation}
where $\alpha^{(k,l)} = \frac{1}{\bracks{1+l-k}\epsilon}$ for $k,l\in\mathbb{N}$ aligns with our ideas. We added a tunable parameter $\epsilon\in\R$ to control the decay and speed of convergence. We also added a component-wise activation function $\phi$. 
We refer to this method as the personalized PageRank graph neural network (PPRGNN).
By setting $\phi$ as the identity function and $\epsilon=1$, we recover Equation~\ref{eq:exp_series} exactly with $\mH = \lim_{l\to\infty} \mH^{(l,l)}$. We formally prove that the sequence $\bracks{\mH^{(l,l)}}_{l\in\mathbb{N}}$ converges to a unique solution when $l\to\infty$ in the following theorem:

\begin{theorem}
\label{pr:pprgnn_converges}
    The sequence $\bracks{\mH^{(l,l)}}_{l\in\mathbb{N}}$ as defined by \Cref{def:pprgnn} converges to a unique solution as $l\to\infty$ for any $\mH^{(0,0)}\in\R^{n\times d}$, $\tilde{\mA}\in\R^{n\times n}$, $\mW\in\R^{d\times d}$, $\mH^{(0)}\in\R^{n\times d}$ and $\alpha^{(k,l)} = \frac{1}{(1+l-k)\epsilon}$ with any $\epsilon>0$, and any Lipschitz continuous activation function $\phi$ with Lipschitz constant $c<\infty$.
\end{theorem}

\begin{proof}
    We first show that $\mH^{(l,l)}$ converges as $l\to\infty$ by proving that 
    \begin{equation}
         \lim_{l\to\infty} \norm{\mH^{(l,l)} - \mH^{(l-1,l-1)}}_F = 0\, .
    \end{equation} 
    We first substitute $\mH^{(l,l)}$ and $\mH^{(l-1,l-1)}$ with their definitions, resulting in
    \begin{multline}
        \norm{\mH^{(l,l)} - \mH^{(l-1,l-1)}}_F \\= \norm{\phi\left(\alpha^{(l,l)}\tilde{\mA}\mH^{(l-1,l)}\mW+\mH^{(0)}\right) - \phi\left(\alpha^{(l-1,l-1)}\tilde{\mA}\mH^{(l-2,l-1)}\mW+\mH^{(0)}\right)}_F\, .
    \end{multline}
    As $\phi$ is Lipschitz continuous for some $c$ by assumption, we obtain
    \begin{equation}
        \norm{\mH^{(l,l)} - \mH^{(l-1,l-1)}}_F \leq c\cdot \norm{\alpha^{(l,l)}\tilde{\mA}\mH^{(l-1,l)}\mW - \alpha^{(l-1,l-1)}\tilde{\mA}\mH^{(l-2,l-1)}\mW}_F\, .
    \end{equation}
    As $\tilde{\mA}$ and $\mW$ are applied to both terms, these can be factored out jointly using the Kronecker product $\mT=\mW^\top\otimes\tilde{\mA}$ and the vectorized form, resulting in
    \begin{equation}
        \norm{\mH^{(l,l)} - \mH^{(l-1,l-1)}}_F \leq c\cdot\norm{\mT\left(\alpha^{(l,l)}\tvec(\mH^{(l-1,l)})- \alpha^{(l-1,l-1)}\tvec(\mH^{(l-2,l-1)})\right)}_F\, .
    \end{equation}
    As we have $\alpha^{(l,l)} = \alpha^{(l-1,l-1)}$, we further have
    \begin{equation}
        \norm{\mH^{(l,l)} - \mH^{(l-1,l-1)}}_F \leq c\cdot\alpha^{(l,l)}\cdot\norm{\mT}_F\cdot\norm{\mH^{(l-1,l)}-\mH^{(l-2,l-1)}}_F
    \end{equation}
    More generally, we also have $\alpha^{(l-k,l)} = \alpha^{(l-k-1,l-1)} = \frac{1}{(1+k)\epsilon}$. By iterative application, we obtain
    \begin{equation}
        \norm{\mH^{(l,l)} - \mH^{(l-1,l-1)}}_F \leq c^{l-1}\cdot\prod^{l-2}_{k=0}\alpha^{(l-k,l)}\cdot\norm{\mT}_F^{l-1}\cdot\norm{\mH^{(1,l)}-\mH^{(0,l-1)}}_F\, .
    \end{equation}
    As $\lim_{l\to\infty}\prod^{l-2}_{k=0} \alpha^{(l-k,l)}\norm{\mT}_F^{l-1} = \lim_{l\to\infty}\frac{1}{(l-1)!\epsilon^l}\norm{\mT}_F^{l-1} = 0$ and for any $\mH^{(0,l-1)}$ and $\mH^{(0,l)}$ there exists a $c_2\in\R$ for which $\norm{\mH^{(1,l)}-\mH^{(0,l-1)}}_F\leq c_2$. This proves the convergence of the sequence. 

    To prove the uniqueness of the solution, it suffices to note that the above proof of convergence holds independently of the initial states $\mH^{(0,l-1)}$ and $\mH^{(0,l)}$.
\end{proof}

As we have established convergence of the PPRGNN when taking infinitely many iterations, there exists a limit representation $\mH = \lim_{l\to\infty} \mH^{(l,l)}$. We can obtain $\mH$ using fixed-point iteration with an absolute stopping criterion as the representation $\mH^{(l,l)}$ with index
\begin{equation}
    l = \min_{k\in\mathbb{N}}\norm{\mH^{(k,k)} - \mH^{(k-1,k-1)}} < \gamma
\end{equation}
for some convergence threshold $\gamma>0$. However, we cannot utilize any intermediate results of computing $\mH^{(k-1,k-1)}$ to compute $\mH^{(k,k)}$, as it requires a full recomputation, each $\mH^{(k,k)}$ having complexity linear in $k$. 
Instead, we approximate $l$ by determining at which length a term $\frac{\mT^k}{k!}\tvec(\mX^{(0)})$ has a negligible impact on the result. We do this by ignoring the restart term $\mH^{(0)}$ and considering the recursive equation
\begin{equation}
\label{eq:5:single_walk}
    \mG^{(k)} = \phi\bracks{\alpha^{(1,k)}\tilde{\mA}\mG^{(k-1)}\mW}
\end{equation}
with $\mG^{(0)} = \mH^{(0)}$. The benefit of this is that it can be computed independently of $l$ and thus requires a single forward pass. We formally prove the convergence of $\mG^{(k)}$ as $k\to\infty$ in the following proposition:

\begin{proposition}
Let $\mG^{(k)} = \phi\bracks{\alpha^{(1,k)}\tilde{\mA}\mG^{(k-1)}\mW}$ with $\alpha^{(1,k)} = \frac{1}{k\epsilon}$ for any $\epsilon>0$, any $\tilde{\mA}\in\R^{n\times n}$, $\mW\in\R^{d\times d}$, $\mG^{(0)}\in\R^{n\times d}$ and component-wise non-expansive activation function $\phi$. 
Then,
    \begin{equation}
        \lim_{l\to\infty} \norm{\mG^{(l)}}_F = 0
    \end{equation}
\end{proposition}

\begin{proof}
    Based on the definition $\mG^{(l)} = \phi\bracks{\alpha^{(1,l)}\tilde{\mA}\mG^{(l-1)}\mW}$, we first use the fact that $\norm{\phi(\mX)}_F\leq \norm{\mX}_F$ for any component-wise non-expansive function $\phi$, resulting in
    \begin{equation}
        \norm{\mG^{(l)}}_F \leq \norm{\alpha^{(1,l)}\tilde{\mA}\mG^{(l-1)}\mW}_F\, .
    \end{equation}
    Using the sub-multiplicativity of the Frobenius norm, we obtain
    \begin{equation}
        \norm{\mG^{(l)}}_F \leq \alpha^{(1,l)}\cdot\norm{\mW}_F\cdot\norm{\tilde{\mA}}_F\cdot\norm{\mG^{(l-1)}}_F\, .
    \end{equation}
    By iterative application, we have
    \begin{equation}
        \norm{\mG^{(l)}}_F \leq \prod_{k=1}^l\alpha^{(1,k)} \cdot \norm{\mW}_F^l\cdot\norm{\tilde{\mA}}_F^l\cdot\norm{\mG^{(0)}}_F\, .
    \end{equation}
    By definition of $\alpha^{(1,k)}$, their product is given as $\prod_{k=1}^l\alpha^{(1,k)} = \frac{1}{l!\,\epsilon^l}$. In the limit case, this results in
    \begin{equation}
        \lim_{l\to\infty} \frac{\norm{\mW}_F^l\cdot\norm{\tilde{\mA}}_F^l}{l!\,\epsilon^l}\cdot\norm{\mG^{(0)}}_F = 0
    \end{equation}
    for any $\mG^{(0)},\mW,\tilde{\mA}$ and $\epsilon>0$ due to the factorial growing faster than the exponential.
\end{proof}

Based on the adapted random walk $\mG^{(l)}$ of length $l$ having negligible impact on the resulting representations, we approximate the index $l$ used within the fixed-point iteration point as
\begin{equation}
\label{eq:5:zero_convergence}
    l \approx \min_{k\in\mathbb{N}} \norm{\mG^{(k)}} < \gamma\, .
\end{equation}
In combination, we approximate $\mH$ by iterating \Cref{eq:5:single_walk} to approximate $l$ based on \Cref{eq:5:zero_convergence}, we then iterate \Cref{def:pprgnn} to obtain $\mH^{(l,l)}$.

\subsection{Efficient Optimization}
While our method to approximate $\mH$ requires iterating $\mG^{(l)}$ until convergence and computing $\mH^{(l,l)}$, the optimization process can be performed efficiently.  
Since the state $\mG^{(l)}$ is not used for predictions, we do not need to track the gradients of this computation, making the training process significantly more efficient. 
However, as the number of iterations $l$ is not known or constrained, the required memory for computing gradients for $\mH$ is unconstrained and may be time-consuming. We will now propose a method to obtain gradients efficiently with bounded memory consumption.
We are interested in the partial derivative of the resulting error $\ell(\mH)\in\R$ obtained by some loss function $\ell$ with respect to the parameter matrix $\mW$ and the initial state $\mH^{(0)}$. Based on the approximation $\mH^{(l,l)} \approx \mH$, we obtain the partial derivative with respect to the input $\mH^{(0)}$ as
\begin{equation}
\label{eq:partial_deriv_h0}
    \frac{\partial \ell\left(\mH^{(l,l)}\right)}{\partial \mH^{(0)}} =  \sum_{k=0}^l \frac{\partial \ell(\mH^{(l,l)})}{\partial \mZ^{(l-k,l)}}
\end{equation}
with $\mZ^{(0,l)} = \mH^{(0)}$ and $\mZ^{(k,l)} = \alpha^{(k,l)}\tilde{\mA}\mH^{(k-1,l)}\mW + \mH^{(0)}$ so that $\mH^{(k,l)} = \phi(\mZ^{(k,l)})$. We obtained the partial derivative with respect to the parameters $\mW$ as
\begin{equation}
\label{eq:partial_deriv_w}
    \frac{\partial \ell\bracks{\mH^{(l,l)}}}{\partial \mW} = \sum_{k=0}^l\bracks{\alpha^{(l-k,l)}\bracks{\mH^{(l-k-1,l)}}^\top\tilde{\mA}^\top}\frac{\partial \ell\bracks{\mH^{(l,l)}}}{\partial \mZ^{(l-k,l)}}\, .
\end{equation}

To obtain these, we further need to compute the partial derivative $\frac{\partial \ell\bracks{\mH^{(l,l)}}}{\partial \mZ^{(k,l)}}$, which we obtained as

\begin{equation}
    \frac{\partial \ell\bracks{\mH^{(l,l)}}}{\partial \mZ^{(k,l)}} =  \frac{\partial \ell\bracks{\mH^{(l,l)}}}{\partial \mH^{(k,l)}}\odot \phi^\prime\bracks{\mZ^{(k,l)}}\, ,
\end{equation}
where $\phi^\prime(\cdot)$ is the element-wise derivative of the activation function $\phi$ applied to its input.
This in turn requires the partial derivative $\frac{\partial \ell\bracks{\mH^{(l,l)}}}{\partial \mH^{(k-1,l)}}$, which requires the following two cases:
\begin{equation}
    \frac{\partial \ell\bracks{\mH^{(l,l)}}}{\partial \mH^{(k,l)}} = \begin{cases}\alpha^{(k+1,l)}\tilde{\mA}^\top\bracks{\frac{\partial \ell\bracks{\mH^{(l,l)}}}{\partial \mZ^{(k+1,l)}}}\mW^\top, & \text{if } k<l\\
    \frac{\partial \ell\bracks{\mH^{(l,l)}}}{\partial \mH^{(l,l)}}, & \text{if } k=l\, .
    \end{cases}
\end{equation}

For $l\to\infty$, both Equation~\ref{eq:partial_deriv_h0} and Equation~\ref{eq:partial_deriv_w} converge to a unique solution, as with the forward pass. For efficient computation of these partial derivatives, we use gradient checkpointing~\parencite{chen2016training} to store intermediate results $\mH^{(k,l)}$ and $\mZ^{(k,l)}$ during the forward pass. To additionally constrain the memory consumption during training for the intermediate states for backpropagation, we further constrain the maximal $k$ used within the sums by a tunable number of iterations $m$ instead of summing over all $l$ terms. The number $m$ can be selected based on the available memory and runtime constraints. This is similar to truncated backpropagation through time (TBPTT)~\parencite{elman1990finding}, which was proposed as a memory-efficient optimization procedure for recurrent neural networks. This truncation also reduces the runtime of the backpropagation.
As the representations of each node in $\mH^{(l,l)}$ are unaffected by the representations of other nodes with distance larger than $l$ hops in the graph, the partial derivatives would be unaware of this potential information. To account for this, we compute the partial derivatives for $l^\prime = l + j$ instead of setting $l$ as the maximal index. This allows optimization to account for further distant representations that would otherwise be unnoticeable. Thus, we compute the partial derivative $\frac{\partial \ell(\mH^{(l^\prime,l^\prime)})}{\partial \mH^{(0)}}$ and $\frac{\partial \ell(\mH^{(l^\prime,l^\prime)})}{\partial \mW}$.

In summary, these approximations enable a more efficient optimization process in terms of runtime and memory usage by truncating the number of backpropagation steps and utilizing gradient checkpointing.
\begin{table}[tb]
\centering
  \begin{tabular}{ccccc}
\toprule
    Dataset & \# of graphs & avg. \# of nodes & \# of classes & Task \\
    \midrule
    PPI & $22$ & $2373$ & $121$ & \makecell{Inductive \\Node classification}\\
    Amazon & $1$ & \num{334863} & $58$ & \makecell{Transductive \\Node classification} \\
    MUTAG & $188$ & $17.9$ & $2$ & Graph classification\\
    PTC & $344$ & $25.5$ & $2$ & Graph classification \\
    COX2 & $467$ & $41.2$ & $2$ & Graph classification \\
    PROTEINS & $1113$ & $39.1$ & $2$ & Graph classification \\
    NCI1 & $4110$ & $29.8$ & $2$ & Graph classification \\
    \bottomrule
\end{tabular}
    \caption[Properties of the graphs used within the considered benchmark tasks.]{Key properties of the graphs used for our evaluation.}\label{tab:properties}
    \end{table}
\begin{table}
\centering
    \begin{tabular}{cc}
\toprule
     Method & Micro-$F_1$-Score \\
     \midrule
     MLP & $46.2$ \\
     GCN & $59.2$ \\
     SSE & $83.6$ \\
     GAT & $97.3$ \\
    APPNP & $44.8$ \\
     IGNN & $\underline{97.6}$ \\
     PPRGNN & $\mathbf{98.9}$ \\
     \bottomrule
\end{tabular}
    \caption[Evaluation results for the PPRGNN on the PPI task.]{Obtained micro-$F_1$-scores on the testing set. Best result in \textbf{bold}, second-best result \underline{underlined}.}\label{tab:ppi}
\end{table}%

\begin{table}[tb]
\centering
\caption[Evaluation of runtimes and the required number of epochs for the PPRGNN.]{Runtimes and required number of optimization steps for the IGNN and PPRGNN on the PPI and Amazon tasks. Both methods use the same architecture and the same number of parameters. The $\#$ of epochs for PPRGNN is the epoch after which PPRGNN surpasses the IGNN in performance. Total time refers to the runtime required for this performance.}
\begin{tabular}{ccccc}
\toprule
     Dataset & Method & \# of Epochs & Avg. time per epoch & Total time \\
     \midrule
     \multirow{2}{*}{Amazon (0.05)} & IGNN & 872 & 14s & 3h 21m  \\
     & PPRGNN & \textbf{175} & \textbf{11s} & \textbf{32m} \\
     \midrule
     \multirow{2}{*}{PPI} & IGNN & 58 & 26s & 25m \\
     & PPRGNN & \textbf{47} & \textbf{18s}  & \textbf{14m} \\
     \bottomrule
\end{tabular}
\label{tab:time}
\end{table}

\begin{figure}[tb]
  \begin{subfigure}[b]{0.49\textwidth}
    \def\svgwidth{\textwidth}
    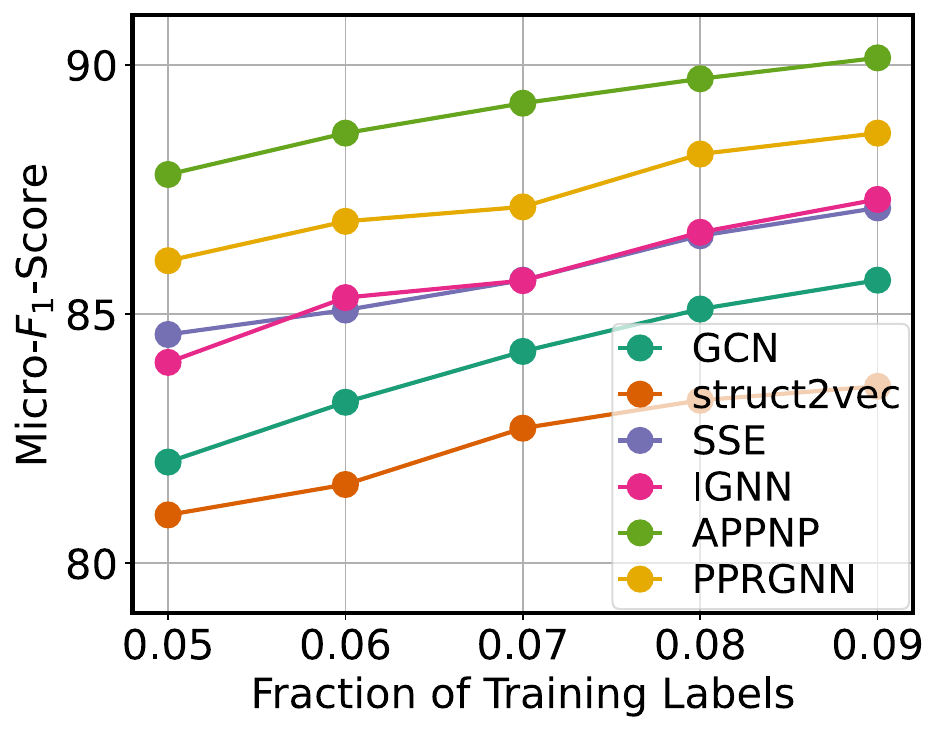
    \caption{Micro-$F_1$-Scores.}
    \label{fig:f1}
  \end{subfigure}
  \hfill
  \begin{subfigure}[b]{0.49\textwidth}
    \def\svgwidth{\textwidth}
    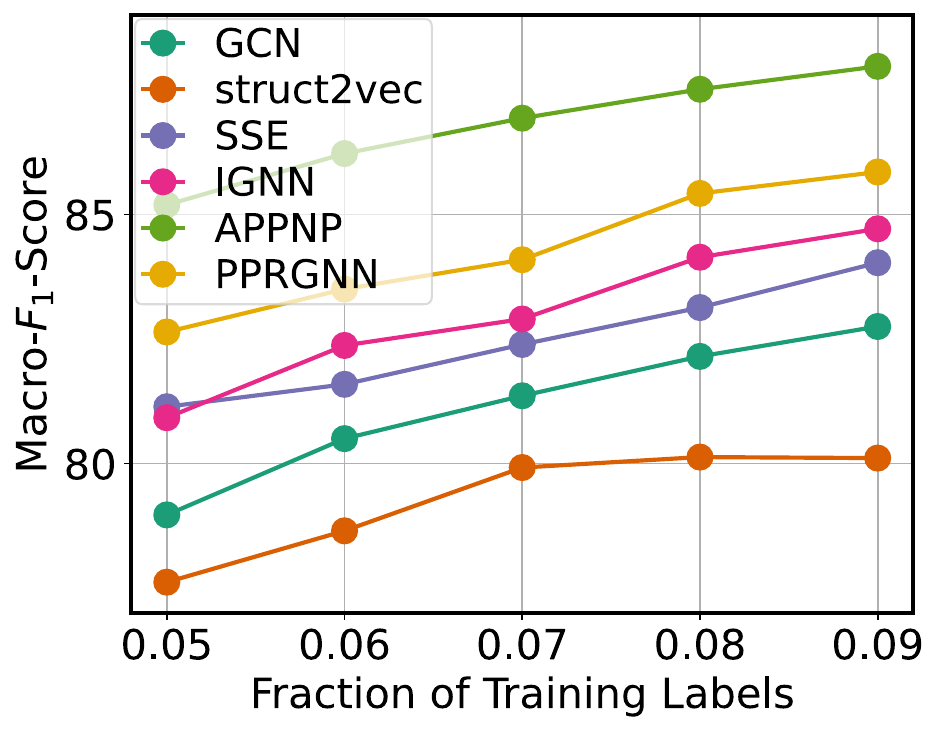
    \caption{Macro-$F_1$-Scores.}
    \label{fig:f2}
  \end{subfigure}
  \caption[Evaluation of the PPRGNN for the Amazon task.]{Testing prediction results for the Amazon task using micro-$F_1$-scores (a) and macro-$F_1$-scores (b). The fraction indicates the share of labeled nodes available for optimization.}
  \label{fig:amazon}
\end{figure}

\begin{table}[tb]
\caption[Evaluation of the PPRGNN for five node classification tasks.]{Evaluation of the PPRGNN for five node classification tasks, following \parencite{gu2020implicit}. Best results in \textbf{bold}, second-best \underline{underlined}.}
\centering
\begin{tabular}{cccccc}
\toprule
     Dataset & MUTAG & PTC & COX2 & PROTEINS & NCI1  \\
     \midrule
     GK & $81.4\pm1.7$ & $55.7\pm0.5$ & - & $71.4\pm0.3$ & $62.5\pm0.3$ \\
     RW & $79.2\pm2.1$ & $55.9\pm0.3$ & - & $59.6\pm0.1$ & - \\
     WL & $84.1\pm1.9$ & $58.0\pm2.5$ & $83.2\pm0.2$ & $74.7\pm0.5$ & $\mathbf{84.5}\pm0.5$ \\
     DGCNN & $85.8$ & $58.6$ & - & $75.5$ & $74.4$ \\
     GCN & $85.6\pm5.8$ & $64.2\pm4.3$ & - & $76.0\pm3.2$ & $80.2\pm2.0$ \\
     GIN & $89.0\pm6.0$ & $63.7\pm8.2$ & - & $75.9\pm3.8$ & $82.7\pm1.6$ \\
     IGNN & $\underline{89.3}\pm6.7$ & $\underline{70.1}\pm5.6$ & $\underline{86.9}\pm4.0$ & $77.7\pm3.4$ & $80.5\pm1.9$ \\
     APPNP & $87.7\pm8.6$ & $64.5\pm5.1$ & $82.2\pm5.5$ & $\underline{78.7}\pm4.8$ & $65.9\pm2.7$ \\
     \midrule
     PPRGNN & $\mathbf{90.4}\pm7.2$ & $\mathbf{75.0}\pm5.7$ & $\mathbf{89.1}\pm3.9$ & $\mathbf{80.2}\pm3.2$ & $\underline{83.5}\pm1.5$ \\
     \bottomrule
\end{tabular}
\label{tab:gc}

\end{table}

\section{Evaluation}
\label{sec:experiments}
We now empirically validate the effectiveness of the PPRGNN using several commonly used benchmark datasets. We follow the experiments conducted for evaluating the IGNN~\parencite{gu2020implicit}. We utilize the implementation provided with the IGNN and solely replace their infinite-depth formulation with our proposed formulation. Equivalently to their approach, we evaluate a model based on the GCN and set the aggregation function $\tilde{\mA} = \mA_\tsym$ to the symmetrically normalized adjacency matrix. PPRGNN and IGNN utilize the same number of parameters. We additionally evaluate APPNP~\parencite{klicpera2019predict} by equivalently replacing the IGNN with their proposed formulation given in \Cref{eq:appnp}. Following the experiments conducted to evaluate the IGNN, we consider the protein-protein interactions (PPI) benchmark dataset~\parencite{zitnik2017predicting}, the Amazon benchmark dataset~\parencite{yang2015defining}, and five graph classification datasets, namely MUTAG, PTC, COX2, PROTEINS, and NCI1~\parencite{yanardag2015deep}. 
For the backward pass, we use a maximum number of $m=5$ iterations. All experiments use a fixed initial learning rate, which we decay when the training loss plateaus for $24$ epochs. All experiments were executed on an Nvidia P100 GPU. Our implementation is publicly available~\onlineresource{https://github.com/roth-andreas/pprgnn}{https://github.com/roth-andreas/pprgnn}.

\subsubsection{PPI}
As the first task, we consider the protein-protein interactions (PPI) benchmark dataset~\parencite{zitnik2017predicting}. Each of the 22 graphs corresponds to a specific human tissue, with each node representing a protein and each edge indicating an interaction between two proteins. 
Based on 50 available features for each node, the task is to perform binary classification for each of 121 available classes representing cellular functions.
Matching the typical evaluation procedure for this task and the implementation of the IGNN, we utilize $18$ graphs for training, two graphs for validation, and two graphs for testing. We evaluate the architecture used by the IGNN, which applies five separate infinite-depth iterations in sequence. We find that removing self-loops is beneficial for this task, and we set $\epsilon=0.25$. Based on the results from IGNN, we also provide their results for an MLP, the graph convolutional network~\parencite{kipf2017semi}, steady-state embedding (SSE)~\parencite{dai2018learning}, and the graph attention network~\parencite{velickovic2017graph}.

The micro-$F_1$-scores for the testing graphs are presented in \Cref{tab:ppi}. While the IGNN achieved an $F_1$-score of $97.6$, the PPRGNN was able to improve this result to $98.9$. The APPNP underfits the data and achieves an $F_1$-score of $44.8$. This holds even when the initial feature transformation before applying the APPNP uses significantly more parameters than the PPRGNN and IGNN. This improvement supports the need for a more complex model with task-specific parameters. GAT achieves a slightly lower $F_1$-score compared to the IGNN at $97.3$, but significantly better than the GCN. In \Cref{tab:time}, we additionally compare runtimes for the IGNN and the PPRGNN. The average runtime per epoch for the PPRGNN is around $18$ seconds compared to $26$ seconds for the IGNN, an improvement of more than $20\%$. It also achieves the lowest validation error after $47$ epochs, an improvement of around $15\%$ compared to $58$ epochs for the IGNN. The optimized PPRGNN utilizes a total of $82$ message-passing iterations for the testing graphs over the five sequential runs until convergence.

\subsubsection{Amazon}
Following the evaluation of the IGNN, we now assess the scalability of the PPRGNN on the Amazon product co-purchasing dataset~\parencite{yang2012defining}, which has been preprocessed according to \textcite{dai2018learning}. Each of the $\num{334863}$ nodes in the graph represents a product, and every edge indicates that the two corresponding products have been purchased together. In this transductive node classification task, the task is to predict one out of $58$ product types for each node based on a small number of labeled products. There are no node features provided. The fraction of labeled products varies between $5\%$ and $9\%$, as proposed by~\textcite{dai2018learning}. Following the model structure of the IGNN, a single PPRGNN layer is used, with a linear encoder and linear decoder used before and after.
As in the IGNN experiments, we also present results of the SSE~\parencite{dai2018learning}, struc2vec~\parencite{ribeiro2017struc2vec}, GCN~\parencite{kipf2017semi}. We additionally evaluate the APPNP for this task.
Micro $F_1$ and Macro $F_1$ scores are visualized in \Cref{fig:amazon} for label fractions between $5\%$ and $9\%$. PPRGNN outperforms the IGNN, SSE, GCN, and struc2vec by at least $1\%$ in all cases. In contrast to the more complex PPI dataset, APPNP outperforms PPRGNN and all other methods. This indicates that the learnable parameters are less critical for this task compared to the PPI dataset. In \Cref{tab:time}, we also present runtimes for the IGNN and the PPRGNN on this dataset. As with the runtimes for the PPI task, the average time per epoch is reduced from $14$ seconds to $11$ seconds. Even more critically, the optimization requires fewer steps, as the minimal validation error is achieved after $175$ epochs for the PPRGNN, compared to $872$ epochs for the IGNN. This may be caused by the inefficiency of the projected gradient descent used within the IGNN. Both factors combined lead to a significantly reduced total training time.

\subsubsection{Graph Classification}

We now consider the five graph classification tasks MUTAG, PTC, COX2, PROTEINS, and NCI1~\parencite{morris2020tudataset}. These are relatively small standard benchmark datasets for binary classification of graphs. We provide key properties of these datasets in \Cref{tab:properties}.
Following the setup of the IGNN, we conduct a ten-fold cross-validation and report the mean and standard deviation across the validation sets. Based on their experimental setup, the employed architecture applies three infinite-depth iterations in sequence.
We set some joint hyperparameters across all five datasets: The convergence threshold is set to $\epsilon=1$, weight decay is set to $1e-6$, and gradient clipping is set to $25$. Apart from NCI1, self-loops were found to be generally beneficial and added to the model. As in IGNN, we present several baselines, including classical kernel methods. These include the random walk (RW) kernel~\parencite{gartner2003on}, the Weisfeiler-Leman (WL) kernel~\parencite{shervashidze2011weisfeiler}, and the graphlet kernel (GK)~\parencite{shervashidze2009efficient}. We also present results for the GCN~\parencite{kipf2017semi}, the graph isomorphism network (GIN)~\parencite{xu2019how}, and the deep graph convolutional neural network (DGCNN)~\parencite{zhang2018an}.

The average results are presented in \Cref{tab:gc}. The PPRGNN achieves the best results for the MUTAG, PTC, COX2, and PROTEINS datasets. For the NCI1 dataset, it achieves the second-best result, as the WL kernel achieves a higher accuracy by $1\%$. After training, the utilized depth of the PPRGNN varies between $22$ and $41$ iterations before the convergence threshold is satisfied. These experiments confirm that all methods avoiding CD, PPRGNN, IGNN, and APPNP, improve results compared to the GCN significantly for many tasks. Compared to the IGNN and APPNP, the PPRGNN also consistently achieves promising results.
    \cleardoublepage

\clearpage
\chapter{Conclusion and Future Work}
\label{chap:6}
In this thesis, we studied properties exhibited by message-passing neural networks (MPNNs) theoretically and empirically. We extended existing insights on MPNNs by identifying several properties exhibited by many established methods. These split the previously identified phenomenon of over-smoothing into separate, finer-grained properties. These findings are not only beneficial for an improved understanding of MPNNs but also allow the streamlined design of methods that exhibit desired properties. We also propose several frameworks for MPNNs, each specifically targeted to avoid one of the properties.

With these findings, we advance the field of graph machine learning in several ways. Our improved understanding of MPNNs can lead to more targeted and theoretically grounded future research. Our research seeks to assist in addressing some of the ongoing challenges and open up new opportunities.

In \Cref{sec:6:conclusion}, we summarize our main findings and relate them to the initial research question stated in \Cref{sec:1:rq}. In \Cref{sec:6:future}, we outline how our contributions can affect future work on graph machine learning and present essential follow-up questions.

\section{Conclusion}
\label{sec:6:conclusion}
This thesis presented several contributions related to two research questions outlined in \Cref{sec:1:rq}, structured into three main chapters.
Following \Cref{rq:1}, we studied the effect of message-passing operations on given node representations. We demonstrated that the commonly observed phenomenon of over-smoothing in MPNNs can be decomposed into multiple underlying, finer-grained properties. 
First, we identified that the norm of the representations can vanish under certain conditions of the feature transformation, a phenomenon that overlaps with over-smoothing. This causes all node representation vectors to converge towards the zero vector, which can lead to over-smoothing under specific definitions and metrics. 
By further studying the effect of message-passing iterations, we identified that for many MPNNs, each single message-passing operation exhibits the shared component amplification (SCA) property. SCA refers to each message-passing iteration amplifying the same component in the data for every feature channel, independently of the parameter instantiation. This constrains the set of representable node representations when applying a message-passing iteration, and limits the effect of the learnable parameters and the optimization. While closely related to over-smoothing, SCA underlies it and describes a part that occurs for each individual message-passing iteration, rather than only for multiple iterations or in the limit. 
For the limit case of infinitely many message-passing iterations, we identified that a single component in the data is dominantly amplified. In contrast, all other components are relatively dampened increasingly, a property we refer to as component dominance (CD). MPNNs exhibiting the CD property yield node representations that become increasingly less informative and independent of the initially provided node representations.
We additionally provide spectral intuition of SCA and CD by showing that message-passing iterations can be interpreted as filters of the spectral graph convolution in the graph Fourier domain. We also show that MPNNs can be interpreted as a special case of the power iteration algorithm. Similarly to CD, power iteration describes the convergence of a vector to the dominant eigenvector of a matrix when repeatedly applying that matrix.

Following our novel insights into the effect of message-passing on representations, we then considered our other challenge. 
With \Cref{rq:2}, our goal was to design MPNNs with beneficial properties based on our improved understanding of these methods. Specifically, we developed frameworks for MPNNs that do not exhibit either the SCA or the CD property.

To design MPNNs that do not exhibit SCA, we proved the necessity of operating on multiple computational graphs. Equivalently, we require multiple edge relations in a graph, e.g., a multigraph or a multi-relational graph. We proposed two approaches for deriving such graphs. First, we introduced the multi-relational split (MRS) framework, with which we proposed converting the computational graph of any existing MPNN into multiple edge relations. Each edge is assigned to one of multiple edge relation types, for each of which a distinct feature transformation is applied to the corresponding message. We introduced structural dependence and independence of nodes, which allowed us to identify cases in which the SCA property can be avoided. While the construction of a method for splitting the computational graph is highly task-dependent, our identified property holds independently of the specific method employed. As a general method that always ensures the structural independence of nodes, we propose splitting a graph into two directed acyclic relations using a strict partial ordering. We demonstrate that this can be efficiently performed, for example, by utilizing the node degree for ordering and distinguishing between edges from higher-degree nodes to lower-degree nodes and vice versa.

As an alternative derivation of multiple computational graphs, we proposed a novel approximation of message-passing methods based on the spectral graph convolution. 
While previously the graph convolution has only been defined in the single-input single-output (SISO) case, we obtained the multi-input multi-output graph convolution (MIMO-GC) using the convolution theorem and the graph Fourier transform. We found that the MIMO-GC naturally utilizes multiple computational graphs. Each computational graph amplifies a distinct component in the representation, matching our theoretical findings on the SCA property. 
We proposed the localized MIMO-GC (LMGC) as a framework for message-passing, which approximates the MIMO-GC by localizing the aggregation step while allowing for multiple computational graphs. The LMGC framework only requires the construction of edge weights, streamlining the design of message-passing methods and the study of their properties. We present two key properties of the LMGC framework. For almost every choice of edge weights, the LMGC is injective on multisets, similar to the graph isomorphism network (GIN)~\parencite{xu2019how}. This property holds even for a single computational graph. For multiple computational graphs, the resulting representations are linearly independent, a property that subsumes injectivity. Our findings also highlighted the close connections between the expressivity of message-passing iterations and avoiding SCA as one key property underlying over-smoothing.

To develop methods that avoid CD, we established a close connection between this phenomenon and the PageRank algorithm. This connection helps to provide a better intuition for CD, as message-passing iterations can be seen as random walk steps in PageRank. It also allows us to adopt modifications that avoid this phenomenon. As personalized PageRank (PPR) was proposed as a modification to PageRank, we proposed adapting PPR for MPNNs and constructing a method that follows its ideas. 
As PPR ensures the localization of the resulting state by combining multiple random walks while guaranteeing the convergence of each random walk, we proposed to ensure the convergence of message-passing iterations.
We refer to the resulting method as the personalized PageRank graph neural network (PPRGNN), whic avoids CD by similarly ensuring convergence of MPNNs to a fixed point. We also proposed an efficient method for obtaining the fixed point, both in terms of runtime and memory consumption.

\section{Future Work}
\label{sec:6:future}
The differences between most recent methods on established benchmark tasks are relatively minor~\parencite{bechler2025position}. This lack of challenging benchmark tasks has limited the informativeness of our empirical evaluations. These datasets do not allow an effective examination of the importance of avoiding specific properties in message-passing. To enable more effective empirical evaluations, future work is needed to develop meaningful real-world benchmark applications for graph machine learning methods. When performance improvements better reflect the practical relevance of a method, promising research directions can be identified more effectively. Currently, weighing the importance of each property is a challenging task.
While various phenomena of MPNNs have been studied both empirically and theoretically, the relevance of addressing any particular issue remains uncertain, as also noted by \textcite{arnaiz2025oversmoothing}. 
This challenge is intensified by the fact that many of these phenomena lack clear definitions and refer to broad aspects. Disentangling observed phenomena into fine-grained properties is crucial for aligning research efforts, identifying critical research directions, and developing more targeted solutions.

Following our novel insights on message-passing iterations, future work can focus on targeting fine-grained properties that underlie over-smoothing, rather than attempting to avoid the broad phenomenon as a whole. Previous work on over-smoothing has addressed different aspects of the property, without recognizing that these aspects are distinct. Our identification of over-smoothing as composed of multiple overlapping phenomena allows the separation of these methods.
The broad scope of over-smoothing not only complicates efforts to establish a unified definition and understanding of the phenomenon, but it also makes it difficult to prioritize which aspects are most critical to address. 
By splitting over-smoothing into multiple separate properties, it becomes easier to design specific methods targeting particular fine-grained properties. It also makes it easier to study which underlying aspects of over-smoothing are most relevant to a given task. We see considerable potential for future work to further identify finer-grained properties of message-passing methods, enabling the field to pinpoint the most critical properties. Our theoretical insights can be further extended by identifying cases in which avoiding SCA, CD, or other aspects of message-passing is most beneficial. The theoretical insights can also be extended to more methods, including nonlinear activation functions for properties that consider iterated message-passing.

Regarding SCA, further research could explore the design of methods using multiple computational graphs. In addition to our identified general properties of such frameworks, additional relevant aspects can be identified. 
As tasks may benefit from different methods depending on the available data, task-specific methods with particular properties can be constructed. Although our empirical results demonstrate the benefits of avoiding SCA for various tasks, its relative importance in contrast to other phenomena requires further investigation.
Similarly, our insights for avoiding CD can be further extended in future work by analyzing its importance in comparison with other properties of MPNNs.
Building on our insights into the connection between MPNNs and PageRank, similar methods can be developed that further mitigate CD.

Overall, the fragmented understanding of various phenomena and their significance has resulted in disjointed research efforts in graph machine learning. By reaching a shared understanding of key phenomena, research efforts could be more streamlined and targeted. Once the community reaches consensus on clearly defined aspects and their relevance in graph machine learning, the field could benefit substantially from more targeted research. This will also require the development of challenging and impactful benchmarks, along with a systematic evaluation of the importance of various properties for performance improvements and research prioritization.
    \cleardoublepage

	\backmatter
	\listoffigures
	\addcontentsline{toc}{chapter}{\listfigurename}
	\cleardoublepage
	
	\listoftables
	\addcontentsline{toc}{chapter}{\listtablename}
	\cleardoublepage
	
	\listofonlineresources
	\addcontentsline{toc}{chapter}{\listofloorname}
	\cleardoublepage
	
	\printbibliography

@article{sandryhaila2013discrete,
  author       = {Aliaksei Sandryhaila and
                  Jos{\'{e}} M. F. Moura},
  title        = {Discrete Signal Processing on Graphs},
  journal      = {{IEEE} Trans. Signal Process.},
  volume       = {61},
  number       = {7},
  pages        = {1644--1656},
  year         = {2013},
  url          = {https://doi.org/10.1109/TSP.2013.2238935},
  doi          = {10.1109/TSP.2013.2238935},
  bibsource    = {dblp computer science bibliography, https://dblp.org}
}

@article{shuman2013the,
  author       = {David I. Shuman and
                  Sunil K. Narang and
                  Pascal Frossard and
                  Antonio Ortega and
                  Pierre Vandergheynst},
  title        = {The Emerging Field of Signal Processing on Graphs: Extending High-Dimensional
                  Data Analysis to Networks and Other Irregular Domains},
  journal      = {{IEEE} Signal Process. Mag.},
  volume       = {30},
  number       = {3},
  pages        = {83--98},
  year         = {2013},
  url          = {https://doi.org/10.1109/MSP.2012.2235192},
  doi          = {10.1109/MSP.2012.2235192},
  bibsource    = {dblp computer science bibliography, https://dblp.org}
}

@book{chung1997spectral,
  author       = {Chung, Fan R. K.},
  title        = {Spectral Graph Theory},
  volume       = {92},
  series       = {CBMS Regional Conference Series in Mathematics},
  year         = {1997},
  publisher    = {Conference Board of the Mathematical Sciences},
  address      = {Washington, DC}
}

@inproceedings{liu2020towards,
  author       = {Meng Liu and
                  Hongyang Gao and
                  Shuiwang Ji},
  editor       = {Rajesh Gupta and
                  Yan Liu and
                  Jiliang Tang and
                  B. Aditya Prakash},
  title        = {Towards Deeper Graph Neural Networks},
  booktitle    = {{KDD} '20: The 26th {ACM} {SIGKDD} Conference on Knowledge Discovery
                  and Data Mining, Virtual Event, CA, USA, August 23-27, 2020},
  pages        = {338--348},
  publisher    = {{ACM}},
  year         = {2020},
  url          = {https://doi.org/10.1145/3394486.3403076},
  doi          = {10.1145/3394486.3403076},
  bibsource    = {dblp computer science bibliography, https://dblp.org}
}

@inproceedings{chen2020measuring,
  author       = {Deli Chen and
                  Yankai Lin and
                  Wei Li and
                  Peng Li and
                  Jie Zhou and
                  Xu Sun},
  title        = {Measuring and Relieving the Over-Smoothing Problem for Graph Neural
                  Networks from the Topological View},
  booktitle    = {The Thirty-Fourth {AAAI} Conference on Artificial Intelligence, {AAAI}
                  2020, The Thirty-Second Innovative Applications of Artificial Intelligence
                  Conference, {IAAI} 2020, The Tenth {AAAI} Symposium on Educational
                  Advances in Artificial Intelligence, {EAAI} 2020, New York, NY, USA,
                  February 7-12, 2020},
  pages        = {3438--3445},
  publisher    = {{AAAI} Press},
  year         = {2020},
  url          = {https://doi.org/10.1609/aaai.v34i04.5747},
  doi          = {10.1609/AAAI.V34I04.5747},
  bibsource    = {dblp computer science bibliography, https://dblp.org}
}

@inproceedings{oono2020graph,
  author       = {Kenta Oono and
                  Taiji Suzuki},
  title        = {Graph Neural Networks Exponentially Lose Expressive Power for Node
                  Classification},
  booktitle    = {8th International Conference on Learning Representations, {ICLR} 2020,
                  Addis Ababa, Ethiopia, April 26-30, 2020},
  publisher    = {OpenReview.net},
  year         = {2020},
  url          = {https://openreview.net/forum?id=S1ldO2EFPr},
  bibsource    = {dblp computer science bibliography, https://dblp.org}
}

@inproceedings{li2018deeper,
  author       = {Qimai Li and
                  Zhichao Han and
                  Xiao{-}Ming Wu},
  editor       = {Sheila A. McIlraith and
                  Kilian Q. Weinberger},
  title        = {Deeper Insights Into Graph Convolutional Networks for Semi-Supervised
                  Learning},
  booktitle    = {Proceedings of the Thirty-Second {AAAI} Conference on Artificial Intelligence,
                  (AAAI-18), the 30th innovative Applications of Artificial Intelligence
                  (IAAI-18), and the 8th {AAAI} Symposium on Educational Advances in
                  Artificial Intelligence (EAAI-18), New Orleans, Louisiana, USA, February
                  2-7, 2018},
  pages        = {3538--3545},
  publisher    = {{AAAI} Press},
  year         = {2018},
  url          = {https://doi.org/10.1609/aaai.v32i1.11604},
  doi          = {10.1609/AAAI.V32I1.11604},
  bibsource    = {dblp computer science bibliography, https://dblp.org}
}

@inproceedings{kipf2017semi,
  author       = {Thomas N. Kipf and
                  Max Welling},
  title        = {Semi-Supervised Classification with Graph Convolutional Networks},
  booktitle    = {5th International Conference on Learning Representations, {ICLR} 2017,
                  Toulon, France, April 24-26, 2017, Conference Track Proceedings},
  publisher    = {OpenReview.net},
  year         = {2017},
  url          = {https://openreview.net/forum?id=SJU4ayYgl},
  bibsource    = {dblp computer science bibliography, https://dblp.org}
}

@inproceedings{keriven2022not,
  author       = {Nicolas Keriven},
  editor       = {Sanmi Koyejo and
                  S. Mohamed and
                  A. Agarwal and
                  Danielle Belgrave and
                  K. Cho and
                  A. Oh},
  title        = {Not too little, not too much: a theoretical analysis of graph (over)smoothing},
  booktitle    = {Advances in Neural Information Processing Systems 35: Annual Conference
                  on Neural Information Processing Systems 2022, NeurIPS 2022, New Orleans,
                  LA, USA, November 28 - December 9, 2022},
  year         = {2022},
  url          = {http://papers.nips.cc/paper\_files/paper/2022/hash/0f956ca6f667c62e0f71511773c86a59-Abstract-Conference.html},
  bibsource    = {dblp computer science bibliography, https://dblp.org}
}

@article{cai2020anote,
  author       = {Chen Cai and
                  Yusu Wang},
  title        = {A Note on Over-Smoothing for Graph Neural Networks},
  journal      = {CoRR},
  volume       = {abs/2006.13318},
  year         = {2020},
  url          = {https://arxiv.org/abs/2006.13318},
  eprinttype    = {arXiv},
  eprint       = {2006.13318},
  bibsource    = {dblp computer science bibliography, https://dblp.org}
}

@inproceedings{zhou2021dirichlet,
  author       = {Kaixiong Zhou and
                  Xiao Huang and
                  Daochen Zha and
                  Rui Chen and
                  Li Li and
                  Soo{-}Hyun Choi and
                  Xia Hu},
  editor       = {Marc'Aurelio Ranzato and
                  Alina Beygelzimer and
                  Yann N. Dauphin and
                  Percy Liang and
                  Jennifer Wortman Vaughan},
  title        = {Dirichlet Energy Constrained Learning for Deep Graph Neural Networks},
  booktitle    = {Advances in Neural Information Processing Systems 34: Annual Conference
                  on Neural Information Processing Systems 2021, NeurIPS 2021, December
                  6-14, 2021, virtual},
  pages        = {21834--21846},
  year         = {2021},
  url          = {https://proceedings.neurips.cc/paper/2021/hash/b6417f112bd27848533e54885b66c288-Abstract.html},
  bibsource    = {dblp computer science bibliography, https://dblp.org}
}

@inproceedings{rusch2023gradient,
  author       = {T. Konstantin Rusch and
                  Benjamin Paul Chamberlain and
                  Michael W. Mahoney and
                  Michael M. Bronstein and
                  Siddhartha Mishra},
  title        = {Gradient Gating for Deep Multi-Rate Learning on Graphs},
  booktitle    = {The Eleventh International Conference on Learning Representations,
                  {ICLR} 2023, Kigali, Rwanda, May 1-5, 2023},
  publisher    = {OpenReview.net},
  year         = {2023},
  url          = {https://openreview.net/forum?id=JpRExTbl1-},
  bibsource    = {dblp computer science bibliography, https://dblp.org}
}

@inproceedings{rusch2022graph,
  author       = {T. Konstantin Rusch and
                  Ben Chamberlain and
                  James Rowbottom and
                  Siddhartha Mishra and
                  Michael M. Bronstein},
  editor       = {Kamalika Chaudhuri and
                  Stefanie Jegelka and
                  Le Song and
                  Csaba Szepesv{\'{a}}ri and
                  Gang Niu and
                  Sivan Sabato},
  title        = {Graph-Coupled Oscillator Networks},
  booktitle    = {International Conference on Machine Learning, {ICML} 2022, 17-23 July
                  2022, Baltimore, Maryland, {USA}},
  series       = {Proceedings of Machine Learning Research},
  volume       = {162},
  pages        = {18888--18909},
  publisher    = {{PMLR}},
  year         = {2022},
  url          = {https://proceedings.mlr.press/v162/rusch22a.html},
  bibsource    = {dblp computer science bibliography, https://dblp.org}
}

@article{rusch2023survey,
  author       = {T. Konstantin Rusch and
                  Michael M. Bronstein and
                  Siddhartha Mishra},
  title        = {A Survey on Oversmoothing in Graph Neural Networks},
  journal      = {CoRR},
  volume       = {abs/2303.10993},
  year         = {2023},
  url          = {https://doi.org/10.48550/arXiv.2303.10993},
  doi          = {10.48550/ARXIV.2303.10993},
  eprinttype    = {arXiv},
  eprint       = {2303.10993},
  bibsource    = {dblp computer science bibliography, https://dblp.org}
}

@inproceedings{wu2023demystifying,
  author       = {Xinyi Wu and
                  Amir Ajorlou and
                  Zihui Wu and
                  Ali Jadbabaie},
  editor       = {Alice Oh and
                  Tristan Naumann and
                  Amir Globerson and
                  Kate Saenko and
                  Moritz Hardt and
                  Sergey Levine},
  title        = {Demystifying Oversmoothing in Attention-Based Graph Neural Networks},
  booktitle    = {Advances in Neural Information Processing Systems 36: Annual Conference
                  on Neural Information Processing Systems 2023, NeurIPS 2023, New Orleans,
                  LA, USA, December 10 - 16, 2023},
  year         = {2023},
  url          = {http://papers.nips.cc/paper\_files/paper/2023/hash/6e4cdfdd909ea4e34bfc85a12774cba0-Abstract-Conference.html},
  bibsource    = {dblp computer science bibliography, https://dblp.org}
}

@article{giovanni2023understanding,
  author       = {Francesco Di Giovanni and
                  James Rowbottom and
                  Benjamin Paul Chamberlain and
                  Thomas Markovich and
                  Michael M. Bronstein},
  title        = {Understanding convolution on graphs via energies},
  journal      = {Trans. Mach. Learn. Res.},
  volume       = {2023},
  year         = {2023},
  url          = {https://openreview.net/forum?id=v5ew3FPTgb},
  bibsource    = {dblp computer science bibliography, https://dblp.org}
}

@inproceedings{maskey2023fractional,
  author       = {Sohir Maskey and
                  Raffaele Paolino and
                  Aras Bacho and
                  Gitta Kutyniok},
  editor       = {Alice Oh and
                  Tristan Naumann and
                  Amir Globerson and
                  Kate Saenko and
                  Moritz Hardt and
                  Sergey Levine},
  title        = {A Fractional Graph Laplacian Approach to Oversmoothing},
  booktitle    = {Advances in Neural Information Processing Systems 36: Annual Conference
                  on Neural Information Processing Systems 2023, NeurIPS 2023, New Orleans,
                  LA, USA, December 10 - 16, 2023},
  year         = {2023},
  url          = {http://papers.nips.cc/paper\_files/paper/2023/hash/2a514213ba899f2911723a38be8d4096-Abstract-Conference.html},
  bibsource    = {dblp computer science bibliography, https://dblp.org}
}

@inproceedings{gu2020implicit,
  author       = {Fangda Gu and
                  Heng Chang and
                  Wenwu Zhu and
                  Somayeh Sojoudi and
                  Laurent El Ghaoui},
  editor       = {Hugo Larochelle and
                  Marc'Aurelio Ranzato and
                  Raia Hadsell and
                  Maria{-}Florina Balcan and
                  Hsuan{-}Tien Lin},
  title        = {Implicit Graph Neural Networks},
  booktitle    = {Advances in Neural Information Processing Systems 33: Annual Conference
                  on Neural Information Processing Systems 2020, NeurIPS 2020, December
                  6-12, 2020, virtual},
  year         = {2020},
  url          = {https://proceedings.neurips.cc/paper/2020/hash/8b5c8441a8ff8e151b191c53c1842a38-Abstract.html},
  bibsource    = {dblp computer science bibliography, https://dblp.org}
}

@inproceedings{roth2022transforming,
  author       = {Andreas Roth and
                  Thomas Liebig},
  editor       = {Massih{-}Reza Amini and
                  St{\'{e}}phane Canu and
                  Asja Fischer and
                  Tias Guns and
                  Petra Kralj Novak and
                  Grigorios Tsoumakas},
  title        = {Transforming PageRank into an Infinite-Depth Graph Neural Network},
  booktitle    = {Machine Learning and Knowledge Discovery in Databases - European Conference,
                  {ECML} {PKDD} 2022, Grenoble, France, September 19-23, 2022, Proceedings,
                  Part {II}},
  series       = {Lecture Notes in Computer Science},
  volume       = {13714},
  pages        = {469--484},
  publisher    = {Springer},
  year         = {2022},
  url          = {https://doi.org/10.1007/978-3-031-26390-3\_27},
  doi          = {10.1007/978-3-031-26390-3\_27},
  bibsource    = {dblp computer science bibliography, https://dblp.org}
}

@inproceedings{velickovic2017graph,
  author       = {Petar Velickovic and
                  Guillem Cucurull and
                  Arantxa Casanova and
                  Adriana Romero and
                  Pietro Li{\`{o}} and
                  Yoshua Bengio},
  title        = {Graph Attention Networks},
  booktitle    = {6th International Conference on Learning Representations, {ICLR} 2018,
                  Vancouver, BC, Canada, April 30 - May 3, 2018, Conference Track Proceedings},
  publisher    = {OpenReview.net},
  year         = {2018},
  url          = {https://openreview.net/forum?id=rJXMpikCZ},
  bibsource    = {dblp computer science bibliography, https://dblp.org}
}

@inproceedings{he2016deep,
  author       = {Kaiming He and
                  Xiangyu Zhang and
                  Shaoqing Ren and
                  Jian Sun},
  title        = {Deep Residual Learning for Image Recognition},
  booktitle    = {2016 {IEEE} Conference on Computer Vision and Pattern Recognition,
                  {CVPR} 2016, Las Vegas, NV, USA, June 27-30, 2016},
  pages        = {770--778},
  publisher    = {{IEEE} Computer Society},
  year         = {2016},
  url          = {https://doi.org/10.1109/CVPR.2016.90},
  doi          = {10.1109/CVPR.2016.90},
  bibsource    = {dblp computer science bibliography, https://dblp.org}
}

@inproceedings{hamilton2017inductive,
  author       = {William L. Hamilton and
                  Zhitao Ying and
                  Jure Leskovec},
  editor       = {Isabelle Guyon and
                  Ulrike von Luxburg and
                  Samy Bengio and
                  Hanna M. Wallach and
                  Rob Fergus and
                  S. V. N. Vishwanathan and
                  Roman Garnett},
  title        = {Inductive Representation Learning on Large Graphs},
  booktitle    = {Advances in Neural Information Processing Systems 30: Annual Conference
                  on Neural Information Processing Systems 2017, December 4-9, 2017,
                  Long Beach, CA, {USA}},
  pages        = {1024--1034},
  year         = {2017},
  url          = {https://proceedings.neurips.cc/paper/2017/hash/5dd9db5e033da9c6fb5ba83c7a7ebea9-Abstract.html},
  bibsource    = {dblp computer science bibliography, https://dblp.org}
}

@inproceedings{yan2022two,
  author       = {Yujun Yan and
                  Milad Hashemi and
                  Kevin Swersky and
                  Yaoqing Yang and
                  Danai Koutra},
  editor       = {Xingquan Zhu and
                  Sanjay Ranka and
                  My T. Thai and
                  Takashi Washio and
                  Xindong Wu},
  title        = {Two Sides of the Same Coin: Heterophily and Oversmoothing in Graph
                  Convolutional Neural Networks},
  booktitle    = {{IEEE} International Conference on Data Mining, {ICDM} 2022, Orlando,
                  FL, USA, November 28 - Dec. 1, 2022},
  pages        = {1287--1292},
  publisher    = {{IEEE}},
  year         = {2022},
  url          = {https://doi.org/10.1109/ICDM54844.2022.00169},
  doi          = {10.1109/ICDM54844.2022.00169},
  bibsource    = {dblp computer science bibliography, https://dblp.org}
}

@inproceedings{ioffe2015batch,
  author       = {Sergey Ioffe and
                  Christian Szegedy},
  editor       = {Francis R. Bach and
                  David M. Blei},
  title        = {Batch Normalization: Accelerating Deep Network Training by Reducing
                  Internal Covariate Shift},
  booktitle    = {Proceedings of the 32nd International Conference on Machine Learning,
                  {ICML} 2015, Lille, France, 6-11 July 2015},
  series       = {{JMLR} Workshop and Conference Proceedings},
  volume       = {37},
  pages        = {448--456},
  publisher    = {JMLR.org},
  year         = {2015},
  url          = {http://proceedings.mlr.press/v37/ioffe15.html},
  bibsource    = {dblp computer science bibliography, https://dblp.org}
}

@inproceedings{zhao2020pairnorm,
  author       = {Lingxiao Zhao and
                  Leman Akoglu},
  title        = {PairNorm: Tackling Oversmoothing in GNNs},
  booktitle    = {8th International Conference on Learning Representations, {ICLR} 2020,
                  Addis Ababa, Ethiopia, April 26-30, 2020},
  publisher    = {OpenReview.net},
  year         = {2020},
  url          = {https://openreview.net/forum?id=rkecl1rtwB},
  bibsource    = {dblp computer science bibliography, https://dblp.org}
}

@inproceedings{xu2019how,
  author       = {Keyulu Xu and
                  Weihua Hu and
                  Jure Leskovec and
                  Stefanie Jegelka},
  title        = {How Powerful are Graph Neural Networks?},
  booktitle    = {7th International Conference on Learning Representations, {ICLR} 2019,
                  New Orleans, LA, USA, May 6-9, 2019},
  publisher    = {OpenReview.net},
  year         = {2019},
  url          = {https://openreview.net/forum?id=ryGs6iA5Km},
  bibsource    = {dblp computer science bibliography, https://dblp.org}
}

@inproceedings{chen2020simple,
  author       = {Ming Chen and
                  Zhewei Wei and
                  Zengfeng Huang and
                  Bolin Ding and
                  Yaliang Li},
  title        = {Simple and Deep Graph Convolutional Networks},
  booktitle    = {Proceedings of the 37th International Conference on Machine Learning,
                  {ICML} 2020, 13-18 July 2020, Virtual Event},
  series       = {Proceedings of Machine Learning Research},
  volume       = {119},
  pages        = {1725--1735},
  publisher    = {{PMLR}},
  year         = {2020},
  url          = {http://proceedings.mlr.press/v119/chen20v.html},
  bibsource    = {dblp computer science bibliography, https://dblp.org}
}

@article{zachary1977information,
  title={An information flow model for conflict and fission in small groups},
  author={Zachary, Wayne W},
  journal={Journal of anthropological research},
  volume={33},
  number={4},
  pages={452--473},
  year={1977},
  publisher={University of New Mexico}
}

@inproceedings{li2016gated,
  author       = {Yujia Li and
                  Daniel Tarlow and
                  Marc Brockschmidt and
                  Richard S. Zemel},
  editor       = {Yoshua Bengio and
                  Yann LeCun},
  title        = {Gated Graph Sequence Neural Networks},
  booktitle    = {4th International Conference on Learning Representations, {ICLR} 2016,
                  San Juan, Puerto Rico, May 2-4, 2016, Conference Track Proceedings},
  year         = {2016},
  url          = {http://arxiv.org/abs/1511.05493},
  bibsource    = {dblp computer science bibliography, https://dblp.org}
}

@techreport{page1999pagerank,
          number = {1999-66},
           month = {11},
          author = {Lawrence Page and Sergey Brin and Rajeev Motwani and Terry Winograd},
            note = {Previous number = SIDL-WP-1999-0120},
           title = {The PageRank Citation Ranking: Bringing Order to the Web.},
            type = {Technical Report},
       publisher = {Stanford InfoLab},
            year = {1999},
     institution = {Stanford InfoLab},
             url = {http://ilpubs.stanford.edu:8090/422/},
}

@inproceedings{klicpera2019predict,
  author       = {Johannes Klicpera and
                  Aleksandar Bojchevski and
                  Stephan G{\"{u}}nnemann},
  title        = {Predict then Propagate: Graph Neural Networks meet Personalized PageRank},
  booktitle    = {7th International Conference on Learning Representations, {ICLR} 2019,
                  New Orleans, LA, USA, May 6-9, 2019},
  publisher    = {OpenReview.net},
  year         = {2019},
  url          = {https://openreview.net/forum?id=H1gL-2A9Ym},
  bibsource    = {dblp computer science bibliography, https://dblp.org}
}

@article{ghaoui2021implicit,
  author       = {Laurent El Ghaoui and
                  Fangda Gu and
                  Bertrand Travacca and
                  Armin Askari and
                  Alicia Y. Tsai},
  title        = {Implicit Deep Learning},
  journal      = {{SIAM} J. Math. Data Sci.},
  volume       = {3},
  number       = {3},
  pages        = {930--958},
  year         = {2021},
  url          = {https://doi.org/10.1137/20M1358517},
  doi          = {10.1137/20M1358517},
  bibsource    = {dblp computer science bibliography, https://dblp.org}
}

@book{berman1994nonnegative,
  author       = {Abraham Berman and
                  Robert J. Plemmons},
  title        = {Nonnegative Matrices in the Mathematical Sciences},
  series       = {Classics in Applied Mathematics},
  volume       = {9},
  publisher    = {{SIAM}},
  year         = {1994},
  url          = {https://doi.org/10.1137/1.9781611971262},
  doi          = {10.1137/1.9781611971262},
  isbn         = {978-0-89871-321-3},
  bibsource    = {dblp computer science bibliography, https://dblp.org}
}

@inproceedings{duchi2008efficient,
  author       = {John C. Duchi and
                  Shai Shalev{-}Shwartz and
                  Yoram Singer and
                  Tushar Chandra},
  editor       = {William W. Cohen and
                  Andrew McCallum and
                  Sam T. Roweis},
  title        = {Efficient projections onto the \emph{l}\({}_{\mbox{1}}\)-ball for
                  learning in high dimensions},
  booktitle    = {Machine Learning, Proceedings of the Twenty-Fifth International Conference
                  {(ICML} 2008), Helsinki, Finland, June 5-9, 2008},
  series       = {{ACM} International Conference Proceeding Series},
  volume       = {307},
  pages        = {272--279},
  publisher    = {{ACM}},
  year         = {2008},
  url          = {https://doi.org/10.1145/1390156.1390191},
  doi          = {10.1145/1390156.1390191},
  bibsource    = {dblp computer science bibliography, https://dblp.org}
}

@article{goldstein1964convex,
  title={Convex programming in Hilbert space},
  author={Goldstein, AA},
  journal={Bulletin of the American Mathematical Society},
  volume={70},
  number={5},
  pages={709--710},
  year={1964}
}

@article{chen2016training,
  author       = {Tianqi Chen and
                  Bing Xu and
                  Chiyuan Zhang and
                  Carlos Guestrin},
  title        = {Training Deep Nets with Sublinear Memory Cost},
  journal      = {CoRR},
  volume       = {abs/1604.06174},
  year         = {2016},
  url          = {http://arxiv.org/abs/1604.06174},
  eprinttype    = {arXiv},
  eprint       = {1604.06174},
  bibsource    = {dblp computer science bibliography, https://dblp.org}
}

@article{elman1990finding,
  author       = {Jeffrey L. Elman},
  title        = {Finding Structure in Time},
  journal      = {Cogn. Sci.},
  volume       = {14},
  number       = {2},
  pages        = {179--211},
  year         = {1990},
  url          = {https://doi.org/10.1207/s15516709cog1402\_1},
  doi          = {10.1207/S15516709COG1402\_1},
  bibsource    = {dblp computer science bibliography, https://dblp.org}
}

@article{zitnik2017predicting,
  author       = {Marinka Zitnik and
                  Jure Leskovec},
  title        = {Predicting multicellular function through multi-layer tissue networks},
  journal      = {Bioinform.},
  volume       = {33},
  number       = {14},
  pages        = {i190--i198},
  year         = {2017},
  url          = {https://doi.org/10.1093/bioinformatics/btx252},
  doi          = {10.1093/BIOINFORMATICS/BTX252},
  bibsource    = {dblp computer science bibliography, https://dblp.org}
}

@inproceedings{dai2018learning,
  author       = {Hanjun Dai and
                  Zornitsa Kozareva and
                  Bo Dai and
                  Alexander J. Smola and
                  Le Song},
  editor       = {Jennifer G. Dy and
                  Andreas Krause},
  title        = {Learning Steady-States of Iterative Algorithms over Graphs},
  booktitle    = {Proceedings of the 35th International Conference on Machine Learning,
                  {ICML} 2018, Stockholmsm{\"{a}}ssan, Stockholm, Sweden, July
                  10-15, 2018},
  series       = {Proceedings of Machine Learning Research},
  volume       = {80},
  pages        = {1114--1122},
  publisher    = {{PMLR}},
  year         = {2018},
  url          = {http://proceedings.mlr.press/v80/dai18a.html},
  bibsource    = {dblp computer science bibliography, https://dblp.org}
}

@inproceedings{yang2012defining,
  author       = {Jaewon Yang and
                  Jure Leskovec},
  editor       = {Mohammed Javeed Zaki and
                  Arno Siebes and
                  Jeffrey Xu Yu and
                  Bart Goethals and
                  Geoffrey I. Webb and
                  Xindong Wu},
  title        = {Defining and Evaluating Network Communities Based on Ground-Truth},
  booktitle    = {12th {IEEE} International Conference on Data Mining, {ICDM} 2012,
                  Brussels, Belgium, December 10-13, 2012},
  pages        = {745--754},
  publisher    = {{IEEE} Computer Society},
  year         = {2012},
  url          = {https://doi.org/10.1109/ICDM.2012.138},
  doi          = {10.1109/ICDM.2012.138},
  bibsource    = {dblp computer science bibliography, https://dblp.org}
}

@inproceedings{ribeiro2017struc2vec,
  author       = {Leonardo Filipe Rodrigues Ribeiro and
                  Pedro H. P. Saverese and
                  Daniel R. Figueiredo},
  title        = {\emph{struc2vec}: Learning Node Representations from Structural Identity},
  booktitle    = {Proceedings of the 23rd {ACM} {SIGKDD} International Conference on
                  Knowledge Discovery and Data Mining, Halifax, NS, Canada, August 13
                  - 17, 2017},
  pages        = {385--394},
  publisher    = {{ACM}},
  year         = {2017},
  url          = {https://doi.org/10.1145/3097983.3098061},
  doi          = {10.1145/3097983.3098061},
  bibsource    = {dblp computer science bibliography, https://dblp.org}
}

@inproceedings{gartner2003on,
  author       = {Thomas G{\"{a}}rtner and
                  Peter A. Flach and
                  Stefan Wrobel},
  editor       = {Bernhard Sch{\"{o}}lkopf and
                  Manfred K. Warmuth},
  title        = {On Graph Kernels: Hardness Results and Efficient Alternatives},
  booktitle    = {Computational Learning Theory and Kernel Machines, 16th Annual Conference
                  on Computational Learning Theory and 7th Kernel Workshop, COLT/Kernel
                  2003, Washington, DC, USA, August 24-27, 2003, Proceedings},
  series       = {Lecture Notes in Computer Science},
  volume       = {2777},
  pages        = {129--143},
  publisher    = {Springer},
  year         = {2003},
  url          = {https://doi.org/10.1007/978-3-540-45167-9\_11},
  doi          = {10.1007/978-3-540-45167-9\_11},
  bibsource    = {dblp computer science bibliography, https://dblp.org}
}

@article{shervashidze2011weisfeiler,
  author       = {Nino Shervashidze and
                  Pascal Schweitzer and
                  Erik Jan van Leeuwen and
                  Kurt Mehlhorn and
                  Karsten M. Borgwardt},
  title        = {Weisfeiler-Lehman Graph Kernels},
  journal      = {J. Mach. Learn. Res.},
  volume       = {12},
  pages        = {2539--2561},
  year         = {2011},
  url          = {https://dl.acm.org/doi/10.5555/1953048.2078187},
  doi          = {10.5555/1953048.2078187},
  bibsource    = {dblp computer science bibliography, https://dblp.org}
}

@inproceedings{shervashidze2009efficient,
  author       = {Nino Shervashidze and
                  S. V. N. Vishwanathan and
                  Tobias Petri and
                  Kurt Mehlhorn and
                  Karsten M. Borgwardt},
  editor       = {David A. Van Dyk and
                  Max Welling},
  title        = {Efficient graphlet kernels for large graph comparison},
  booktitle    = {Proceedings of the Twelfth International Conference on Artificial
                  Intelligence and Statistics, {AISTATS} 2009, Clearwater Beach, Florida,
                  USA, April 16-18, 2009},
  series       = {{JMLR} Proceedings},
  volume       = {5},
  pages        = {488--495},
  publisher    = {JMLR.org},
  year         = {2009},
  url          = {http://proceedings.mlr.press/v5/shervashidze09a.html},
  bibsource    = {dblp computer science bibliography, https://dblp.org}
}

@inproceedings{zhang2018an,
  author       = {Muhan Zhang and
                  Zhicheng Cui and
                  Marion Neumann and
                  Yixin Chen},
  editor       = {Sheila A. McIlraith and
                  Kilian Q. Weinberger},
  title        = {An End-to-End Deep Learning Architecture for Graph Classification},
  booktitle    = {Proceedings of the Thirty-Second {AAAI} Conference on Artificial Intelligence,
                  (AAAI-18), the 30th innovative Applications of Artificial Intelligence
                  (IAAI-18), and the 8th {AAAI} Symposium on Educational Advances in
                  Artificial Intelligence (EAAI-18), New Orleans, Louisiana, USA, February
                  2-7, 2018},
  pages        = {4438--4445},
  publisher    = {{AAAI} Press},
  year         = {2018},
  url          = {https://doi.org/10.1609/aaai.v32i1.11782},
  doi          = {10.1609/AAAI.V32I1.11782},
  bibsource    = {dblp computer science bibliography, https://dblp.org}
}

@inproceedings{bruna2014spectral,
  author       = {Joan Bruna and
                  Wojciech Zaremba and
                  Arthur Szlam and
                  Yann LeCun},
  editor       = {Yoshua Bengio and
                  Yann LeCun},
  title        = {Spectral Networks and Locally Connected Networks on Graphs},
  booktitle    = {2nd International Conference on Learning Representations, {ICLR} 2014,
                  Banff, AB, Canada, April 14-16, 2014, Conference Track Proceedings},
  year         = {2014},
  url          = {http://arxiv.org/abs/1312.6203},
  bibsource    = {dblp computer science bibliography, https://dblp.org}
}

@inproceedings{babai2016graph,
  author       = {L{\'{a}}szl{\'{o}} Babai},
  editor       = {Daniel Wichs and
                  Yishay Mansour},
  title        = {Graph isomorphism in quasipolynomial time [extended abstract]},
  booktitle    = {Proceedings of the 48th Annual {ACM} {SIGACT} Symposium on Theory
                  of Computing, {STOC} 2016, Cambridge, MA, USA, June 18-21, 2016},
  pages        = {684--697},
  publisher    = {{ACM}},
  year         = {2016},
  url          = {https://doi.org/10.1145/2897518.2897542},
  doi          = {10.1145/2897518.2897542},
  bibsource    = {dblp computer science bibliography, https://dblp.org}
}

@article{luxburg2007a,
  author       = {Ulrike von Luxburg},
  title        = {A tutorial on spectral clustering},
  journal      = {Stat. Comput.},
  volume       = {17},
  number       = {4},
  pages        = {395--416},
  year         = {2007},
  url          = {https://doi.org/10.1007/s11222-007-9033-z},
  doi          = {10.1007/S11222-007-9033-Z},
  bibsource    = {dblp computer science bibliography, https://dblp.org}
}

@article{fiedler1973algebraic,
author = {Fiedler, Miroslav},
journal = {Czechoslovak Mathematical Journal},
number = {2},
pages = {298-305},
publisher = {Institute of Mathematics, Academy of Sciences of the Czech Republic},
title = {Algebraic connectivity of graphs},
volume = {23},
year = {1973},
}

@article{hammond2011wavelets,
title = {Wavelets on graphs via spectral graph theory},
journal = {Applied and Computational Harmonic Analysis},
volume = {30},
number = {2},
pages = {129-150},
year = {2011},
doi = {https://doi.org/10.1016/j.acha.2010.04.005},
author = {David K. Hammond and Pierre Vandergheynst and Rémi Gribonval},
}

@article{kriege2020a,
  author       = {Nils M. Kriege and
                  Fredrik D. Johansson and
                  Christopher Morris},
  title        = {A survey on graph kernels},
  journal      = {Appl. Netw. Sci.},
  volume       = {5},
  number       = {1},
  pages        = {6},
  year         = {2020},
  url          = {https://doi.org/10.1007/s41109-019-0195-3},
  doi          = {10.1007/S41109-019-0195-3},
  bibsource    = {dblp computer science bibliography, https://dblp.org}
}

@book{oppenheim2013discrete,
  author   = {Oppenheim, Alan and Schafer, Ronald},
  title    = {{Discrete-Time Signal Processing }},
  pages    = {1056},
  publisher = {Pearson Deutschland},
  year     = {2013},
  doi      = {},
}

@inproceedings{vaswani2017attention,
  author       = {Ashish Vaswani and
                  Noam Shazeer and
                  Niki Parmar and
                  Jakob Uszkoreit and
                  Llion Jones and
                  Aidan N. Gomez and
                  Lukasz Kaiser and
                  Illia Polosukhin},
  editor       = {Isabelle Guyon and
                  Ulrike von Luxburg and
                  Samy Bengio and
                  Hanna M. Wallach and
                  Rob Fergus and
                  S. V. N. Vishwanathan and
                  Roman Garnett},
  title        = {Attention is All you Need},
  booktitle    = {Advances in Neural Information Processing Systems 30: Annual Conference
                  on Neural Information Processing Systems 2017, December 4-9, 2017,
                  Long Beach, CA, {USA}},
  pages        = {5998--6008},
  year         = {2017},
  url          = {https://proceedings.neurips.cc/paper/2017/hash/3f5ee243547dee91fbd053c1c4a845aa-Abstract.html},
  bibsource    = {dblp computer science bibliography, https://dblp.org}
}

@article{nt2019revisiting,
  author       = {Hoang NT and
                  Takanori Maehara},
  title        = {Revisiting Graph Neural Networks: All We Have is Low-Pass Filters},
  journal      = {CoRR},
  volume       = {abs/1905.09550},
  year         = {2019},
  url          = {http://arxiv.org/abs/1905.09550},
  eprinttype    = {arXiv},
  eprint       = {1905.09550},
  bibsource    = {dblp computer science bibliography, https://dblp.org}
}

@inproceedings{kothapalli2023a,
  author       = {Vignesh Kothapalli and
                  Tom Tirer and
                  Joan Bruna},
  editor       = {Alice Oh and
                  Tristan Naumann and
                  Amir Globerson and
                  Kate Saenko and
                  Moritz Hardt and
                  Sergey Levine},
  title        = {A Neural Collapse Perspective on Feature Evolution in Graph Neural
                  Networks},
  booktitle    = {Advances in Neural Information Processing Systems 36: Annual Conference
                  on Neural Information Processing Systems 2023, NeurIPS 2023, New Orleans,
                  LA, USA, December 10 - 16, 2023},
  year         = {2023},
  url          = {http://papers.nips.cc/paper\_files/paper/2023/hash/2dd8a2a8685602586c1173f0b644d0e3-Abstract-Conference.html},
  bibsource    = {dblp computer science bibliography, https://dblp.org}
}

@inproceedings{jin2022feature,
  author       = {Wei Jin and
                  Xiaorui Liu and
                  Yao Ma and
                  Charu C. Aggarwal and
                  Jiliang Tang},
  editor       = {Aidong Zhang and
                  Huzefa Rangwala},
  title        = {Feature Overcorrelation in Deep Graph Neural Networks: {A} New Perspective},
  booktitle    = {{KDD} '22: The 28th {ACM} {SIGKDD} Conference on Knowledge Discovery
                  and Data Mining, Washington, DC, USA, August 14 - 18, 2022},
  pages        = {709--719},
  publisher    = {{ACM}},
  year         = {2022},
  url          = {https://doi.org/10.1145/3534678.3539445},
  doi          = {10.1145/3534678.3539445},
  bibsource    = {dblp computer science bibliography, https://dblp.org}
}

@inproceedings{guo2023contranorm,
  author       = {Xiaojun Guo and
                  Yifei Wang and
                  Tianqi Du and
                  Yisen Wang},
  title        = {ContraNorm: {A} Contrastive Learning Perspective on Oversmoothing
                  and Beyond},
  booktitle    = {The Eleventh International Conference on Learning Representations,
                  {ICLR} 2023, Kigali, Rwanda, May 1-5, 2023},
  publisher    = {OpenReview.net},
  year         = {2023},
  url          = {https://openreview.net/forum?id=SM7XkJouWHm},
  bibsource    = {dblp computer science bibliography, https://dblp.org}
}

@inproceedings{gao2019representation,
  author       = {Jun Gao and
                  Di He and
                  Xu Tan and
                  Tao Qin and
                  Liwei Wang and
                  Tie{-}Yan Liu},
  title        = {Representation Degeneration Problem in Training Natural Language Generation
                  Models},
  booktitle    = {7th International Conference on Learning Representations, {ICLR} 2019,
                  New Orleans, LA, USA, May 6-9, 2019},
  publisher    = {OpenReview.net},
  year         = {2019},
  url          = {https://openreview.net/forum?id=SkEYojRqtm},
  bibsource    = {dblp computer science bibliography, https://dblp.org}
}

@inproceedings{jing2022understanding,
  author       = {Li Jing and
                  Pascal Vincent and
                  Yann LeCun and
                  Yuandong Tian},
  title        = {Understanding Dimensional Collapse in Contrastive Self-supervised
                  Learning},
  booktitle    = {The Tenth International Conference on Learning Representations, {ICLR}
                  2022, Virtual Event, April 25-29, 2022},
  publisher    = {OpenReview.net},
  year         = {2022},
  url          = {https://openreview.net/forum?id=YevsQ05DEN7},
  bibsource    = {dblp computer science bibliography, https://dblp.org}
}

@inproceedings{roth2025what,
  author       = {Andreas Roth and
                  Thomas Liebig},
  title        = {What Can We Learn From {MIMO} Graph Convolutions?},
  booktitle    = {Proceedings of the Thirty-Fourth International Joint Conference on
                  Artificial Intelligence, {IJCAI} 2025, Montreal, Canada, August 16-22,
                  2025},
  pages        = {6120--6128},
  publisher    = {ijcai.org},
  year         = {2025},
  url          = {https://doi.org/10.24963/ijcai.2025/681},
  doi          = {10.24963/IJCAI.2025/681},
  bibsource    = {dblp computer science bibliography, https://dblp.org}
}

@inproceedings{he2021bernnet,
  author       = {Mingguo He and
                  Zhewei Wei and
                  Zengfeng Huang and
                  Hongteng Xu},
  editor       = {Marc'Aurelio Ranzato and
                  Alina Beygelzimer and
                  Yann N. Dauphin and
                  Percy Liang and
                  Jennifer Wortman Vaughan},
  title        = {BernNet: Learning Arbitrary Graph Spectral Filters via Bernstein Approximation},
  booktitle    = {Advances in Neural Information Processing Systems 34: Annual Conference
                  on Neural Information Processing Systems 2021, NeurIPS 2021, December
                  6-14, 2021, virtual},
  pages        = {14239--14251},
  year         = {2021},
  url          = {https://proceedings.neurips.cc/paper/2021/hash/76f1cfd7754a6e4fc3281bcccb3d0902-Abstract.html},
  bibsource    = {dblp computer science bibliography, https://dblp.org}
}

@article{mccallum2000automating,
  author       = {Andrew Kachites McCallum and
                  Kamal Nigam and
                  Jason Rennie and
                  Kristie Seymore},
  title        = {Automating the Construction of Internet Portals with Machine Learning},
  journal      = {Inf. Retr.},
  volume       = {3},
  number       = {2},
  pages        = {127--163},
  year         = {2000},
  url          = {https://doi.org/10.1023/A:1009953814988},
  doi          = {10.1023/A:1009953814988},
  bibsource    = {dblp computer science bibliography, https://dblp.org}
}

@inproceedings{roth2024preventing,
  author       = {Andreas Roth and
                  Franka Bause and
                  Nils Morten Kriege and
                  Thomas Liebig},
  editor       = {Guy Wolf and
                  Smita Krishnaswamy},
  title        = {Preventing Representational Rank Collapse in MPNNs by Splitting the
                  Computational Graph},
  booktitle    = {Learning on Graphs Conference, 26-29 November 2024, Virtual},
  series       = {Proceedings of Machine Learning Research},
  volume       = {269},
  pages        = {14},
  publisher    = {{PMLR}},
  year         = {2024},
  url          = {https://proceedings.mlr.press/v269/roth25a.html},
  bibsource    = {dblp computer science bibliography, https://dblp.org}
}

@inproceedings{roth2023rank,
  author       = {Andreas Roth and
                  Thomas Liebig},
  editor       = {Soledad Villar and
                  Benjamin Chamberlain},
  title        = {Rank Collapse Causes Over-Smoothing and Over-Correlation in Graph
                  Neural Networks},
  booktitle    = {Learning on Graphs Conference, 27-30 November 2023, Virtual Event},
  series       = {Proceedings of Machine Learning Research},
  volume       = {231},
  pages        = {35},
  publisher    = {{PMLR}},
  year         = {2023},
  url          = {https://proceedings.mlr.press/v231/roth24a.html},
  bibsource    = {dblp computer science bibliography, https://dblp.org}
}

@article{roth2024simplifying,
  author       = {Andreas Roth},
  title        = {Simplifying the Theory on Over-Smoothing},
  journal      = {CoRR},
  volume       = {abs/2407.11876},
  year         = {2024},
  url          = {https://doi.org/10.48550/arXiv.2407.11876},
  doi          = {10.48550/ARXIV.2407.11876},
  eprinttype    = {arXiv},
  eprint       = {2407.11876},
  bibsource    = {dblp computer science bibliography, https://dblp.org}
}

@inproceedings{brody2022how,
  author       = {Shaked Brody and
                  Uri Alon and
                  Eran Yahav},
  title        = {How Attentive are Graph Attention Networks?},
  booktitle    = {The Tenth International Conference on Learning Representations, {ICLR}
                  2022, Virtual Event, April 25-29, 2022},
  publisher    = {OpenReview.net},
  year         = {2022},
  url          = {https://openreview.net/forum?id=F72ximsx7C1},
  bibsource    = {dblp computer science bibliography, https://dblp.org}
}

@inproceedings{shi2021masked,
  author       = {Yunsheng Shi and
                  Zhengjie Huang and
                  Shikun Feng and
                  Hui Zhong and
                  Wenjing Wang and
                  Yu Sun},
  editor       = {Zhi{-}Hua Zhou},
  title        = {Masked Label Prediction: Unified Message Passing Model for Semi-Supervised
                  Classification},
  booktitle    = {Proceedings of the Thirtieth International Joint Conference on Artificial
                  Intelligence, {IJCAI} 2021, Virtual Event / Montreal, Canada, 19-27
                  August 2021},
  pages        = {1548--1554},
  publisher    = {ijcai.org},
  year         = {2021},
  url          = {https://doi.org/10.24963/ijcai.2021/214},
  doi          = {10.24963/IJCAI.2021/214},
  bibsource    = {dblp computer science bibliography, https://dblp.org}
}

@inproceedings{rong2020self,
  author       = {Yu Rong and
                  Yatao Bian and
                  Tingyang Xu and
                  Weiyang Xie and
                  Ying Wei and
                  Wenbing Huang and
                  Junzhou Huang},
  editor       = {Hugo Larochelle and
                  Marc'Aurelio Ranzato and
                  Raia Hadsell and
                  Maria{-}Florina Balcan and
                  Hsuan{-}Tien Lin},
  title        = {Self-Supervised Graph Transformer on Large-Scale Molecular Data},
  booktitle    = {Advances in Neural Information Processing Systems 33: Annual Conference
                  on Neural Information Processing Systems 2020, NeurIPS 2020, December
                  6-12, 2020, virtual},
  year         = {2020},
  url          = {https://proceedings.neurips.cc/paper/2020/hash/94aef38441efa3380a3bed3faf1f9d5d-Abstract.html},
  bibsource    = {dblp computer science bibliography, https://dblp.org}
}

@inproceedings{rampasek2022recipe,
  author       = {Ladislav Ramp{\'{a}}sek and
                  Michael Galkin and
                  Vijay Prakash Dwivedi and
                  Anh Tuan Luu and
                  Guy Wolf and
                  Dominique Beaini},
  editor       = {Sanmi Koyejo and
                  S. Mohamed and
                  A. Agarwal and
                  Danielle Belgrave and
                  K. Cho and
                  A. Oh},
  title        = {Recipe for a General, Powerful, Scalable Graph Transformer},
  booktitle    = {Advances in Neural Information Processing Systems 35: Annual Conference
                  on Neural Information Processing Systems 2022, NeurIPS 2022, New Orleans,
                  LA, USA, November 28 - December 9, 2022},
  year         = {2022},
  url          = {http://papers.nips.cc/paper\_files/paper/2022/hash/5d4834a159f1547b267a05a4e2b7cf5e-Abstract-Conference.html},
  bibsource    = {dblp computer science bibliography, https://dblp.org}
}

@Inbook{knabner2017lineare,
author="Knabner, Peter
and Barth, Wolf",
title="Lineare Algebra und Analysis",
bookTitle="Lineare Algebra: Aufgaben und L{\"o}sungen",
year="2017",
publisher="Springer Berlin Heidelberg",
address="Berlin, Heidelberg",
pages="229--242",
doi="10.1007/978-3-662-54991-9_15",
url="https://doi.org/10.1007/978-3-662-54991-9_15"
}

@book{tao2012topics,
  title={Topics in random matrix theory},
  author={Tao, Terence},
  volume={132},
  year={2012},
  publisher={American Mathematical Soc.}
}

@article{cao2021sum,
  author       = {Jian Cao and
                  Marc G. Genton and
                  David E. Keyes and
                  George M. Turkiyyah},
  title        = {Sum of Kronecker products representation and its Cholesky factorization
                  for spatial covariance matrices from large grids},
  journal      = {Comput. Stat. Data Anal.},
  volume       = {157},
  pages        = {107165},
  year         = {2021},
  url          = {https://doi.org/10.1016/j.csda.2020.107165},
  doi          = {10.1016/J.CSDA.2020.107165},
  bibsource    = {dblp computer science bibliography, https://dblp.org}
}

@inproceedings{kingma2015adam,
  author       = {Diederik P. Kingma and
                  Jimmy Ba},
  editor       = {Yoshua Bengio and
                  Yann LeCun},
  title        = {Adam: {A} Method for Stochastic Optimization},
  booktitle    = {3rd International Conference on Learning Representations, {ICLR} 2015,
                  San Diego, CA, USA, May 7-9, 2015, Conference Track Proceedings},
  year         = {2015},
  url          = {http://arxiv.org/abs/1412.6980},
  bibsource    = {dblp computer science bibliography, https://dblp.org}
}

@inproceedings{chien2021adaptive,
  author       = {Eli Chien and
                  Jianhao Peng and
                  Pan Li and
                  Olgica Milenkovic},
  title        = {Adaptive Universal Generalized PageRank Graph Neural Network},
  booktitle    = {9th International Conference on Learning Representations, {ICLR} 2021,
                  Virtual Event, Austria, May 3-7, 2021},
  publisher    = {OpenReview.net},
  year         = {2021},
  url          = {https://openreview.net/forum?id=n6jl7fLxrP},
  bibsource    = {dblp computer science bibliography, https://dblp.org}
}

@inproceedings{bo2021beyond,
  author       = {Deyu Bo and
                  Xiao Wang and
                  Chuan Shi and
                  Huawei Shen},
  title        = {Beyond Low-frequency Information in Graph Convolutional Networks},
  booktitle    = {Thirty-Fifth {AAAI} Conference on Artificial Intelligence, {AAAI}
                  2021, Thirty-Third Conference on Innovative Applications of Artificial
                  Intelligence, {IAAI} 2021, The Eleventh Symposium on Educational Advances
                  in Artificial Intelligence, {EAAI} 2021, Virtual Event, February 2-9,
                  2021},
  pages        = {3950--3957},
  publisher    = {{AAAI} Press},
  year         = {2021},
  url          = {https://doi.org/10.1609/aaai.v35i5.16514},
  doi          = {10.1609/AAAI.V35I5.16514},
  bibsource    = {dblp computer science bibliography, https://dblp.org}
}

@inproceedings{yang2016revisiting,
  author       = {Zhilin Yang and
                  William W. Cohen and
                  Ruslan Salakhutdinov},
  editor       = {Maria{-}Florina Balcan and
                  Kilian Q. Weinberger},
  title        = {Revisiting Semi-Supervised Learning with Graph Embeddings},
  booktitle    = {Proceedings of the 33nd International Conference on Machine Learning,
                  {ICML} 2016, New York City, NY, USA, June 19-24, 2016},
  series       = {{JMLR} Workshop and Conference Proceedings},
  volume       = {48},
  pages        = {40--48},
  publisher    = {JMLR.org},
  year         = {2016},
  url          = {http://proceedings.mlr.press/v48/yanga16.html},
  bibsource    = {dblp computer science bibliography, https://dblp.org}
}

@inproceedings{pei2020geom,
  author       = {Hongbin Pei and
                  Bingzhe Wei and
                  Kevin Chen{-}Chuan Chang and
                  Yu Lei and
                  Bo Yang},
  title        = {Geom-GCN: Geometric Graph Convolutional Networks},
  booktitle    = {8th International Conference on Learning Representations, {ICLR} 2020,
                  Addis Ababa, Ethiopia, April 26-30, 2020},
  publisher    = {OpenReview.net},
  year         = {2020},
  url          = {https://openreview.net/forum?id=S1e2agrFvS},
  bibsource    = {dblp computer science bibliography, https://dblp.org}
}

@article{rozemberczki2021multi,
  author       = {Benedek Rozemberczki and
                  Carl Allen and
                  Rik Sarkar},
  title        = {Multi-Scale attributed node embedding},
  journal      = {J. Complex Networks},
  volume       = {9},
  number       = {2},
  year         = {2021},
  url          = {https://doi.org/10.1093/comnet/cnab014},
  doi          = {10.1093/COMNET/CNAB014},
  bibsource    = {dblp computer science bibliography, https://dblp.org}
}

@inproceedings{tang2009social,
  author       = {Jie Tang and
                  Jimeng Sun and
                  Chi Wang and
                  Zi Yang},
  editor       = {John F. Elder IV and
                  Fran{\c{c}}oise Fogelman{-}Souli{\'{e}} and
                  Peter A. Flach and
                  Mohammed Javeed Zaki},
  title        = {Social influence analysis in large-scale networks},
  booktitle    = {Proceedings of the 15th {ACM} {SIGKDD} International Conference on
                  Knowledge Discovery and Data Mining, Paris, France, June 28 - July
                  1, 2009},
  pages        = {807--816},
  publisher    = {{ACM}},
  year         = {2009},
  url          = {https://doi.org/10.1145/1557019.1557108},
  doi          = {10.1145/1557019.1557108},
  bibsource    = {dblp computer science bibliography, https://dblp.org}
}

@inproceedings{roth2022forecasting,
  author       = {Andreas Roth and
                  Thomas Liebig},
  editor       = {K. Sel{\c{c}}uk Candan and
                  Thang N. Dinh and
                  My T. Thai and
                  Takashi Washio},
  title        = {Forecasting Unobserved Node States with spatio-temporal Graph Neural
                  Networks},
  booktitle    = {{IEEE} International Conference on Data Mining Workshops, {ICDM} 2022
                  - Workshops, Orlando, FL, USA, November 28 - Dec. 1, 2022},
  pages        = {740--747},
  publisher    = {{IEEE}},
  year         = {2022},
  url          = {https://doi.org/10.1109/ICDMW58026.2022.00101},
  doi          = {10.1109/ICDMW58026.2022.00101},
  bibsource    = {dblp computer science bibliography, https://dblp.org}
}

@inproceedings{stark2022equibind,
  author       = {Hannes St{\"{a}}rk and
                  Octavian Ganea and
                  Lagnajit Pattanaik and
                  Regina Barzilay and
                  Tommi S. Jaakkola},
  editor       = {Kamalika Chaudhuri and
                  Stefanie Jegelka and
                  Le Song and
                  Csaba Szepesv{\'{a}}ri and
                  Gang Niu and
                  Sivan Sabato},
  title        = {EquiBind: Geometric Deep Learning for Drug Binding Structure Prediction},
  booktitle    = {International Conference on Machine Learning, {ICML} 2022, 17-23 July
                  2022, Baltimore, Maryland, {USA}},
  series       = {Proceedings of Machine Learning Research},
  volume       = {162},
  pages        = {20503--20521},
  publisher    = {{PMLR}},
  year         = {2022},
  url          = {https://proceedings.mlr.press/v162/stark22b.html},
  bibsource    = {dblp computer science bibliography, https://dblp.org}
}

@inproceedings{gilmer2017neural,
  author       = {Justin Gilmer and
                  Samuel S. Schoenholz and
                  Patrick F. Riley and
                  Oriol Vinyals and
                  George E. Dahl},
  editor       = {Doina Precup and
                  Yee Whye Teh},
  title        = {Neural Message Passing for Quantum Chemistry},
  booktitle    = {Proceedings of the 34th International Conference on Machine Learning,
                  {ICML} 2017, Sydney, NSW, Australia, 6-11 August 2017},
  series       = {Proceedings of Machine Learning Research},
  volume       = {70},
  pages        = {1263--1272},
  publisher    = {{PMLR}},
  year         = {2017},
  url          = {http://proceedings.mlr.press/v70/gilmer17a.html},
  bibsource    = {dblp computer science bibliography, https://dblp.org}
}

@inproceedings{alon2021on,
  author       = {Uri Alon and
                  Eran Yahav},
  title        = {On the Bottleneck of Graph Neural Networks and its Practical Implications},
  booktitle    = {9th International Conference on Learning Representations, {ICLR} 2021,
                  Virtual Event, Austria, May 3-7, 2021},
  publisher    = {OpenReview.net},
  year         = {2021},
  url          = {https://openreview.net/forum?id=i80OPhOCVH2},
  bibsource    = {dblp computer science bibliography, https://dblp.org}
}

@inproceedings{luan2022revisiting,
  author       = {Sitao Luan and
                  Chenqing Hua and
                  Qincheng Lu and
                  Jiaqi Zhu and
                  Mingde Zhao and
                  Shuyuan Zhang and
                  Xiao{-}Wen Chang and
                  Doina Precup},
  editor       = {Sanmi Koyejo and
                  S. Mohamed and
                  A. Agarwal and
                  Danielle Belgrave and
                  K. Cho and
                  A. Oh},
  title        = {Revisiting Heterophily For Graph Neural Networks},
  booktitle    = {Advances in Neural Information Processing Systems 35: Annual Conference
                  on Neural Information Processing Systems 2022, NeurIPS 2022, New Orleans,
                  LA, USA, November 28 - December 9, 2022},
  year         = {2022},
  url          = {http://papers.nips.cc/paper\_files/paper/2022/hash/092359ce5cf60a80e882378944bf1be4-Abstract-Conference.html},
  bibsource    = {dblp computer science bibliography, https://dblp.org}
}

@book {griffiths1997principles,
    AUTHOR = {Griffiths, Phillip and Harris, Joseph},
     TITLE = {Principles of algebraic geometry},
    SERIES = {Wiley Classics Library},
      NOTE = {Reprint of the 1978 original},
 PUBLISHER = {John Wiley \& Sons, Inc., New York},
      YEAR = {1994},
     PAGES = {xiv+813},
   MRCLASS = {14-01},
  MRNUMBER = {1288523},
       DOI = {10.1002/9781118032527},
       URL = {https://doi.org/10.1002/9781118032527},
}

@book{gallager1996discrete,
  title        = {Discrete Stochastic Processes},
  author       = {Gallager, Robert G.},
  series       = {The Springer International Series in Engineering and Computer Science},
  publisher    = {Springer},
  address      = {New York, NY},
  year         = {1996},
  edition      = {1},
  doi          = {10.1007/978-1-4615-2329-1},
}

@book{kowalewski1909einfuehrung,
url = {https://doi.org/10.1515/9783112342442},
title = {Einführung in die Determinantentheorie},
author = {Gerhard Kowalewski},
publisher = {De Gruyter},
address = {Berlin, Boston},
doi = {doi:10.1515/9783112342442},
year = {1909},
}

@article{mises1929praktische,
author = {Mises, R. V. and Pollaczek-Geiringer, H.},
title = {Praktische Verfahren der Gleichungsauflösung .},
journal = {ZAMM - Journal of Applied Mathematics and Mechanics / Zeitschrift für Angewandte Mathematik und Mechanik},
volume = {9},
number = {2},
pages = {152-164},
doi = {https://doi.org/10.1002/zamm.19290090206},
year = {1929}
}

@Article{pérez2015activity,
author={P{\'e}rez-Villanueva, Jaime
and M{\'e}ndez-Lucio, Oscar
and Soria-Arteche, Olivia
and Medina-Franco, Jos{\'e} L.},
title={Activity cliffs and activity cliff generators based on chemotype-related activity landscapes},
journal={Molecular Diversity},
year={2015},
month={11},
day={01},
volume={19},
number={4},
pages={1021-1035},
doi={10.1007/s11030-015-9609-z},
}

@inproceedings{morris2019weisfeiler,
  author       = {Christopher Morris and
                  Martin Ritzert and
                  Matthias Fey and
                  William L. Hamilton and
                  Jan Eric Lenssen and
                  Gaurav Rattan and
                  Martin Grohe},
  title        = {Weisfeiler and Leman Go Neural: Higher-Order Graph Neural Networks},
  booktitle    = {The Thirty-Third {AAAI} Conference on Artificial Intelligence, {AAAI}
                  2019, The Thirty-First Innovative Applications of Artificial Intelligence
                  Conference, {IAAI} 2019, The Ninth {AAAI} Symposium on Educational
                  Advances in Artificial Intelligence, {EAAI} 2019, Honolulu, Hawaii,
                  USA, January 27 - February 1, 2019},
  pages        = {4602--4609},
  publisher    = {{AAAI} Press},
  year         = {2019},
  url          = {https://doi.org/10.1609/aaai.v33i01.33014602},
  doi          = {10.1609/AAAI.V33I01.33014602},
  bibsource    = {dblp computer science bibliography, https://dblp.org}
}

@inproceedings{kashima2003marginalized,
  author       = {Hisashi Kashima and
                  Koji Tsuda and
                  Akihiro Inokuchi},
  editor       = {Tom Fawcett and
                  Nina Mishra},
  title        = {Marginalized Kernels Between Labeled Graphs},
  booktitle    = {Machine Learning, Proceedings of the Twentieth International Conference
                  {(ICML} 2003), August 21-24, 2003, Washington, DC, {USA}},
  pages        = {321--328},
  publisher    = {{AAAI} Press},
  year         = {2003},
  url          = {http://www.aaai.org/Library/ICML/2003/icml03-044.php},
  bibsource    = {dblp computer science bibliography, https://dblp.org}
}

@inproceedings{hu2021ogblsc,
  author       = {Weihua Hu and
                  Matthias Fey and
                  Hongyu Ren and
                  Maho Nakata and
                  Yuxiao Dong and
                  Jure Leskovec},
  editor       = {Joaquin Vanschoren and
                  Sai{-}Kit Yeung},
  title        = {{OGB-LSC:} {A} Large-Scale Challenge for Machine Learning on Graphs},
  booktitle    = {Proceedings of the Neural Information Processing Systems Track on
                  Datasets and Benchmarks 1, NeurIPS Datasets and Benchmarks 2021, December
                  2021, virtual},
  year         = {2021},
  url          = {https://datasets-benchmarks-proceedings.neurips.cc/paper/2021/hash/db8e1af0cb3aca1ae2d0018624204529-Abstract-round2.html},
  bibsource    = {dblp computer science bibliography, https://dblp.org}
}

@article{lecun2015deep,
  author       = {Yann LeCun and
                  Yoshua Bengio and
                  Geoffrey E. Hinton},
  title        = {Deep learning},
  journal      = {Nat.},
  volume       = {521},
  number       = {7553},
  pages        = {436--444},
  year         = {2015},
  url          = {https://doi.org/10.1038/nature14539},
  doi          = {10.1038/NATURE14539},
  bibsource    = {dblp computer science bibliography, https://dblp.org}
}

@article{amari1967a,
  author       = {Shun{-}ichi Amari},
  title        = {A Theory of Adaptive Pattern Classifiers},
  journal      = {{IEEE} Trans. Electron. Comput.},
  volume       = {16},
  number       = {3},
  pages        = {299--307},
  year         = {1967},
  url          = {https://doi.org/10.1109/PGEC.1967.264666},
  doi          = {10.1109/PGEC.1967.264666},
  bibsource    = {dblp computer science bibliography, https://dblp.org}
}

@inproceedings{defferrard2016convolutional,
  author       = {Micha{\"{e}}l Defferrard and
                  Xavier Bresson and
                  Pierre Vandergheynst},
  editor       = {Daniel D. Lee and
                  Masashi Sugiyama and
                  Ulrike von Luxburg and
                  Isabelle Guyon and
                  Roman Garnett},
  title        = {Convolutional Neural Networks on Graphs with Fast Localized Spectral
                  Filtering},
  booktitle    = {Advances in Neural Information Processing Systems 29: Annual Conference
                  on Neural Information Processing Systems 2016, December 5-10, 2016,
                  Barcelona, Spain},
  pages        = {3837--3845},
  year         = {2016},
  url          = {https://proceedings.neurips.cc/paper/2016/hash/04df4d434d481c5bb723be1b6df1ee65-Abstract.html},
  bibsource    = {dblp computer science bibliography, https://dblp.org}
}

@article{neil1963convolution,
  author = {Richard O’Neil},
  title = {Convolution operators and l(p,q) spaces},
  journal = {Duke Mathematical Journal},
  year = {1963},
  volume = {30},
  issue = {1},
  doi = {10.1215/s0012-7094-63-03015-1}
}

@article{hansen2020sheaf,
  author       = {Jakob Hansen and
                  Thomas Gebhart},
  title        = {Sheaf Neural Networks},
  journal      = {CoRR},
  volume       = {abs/2012.06333},
  year         = {2020},
  url          = {https://arxiv.org/abs/2012.06333},
  eprinttype    = {arXiv},
  eprint       = {2012.06333},
  bibsource    = {dblp computer science bibliography, https://dblp.org}
}

@inproceedings{bodnar2022neural,
  author       = {Cristian Bodnar and
                  Francesco Di Giovanni and
                  Benjamin Paul Chamberlain and
                  Pietro Li{\'{o}} and
                  Michael M. Bronstein},
  editor       = {Sanmi Koyejo and
                  S. Mohamed and
                  A. Agarwal and
                  Danielle Belgrave and
                  K. Cho and
                  A. Oh},
  title        = {Neural Sheaf Diffusion: {A} Topological Perspective on Heterophily
                  and Oversmoothing in GNNs},
  booktitle    = {Advances in Neural Information Processing Systems 35: Annual Conference
                  on Neural Information Processing Systems 2022, NeurIPS 2022, New Orleans,
                  LA, USA, November 28 - December 9, 2022},
  year         = {2022},
  url          = {http://papers.nips.cc/paper\_files/paper/2022/hash/75c45fca2aa416ada062b26cc4fb7641-Abstract-Conference.html},
  bibsource    = {dblp computer science bibliography, https://dblp.org}
}

@article{dwivedi2023benchmarking,
  author       = {Vijay Prakash Dwivedi and
                  Chaitanya K. Joshi and
                  Anh Tuan Luu and
                  Thomas Laurent and
                  Yoshua Bengio and
                  Xavier Bresson},
  title        = {Benchmarking Graph Neural Networks},
  journal      = {J. Mach. Learn. Res.},
  volume       = {24},
  pages        = {43:1--43:48},
  year         = {2023},
  url          = {https://jmlr.org/papers/v24/22-0567.html},
  bibsource    = {dblp computer science bibliography, https://dblp.org}
}

@article{weisfeiler1968reduction,
  author = {Weisfeiler, Boris and Lehman, A. A.},
  journal = {Nauchno-Technicheskaya Informatsia},
  number = {N9},
  pages = {12--16},
  title = {{A Reduction of a Graph to a Canonical Form and an Algebra Arising During This Reduction}},
  volume = {Ser. 2},
  year = 1968
}

@inproceedings{suresh2021breaking,
  author       = {Susheel Suresh and
                  Vinith Budde and
                  Jennifer Neville and
                  Pan Li and
                  Jianzhu Ma},
  editor       = {Feida Zhu and
                  Beng Chin Ooi and
                  Chunyan Miao},
  title        = {Breaking the Limit of Graph Neural Networks by Improving the Assortativity
                  of Graphs with Local Mixing Patterns},
  booktitle    = {{KDD} '21: The 27th {ACM} {SIGKDD} Conference on Knowledge Discovery
                  and Data Mining, Virtual Event, Singapore, August 14-18, 2021},
  pages        = {1541--1551},
  publisher    = {{ACM}},
  year         = {2021},
  url          = {https://doi.org/10.1145/3447548.3467373},
  doi          = {10.1145/3447548.3467373},
  bibsource    = {dblp computer science bibliography, https://dblp.org}
}

@inproceedings{yang2020factorizable,
  author       = {Yiding Yang and
                  Zunlei Feng and
                  Mingli Song and
                  Xinchao Wang},
  editor       = {Hugo Larochelle and
                  Marc'Aurelio Ranzato and
                  Raia Hadsell and
                  Maria{-}Florina Balcan and
                  Hsuan{-}Tien Lin},
  title        = {Factorizable Graph Convolutional Networks},
  booktitle    = {Advances in Neural Information Processing Systems 33: Annual Conference
                  on Neural Information Processing Systems 2020, NeurIPS 2020, December
                  6-12, 2020, virtual},
  year         = {2020},
  url          = {https://proceedings.neurips.cc/paper/2020/hash/ea3502c3594588f0e9d5142f99c66627-Abstract.html},
  bibsource    = {dblp computer science bibliography, https://dblp.org}
}

@article{guo2024esgnn,
  author       = {Jingwei Guo and
                  Kaizhu Huang and
                  Rui Zhang and
                  Xinping Yi},
  title        = {{ES-GNN:} Generalizing Graph Neural Networks Beyond Homophily With
                  Edge Splitting},
  journal      = {{IEEE} Trans. Pattern Anal. Mach. Intell.},
  volume       = {46},
  number       = {12},
  pages        = {11345--11360},
  year         = {2024},
  url          = {https://doi.org/10.1109/TPAMI.2024.3459932},
  doi          = {10.1109/TPAMI.2024.3459932},
  bibsource    = {dblp computer science bibliography, https://dblp.org}
}

@inproceedings{giunchiglia2022towards,
  author       = {Valentina Giunchiglia and
                  Chirag Varun Shukla and
                  Guadalupe Gonzalez and
                  Chirag Agarwal},
  editor       = {Bastian Rieck and
                  Razvan Pascanu},
  title        = {Towards Training GNNs Using Explanation Directed Message Passing},
  booktitle    = {Learning on Graphs Conference, LoG 2022, 9-12 December 2022, Virtual
                  Event},
  series       = {Proceedings of Machine Learning Research},
  volume       = {198},
  pages        = {28},
  publisher    = {{PMLR}},
  year         = {2022},
  url          = {https://proceedings.mlr.press/v198/giunchiglia22a.html},
  bibsource    = {dblp computer science bibliography, https://dblp.org}
}

@inproceedings{eliasof2024feature,
  author       = {Moshe Eliasof and
                  Eldad Haber and
                  Eran Treister},
  editor       = {Michael J. Wooldridge and
                  Jennifer G. Dy and
                  Sriraam Natarajan},
  title        = {Feature Transportation Improves Graph Neural Networks},
  booktitle    = {Thirty-Eighth {AAAI} Conference on Artificial Intelligence, {AAAI}
                  2024, Thirty-Sixth Conference on Innovative Applications of Artificial
                  Intelligence, {IAAI} 2024, Fourteenth Symposium on Educational Advances
                  in Artificial Intelligence, {EAAI} 2014, February 20-27, 2024, Vancouver,
                  Canada},
  pages        = {11874--11882},
  publisher    = {{AAAI} Press},
  year         = {2024},
  url          = {https://doi.org/10.1609/aaai.v38i11.29073},
  doi          = {10.1609/AAAI.V38I11.29073},
  bibsource    = {dblp computer science bibliography, https://dblp.org}
}

@inproceedings{rossi2023edge,
  author       = {Emanuele Rossi and
                  Bertrand Charpentier and
                  Francesco Di Giovanni and
                  Fabrizio Frasca and
                  Stephan G{\"{u}}nnemann and
                  Michael M. Bronstein},
  editor       = {Soledad Villar and
                  Benjamin Chamberlain},
  title        = {Edge Directionality Improves Learning on Heterophilic Graphs},
  booktitle    = {Learning on Graphs Conference, 27-30 November 2023, Virtual Event},
  series       = {Proceedings of Machine Learning Research},
  volume       = {231},
  pages        = {25},
  publisher    = {{PMLR}},
  year         = {2023},
  url          = {https://proceedings.mlr.press/v231/rossi24a.html},
  bibsource    = {dblp computer science bibliography, https://dblp.org}
}

@article{bresson2017residual,
  author       = {Xavier Bresson and
                  Thomas Laurent},
  title        = {Residual Gated Graph ConvNets},
  journal      = {CoRR},
  volume       = {abs/1711.07553},
  year         = {2017},
  url          = {http://arxiv.org/abs/1711.07553},
  eprinttype    = {arXiv},
  eprint       = {1711.07553},
  bibsource    = {dblp computer science bibliography, https://dblp.org}
}

@inproceedings{scholkemper2025residual,
  author       = {Michael Scholkemper and
                  Xinyi Wu and
                  Ali Jadbabaie and
                  Michael T. Schaub},
  title        = {Residual Connections and Normalization Can Provably Prevent Oversmoothing
                  in GNNs},
  booktitle    = {The Thirteenth International Conference on Learning Representations,
                  {ICLR} 2025, Singapore, April 24-28, 2025},
  publisher    = {OpenReview.net},
  year         = {2025},
  url          = {https://openreview.net/forum?id=i8vPRlsrYu},
  bibsource    = {dblp computer science bibliography, https://dblp.org}
}

@inproceedings{li2021training,
  author       = {Guohao Li and
                  Matthias M{\"{u}}ller and
                  Bernard Ghanem and
                  Vladlen Koltun},
  editor       = {Marina Meila and
                  Tong Zhang},
  title        = {Training Graph Neural Networks with 1000 Layers},
  booktitle    = {Proceedings of the 38th International Conference on Machine Learning,
                  {ICML} 2021, 18-24 July 2021, Virtual Event},
  series       = {Proceedings of Machine Learning Research},
  volume       = {139},
  pages        = {6437--6449},
  publisher    = {{PMLR}},
  year         = {2021},
  url          = {http://proceedings.mlr.press/v139/li21o.html},
  bibsource    = {dblp computer science bibliography, https://dblp.org}
}

@inproceedings{xu2018representation,
  author       = {Keyulu Xu and
                  Chengtao Li and
                  Yonglong Tian and
                  Tomohiro Sonobe and
                  Ken{-}ichi Kawarabayashi and
                  Stefanie Jegelka},
  editor       = {Jennifer G. Dy and
                  Andreas Krause},
  title        = {Representation Learning on Graphs with Jumping Knowledge Networks},
  booktitle    = {Proceedings of the 35th International Conference on Machine Learning,
                  {ICML} 2018, Stockholmsm{\"{a}}ssan, Stockholm, Sweden, July
                  10-15, 2018},
  series       = {Proceedings of Machine Learning Research},
  volume       = {80},
  pages        = {5449--5458},
  publisher    = {{PMLR}},
  year         = {2018},
  url          = {http://proceedings.mlr.press/v80/xu18c.html},
  bibsource    = {dblp computer science bibliography, https://dblp.org}
}

@article{fey2019just,
  author       = {Matthias Fey},
  title        = {Just Jump: Dynamic Neighborhood Aggregation in Graph Neural Networks},
  journal      = {CoRR},
  volume       = {abs/1904.04849},
  year         = {2019},
  url          = {http://arxiv.org/abs/1904.04849},
  eprinttype    = {arXiv},
  eprint       = {1904.04849},
  bibsource    = {dblp computer science bibliography, https://dblp.org}
}

@inproceedings{finkelshtein2024cooperative,
  author       = {Ben Finkelshtein and
                  Xingyue Huang and
                  Michael M. Bronstein and
                  {\.I}smail {\.I}lkan Ceylan},
  title        = {Cooperative Graph Neural Networks},
  booktitle    = {Forty-first International Conference on Machine Learning, {ICML} 2024,
                  Vienna, Austria, July 21-27, 2024},
  publisher    = {OpenReview.net},
  year         = {2024},
  url          = {https://openreview.net/forum?id=ZQcqXCuoxD},
  bibsource    = {dblp computer science bibliography, https://dblp.org}
}

@article{li2020deepergcn,
  author       = {Guohao Li and
                  Chenxin Xiong and
                  Ali K. Thabet and
                  Bernard Ghanem},
  title        = {DeeperGCN: All You Need to Train Deeper GCNs},
  journal      = {CoRR},
  volume       = {abs/2006.07739},
  year         = {2020},
  url          = {https://arxiv.org/abs/2006.07739},
  eprinttype    = {arXiv},
  eprint       = {2006.07739},
  bibsource    = {dblp computer science bibliography, https://dblp.org}
}

@inproceedings{eliasof2023improving,
  author       = {Moshe Eliasof and
                  Lars Ruthotto and
                  Eran Treister},
  editor       = {Andreas Krause and
                  Emma Brunskill and
                  Kyunghyun Cho and
                  Barbara Engelhardt and
                  Sivan Sabato and
                  Jonathan Scarlett},
  title        = {Improving Graph Neural Networks with Learnable Propagation Operators},
  booktitle    = {International Conference on Machine Learning, {ICML} 2023, 23-29 July
                  2023, Honolulu, Hawaii, {USA}},
  series       = {Proceedings of Machine Learning Research},
  volume       = {202},
  pages        = {9224--9245},
  publisher    = {{PMLR}},
  year         = {2023},
  url          = {https://proceedings.mlr.press/v202/eliasof23b.html},
  bibsource    = {dblp computer science bibliography, https://dblp.org}
}

@inproceedings{corso2020principal,
  author       = {Gabriele Corso and
                  Luca Cavalleri and
                  Dominique Beaini and
                  Pietro Li{\`{o}} and
                  Petar Velickovic},
  editor       = {Hugo Larochelle and
                  Marc'Aurelio Ranzato and
                  Raia Hadsell and
                  Maria{-}Florina Balcan and
                  Hsuan{-}Tien Lin},
  title        = {Principal Neighbourhood Aggregation for Graph Nets},
  booktitle    = {Advances in Neural Information Processing Systems 33: Annual Conference
                  on Neural Information Processing Systems 2020, NeurIPS 2020, December
                  6-12, 2020, virtual},
  year         = {2020},
  url          = {https://proceedings.neurips.cc/paper/2020/hash/99cad265a1768cc2dd013f0e740300ae-Abstract.html},
  bibsource    = {dblp computer science bibliography, https://dblp.org}
}

@inproceedings{tailor2022do,
  author       = {Shyam A. Tailor and
                  Felix L. Opolka and
                  Pietro Li{\`{o}} and
                  Nicholas Donald Lane},
  title        = {Do We Need Anisotropic Graph Neural Networks?},
  booktitle    = {The Tenth International Conference on Learning Representations, {ICLR}
                  2022, Virtual Event, April 25-29, 2022},
  publisher    = {OpenReview.net},
  year         = {2022},
  url          = {https://openreview.net/forum?id=hl9ePdHO4\_s},
  bibsource    = {dblp computer science bibliography, https://dblp.org}
}

@inproceedings{rosenbluth2023some,
  author       = {Eran Rosenbluth and
                  Jan T{\"{o}}nshoff and
                  Martin Grohe},
  title        = {Some Might Say All You Need Is Sum},
  booktitle    = {Proceedings of the Thirty-Second International Joint Conference on
                  Artificial Intelligence, {IJCAI} 2023, 19th-25th August 2023, Macao,
                  SAR, China},
  pages        = {4172--4179},
  publisher    = {ijcai.org},
  year         = {2023},
  url          = {https://doi.org/10.24963/ijcai.2023/464},
  doi          = {10.24963/IJCAI.2023/464},
  bibsource    = {dblp computer science bibliography, https://dblp.org}
}

@inproceedings{vashishth2020composition,
  author       = {Shikhar Vashishth and
                  Soumya Sanyal and
                  Vikram Nitin and
                  Partha P. Talukdar},
  title        = {Composition-based Multi-Relational Graph Convolutional Networks},
  booktitle    = {8th International Conference on Learning Representations, {ICLR} 2020,
                  Addis Ababa, Ethiopia, April 26-30, 2020},
  publisher    = {OpenReview.net},
  year         = {2020},
  url          = {https://openreview.net/forum?id=BylA\_C4tPr},
  bibsource    = {dblp computer science bibliography, https://dblp.org}
}

@inproceedings{schlichtkrull2018modeling,
  author       = {Michael Sejr Schlichtkrull and
                  Thomas N. Kipf and
                  Peter Bloem and
                  Rianne van den Berg and
                  Ivan Titov and
                  Max Welling},
  editor       = {Aldo Gangemi and
                  Roberto Navigli and
                  Maria{-}Esther Vidal and
                  Pascal Hitzler and
                  Rapha{\"{e}}l Troncy and
                  Laura Hollink and
                  Anna Tordai and
                  Mehwish Alam},
  title        = {Modeling Relational Data with Graph Convolutional Networks},
  booktitle    = {The Semantic Web - 15th International Conference, {ESWC} 2018, Heraklion,
                  Crete, Greece, June 3-7, 2018, Proceedings},
  series       = {Lecture Notes in Computer Science},
  volume       = {10843},
  pages        = {593--607},
  publisher    = {Springer},
  year         = {2018},
  url          = {https://doi.org/10.1007/978-3-319-93417-4\_38},
  doi          = {10.1007/978-3-319-93417-4\_38},
  bibsource    = {dblp computer science bibliography, https://dblp.org}
}

@article{butler2023convolutional,
  author       = {Landon Butler and
                  Alejandro Parada{-}Mayorga and
                  Alejandro Ribeiro},
  title        = {Convolutional Learning on Multigraphs},
  journal      = {{IEEE} Trans. Signal Process.},
  volume       = {71},
  pages        = {933--946},
  year         = {2023},
  url          = {https://doi.org/10.1109/TSP.2023.3259144},
  doi          = {10.1109/TSP.2023.3259144},
  bibsource    = {dblp computer science bibliography, https://dblp.org}
}

@inproceedings{barbero2024locality,
  author       = {Federico Barbero and
                  Ameya Velingker and
                  Amin Saberi and
                  Michael M. Bronstein and
                  Francesco Di Giovanni},
  title        = {Locality-Aware Graph Rewiring in GNNs},
  booktitle    = {The Twelfth International Conference on Learning Representations,
                  {ICLR} 2024, Vienna, Austria, May 7-11, 2024},
  publisher    = {OpenReview.net},
  year         = {2024},
  url          = {https://openreview.net/forum?id=4Ua4hKiAJX},
  bibsource    = {dblp computer science bibliography, https://dblp.org}
}

@inproceedings{abboud2022shortest,
  author       = {Ralph Abboud and
                  Radoslav Dimitrov and
                  {\.I}smail {\.I}lkan Ceylan},
  editor       = {Bastian Rieck and
                  Razvan Pascanu},
  title        = {Shortest Path Networks for Graph Property Prediction},
  booktitle    = {Learning on Graphs Conference, LoG 2022, 9-12 December 2022, Virtual
                  Event},
  series       = {Proceedings of Machine Learning Research},
  volume       = {198},
  pages        = {5},
  publisher    = {{PMLR}},
  year         = {2022},
  url          = {https://proceedings.mlr.press/v198/abboud22a.html},
  bibsource    = {dblp computer science bibliography, https://dblp.org}
}

@inproceedings{topping2022understanding,
  author       = {Jake Topping and
                  Francesco Di Giovanni and
                  Benjamin Paul Chamberlain and
                  Xiaowen Dong and
                  Michael M. Bronstein},
  title        = {Understanding over-squashing and bottlenecks on graphs via curvature},
  booktitle    = {The Tenth International Conference on Learning Representations, {ICLR}
                  2022, Virtual Event, April 25-29, 2022},
  publisher    = {OpenReview.net},
  year         = {2022},
  url          = {https://openreview.net/forum?id=7UmjRGzp-A},
  bibsource    = {dblp computer science bibliography, https://dblp.org}
}

@inproceedings{nguyen2023revisiting,
  author       = {Khang Nguyen and
                  Nong Minh Hieu and
                  Vinh Duc Nguyen and
                  Nhat Ho and
                  Stanley J. Osher and
                  Tan Minh Nguyen},
  editor       = {Andreas Krause and
                  Emma Brunskill and
                  Kyunghyun Cho and
                  Barbara Engelhardt and
                  Sivan Sabato and
                  Jonathan Scarlett},
  title        = {Revisiting Over-smoothing and Over-squashing Using Ollivier-Ricci
                  Curvature},
  booktitle    = {International Conference on Machine Learning, {ICML} 2023, 23-29 July
                  2023, Honolulu, Hawaii, {USA}},
  series       = {Proceedings of Machine Learning Research},
  volume       = {202},
  pages        = {25956--25979},
  publisher    = {{PMLR}},
  year         = {2023},
  url          = {https://proceedings.mlr.press/v202/nguyen23c.html},
  bibsource    = {dblp computer science bibliography, https://dblp.org}
}

@article{sterling2015zinc,
  author       = {Teague Sterling and
                  John J. Irwin},
  title        = {{ZINC} 15 - Ligand Discovery for Everyone},
  journal      = {J. Chem. Inf. Model.},
  volume       = {55},
  number       = {11},
  pages        = {2324--2337},
  year         = {2015},
  doi          = {10.1021/ACS.JCIM.5B00559},
}

@inproceedings{dwivedi2022long,
  author       = {Vijay Prakash Dwivedi and
                  Ladislav Ramp{\'{a}}sek and
                  Michael Galkin and
                  Ali Parviz and
                  Guy Wolf and
                  Anh Tuan Luu and
                  Dominique Beaini},
  editor       = {Sanmi Koyejo and
                  S. Mohamed and
                  A. Agarwal and
                  Danielle Belgrave and
                  K. Cho and
                  A. Oh},
  title        = {Long Range Graph Benchmark},
  booktitle    = {Advances in Neural Information Processing Systems 35: Annual Conference
                  on Neural Information Processing Systems 2022, NeurIPS 2022, New Orleans,
                  LA, USA, November 28 - December 9, 2022},
  year         = {2022},
  url          = {http://papers.nips.cc/paper\_files/paper/2022/hash/8c3c666820ea055a77726d66fc7d447f-Abstract-Datasets\_and\_Benchmarks.html},
  bibsource    = {dblp computer science bibliography, https://dblp.org}
}

@article{tonshoff2024where,
  author       = {Jan T{\"{o}}nshoff and
                  Martin Ritzert and
                  Eran Rosenbluth and
                  Martin Grohe},
  title        = {Where Did the Gap Go? Reassessing the Long-Range Graph Benchmark},
  journal      = {Trans. Mach. Learn. Res.},
  volume       = {2024},
  year         = {2024},
  url          = {https://openreview.net/forum?id=Nm0WX86sKv},
  bibsource    = {dblp computer science bibliography, https://dblp.org}
}

@inproceedings{loshchilov2019decoupled,
  author       = {Ilya Loshchilov and
                  Frank Hutter},
  title        = {Decoupled Weight Decay Regularization},
  booktitle    = {7th International Conference on Learning Representations, {ICLR} 2019,
                  New Orleans, LA, USA, May 6-9, 2019},
  publisher    = {OpenReview.net},
  year         = {2019},
  url          = {https://openreview.net/forum?id=Bkg6RiCqY7},
  bibsource    = {dblp computer science bibliography, https://dblp.org}
}

@inproceedings{lim2021large,
  author       = {Derek Lim and
                  Felix Hohne and
                  Xiuyu Li and
                  Sijia Linda Huang and
                  Vaishnavi Gupta and
                  Omkar Bhalerao and
                  Ser{-}Nam Lim},
  editor       = {Marc'Aurelio Ranzato and
                  Alina Beygelzimer and
                  Yann N. Dauphin and
                  Percy Liang and
                  Jennifer Wortman Vaughan},
  title        = {Large Scale Learning on Non-Homophilous Graphs: New Benchmarks and
                  Strong Simple Methods},
  booktitle    = {Advances in Neural Information Processing Systems 34: Annual Conference
                  on Neural Information Processing Systems 2021, NeurIPS 2021, December
                  6-14, 2021, virtual},
  pages        = {20887--20902},
  year         = {2021},
  url          = {https://proceedings.neurips.cc/paper/2021/hash/ae816a80e4c1c56caa2eb4e1819cbb2f-Abstract.html},
  bibsource    = {dblp computer science bibliography, https://dblp.org}
}

@inproceedings{platonov2023a,
  author       = {Oleg Platonov and
                  Denis Kuznedelev and
                  Michael Diskin and
                  Artem Babenko and
                  Liudmila Prokhorenkova},
  title        = {A critical look at the evaluation of GNNs under heterophily: Are we
                  really making progress?},
  booktitle    = {The Eleventh International Conference on Learning Representations,
                  {ICLR} 2023, Kigali, Rwanda, May 1-5, 2023},
  publisher    = {OpenReview.net},
  year         = {2023},
  url          = {https://openreview.net/forum?id=tJbbQfw-5wv},
  bibsource    = {dblp computer science bibliography, https://dblp.org}
}

@inproceedings{kreuzer2021rethinking,
  author       = {Devin Kreuzer and
                  Dominique Beaini and
                  William L. Hamilton and
                  Vincent L{\'{e}}tourneau and
                  Prudencio Tossou},
  editor       = {Marc'Aurelio Ranzato and
                  Alina Beygelzimer and
                  Yann N. Dauphin and
                  Percy Liang and
                  Jennifer Wortman Vaughan},
  title        = {Rethinking Graph Transformers with Spectral Attention},
  booktitle    = {Advances in Neural Information Processing Systems 34: Annual Conference
                  on Neural Information Processing Systems 2021, NeurIPS 2021, December
                  6-14, 2021, virtual},
  pages        = {21618--21629},
  year         = {2021},
  url          = {https://proceedings.neurips.cc/paper/2021/hash/b4fd1d2cb085390fbbadae65e07876a7-Abstract.html},
  bibsource    = {dblp computer science bibliography, https://dblp.org}
}

@article{erdos1959on,
  author = {Erdős, P and R\'enyi, A},
  journal = {Publicationes Mathematicae Debrecen},
  pages = {290--297},
  title = {On Random Graphs I},
  volume = 6,
  year = 1959
}

@article{yang2015defining,
  author       = {Jaewon Yang and
                  Jure Leskovec},
  title        = {Defining and evaluating network communities based on ground-truth},
  journal      = {Knowl. Inf. Syst.},
  volume       = {42},
  number       = {1},
  pages        = {181--213},
  year         = {2015},
  url          = {https://doi.org/10.1007/s10115-013-0693-z},
  doi          = {10.1007/S10115-013-0693-Z},
  bibsource    = {dblp computer science bibliography, https://dblp.org}
}

@inproceedings{yanardag2015deep,
  author       = {Pinar Yanardag and
                  S. V. N. Vishwanathan},
  editor       = {Longbing Cao and
                  Chengqi Zhang and
                  Thorsten Joachims and
                  Geoffrey I. Webb and
                  Dragos D. Margineantu and
                  Graham Williams},
  title        = {Deep Graph Kernels},
  booktitle    = {Proceedings of the 21th {ACM} {SIGKDD} International Conference on
                  Knowledge Discovery and Data Mining, Sydney, NSW, Australia, August
                  10-13, 2015},
  pages        = {1365--1374},
  publisher    = {{ACM}},
  year         = {2015},
  url          = {https://doi.org/10.1145/2783258.2783417},
  doi          = {10.1145/2783258.2783417},
  bibsource    = {dblp computer science bibliography, https://dblp.org}
}

@inproceedings{morris2020tudataset,
    title={TUDataset: A collection of benchmark datasets for learning with graphs},
    author={Christopher Morris and Nils M. Kriege and Franka Bause and Kristian Kersting and Petra Mutzel and Marion Neumann},
    booktitle={ICML 2020 Workshop on Graph Representation Learning and Beyond (GRL+ 2020)},
    archivePrefix={arXiv},
    eprint={2007.08663},
    url={www.graphlearning.io},
    year={2020}
}

@article{stachenfeld2021graph,
  author       = {Kimberly L. Stachenfeld and
                  Jonathan Godwin and
                  Peter W. Battaglia},
  title        = {Graph Networks with Spectral Message Passing},
  journal      = {CoRR},
  volume       = {abs/2101.00079},
  year         = {2021},
  url          = {https://arxiv.org/abs/2101.00079},
  eprinttype    = {arXiv},
  eprint       = {2101.00079},
  bibsource    = {dblp computer science bibliography, https://dblp.org}
}

@inproceedings{rampasek2021hierarchical,
  author       = {Ladislav Ramp{\'{a}}sek and
                  Guy Wolf},
  title        = {Hierarchical Graph Neural Nets can Capture Long-Range Interactions},
  booktitle    = {2021 {IEEE} 31st International Workshop on Machine Learning for Signal
                  Processing (MLSP), Gold Coast, Australia, October 25-28, 2021},
  pages        = {1--6},
  publisher    = {{IEEE}},
  year         = {2021},
  url          = {https://doi.org/10.1109/MLSP52302.2021.9596069},
  doi          = {10.1109/MLSP52302.2021.9596069},
  bibsource    = {dblp computer science bibliography, https://dblp.org}
}

@inproceedings{freitas2021a,
  author       = {Scott Freitas and
                  Yuxiao Dong and
                  Joshua Neil and
                  Duen Horng Chau},
  editor       = {Joaquin Vanschoren and
                  Sai{-}Kit Yeung},
  title        = {A Large-Scale Database for Graph Representation Learning},
  booktitle    = {Proceedings of the Neural Information Processing Systems Track on
                  Datasets and Benchmarks 1, NeurIPS Datasets and Benchmarks 2021, December
                  2021, virtual},
  year         = {2021},
  url          = {https://datasets-benchmarks-proceedings.neurips.cc/paper/2021/hash/5fd0b37cd7dbbb00f97ba6ce92bf5add-Abstract-round1.html},
  bibsource    = {dblp computer science bibliography, https://dblp.org}
}

@inproceedings{fey2024position,
  author       = {Matthias Fey and
                  Weihua Hu and
                  Kexin Huang and
                  Jan Eric Lenssen and
                  Rishabh Ranjan and
                  Joshua Robinson and
                  Rex Ying and
                  Jiaxuan You and
                  Jure Leskovec},
  title        = {Position: Relational Deep Learning - Graph Representation Learning
                  on Relational Databases},
  booktitle    = {Forty-first International Conference on Machine Learning, {ICML} 2024,
                  Vienna, Austria, July 21-27, 2024},
  publisher    = {OpenReview.net},
  year         = {2024},
  url          = {https://openreview.net/forum?id=BIMSHniyCP},
  bibsource    = {dblp computer science bibliography, https://dblp.org}
}

@article{openai2023gpt4,
  author       = {OpenAI},
  title        = {{GPT-4} Technical Report},
  journal      = {CoRR},
  volume       = {abs/2303.08774},
  year         = {2023},
  url          = {https://doi.org/10.48550/arXiv.2303.08774},
  doi          = {10.48550/ARXIV.2303.08774},
  eprinttype    = {arXiv},
  eprint       = {2303.08774},
  bibsource    = {dblp computer science bibliography, https://dblp.org}
}

@article{touvron2023llama,
  author       = {Hugo Touvron and
                  Thibaut Lavril and
                  Gautier Izacard and
                  Xavier Martinet and
                  Marie{-}Anne Lachaux and
                  Timoth{\'{e}}e Lacroix and
                  Baptiste Rozi{\`{e}}re and
                  Naman Goyal and
                  Eric Hambro and
                  Faisal Azhar and
                  Aur{\'{e}}lien Rodriguez and
                  Armand Joulin and
                  Edouard Grave and
                  Guillaume Lample},
  title        = {LLaMA: Open and Efficient Foundation Language Models},
  journal      = {CoRR},
  volume       = {abs/2302.13971},
  year         = {2023},
  url          = {https://doi.org/10.48550/arXiv.2302.13971},
  doi          = {10.48550/ARXIV.2302.13971},
  eprinttype    = {arXiv},
  eprint       = {2302.13971},
  bibsource    = {dblp computer science bibliography, https://dblp.org}
}

@article{schmidhuber2015deep,
  author       = {J{\"{u}}rgen Schmidhuber},
  title        = {Deep learning in neural networks: An overview},
  journal      = {Neural Networks},
  volume       = {61},
  pages        = {85--117},
  year         = {2015},
  url          = {https://doi.org/10.1016/j.neunet.2014.09.003},
  doi          = {10.1016/J.NEUNET.2014.09.003},
  bibsource    = {dblp computer science bibliography, https://dblp.org}
}

@inproceedings{you2022roland,
  author       = {Jiaxuan You and
                  Tianyu Du and
                  Jure Leskovec},
  editor       = {Aidong Zhang and
                  Huzefa Rangwala},
  title        = {{ROLAND:} Graph Learning Framework for Dynamic Graphs},
  booktitle    = {{KDD} '22: The 28th {ACM} {SIGKDD} Conference on Knowledge Discovery
                  and Data Mining, Washington, DC, USA, August 14 - 18, 2022},
  pages        = {2358--2366},
  publisher    = {{ACM}},
  year         = {2022},
  url          = {https://doi.org/10.1145/3534678.3539300},
  doi          = {10.1145/3534678.3539300},
  bibsource    = {dblp computer science bibliography, https://dblp.org}
}

@book{hogan2021knowledge,
  author       = {Aidan Hogan and
                  Eva Blomqvist and
                  Michael Cochez and
                  Claudia d'Amato and
                  Gerard de Melo and
                  Claudio Gutierrez and
                  Sabrina Kirrane and
                  Jos{\'{e}} Emilio Labra Gayo and
                  Roberto Navigli and
                  Sebastian Neumaier and
                  Axel{-}Cyrille Ngonga Ngomo and
                  Axel Polleres and
                  Sabbir M. Rashid and
                  Anisa Rula and
                  Lukas Schmelzeisen and
                  Juan Sequeda and
                  Steffen Staab and
                  Antoine Zimmermann},
  title        = {Knowledge Graphs},
  series       = {Synthesis Lectures on Data, Semantics, and Knowledge},
  publisher    = {Morgan {\&} Claypool Publishers},
  year         = {2021},
  url          = {https://doi.org/10.2200/S01125ED1V01Y202109DSK022},
  doi          = {10.2200/S01125ED1V01Y202109DSK022},
  isbn         = {978-3-031-00790-3},
  bibsource    = {dblp computer science bibliography, https://dblp.org}
}

@article{bechler2025position,
  author       = {Maya Bechler{-}Speicher and
                  Ben Finkelshtein and
                  Fabrizio Frasca and
                  Luis M{\"{u}}ller and
                  Jan T{\"{o}}nshoff and
                  Antoine Siraudin and
                  Viktor Zaverkin and
                  Michael M. Bronstein and
                  Mathias Niepert and
                  Bryan Perozzi and
                  Mikhail Galkin and
                  Christopher Morris},
  title        = {Position: Graph Learning Will Lose Relevance Due To Poor Benchmarks},
  journal      = {CoRR},
  volume       = {abs/2502.14546},
  year         = {2025},
  url          = {https://doi.org/10.48550/arXiv.2502.14546},
  doi          = {10.48550/ARXIV.2502.14546},
  eprinttype    = {arXiv},
  eprint       = {2502.14546},
  bibsource    = {dblp computer science bibliography, https://dblp.org}
}

@inproceedings{rong2020dropedge,
  author       = {Yu Rong and
                  Wenbing Huang and
                  Tingyang Xu and
                  Junzhou Huang},
  title        = {DropEdge: Towards Deep Graph Convolutional Networks on Node Classification},
  booktitle    = {8th International Conference on Learning Representations, {ICLR} 2020,
                  Addis Ababa, Ethiopia, April 26-30, 2020},
  publisher    = {OpenReview.net},
  year         = {2020},
  url          = {https://openreview.net/forum?id=Hkx1qkrKPr},
  bibsource    = {dblp computer science bibliography, https://dblp.org}
}

@inproceedings{jamadandi2024spectral,
  author       = {Adarsh Jamadandi and
                  Celia Rubio{-}Madrigal and
                  Rebekka Burkholz},
  editor       = {Amir Globersons and
                  Lester Mackey and
                  Danielle Belgrave and
                  Angela Fan and
                  Ulrich Paquet and
                  Jakub M. Tomczak and
                  Cheng Zhang},
  title        = {Spectral Graph Pruning Against Over-Squashing and Over-Smoothing},
  booktitle    = {Advances in Neural Information Processing Systems 38: Annual Conference
                  on Neural Information Processing Systems 2024, NeurIPS 2024, Vancouver,
                  BC, Canada, December 10 - 15, 2024},
  year         = {2024},
  url          = {http://papers.nips.cc/paper\_files/paper/2024/hash/140aac600566125915df7e74ff538f66-Abstract-Conference.html},
  bibsource    = {dblp computer science bibliography, https://dblp.org}
}

@inproceedings{rubio2025gnns,
  author       = {Celia Rubio{-}Madrigal and
                  Adarsh Jamadandi and
                  Rebekka Burkholz},
  title        = {GNNs Getting ComFy: Community and Feature Similarity Guided Rewiring},
  booktitle    = {The Thirteenth International Conference on Learning Representations,
                  {ICLR} 2025, Singapore, April 24-28, 2025},
  publisher    = {OpenReview.net},
  year         = {2025},
  url          = {https://openreview.net/forum?id=g6v09VxgFw},
  bibsource    = {dblp computer science bibliography, https://dblp.org}
}

@book{cormen1989introduction,
  author       = {Thomas H. Cormen and
                  Charles E. Leiserson and
                  Ronald L. Rivest},
  title        = {Introduction to Algorithms},
  publisher    = {The {MIT} Press and McGraw-Hill Book Company},
  year         = {1989},
  isbn         = {0-262-03141-8},
  bibsource    = {dblp computer science bibliography, https://dblp.org}
}

@book{garey1979computers,
  author       = {M. R. Garey and
                  David S. Johnson},
  title        = {Computers and Intractability: {A} Guide to the Theory of NP-Completeness},
  publisher    = {W. H. Freeman},
  year         = {1979},
  isbn         = {0-7167-1044-7},
  bibsource    = {dblp computer science bibliography, https://dblp.org}
}

@book{wells2012structural,
  title={Structural inorganic chemistry},
  author={Wells, Alexander Frank},
  year={2012},
  publisher={Oxford University Press, USA}
}

@inproceedings{gama2018mimo,
  author       = {Fernando Gama and
                  Antonio G. Marques and
                  Alejandro Ribeiro and
                  Geert Leus},
  title        = {{MIMO} Graph Filters for Convolutional Neural Networks},
  booktitle    = {19th {IEEE} International Workshop on Signal Processing Advances in
                  Wireless Communications, {SPAWC} 2018, Kalamata, Greece, June 25-28,
                  2018},
  pages        = {1--5},
  publisher    = {{IEEE}},
  year         = {2018},
  url          = {https://doi.org/10.1109/SPAWC.2018.8445934},
  doi          = {10.1109/SPAWC.2018.8445934},
  bibsource    = {dblp computer science bibliography, https://dblp.org}
}

@inproceedings{sun2022position,
  author       = {Qingyun Sun and
                  Jianxin Li and
                  Haonan Yuan and
                  Xingcheng Fu and
                  Hao Peng and
                  Cheng Ji and
                  Qian Li and
                  Philip S. Yu},
  editor       = {Mohammad Al Hasan and
                  Li Xiong},
  title        = {Position-aware Structure Learning for Graph Topology-imbalance by
                  Relieving Under-reaching and Over-squashing},
  booktitle    = {Proceedings of the 31st {ACM} International Conference on Information
                  {\&} Knowledge Management, Atlanta, GA, USA, October 17-21, 2022},
  pages        = {1848--1857},
  publisher    = {{ACM}},
  year         = {2022},
  url          = {https://doi.org/10.1145/3511808.3557419},
  doi          = {10.1145/3511808.3557419},
  bibsource    = {dblp computer science bibliography, https://dblp.org}
}

@article{arnaiz2025oversmoothing,
  author       = {Adri{\'{a}}n Arnaiz{-}Rodr{\'{\i}}guez and
                  Federico Errica},
  title        = {Oversmoothing, "Oversquashing", Heterophily, Long-Range,
                  and more: Demystifying Common Beliefs in Graph Machine Learning},
  journal      = {CoRR},
  volume       = {abs/2505.15547},
  year         = {2025},
  url          = {https://doi.org/10.48550/arXiv.2505.15547},
  doi          = {10.48550/ARXIV.2505.15547},
  eprinttype    = {arXiv},
  eprint       = {2505.15547},
  bibsource    = {dblp computer science bibliography, https://dblp.org}
}

@article{sen2008collective,
  author       = {Prithviraj Sen and
                  Galileo Namata and
                  Mustafa Bilgic and
                  Lise Getoor and
                  Brian Gallagher and
                  Tina Eliassi{-}Rad},
  title        = {Collective Classification in Network Data},
  journal      = {{AI} Mag.},
  volume       = {29},
  number       = {3},
  pages        = {93--106},
  year         = {2008},
  url          = {https://doi.org/10.1609/aimag.v29i3.2157},
  doi          = {10.1609/AIMAG.V29I3.2157},
  bibsource    = {dblp computer science bibliography, https://dblp.org}
}

@inproceedings{kriege2016on,
  author       = {Nils M. Kriege and
                  Pierre{-}Louis Giscard and
                  Richard C. Wilson},
  editor       = {Daniel D. Lee and
                  Masashi Sugiyama and
                  Ulrike von Luxburg and
                  Isabelle Guyon and
                  Roman Garnett},
  title        = {On Valid Optimal Assignment Kernels and Applications to Graph Classification},
  booktitle    = {Advances in Neural Information Processing Systems 29: Annual Conference
                  on Neural Information Processing Systems 2016, December 5-10, 2016,
                  Barcelona, Spain},
  pages        = {1615--1623},
  year         = {2016},
  url          = {https://proceedings.neurips.cc/paper/2016/hash/0efe32849d230d7f53049ddc4a4b0c60-Abstract.html},
  bibsource    = {dblp computer science bibliography, https://dblp.org}
}

@inproceedings{frohlich2005optimal,
  author       = {Holger Fr{\"{o}}hlich and
                  J{\"{o}}rg K. Wegner and
                  Florian Sieker and
                  Andreas Zell},
  editor       = {Luc De Raedt and
                  Stefan Wrobel},
  title        = {Optimal assignment kernels for attributed molecular graphs},
  booktitle    = {Machine Learning, Proceedings of the Twenty-Second International Conference
                  {(ICML} 2005), Bonn, Germany, August 7-11, 2005},
  series       = {{ACM} International Conference Proceeding Series},
  volume       = {119},
  pages        = {225--232},
  publisher    = {{ACM}},
  year         = {2005},
  url          = {https://doi.org/10.1145/1102351.1102380},
  doi          = {10.1145/1102351.1102380},
  bibsource    = {dblp computer science bibliography, https://dblp.org}
}

@inproceedings{kriege2012subgraph,
  author       = {Nils M. Kriege and
                  Petra Mutzel},
  title        = {Subgraph Matching Kernels for Attributed Graphs},
  booktitle    = {Proceedings of the 29th International Conference on Machine Learning,
                  {ICML} 2012, Edinburgh, Scotland, UK, June 26 - July 1, 2012},
  publisher    = {icml.cc / Omnipress},
  year         = {2012},
  url          = {http://icml.cc/2012/papers/542.pdf},
  bibsource    = {dblp computer science bibliography, https://dblp.org}
}

@inproceedings{feragen2013scalable,
  author       = {Aasa Feragen and
                  Niklas Kasenburg and
                  Jens Petersen and
                  Marleen de Bruijne and
                  Karsten M. Borgwardt},
  editor       = {Christopher J. C. Burges and
                  L{\'{e}}on Bottou and
                  Zoubin Ghahramani and
                  Kilian Q. Weinberger},
  title        = {Scalable kernels for graphs with continuous attributes},
  booktitle    = {Advances in Neural Information Processing Systems 26: 27th Annual
                  Conference on Neural Information Processing Systems 2013. Proceedings
                  of a meeting held December 5-8, 2013, Lake Tahoe, Nevada, United States},
  pages        = {216--224},
  year         = {2013},
  url          = {https://proceedings.neurips.cc/paper/2013/hash/a2557a7b2e94197ff767970b67041697-Abstract.html},
  bibsource    = {dblp computer science bibliography, https://dblp.org}
}

@inproceedings{orsini2015graph,
  author       = {Francesco Orsini and
                  Paolo Frasconi and
                  Luc De Raedt},
  editor       = {Qiang Yang and
                  Michael J. Wooldridge},
  title        = {Graph Invariant Kernels},
  booktitle    = {Proceedings of the Twenty-Fourth International Joint Conference on
                  Artificial Intelligence, {IJCAI} 2015, Buenos Aires, Argentina, July
                  25-31, 2015},
  pages        = {3756--3762},
  publisher    = {{AAAI} Press},
  year         = {2015},
  url          = {http://ijcai.org/Abstract/15/528},
  bibsource    = {dblp computer science bibliography, https://dblp.org}
}

@inproceedings{morris2016faster,
  author       = {Christopher Morris and
                  Nils M. Kriege and
                  Kristian Kersting and
                  Petra Mutzel},
  editor       = {Francesco Bonchi and
                  Josep Domingo{-}Ferrer and
                  Ricardo Baeza{-}Yates and
                  Zhi{-}Hua Zhou and
                  Xindong Wu},
  title        = {Faster Kernels for Graphs with Continuous Attributes via Hashing},
  booktitle    = {{IEEE} 16th International Conference on Data Mining, {ICDM} 2016,
                  December 12-15, 2016, Barcelona, Spain},
  pages        = {1095--1100},
  publisher    = {{IEEE} Computer Society},
  year         = {2016},
  url          = {https://doi.org/10.1109/ICDM.2016.0142},
  doi          = {10.1109/ICDM.2016.0142},
  bibsource    = {dblp computer science bibliography, https://dblp.org}
}

@inproceedings{kriege2018a,
  author       = {Nils M. Kriege and
                  Christopher Morris and
                  Anja Rey and
                  Christian Sohler},
  editor       = {J{\'{e}}r{\^{o}}me Lang},
  title        = {A Property Testing Framework for the Theoretical Expressivity of Graph
                  Kernels},
  booktitle    = {Proceedings of the Twenty-Seventh International Joint Conference on
                  Artificial Intelligence, {IJCAI} 2018, July 13-19, 2018, Stockholm,
                  Sweden},
  pages        = {2348--2354},
  publisher    = {ijcai.org},
  year         = {2018},
  url          = {https://doi.org/10.24963/ijcai.2018/325},
  doi          = {10.24963/IJCAI.2018/325},
  bibsource    = {dblp computer science bibliography, https://dblp.org}
}

@inproceedings{kriege2019computing,
  author       = {Nils M. Kriege and
                  Pierre{-}Louis Giscard and
                  Franka Bause and
                  Richard C. Wilson},
  editor       = {Jianyong Wang and
                  Kyuseok Shim and
                  Xindong Wu},
  title        = {Computing Optimal Assignments in Linear Time for Approximate Graph
                  Matching},
  booktitle    = {2019 {IEEE} International Conference on Data Mining, {ICDM} 2019,
                  Beijing, China, November 8-11, 2019},
  pages        = {349--358},
  publisher    = {{IEEE}},
  year         = {2019},
  url          = {https://doi.org/10.1109/ICDM.2019.00045},
  doi          = {10.1109/ICDM.2019.00045},
  bibsource    = {dblp computer science bibliography, https://dblp.org}
}

@inproceedings{johansson2015learning,
  author       = {Fredrik D. Johansson and
                  Devdatt P. Dubhashi},
  editor       = {Longbing Cao and
                  Chengqi Zhang and
                  Thorsten Joachims and
                  Geoffrey I. Webb and
                  Dragos D. Margineantu and
                  Graham Williams},
  title        = {Learning with Similarity Functions on Graphs using Matchings of Geometric
                  Embeddings},
  booktitle    = {Proceedings of the 21th {ACM} {SIGKDD} International Conference on
                  Knowledge Discovery and Data Mining, Sydney, NSW, Australia, August
                  10-13, 2015},
  pages        = {467--476},
  publisher    = {{ACM}},
  year         = {2015},
  url          = {https://doi.org/10.1145/2783258.2783341},
  doi          = {10.1145/2783258.2783341},
  bibsource    = {dblp computer science bibliography, https://dblp.org}
}

@inproceedings{johansson2014global,
  author       = {Fredrik D. Johansson and
                  Vinay Jethava and
                  Devdatt P. Dubhashi and
                  Chiranjib Bhattacharyya},
  title        = {Global graph kernels using geometric embeddings},
  booktitle    = {Proceedings of the 31th International Conference on Machine Learning,
                  {ICML} 2014, Beijing, China, 21-26 June 2014},
  series       = {{JMLR} Workshop and Conference Proceedings},
  volume       = {32},
  pages        = {694--702},
  publisher    = {JMLR.org},
  year         = {2014},
  url          = {http://proceedings.mlr.press/v32/johansson14.html},
  bibsource    = {dblp computer science bibliography, https://dblp.org}
}

@article{oneto2017measuring,
  author       = {Luca Oneto and
                  Nicol{\`{o}} Navarin and
                  Michele Donini and
                  Alessandro Sperduti and
                  Fabio Aiolli and
                  Davide Anguita},
  title        = {Measuring the expressivity of graph kernels through Statistical Learning
                  Theory},
  journal      = {Neurocomputing},
  volume       = {268},
  pages        = {4--16},
  year         = {2017},
  url          = {https://doi.org/10.1016/j.neucom.2017.02.088},
  doi          = {10.1016/J.NEUCOM.2017.02.088},
  bibsource    = {dblp computer science bibliography, https://dblp.org}
}

@inproceedings{morris2017glocalized,
  author       = {Christopher Morris and
                  Kristian Kersting and
                  Petra Mutzel},
  editor       = {Vijay Raghavan and
                  Srinivas Aluru and
                  George Karypis and
                  Lucio Miele and
                  Xindong Wu},
  title        = {Glocalized Weisfeiler-Lehman Graph Kernels: Global-Local Feature Maps
                  of Graphs},
  booktitle    = {2017 {IEEE} International Conference on Data Mining, {ICDM} 2017,
                  New Orleans, LA, USA, November 18-21, 2017},
  pages        = {327--336},
  publisher    = {{IEEE} Computer Society},
  year         = {2017},
  url          = {https://doi.org/10.1109/ICDM.2017.42},
  doi          = {10.1109/ICDM.2017.42},
  bibsource    = {dblp computer science bibliography, https://dblp.org}
}

@inproceedings{hido2009a,
  author       = {Shohei Hido and
                  Hisashi Kashima},
  editor       = {Wei Wang and
                  Hillol Kargupta and
                  Sanjay Ranka and
                  Philip S. Yu and
                  Xindong Wu},
  title        = {A Linear-Time Graph Kernel},
  booktitle    = {{ICDM} 2009, The Ninth {IEEE} International Conference on Data Mining,
                  Miami, Florida, USA, 6-9 December 2009},
  pages        = {179--188},
  publisher    = {{IEEE} Computer Society},
  year         = {2009},
  url          = {https://doi.org/10.1109/ICDM.2009.30},
  doi          = {10.1109/ICDM.2009.30},
  bibsource    = {dblp computer science bibliography, https://dblp.org}
}

@inproceedings{woznica2010adaptive,
  author       = {Adam Woznica and
                  Alexandros Kalousis and
                  Melanie Hilario},
  editor       = {Mohammed Javeed Zaki and
                  Jeffrey Xu Yu and
                  Balaraman Ravindran and
                  Vikram Pudi},
  title        = {Adaptive Matching Based Kernels for Labelled Graphs},
  booktitle    = {Advances in Knowledge Discovery and Data Mining, 14th Pacific-Asia
                  Conference, {PAKDD} 2010, Hyderabad, India, June 21-24, 2010. Proceedings.
                  Part {II}},
  series       = {Lecture Notes in Computer Science},
  volume       = {6119},
  pages        = {374--385},
  publisher    = {Springer},
  year         = {2010},
  url          = {https://doi.org/10.1007/978-3-642-13672-6\_37},
  doi          = {10.1007/978-3-642-13672-6\_37},
  bibsource    = {dblp computer science bibliography, https://dblp.org}
}

@inproceedings{horvath2004cyclic,
  author       = {Tam{\'{a}}s Horv{\'{a}}th and
                  Thomas G{\"{a}}rtner and
                  Stefan Wrobel},
  editor       = {Won Kim and
                  Ron Kohavi and
                  Johannes Gehrke and
                  William DuMouchel},
  title        = {Cyclic pattern kernels for predictive graph mining},
  booktitle    = {Proceedings of the Tenth {ACM} {SIGKDD} International Conference on
                  Knowledge Discovery and Data Mining, Seattle, Washington, USA, August
                  22-25, 2004},
  pages        = {158--167},
  publisher    = {{ACM}},
  year         = {2004},
  url          = {https://doi.org/10.1145/1014052.1014072},
  doi          = {10.1145/1014052.1014072},
  bibsource    = {dblp computer science bibliography, https://dblp.org}
}

@inproceedings{costa2010fast,
  author       = {Fabrizio Costa and
                  Kurt De Grave},
  editor       = {Johannes F{\"{u}}rnkranz and
                  Thorsten Joachims},
  title        = {Fast Neighborhood Subgraph Pairwise Distance Kernel},
  booktitle    = {Proceedings of the 27th International Conference on Machine Learning
                  (ICML-10), June 21-24, 2010, Haifa, Israel},
  pages        = {255--262},
  publisher    = {Omnipress},
  year         = {2010},
  url          = {https://icml.cc/Conferences/2010/papers/347.pdf},
  bibsource    = {dblp computer science bibliography, https://dblp.org}
}

@inproceedings{borgwardt2005shortest,
  author       = {Karsten M. Borgwardt and
                  Hans{-}Peter Kriegel},
  title        = {Shortest-Path Kernels on Graphs},
  booktitle    = {Proceedings of the 5th {IEEE} International Conference on Data Mining
                  {(ICDM} 2005), 27-30 November 2005, Houston, Texas, {USA}},
  pages        = {74--81},
  publisher    = {{IEEE} Computer Society},
  year         = {2005},
  url          = {https://doi.org/10.1109/ICDM.2005.132},
  doi          = {10.1109/ICDM.2005.132},
  bibsource    = {dblp computer science bibliography, https://dblp.org}
}

@inproceedings{hermansson2015generalized,
  author       = {Linus Hermansson and
                  Fredrik D. Johansson and
                  Osamu Watanabe},
  editor       = {Nathalie Japkowicz and
                  Stan Matwin},
  title        = {Generalized Shortest Path Kernel on Graphs},
  booktitle    = {Discovery Science - 18th International Conference, {DS} 2015, Banff,
                  AB, Canada, October 4-6, 2015, Proceedings},
  series       = {Lecture Notes in Computer Science},
  volume       = {9356},
  pages        = {78--85},
  publisher    = {Springer},
  year         = {2015},
  url          = {https://doi.org/10.1007/978-3-319-24282-8\_8},
  doi          = {10.1007/978-3-319-24282-8\_8},
  bibsource    = {dblp computer science bibliography, https://dblp.org}
}

@article{vishwanathan2010graph,
  author       = {S. V. N. Vishwanathan and
                  Nicol N. Schraudolph and
                  Risi Kondor and
                  Karsten M. Borgwardt},
  title        = {Graph Kernels},
  journal      = {J. Mach. Learn. Res.},
  volume       = {11},
  pages        = {1201--1242},
  year         = {2010},
  url          = {https://dl.acm.org/doi/10.5555/1756006.1859891},
  doi          = {10.5555/1756006.1859891},
  bibsource    = {dblp computer science bibliography, https://dblp.org}
}

@inproceedings{kriege2014explicit,
  author       = {Nils M. Kriege and
                  Marion Neumann and
                  Kristian Kersting and
                  Petra Mutzel},
  editor       = {Ravi Kumar and
                  Hannu Toivonen and
                  Jian Pei and
                  Joshua Zhexue Huang and
                  Xindong Wu},
  title        = {Explicit Versus Implicit Graph Feature Maps: {A} Computational Phase
                  Transition for Walk Kernels},
  booktitle    = {2014 {IEEE} International Conference on Data Mining, {ICDM} 2014,
                  Shenzhen, China, December 14-17, 2014},
  pages        = {881--886},
  publisher    = {{IEEE} Computer Society},
  year         = {2014},
  url          = {https://doi.org/10.1109/ICDM.2014.129},
  doi          = {10.1109/ICDM.2014.129},
  bibsource    = {dblp computer science bibliography, https://dblp.org}
}

@article{kriege2019aunifying,
  author       = {Nils M. Kriege and
                  Marion Neumann and
                  Christopher Morris and
                  Kristian Kersting and
                  Petra Mutzel},
  title        = {A unifying view of explicit and implicit feature maps of graph kernels},
  journal      = {Data Min. Knowl. Discov.},
  volume       = {33},
  number       = {6},
  pages        = {1505--1547},
  year         = {2019},
  url          = {https://doi.org/10.1007/s10618-019-00652-0},
  doi          = {10.1007/S10618-019-00652-0},
  bibsource    = {dblp computer science bibliography, https://dblp.org}
}

@article{cai2018asimple,
  author       = {Chen Cai and
                  Yusu Wang},
  title        = {A simple yet effective baseline for non-attribute graph classification},
  journal      = {CoRR},
  volume       = {abs/1811.03508},
  year         = {2018},
  url          = {http://arxiv.org/abs/1811.03508},
  eprinttype    = {arXiv},
  eprint       = {1811.03508},
  bibsource    = {dblp computer science bibliography, https://dblp.org}
}

@article{hornik1989multilayer,
  author       = {Kurt Hornik and
                  Maxwell B. Stinchcombe and
                  Halbert White},
  title        = {Multilayer feedforward networks are universal approximators},
  journal      = {Neural Networks},
  volume       = {2},
  number       = {5},
  pages        = {359--366},
  year         = {1989},
  url          = {https://doi.org/10.1016/0893-6080(89)90020-8},
  doi          = {10.1016/0893-6080(89)90020-8},
  bibsource    = {dblp computer science bibliography, https://dblp.org}
}

@article{fey2019fast,
  author       = {Matthias Fey and
                  Jan Eric Lenssen},
  title        = {Fast Graph Representation Learning with PyTorch Geometric},
  journal      = {CoRR},
  volume       = {abs/1903.02428},
  year         = {2019},
  url          = {http://arxiv.org/abs/1903.02428},
  eprinttype    = {arXiv},
  eprint       = {1903.02428},
  bibsource    = {dblp computer science bibliography, https://dblp.org}
}

@book{horn1991topics,
  author       = {Roger A. Horn and
                  Charles R. Johnson},
  title        = {Topics in Matrix Analysis},
  publisher    = {Cambridge University Press},
  year         = {1991},
  url          = {https://doi.org/10.1017/CBO9780511840371},
  doi          = {10.1017/CBO9780511840371},
  isbn         = {978-0-521-46713-1},
  bibsource    = {dblp computer science bibliography, https://dblp.org}
}

@inproceedings{sanders2023curvature,
  author       = {Cedric Sanders and
                  Andreas Roth and
                  Thomas Liebig},
  editor       = {Rosa Meo and
                  Fabrizio Silvestri},
  title        = {Curvature-Based Pooling Within Graph Neural Networks},
  booktitle    = {Machine Learning and Principles and Practice of Knowledge Discovery
                  in Databases - International Workshops of {ECML} {PKDD} 2023, Turin,
                  Italy, September 18-22, 2023, Revised Selected Papers, Part {III}},
  series       = {Communications in Computer and Information Science},
  volume       = {2135},
  pages        = {471--485},
  publisher    = {Springer},
  year         = {2023},
  url          = {https://doi.org/10.1007/978-3-031-74633-8\_35},
  doi          = {10.1007/978-3-031-74633-8\_35},
  bibsource    = {dblp computer science bibliography, https://dblp.org}
}

@article{lachi2025expressive,
title={Expressive Pooling for Graph Neural Networks},
author={Veronica Lachi and Alice Moallemy-Oureh and Andreas Roth and Pascal Welke},
journal={Transactions on Machine Learning Research},
year={2025},
url={https://openreview.net/forum?id=xGADInGWMt},
note={}
}

@article{roth2021adata,
  author       = {Andreas Roth and
                  Konstantin W{\"{u}}stefeld and
                  Frank Weichert},
  title        = {A Data-Centric Augmentation Approach for Disturbed Sensor Image Segmentation},
  journal      = {J. Imaging},
  volume       = {7},
  number       = {10},
  pages        = {206},
  year         = {2021},
  url          = {https://doi.org/10.3390/jimaging7100206},
  doi          = {10.3390/JIMAGING7100206},
  bibsource    = {dblp computer science bibliography, https://dblp.org}
}

@article{bouritsas2023improving,
  author       = {Giorgos Bouritsas and
                  Fabrizio Frasca and
                  Stefanos Zafeiriou and
                  Michael M. Bronstein},
  title        = {Improving Graph Neural Network Expressivity via Subgraph Isomorphism
                  Counting},
  journal      = {{IEEE} Trans. Pattern Anal. Mach. Intell.},
  volume       = {45},
  number       = {1},
  pages        = {657--668},
  year         = {2023},
  url          = {https://doi.org/10.1109/TPAMI.2022.3154319},
  doi          = {10.1109/TPAMI.2022.3154319},
  bibsource    = {dblp computer science bibliography, https://dblp.org}
}

@article{bengio1994learning,
  author       = {Yoshua Bengio and
                  Patrice Y. Simard and
                  Paolo Frasconi},
  title        = {Learning long-term dependencies with gradient descent is difficult},
  journal      = {{IEEE} Trans. Neural Networks},
  volume       = {5},
  number       = {2},
  pages        = {157--166},
  year         = {1994},
  url          = {https://doi.org/10.1109/72.279181},
  doi          = {10.1109/72.279181},
  bibsource    = {dblp computer science bibliography, https://dblp.org}
}

@inproceedings{pascanu2013on,
  author       = {Razvan Pascanu and
                  Tom{\'{a}}s Mikolov and
                  Yoshua Bengio},
  title        = {On the difficulty of training recurrent neural networks},
  booktitle    = {Proceedings of the 30th International Conference on Machine Learning,
                  {ICML} 2013, Atlanta, GA, USA, 16-21 June 2013},
  series       = {{JMLR} Workshop and Conference Proceedings},
  volume       = {28},
  pages        = {1310--1318},
  publisher    = {JMLR.org},
  year         = {2013},
  url          = {http://proceedings.mlr.press/v28/pascanu13.html},
  bibsource    = {dblp computer science bibliography, https://dblp.org}
}

@article{hochreiter1997long,
  author       = {Sepp Hochreiter and
                  J{\"{u}}rgen Schmidhuber},
  title        = {Long Short-Term Memory},
  journal      = {Neural Comput.},
  volume       = {9},
  number       = {8},
  pages        = {1735--1780},
  year         = {1997},
  url          = {https://doi.org/10.1162/neco.1997.9.8.1735},
  doi          = {10.1162/NECO.1997.9.8.1735},
  bibsource    = {dblp computer science bibliography, https://dblp.org}
}

@book{goodfellow2016deep,
  author       = {Ian J. Goodfellow and
                  Yoshua Bengio and
                  Aaron C. Courville},
  title        = {Deep Learning},
  series       = {Adaptive computation and machine learning},
  publisher    = {{MIT} Press},
  year         = {2016},
  url          = {http://www.deeplearningbook.org/},
  isbn         = {978-0-262-03561-3},
  bibsource    = {dblp computer science bibliography, https://dblp.org}
}

@article{luks1982isomorphism,
  author       = {Eugene M. Luks},
  title        = {Isomorphism of Graphs of Bounded Valence can be Tested in Polynomial
                  Time},
  journal      = {J. Comput. Syst. Sci.},
  volume       = {25},
  number       = {1},
  pages        = {42--65},
  year         = {1982},
  url          = {https://doi.org/10.1016/0022-0000(82)90009-5},
  doi          = {10.1016/0022-0000(82)90009-5},
  bibsource    = {dblp computer science bibliography, https://dblp.org}
}

@article{cortes1995support,
  author       = {Corinna Cortes and
                  Vladimir Vapnik},
  title        = {Support-Vector Networks},
  journal      = {Mach. Learn.},
  volume       = {20},
  number       = {3},
  pages        = {273--297},
  year         = {1995},
  url          = {https://doi.org/10.1007/BF00994018},
  doi          = {10.1007/BF00994018},
  bibsource    = {dblp computer science bibliography, https://dblp.org}
}

@article{henaff2015deep,
  author       = {Mikael Henaff and
                  Joan Bruna and
                  Yann LeCun},
  title        = {Deep Convolutional Networks on Graph-Structured Data},
  journal      = {CoRR},
  volume       = {abs/1506.05163},
  year         = {2015},
  url          = {http://arxiv.org/abs/1506.05163},
  eprinttype    = {arXiv},
  eprint       = {1506.05163},
  bibsource    = {dblp computer science bibliography, https://dblp.org}
}

@article{lecun1989backpropagation,
  author       = {Yann LeCun and
                  Bernhard E. Boser and
                  John S. Denker and
                  Donnie Henderson and
                  Richard E. Howard and
                  Wayne E. Hubbard and
                  Lawrence D. Jackel},
  title        = {Backpropagation Applied to Handwritten Zip Code Recognition},
  journal      = {Neural Comput.},
  volume       = {1},
  number       = {4},
  pages        = {541--551},
  year         = {1989},
  url          = {https://doi.org/10.1162/neco.1989.1.4.541},
  doi          = {10.1162/NECO.1989.1.4.541},
  bibsource    = {dblp computer science bibliography, https://dblp.org}
}

@Article{fukushima1980neocognitron,
author={Fukushima, Kunihiko},
title={Neocognitron: A self-organizing neural network model for a mechanism of pattern recognition unaffected by shift in position},
journal={Biological Cybernetics},
year={1980},
month={4},
day={01},
volume={36},
number={4},
pages={193-202},
doi={10.1007/BF00344251},
}

@book{dudley2022real, place={Cambridge}, edition={2}, series={Cambridge Studies in Advanced Mathematics}, title={Real Analysis and Probability}, publisher={Cambridge University Press}, author={Dudley, R. M.}, year={2002}, collection={Cambridge Studies in Advanced Mathematics}}
\newpage\hbox{}\thispagestyle{empty}\newpage
\cleardoublepage
	
\end{document}